\documentclass[final,3p,times,compress,linktoc=all]{elsarticle}

\graphicspath{{./images/}{../images/}}

\renewcommand{\arraystretch}{1.2}   

\makeatletter
\def\ps@pprintTitle{%
  \let\@oddhead\@empty
  \let\@evenhead\@empty
  \def\@oddfoot{\reset@font\hfil\thepage\hfil}
  \let\@evenfoot\@oddfoot
}
\makeatother

\usepackage[T1]{fontenc}
\usepackage[utf8]{inputenc}
\usepackage[english]{babel}

\usepackage{amsmath} 
\usepackage{amssymb} 
\usepackage{bm}      
\usepackage{bbm}     

\usepackage{array}
\usepackage{booktabs}
\usepackage{tabularx}  
\usepackage[flushleft]{threeparttable} 
\usepackage{multirow}
\newcolumntype{C}[1]{>{\centering\arraybackslash}p{#1}}
\newcolumntype{L}[1]{>{\raggedright\arraybackslash}p{#1}}
\newcolumntype{R}[1]{>{\raggedleft\arraybackslash}p{#1}}

\usepackage{lineno}
\usepackage{multicol}
\usepackage[most]{tcolorbox} 
\usepackage{enumitem}     
\usepackage{acronym}     

\usepackage{caption}

\usepackage{changepage} 

\usepackage{graphicx}
\usepackage[section]{placeins}

\usepackage{multibib}
\usepackage{orcidlink}
\usepackage{setspace}

\usepackage{subcaption}

\usepackage[normalem]{ulem}
\useunder{\uline}{\ul}{}

\usepackage{relsize}

\usepackage{xcolor}

\definecolor{mybg}{HTML}{f1f2f6}   
\definecolor{mybg2}{HTML}{ffffff}   

\definecolor{primordial}{HTML}{d1495b}
\definecolor{mytitleAS}{HTML}{d1495b}
\definecolor{mytitleCE}{HTML}{d1495b}
\definecolor{mytitleCL}{HTML}{d1495b}
\definecolor{mytitleMC}{HTML}{d1495b}
\definecolor{mytitleMS}{HTML}{d1495b}
\definecolor{mytitleQL}{HTML}{d1495b}
\definecolor{mytitleSN}{HTML}{d1495b}
\definecolor{knowledge}{HTML}{00798c}
\definecolor{mytitleKN}{HTML}{00798c}
\definecolor{others}{HTML}{30638e}
\definecolor{mytitleVO}{HTML}{30638e}
\definecolor{mytitleAT}{HTML}{30638e}
\definecolor{nd}{HTML}{edae49}
\definecolor{mytitleUG}{HTML}{edae49}
\definecolor{RQ}{HTML}{456990}

\usepackage{xspace}

\newcommand{\xaxis}{{$x$-axis}\xspace}
\newcommand{\yaxis}{{$y$-axis}\xspace}

\usepackage{relsize}

\newcommand{\model}[1]{{\small\textsf{#1}}}
\newcommand{\modelc}[1]{{\scriptsize\textsf{#1}}}
\newcommand{\ds}[1]{{\small\textsf{#1}}}
\newcommand{\dsc}[1]{{\scriptsize\textsf{#1}}}
\newcommand{\dm}[1]{{\tt\small{#1}}}
\newcommand{\dmc}[1]{{\tt\scriptsize{#1}}}

\newcommand{\rubricfont}{\fontsize{6pt}{8pt}\selectfont}

\definecolor{darkblue}{RGB}{0,0,139}
\definecolor{darkred}{RGB}{139,0,0}
\definecolor{grey}{RGB}{128,128,128}
\definecolor{darkgreen}{RGB}{0,139,0}

\newcommand{\removed}[1]{}

\newcommand{\DeLeAn}{{\sf\relsize{-1}DeLeAn}\xspace} 
\newcommand{\ADeLe}{{\sf\relsize{-1}ADeLe}\xspace}
\newcommand{\ADeLeLight}{{\sf\relsize{-1}ADeLe-Light}\xspace}

\newcommand{\dimension}[1]{{\tt\relsize{-0.5}{#1}}\xspace}

\usepackage[utf8]{inputenc}
\usepackage{tocloft}

\title{General Scales Unlock AI Evaluation with Explanatory and Predictive Power}
\author{%
\textbf{Lexin Zhou}$^{1,2,3}$ \quad 
\textbf{Lorenzo Pacchiardi}$^{1}$ \quad 
\textbf{Fernando Martínez-Plumed}$^{3}$ \quad 
\textbf{Katherine M. Collins}$^{4}$\\[0.8ex]
\textbf{Yael Moros-Daval}$^{3}$ \quad 
\textbf{Seraphina Zhang}$^{1,5}$ \quad
\textbf{Qinlin Zhao}$^{2}$ \quad 
\textbf{Yitian Huang}$^{2}$ \quad 
\textbf{Luning Sun}$^{6}$ \\[0.8ex]
\textbf{Jonathan E. Prunty}$^{1}$ \quad 
\textbf{Zongqian Li}$^{7}$ \quad 
\textbf{Pablo Sánchez-García}$^{8}$\quad 
\textbf{Kexin Jiang Chen}$^{3}$ \\[0.8ex]
\textbf{Pablo A. M. Casares}$^{3}$ \quad 
\textbf{Jiyun Zu}$^{9}$ \quad 
\textbf{John Burden}$^{1}$ \quad 
\textbf{Behzad Mehrbakhsh}$^{3}$ \quad 
\textbf{David Stillwell}$^{6}$ \\[0.8ex]
\textbf{Manuel Cebrian}$^{10}$ \quad 
\textbf{Jindong Wang}$^{11}$\quad 
\textbf{Peter Henderson}$^{12}$ \quad 
\textbf{Sherry Tongshuang Wu}$^{13}$ \\[0.8ex]
\textbf{Patrick C. Kyllonen}$^{9}$ \quad 
\textbf{Lucy Cheke}$^{1,5}$ \quad 
\textbf{Xing Xie}$^{2}$ \quad 
\textbf{José Hernández-Orallo}$^{1,3,14}$
}

\fntext[1]{Leverhulme Centre for the Future of Intelligence, University of Cambridge, UK}
\fntext[2]{Microsoft Research Asia}
\fntext[3]{Valencian Research Institute for Artificial Intelligence (VRAIN), Universitat Politècnica de València, Spain}
\fntext[4]{Department of Engineering, University of Cambridge, UK}
\fntext[5]{Department of Psychology, University of Cambridge, UK}
\fntext[6]{The Psychometrics Centre, University of Cambridge, UK}
\fntext[7]{Department of Theoretical and Applied Linguistics, University of Cambridge, UK}
\fntext[8]{KU Leuven, Belgium}
\fntext[9]{Educational Testing Service, US}
\fntext[10]{Center for Automation and Robotics, Spanish National Research Council, Madrid, Spain}
\fntext[11]{William \& Mary, US}
\fntext[12]{Princeton University, US}
\fntext[13]{Carnegie Mellon University, US}

\fntext[labelC]{Corresponding author, including requests for materials: \emph{jorallo@upv.es}} 

\begin{document}
\begin{frontmatter}
\begin{abstract}
Ensuring safe and effective use of AI requires understanding and anticipating its performance on novel tasks, from advanced scientific challenges to transformed workplace activities.
So far,  
benchmarking has guided progress in AI, but it has offered limited explanatory and predictive power for general-purpose AI systems, given the low transferability across diverse tasks. In this paper, we introduce general scales 
for AI evaluation 
that can explain what common AI benchmarks really measure, extract ability profiles of AI systems, and predict their performance for new task instances, in- and out-of-distribution. 
Our fully-automated methodology builds on 
18 newly-crafted rubrics that place instance demands on general scales that do not saturate. 
Illustrated for 15 large language models and 63 tasks, high explanatory power is unleashed from inspecting the demand and ability profiles, bringing insights 
on the sensitivity and specificity exhibited by different benchmarks, and how knowledge, metacognition and reasoning are affected by model size, chain-of-thought and distillation. 
Surprisingly, 
high predictive power at the instance level becomes possible using these demand levels, providing superior estimates over black-box baseline predictors based on embeddings or finetuning, especially in out-of-distribution settings (new tasks and new benchmarks). 
The scales, rubrics, battery, techniques and results presented here 
represent a major step for 
AI evaluation,  
underpinning the reliable deployment of AI in the years ahead. \textbf{(Collaborative platform: \url{https://kinds-of-intelligence-cfi.github.io/ADELE}.)}


\end{abstract}
\end{frontmatter}

\clearpage

\vspace{-5cm} 
\tableofcontents  

\newpage

\setcounter{footnote}{0}

\section{Introduction}

Current general-purpose AI systems, such as large language models (LLMs) and other foundation models, are highly unreliable and unpredictable; they may 
succeed in solving extremely challenging college-level mathematical problems, yet paradoxically struggle with some basic arithmetic operations  \cite{zhou2024predictable,zhou2024larger}. This places a large burden on AI evaluation in terms of explanatory and predictive power: we need to understand where the AI system is failing and anticipate where it can be applied successfully. Given the diversity of tasks general-purpose AI systems deal with, comprehensive explanatory power cannot be based on  features specific to each individual task (be they based on text, image, or other modality), but must be based instead on a set of {\em general capabilities} that are meaningful for humans. 
Moreover, as each small variation in the instantiation of a task
may lead to very different outcomes, the predictive power in the real world must take place for each new {\em instance}. The development of AI evaluation methodologies that make both explanatory and predictive power compatible has been elusive so far.

The traditional performance-oriented evaluation approach has shown limited predictive power {\em at the instance level}, inside or outside the benchmark \cite{burnell2023rethink,eriksson2025can}. If DeepSeek-R1 achieves 79.8\% average performance \cite{guo2025deepseek} on a popular mathematical benchmark such as AIME\footnote{The American Invitational Mathematics Examination dataset: \url{https://maa.org/maa-invitational-competitions/}.}, we cannot 
 make informed estimates of success on individual items sampled from that benchmark, even if one is known to be very easy and the other to be very difficult. This 79.8\% is hardly informative about individual instances in or out of the distribution, and even the average extrapolates poorly to other mathematical benchmarks. Indeed, these aggregate scores are a function of the benchmark and the AI system, {\em not} capability estimates that really inform us about what the system can or cannot do. 
%
Instead of aggregating performance, other evaluation paradigms estimate some properties of the subject (a human or an AI system)\footnote{We use the term system or subject for what is being evaluated and the term instance or item for each individual question or problem in a test or benchmark. Definitions for these terms, and others such as ability, contamination, ratio scale, etc., are detailed in appendix \ref{sec:terminology}.} which, jointly with some properties of the item (the instance), can predict performance.  
Several techniques from psychometrics and other behavioural sciences have been applied to AI evaluation \cite{wang2023evaluating}, such as factor analysis \cite{burnell2023revealing,ilic2024evidence} and item response theory \cite{martinez2019item}. However, the extracted factors or parameters are not easily interpretable and strongly depend on the employed population of systems and benchmarks.  
`Assessors' \cite{hernandez2022training,schellaert2025}, related to uncertainty estimation and calibration methods \cite{schellaert2025}, are score prediction models that can anticipate performance at the instance level, and can be applied to new tasks, via latent features. However, these features are difficult to interpret, and 
typically extrapolate poorly 
out of distribution \cite{schellaert2025analysing,pacchiardi2024100instancesneedpredicting,ye2021towards,srinivasan2021predicting,ahuja2022beyond}.  
In contrast, when the features are engineered by humans, in more 
cognitively-inspired approaches, we can derive explanatory capability profiles \cite{burden2023inferring}. However, the 
scalability of the approach is compromised by the need for experts that develop the cognitive models and annotate the testing items.

These perspectives differ in their core methodologies regarding what is measured and how \cite{burden2025paradigmsaievaluationmapping}, but they all have grappled with balancing explanatory depth and predictive power. Also, most of these frameworks derive features, parameters or scales that are regularly  saturated by an extremely volatile space of AI systems and benchmarks that is constantly replaced  \cite{schlangen2019language,zellers2019hellaswag}. Aiming for ecologically-valid assessment in the real world  \cite{stahl2023systematic,burnell2023rethink,rauh2024gaps}, would increase the challenge 
but most current issues in AI evaluation \cite{hernandez2020ai,kadian2020sim2real,burden2024evaluating} already occur in a narrow but very common kind of evaluation based on benchmarks \cite{hardy2024more,eriksson2025can,burden2025paradigmsaievaluationmapping}.  Solving these issues is a prerequisite for more robust assessment in the real world, such as interactive, subjective and adaptive evaluations \cite{leeevaluating, collins2024evaluating,cohn2023dialectical}. 

While we do not address all problems in all kinds of evaluation
, we present a new methodology that can accompany, map and inform AI progress, regulation and deployment in the following decades. First, it introduces 18 open scales in the range $(0,\infty)$, whose values---the demand levels for each dimension---are obtained through  18 carefully crafted  
demand-level-annotation (\DeLeAn) rubrics, which humans can understand and apply to any testing instance. For scalability, they can be applied robustly by an LLM to existing or new benchmarks and tasks. The obtained demand levels are robust to scale saturation by progress in AI or to alterations in the difficulty of the instances, a recurrent problem in AI evaluation 
\cite{schlangen2019language,zellers2019hellaswag,srivastava2022beyond,suzgun2022challenging,kazemi2025big}. 
By running the rubrics through a collection of common AI benchmarks, we obtain the annotated-demand-levels (\ADeLe) battery, whose histograms of demand levels reveal the sensitivity and specificity of each benchmark compared to what they should measure. By using \ADeLe on an LLM, we get 18 characteristic curves and hence 18 ability estimates that can be summarised in an ability profile, completely independent of other LLMs. We hence obtain demand levels and ability levels that are not specific to any population of systems and benchmarks. Finally, the demand levels (plus an extra unguessability level) can be leveraged to build an assessor
with high predictive power for unseen instances in- and out-of-distribution (new tasks and benchmarks).

In a nutshell, this allows for {\em causal} explanation and prediction for both instances and systems. For example, if an AI system has a profile with four abilities (1,2,4,2), we can {\em predict} failure at an instance with a demand profile of (0,4,1,0), and {\em explain} the failure by noting that the system's second ability is insufficient. Furthermore, we can perform counterfactual analyses, such as exploring scenarios where the system's ability is enhanced or the task's demand is reduced. 
Extending this illustrative example, our methodology enables the following possibilities:

\begin{enumerate}
\item We can {\bf 
carve 
the space of capabilities 
into a  hierarchical set of general scales}.
The \DeLeAn rubrics (see Table~\ref{tab:dimensions})
are applied systematically to the 16,108 instances of 
the \ADeLe battery (see Table~\ref{tab:selected_benchmarks_and_tasks}), yielding 289,944 annotations.
 The clarity of the rubrics is validated by how well the LLM annotates scores in agreement with humans. The moderate correlations between dimensions (Figure~\ref{fig:demand_correlation}) suggest potentially distinctive capabilities, with instances that differ on any pair of capabilities, and correlations being consistent with the hierarchy. 
\item We can {\bf  explain what common benchmarks really measure}. In particular, we first discover various demands in extraneous dimensions such as the \textit{atypicality} (from common to unique), \textit{volume} (from small to large) and \textit{unguessability} (from multiple-choice to open-ended), suggesting 
contamination\footnote{{Overestimation because similar data was seen during training \cite{roberts2023data}.}}, amalgamation\footnote{{Underestimation because examples are made more difficult by agglomerating more things to the task \citep{Levy2024SameTM}.}} or funnelling\footnote{{Under or over estimation by changing the difficulty of a task by reducing or increasing options \cite{wang2024mmlu} or distractors.}}, 
 respectively 
(see Figure~\ref{fig:demand_correlation} and Table~\ref{tab:main-predictability-results-only-AT-UG-VO}). Beyond these effects, many benchmarks lack either specificity or sensitivity: they do not have a minimum number of instances of all demands for the dimensions their designers claimed they measure, and they include non-zero demands on other dimensions. 
See Figure~\ref{fig:benchmark_demand_profiles}.
\item We can {\bf explain what AI systems can do}. In our experiments with three families of LLMs, we find that the ability scores at knowledge dimensions are mostly determined by model size (scaling and distillation), while quantitative and logical reasoning, learning and abstraction, and perhaps surprisingly, mind modelling and social capabilities are boosted in chain-of-thought, inference-heavy models (such as OpenAI's O1 and DeepSeek's R1-Distill). See Figures~\ref{fig:characteristic_curves_combined} and \ref{fig:model_capability_profiles}. Consistent results can be obtained with a small sample of the \ADeLe battery (\ADeLeLight, Figure \ref{fig:model_capability_profiles_light}) that mostly removes instances with redundant demand profiles.
\item We can {\bf predict performance for new task instances}. High predictive power 
at the instance level is possible, superior than black-box assessor baselines based on embeddings or fine-tuning, especially in out-of-distribution settings (new tasks and new benchmarks). This opens up a range of applications, such as better routing methods to choose what model to use \cite{ong2024routellm}, safety operating areas where assurance is guaranteed \cite{zhou2024predictable} and anticipatory reject rules when harm or cost is anticipated \cite{zhou2022reject,predictaboard}. 
See Tables~\ref{tab:main-id-predictability-results}, ~\ref{tab:main-task-ood-predictability-results} and ~\ref{tab:main-benchmark-ood-predictability-results}, and Figure~\ref{fig:average_feature_importance}.
\end{enumerate}

\noindent 

\begin{figure}[!ht]
  \centering
  \includegraphics[width=0.9\linewidth]{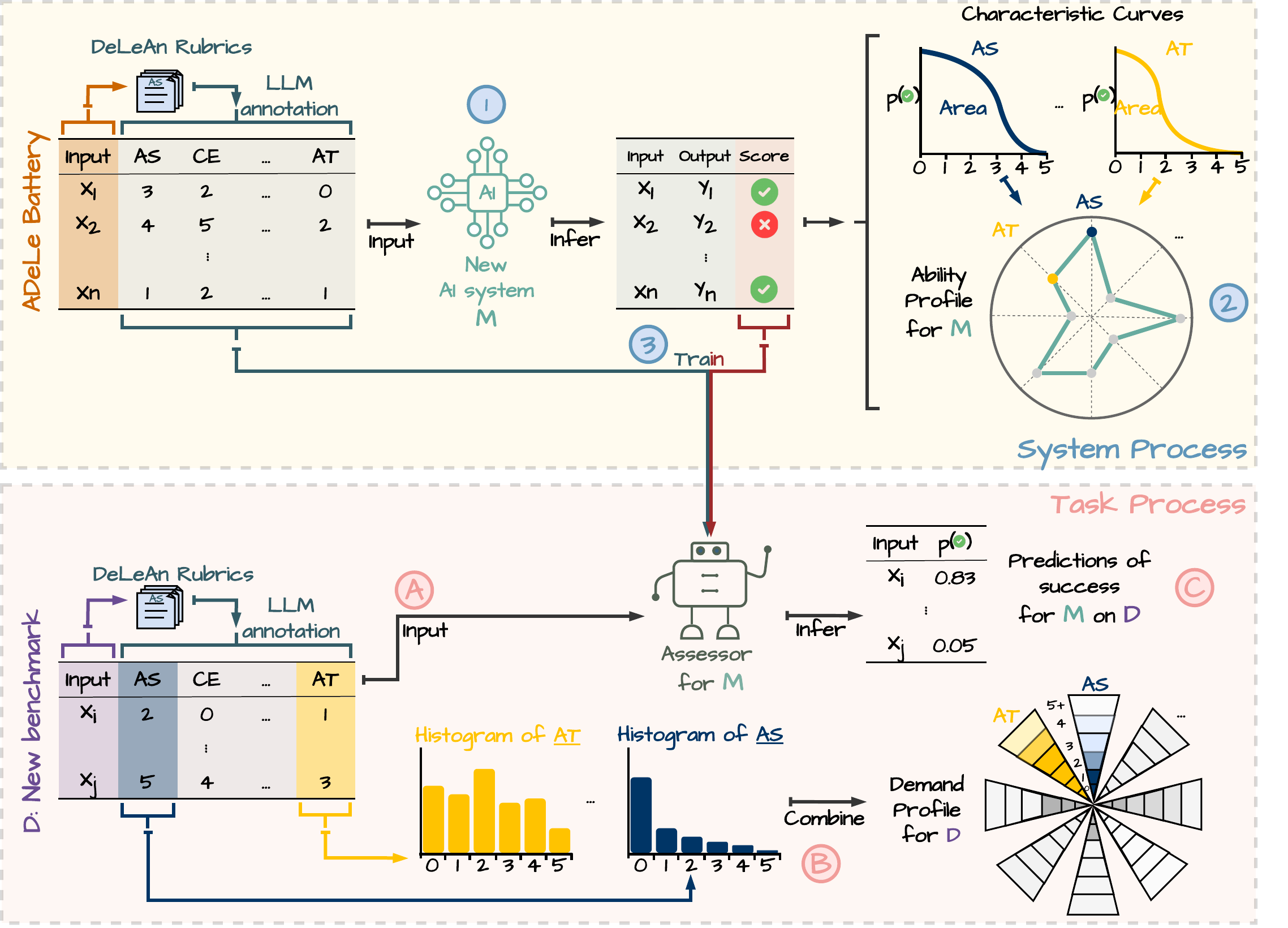}
  \caption{Processes to explain and predict performance for new systems and benchmarks. Top: ``System Process'': Steps for each new AI system: (1) Run the new system on the annotated-demand-levels (\ADeLe) Battery,  
  (2)  Plot characteristic curves for all dimensions and extract the ability profile for the system, and, optionally, (3) Train a simple assessor using the annotated levels as inputs and the score as output. Bottom: ``Task Process'': Steps for each new task or benchmark: (A) Apply the demand-level-annotation (\DeLeAn) rubrics to the new tasks using a standard LLM, 
(B) Get demand histograms and demand profiles that explain what demands the tasks require, and, optionally,  
(C) Predict performance for the new tasks for any system that has built an assessor after the ``System Process''. Assessors based on the demand profile have especially higher predictive power in out-of-distribution settings than other baseline assessors, anticipating validity in novel situations.
}
  \label{fig:methodology}
\end{figure}

\noindent Figure~\ref{fig:methodology} illustrates our methodology, with two processes which can be followed independently. If we have a new AI system we want to explain or predict about, we will undergo the ``System Process'' (top): running the model on the annotated-demand-levels (\ADeLe) battery, plotting characteristic curves (see, e.g., Figure \ref{fig:oneSCC}) and summarising the profile of abilities with a radial plot as 
in Figure \ref{fig:model_capability_profiles}. 
If we want to analyse a new task instance or benchmark, we will take the ``Task process'' (bottom), 
in which the
demand-level-annotation (\DeLeAn) rubrics will be used to automatically obtain a demand profile.
This can be compared with the system capability profile  for any AI system that has previously gone through the ``System Process'' 
to understand how well the system performs for the task (e.g., identifying specific areas of strength and weakness relative to the demands). We can even predict performance at the instance level from this comparison, although the use of a powerful assessor is a better option, which can help us decide whether it is reasonable to employ the AI system in a given situation. 

These processes are fully automated through open-source pipelines, and can be easily customised by AI researchers, policy-makers and regulators, by extending the scale to 
 other capabilities, traits or propensities (e.g., affecting safety or fairness). 
%
%
%
%
%
%


\section{AI Evaluation at Scale} 

The key element for our overhauling of AI evaluation is the configuration of scales that are understandable, general and well-grounded in psychology and measurement theory. We first define these 
scales using rubrics that serve as measurement instruments for instance demands, and then build the methodology around them.

\subsection{General Scales and Automated Annotation}\label{sec:dimensions}

Over millennia, human folk psychology \cite{Andrews2020-ANDITF} has shaped linguistic concepts such as `smart' and `shy' to help us explain and predict people's behaviour. This observation has been the base of the {\em lexical hypothesis}, operationalised more than a century ago \cite{partridge1910outline} for the identification of many {\em constructs} in psychometrics and other social sciences, from capabilities to personality traits, such as `critical thinking' or `conscientiousness'. 
Based on human experimental data from a range of intellectual tasks for over a century, 
psychology has developed many taxonomies of human capabilities and traits, such 
as the Cattell-Horn-Carroll  hierarchical structure of human cognitive abilities 
and the Big Five personality traits \cite{rust2014modern}. Whereas some of these constructs, such as `reasoning' and `comprehension' capabilities, are meaningful for humans when applied to machines, others such as `working memory' or `processing speed' are not very insightful for AI systems \cite{hernandez2017measure}. 
In light of this, a taxonomy of 14 general capabilities inspired by the literature in human psychology, comparative cognition and artificial intelligence \cite{hernandez2017measure} was designed \cite{hernandez2019ai} in such a way that could be understandable by humans but at the same time applicable to explain and predict the behaviour of both humans and machines. 
Here, we will put this design criterion to the test. 
Our work builds upon \citet{tolan2021measuring}, who introduced a rubric for the same taxonomy that was used to assign the presence or absence of the need for each capability in generic tasks extracted from worker surveys, occupational databases and AI benchmarks. 

First, we extend the taxonomy by including knowledge dimensions, as well as 
new control dimensions. Second, we develop new scales and rubrics in a quantitative range between 0 
and 5+, with 0 representing absence of demand and values 1-4 representing increasing {\em demand levels} of the capability, and 5+ representing 5 or above. For instance, the famous 
`Sally-Ann' false-belief task assesses understanding of an individual's false belief regarding an object's properties if those properties change while they are not looking (Sally will look for her marble in the basket where she left it, even though Anne moved it to the box when Sally was away). This may be level 4 for dimension \dm{MS (Mind modelling and social cognition)}, but may be level 0 for dimension \dm{QLq (Quantitative reasoning)}. 
Similarly, the question 
``if all A are B, some B are C, no C are D, and all D are E, what can be inferred about the relationship between A and E?” may be level 4 for \dm{QLl (Logical reasoning)} but level 0 for \dm{MS (Mind modelling and social cognition)}. 

Table \ref{tab:dimensions} shows the set of dimensions we have included in the first version of the demand-level-annotation (\DeLeAn) rubric set. We adapt 7 broad capabilities from \citet{tolan2021measuring}, applicable to LLMs (e.g., `auditory processing' was discarded), and refine some of them hierarchically with subdimensions, making them a group of 11 `proper' cognitive  capabilities that we call `primordial'. Beyond capabilities, we additionally include new dimensions accounting for domain `knowledge', separated into five subdimensions (\dm{KNn}, \dm{KNs}, \dm{KNa}, \dm{KNf}, \dm{KNc}) covering big branches of human knowledge, and three `extraneous' ones, \dm{AT} (\dm{Atypicality}), \dm{VO} (\dm{Volume}),  and \dm{UG} (\dm{Unguessability}), to account for elements that make the task more challenging independently of primordial or knowledge demands. 

In particular, \dm{Atypicality} deals with {\em contamination} \cite{balloccu2024leak,jiang2024investigating} and other familiarisation effects leading to capability overestimation because similar data was seen during training. An AI system may simply succeed because it has memorised the instance. This dimension can be used to explain and predict performance, by identifying \dm{AT}  as a confounder with the other demands. The second extraneous dimension, 
\dm{Volume}, 
represents the use of `collages' to make instances more difficult. For instance, if we put ten simple additions in an exercise and we score whether all of them are correct, then we have increased the difficulty significantly, but the quantitative reasoning demand is the same. We call this phenomenon {\em amalgamation}, and it is a recurrent trick to make instances more difficult, either in benchmarks of increasing hardness \cite{srivastava2022beyond,suzgun2022challenging,kazemi2025big}
or in adversarial testing \cite{kiela2021dynabench}. There is a correlation between the size of the questions (and the answers) and the difficulty you can achieve with it \cite[Figs. 3\&4]{srivastava2022beyond}. 
In the end, amalgamation produces an underestimation of the capabilities, because the subjects fail at tasks that are incorporating many simple things simply because the chances of error accumulate, even if the cognitive load is not necessarily increased \cite{sweller2011cognitive,kalyuga2011cognitive}. Finally, \dimension{Unguessability} captures 
the very usual {\em funnelling} effect to make a question more amenable for scoring but at the same time reducing its difficulty. The obvious case is the use of multiple-choice questions, which have become predominant in most AI benchmarks, despite its issues \cite{balepur2025these}. Reducing or increasing the number of options has been a common practice to change the `difficulty' of a task without modifying its cognitive demands  \cite{wang2024mmlu}. In general, these three extraneous dimensions will account for an important proportion of the predictability in LLM success, and including them helps clarify these confounding effects. 

This makes a total of 19 dimensions, with the first 18 corresponding to proper demands (11 primordial, 5 knowledge and 2 extraneous) that may be met by the subject or not, while \dimension{Unguessability} being a dimension reflecting funnelling in the item design (e.g., multiple-choice questions). Because of that, it is the only dimension expressed between 0 (the correct answer is trivially determined by the question) and 100 (unguessable, i.e., a good open-ended question). 
Each of the 18 demand rubrics includes a general description of the construct to be annotated, followed by a description of each of the levels, from 0 to 5+, with three `anchor' instances each.  More details about the choices of the scales can be found in section \ref{sec:scales}, and the full rubrics can be found in appendix \ref{sec:rubrics}, while related work to the general motivations, techniques and methodology in appendix \ref{sec:relatedwork}. 

\renewcommand{\arraystretch}{1.32}
\begin{table*}[!h]
\centering
 \caption{Dimensions and subdimensions in the demand-level-annotation (\DeLeAn) rubric set. The first 18 (grouped into 11 `primordial' in red, 5 `knowledge' in teal and 2 `extraneous' in dark blue) are demand scales in the range (0, 5), while {\tt\footnotesize UG} ({\tt\footnotesize Unguessability}) is not a demand: it is another extraneous dimension representing the 1 minus the probability of success by random guess or a naive method. For instance, a multiple-choice question with four options would have value 75\%. Full rubrics in appendix~\ref{sec:rubrics}.}
 \resizebox{0.90\textwidth}{!}{ 
  \begin{tabular}{C{0.015\textwidth}p{0.16\textwidth}C{0.022\textwidth}p{0.17\textwidth}p{0.62\textwidth}}
\toprule
& \textbf{Dimension (Broad)} & & \textbf{Dimension\newline(Specific)} & \textbf{Description of Demands}\\ 
\midrule

      \multirow{2}{*}{{\color{primordial}\tt\footnotesize AS}} & \multirow{2}{0cm}{{\color{primordial}\tt\footnotesize Attention and~Scan}} & 
      \multirow{2}{*}{{\color{primordial}\tt\footnotesize AS}} & \multirow{2}{0cm}{{\color{primordial}\tt\footnotesize Attention and~Scan}} &  Focus on or locate specific elements within a given stream of information or environment in the whole process of solving a task.  \\
\midrule
    \multirow{4}{*}{{\color{primordial}\tt\footnotesize CE}}  & \multirow{4}{0cm}{{\color{primordial}\tt\footnotesize Comprehension and~Expression}} & 
     \multirow{2}{*}{{\color{primordial}\tt\footnotesize CEc}}  & \multirow{2}{0cm}{{\color{primordial}\tt\footnotesize Verbal Comprehension}} &  Understand text, stories or the semantic content of other representations of ideas in different formats or modalities.\\      
     & & \multirow{2}{*}{{\color{primordial}\tt\footnotesize CEe}}  & \multirow{2}{0cm}{{\color{primordial}\tt\footnotesize Verbal Expression}} & Generate and articulate ideas, stories, or semantic content in different formats or modalities. \\
\midrule
      \multirow{3}{*}{{\color{primordial}\tt\footnotesize CL}} & \multirow{3}{0cm}{{\color{primordial}\tt\footnotesize Conceptualisation, Learning~and Abstraction}} & 
      \multirow{3}{*}{{\color{primordial}\tt\footnotesize CL}} & \multirow{3}{0cm}{{\color{primordial}\tt\footnotesize Conceptualisation, Learning~and Abstraction}} &  Build new concepts, engage in inductive and analogical reasoning, map relationships between domains, and generate abstractions from concrete examples. \\
\midrule
    \multirow{6}{*}{{\color{primordial}\tt\footnotesize MC}}  & \multirow{6}{0cm}{{\color{primordial}\tt\footnotesize Metacognition~and Critical~Thinking}} & 
     \multirow{2}{*}{{\color{primordial}\tt\footnotesize MCr}}  & \multirow{2}{0cm}{{\color{primordial}\tt\footnotesize Identifying~Relevant Information}} &  Recognise what information helps solve the task or does not, and how this recognition process unfolds as they work toward the solution.\\      
     & & \multirow{2}{*}{{\color{primordial}\tt\footnotesize MCt}}  & \multirow{2}{0cm}{{\color{primordial}\tt\footnotesize Critical~Thinking Processes}} & Monitor or regulate multiple thought processes to answer the question effectively, ranging from simple recall to high-level critical thinking. \\
     & & \multirow{2}{*}{{\color{primordial}\tt\footnotesize MCu}}  & \multirow{2}{0cm}{{\color{primordial}\tt\footnotesize Calibrating~Knowns and~Unknowns}} & Recognise the boundaries of one's knowledge and confidently identify what one knows they know, knows they don't know, or is uncertain about. \\
\midrule
      \multirow{3}{*}{{\color{primordial}\tt\footnotesize MS}} & \multirow{3}{0cm}{{\color{primordial}\tt\footnotesize Mind~Modelling and~Social Cognition}} & 
      \multirow{3}{*}{{\color{primordial}\tt\footnotesize MS}} & \multirow{3}{0cm}{{\color{primordial}\tt\footnotesize Mind~Modelling and~Social Cognition}} &  Model the minds of other agents or reasoning about how the beliefs, desires, intentions, and emotions of multiple other agents might interact to determine future behaviours.  \\
\midrule
    \multirow{4}{*}{{\color{primordial}\tt\footnotesize QL}}  & \multirow{4}{0cm}{{\color{primordial}\tt\footnotesize Quantitative~and Logical~Reasoning}} & 
     \multirow{2}{*}{{\color{primordial}\tt\footnotesize QLl}}  & \multirow{2}{0cm}{{\color{primordial}\tt\footnotesize Logical Reasoning}} &  Match and apply rules, procedures, algorithms or systematic steps to premises to solve problems, derive conclusions and make decisions.\\      
     & & \multirow{2}{*}{{\color{primordial}\tt\footnotesize QLq}}  & \multirow{2}{0cm}{{\color{primordial}\tt\footnotesize Quantitative Reasoning}} & Work with and reason about quantities, numbers, and numerical relationships. \\
\midrule
    \multirow{2}{*}{{\color{primordial}\tt\footnotesize SN}}  & \multirow{2}{0cm}{{\color{primordial}\tt\footnotesize Spatial~Reasoning and~Navigation}} & 
     \multirow{2}{*}{{\color{primordial}\tt\footnotesize SNs}}  & \multirow{2}{0cm}{{\color{primordial}\tt\footnotesize Spatio-physical Reasoning}} &  Understand spatial relationships between objects and predicting physical interactions.\\      
\midrule
    \multirow{10}{*}{{\color{knowledge}\tt\footnotesize KN}}  & \multirow{10}{0cm}{{\color{knowledge}\tt\footnotesize Knowledge}} & 
     \multirow{2}{*}{{\color{knowledge}\tt\footnotesize KNa}}  & \multirow{2}{0cm}{{\color{knowledge}\tt\footnotesize Knowledge~of Applied~Sciences}} &  Knowledge or conceptual understanding in applied sciences (e.g., medicine, law, education, business, agriculture, engineering except IT). \\      
     & & \multirow{2}{*}{{\color{knowledge}\tt\footnotesize KNc}}  & \multirow{2}{0cm}{{\color{knowledge}\tt\footnotesize Customary~Everyday Knowledge}} & Knowledge in information that most people in a given society typically acquire through daily life experiences, social interactions, and media. \\
     & & \multirow{2}{*}{{\color{knowledge}\tt\footnotesize KNf}}  & \multirow{2}{0cm}{{\color{knowledge}\tt\footnotesize Knowledge~of Formal~Sciences}} &  Knowledge or conceptual understanding in  formal sciences (e.g., mathematics, logic, computer science, statistics).  \\      
     & & \multirow{2}{*}{{\color{knowledge}\tt\footnotesize KNn}}  & \multirow{2}{0cm}{{\color{knowledge}\tt\footnotesize Knowledge~of Natural~Sciences}} & Knowledge or conceptual understanding in natural sciences (e.g., physics, chemistry, biology, astronomy, earth sciences, ecology).  \\
     & & \multirow{2}{*}{{\color{knowledge}\tt\footnotesize KNs}}  & \multirow{2}{0cm}{{\color{knowledge}\tt\footnotesize Knowledge~of Social~Sciences}} &  Knowledge or conceptual understanding in social sciences and humanities (e.g., history, psychology, sociology, literature, art, philosophy).  \\ 
\midrule
      \multirow{2}{*}{{\color{others}\tt\footnotesize AT}} & \multirow{2}{0cm}{{\color{others}\tt\footnotesize Atypicality}} & 
      \multirow{2}{*}{{\color{others}\tt\footnotesize AT}} & \multirow{2}{0cm}{{\color{others}\tt\footnotesize Atypicality}} &  How uncommon the task is or how unlikely it is that the instance has appeared in various sources (internet, textbooks, tests).  \\
\midrule
      \multirow{2}{*}{{\color{others}\tt\footnotesize VO}} & \multirow{2}{0cm}{{\color{others}\tt\footnotesize Volume}} & 
      \multirow{2}{*}{{\color{others}\tt\footnotesize VO}} & \multirow{2}{0cm}{{\color{others}\tt\footnotesize Volume}} &  Proportional to the logarithm of the time a fully competent human needs to read and complete the task in ideal conditions, excluding interruptions.  \\
\midrule
    \multirow{2}{*}{{\color{nd}\tt\footnotesize UG}} & \multirow{2}{0cm}{{\color{nd}\tt\footnotesize Unguessability}} & 
      \multirow{2}{*}{{\color{nd}\tt\footnotesize UG}} & \multirow{2}{0cm}{{\color{nd}\tt\footnotesize Unguessability}} &  The chance of error (percentage) of a task if following  obvious cues or by random guess.  \\
\bottomrule
\end{tabular}
}
    \label{tab:dimensions}
\vspace{1cm} 
\end{table*}

\begin{figure}[!ht]
  \centering
  \includegraphics[width=0.95\linewidth]{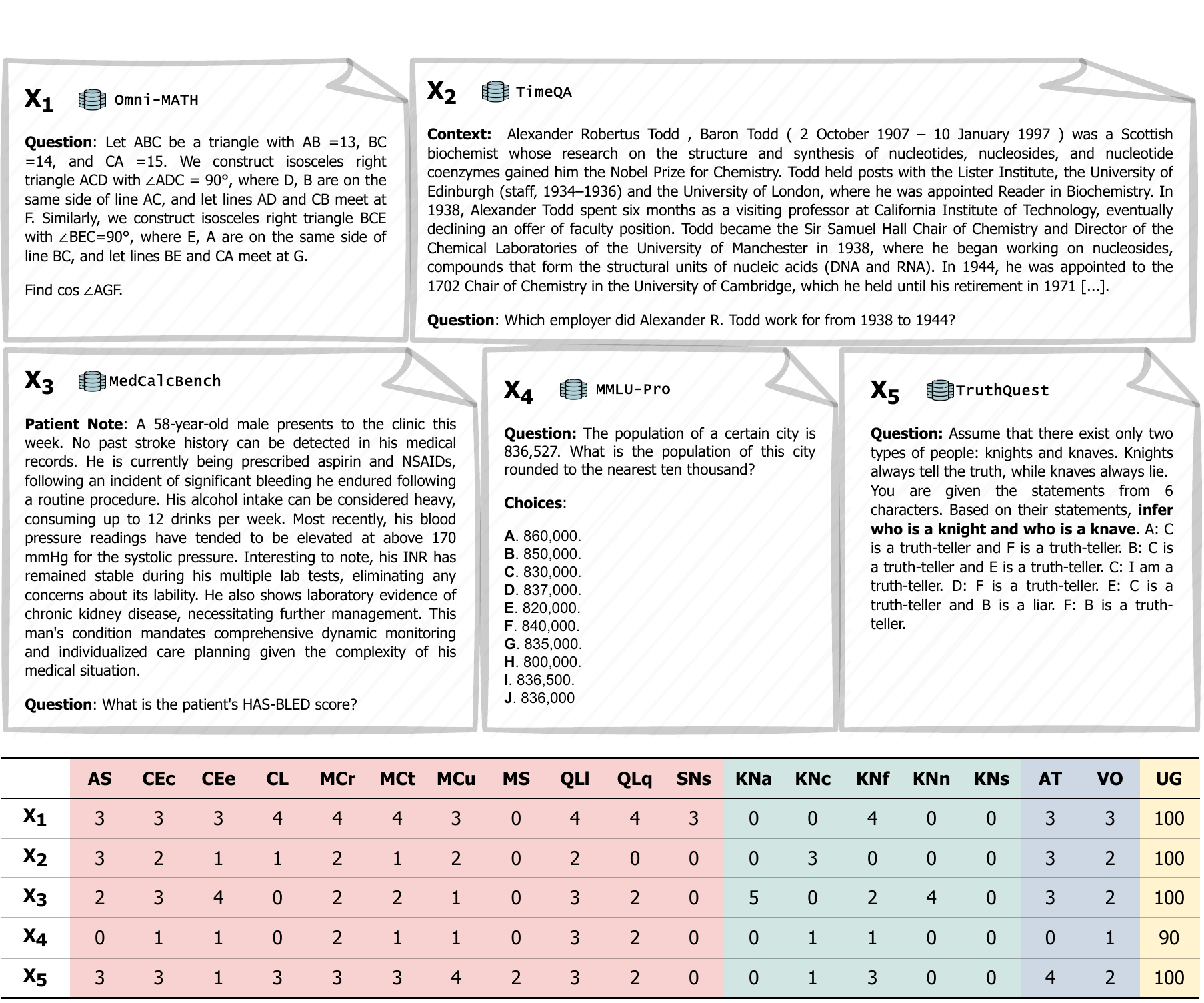}
\caption{Level annotations of five items (from benchmarks OmniMath, TimeQA, MedCalcBench, MMLU-Pro, TruthQuest, respectively) using the \DeLeAn rubric set by GPT-4o. The listed demands in the table (from left to right) follow the same order shown in Table \ref{tab:dimensions} (from top to bottom): 11 being primordial, 5 knowledge and 3 extraneous. 
}
  \label{fig:annotation}
\end{figure}

With this rubric in hand, we need to annotate any new instance along each dimension. Traditionally, we would have humans annotate each dimension and each task; however, recruiting humans to annotate all tasks and dimensions for every new benchmark that we may consider is very costly. Powerful LLMs offer a more scalable alternative to rapidly and flexibly annotate thousands of items, in near real time. Five annotations are illustrated in Figure~\ref{fig:annotation}. Whereas there may be some discordances between LLM and human scores as we discuss in Section \ref{sec:llmannotators}, scalability is critical for the broad and flexible deployment of our evaluation methodology. This can be seen as a trade-off but also as an opportunity to have stable and fully-reproducible annotations using LLMs, which can be improved as LLMs get better or are more aligned with human interpretation. In fact, the three instance anchors per level were very instrumental for the LLMs to perform good ratings (in a few-shot inference fashion) but also for human understanding. In our case, we performed the annotations with GPT-4o 
with which we found $r_{WG}$\footnote{A value of 0.7 for this index is generally regarded as a good agreement threshold \cite{james1984estimating, lebreton2008answers}.} scores well above 0.75 for all dimensions (averaging to 0.86) between the Delphi-agreed demand level from five humans and GPT-4o (details in the Methods section~\ref{sec:methods}). 
The use of comprehensive rubrics in natural language that can be applied automatically is a major advancement to make the explanatory power a reality, especially if humans could interact with the LLM to explain their annotations.



\subsection{Slicing the Demand-Ability Space}\label{sec:methodology}

Annotating instances using these general scales allows us to compare what makes them easy or hard, and provides the same lens of analysis independently of where the instance comes from: human test, AI benchmark or new item design. We can discard or combine instances to build a specific test profile. While this is not new in psychology or AI \cite{zhang2024task}, the scales can be applied to any task, test or collection of benchmarks (\DeLeAn v.1.0 only includes textual modality). Having the same scales makes a comparison of the vast space of tests and benchmarks possible for the first time. 

For instance, in this paper we applied  \DeLeAn  to 16,108 instances from 63 tasks from 20 benchmarks\footnote{Curated from the 2024 proceedings of six AI conferences and other venues, while ensuring both data quality and diversity (details in Methods \ref{sec:benchmark_battery_curation}).}. 
This is unprecedented, as all these tasks are now represented within the same 19-dimensional space of general cognitive demands. After the annotation, we obtain the \ADeLe battery (Table~\ref{tab:selected_benchmarks_and_tasks}, more details and selection criteria in the Methods section). We can observe the distribution of the levels of demand for each dimension, the {\em demand profile}, represented as a polar histogram (see Figure~\ref{fig:benchmark_demand_profiles}). Exploring this for the whole battery and for its benchmarks helps answer the question of 
whether each benchmark can measure what their developers claimed they should measure. We will explore this in the Results section. 

\begin{figure}[!ht]
  \centering
  \includegraphics[width=0.5\linewidth]{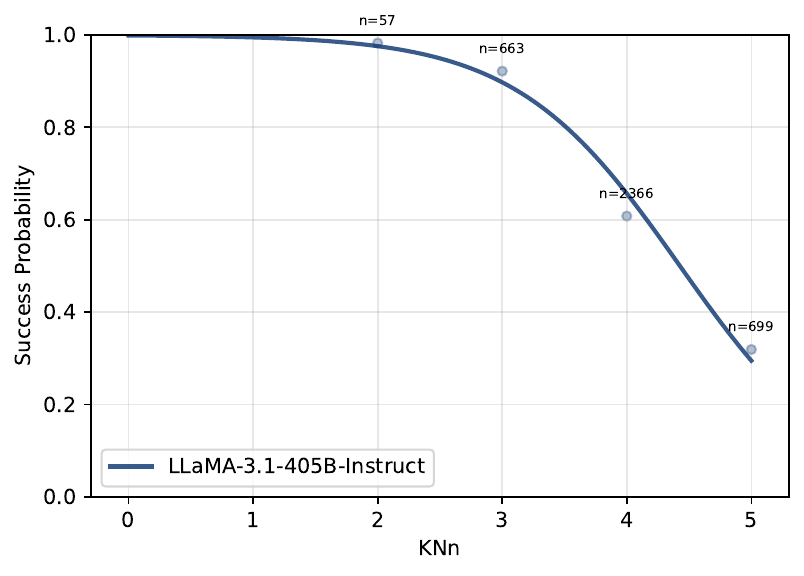}
\caption{The characteristic curve of \modelc{Llama-3.1-405B-Instruct} for dimension 
\dimension{KNn}(\dimension{Knowledge of Natural Sciences})
on the \ADeLe battery. The \xaxis shows the demand levels 0-5 for  \dimension{KNn} and the \yaxis the average performance for that level (probability of success). As usual, level 0 has no points left (0 never dominates), but in this case we see a situation with no point for 1. The curve is a logistic fit in the output range (0,1).} 
  \label{fig:oneSCC} 
\end{figure}

Once instances are annotated, we can do more insightful analyses with them than just calculating one average for a whole dataset. 
When we run an LLM on an annotated benchmark such as the \ADeLe battery, we can analyse each dimension separately. Given that there are correlations between the dimensions and other effects, as will be shown in the Results section, we represent all the instances in the battery according to their annotation level for a specific dimension of interest. For example, Figure~\ref{fig:oneSCC} shows a 
\textit{subject characteristic curve} \cite{lumsden1977person} for
the results of \model{Llama-3.1-405B-Instruct} on 16,108 instances of the \ADeLe battery, but binned by the levels on the dimension \dm{KNn} (\dm{Knowledge of Natural Sciences}). As we will explain in Methods, for each bin $b$ for that dimension, we exclude all points for which the level of any other dimension is greater. In other words, we want the represented dimension to dominate on the instances we are showing (in this case, only 3,785 out of 16,108).

On this plot we can then fit a logistic function and derive the LLM's ability as
 {\em the level of demand where the probability of the subject to succeed is 0.5, assuming all other demand levels are lower}. In Figure~\ref{fig:oneSCC}, this leads to an ability of 4.3. This interpretation of {\em ability} is in accordance with psychometric tradition \cite[p. 249]{thurstone1937ability}, 
and will be followed for the rest of the paper: an ability of 4.3 does not necessarily 
mean that the subject solves all tasks instances of level 4.3 or less, but that it has 50\% chance of succeeding at level 4.3, higher at level 3, much higher at level 2, etc., and of course lower at level 5 and above, in a sigmoidal way, as we see in the figure. The exact estimation of the ability (as the usually equivalent area under the curve) will be explained in the Methods section.

The advantages of these curves and this manner of  interpreting ability are bolstered by the fact that the scale on the \xaxis is absolute rather than relative. With the 3,786 instances in Figure~\ref{fig:oneSCC}, we get an average accuracy of 62\%. If we chose the $n=699$ instances of level 5 and repeat them 500 times, the average accuracy of the LLM would drop dramatically (below 40\%), as we are adding more difficult examples. This is what adversarial testing does~\cite{kiela2021dynabench}, 
especially when benchmarks saturate. In contrast, the average accuracy for the instances {\em at bin 5} would remain the same, and the characteristic curve would not be affected at all. The \textit{ability} would not alter: 4.3. This case neatly represents the difference between performance, which is a measure of a pair subject and task distribution (so changing from 62\% to 40\% when the task distribution changes), and ability, which is an inferred property of a subject that is invariant to the task distribution. While all this is strongly inspired by item response theory (IRT), and the linear logistic test model (LLTM) in particular \cite{fischer1973linear}, it is important to clarify that unlike these and other latent factors approaches, those in AI included \cite{burnell2023revealing,ilic2024evidence,ruan2024observational}, we only use the information of a single LLM for the estimation of its abilities. Accordingly, abilities have a value and a scale that are completely independent of other LLMs. We use the term `non-populational' to refer to an indicator or measurement that does not depend on the rest of the population, only on the individual. On the contrary, many other inferential techniques that are populational, such as IRT, principal component analysis or factor analysis, usually work well with human populations because samples are sufficiently stable across time, but lead to different results for AI system `populations', whenever a new set of LLMs are added to the inferential pool (e.g., see the changes in factors discovered in LLMs from \cite{burnell2023revealing} to \cite{ilic2024evidence} taking place in a few months' time). This volatility does not happen with our approach. Our abilities are not relative to a population of subjects, and the scale is absolute. Even if the evaluation battery were extended with instances of level 7 or 8 for some dimensions to account for more powerful future AI systems, the logistic curve for the old systems would likely have low values on these instances, thus not affecting much the estimate of these less powerful models.

With this procedure on the characteristic curves we can derive {\em ability profiles} as 18-dimensional vectors containing the abilities. The usual way of representing a score profile with many dimensions is a radial plot, and this is common in the behavioural sciences, and more recently in AI. However, if we look at these plots in AI papers (e.g., \cite{balachandran2024eureka,fountas2024human}),  
we see that what they represent in each dimension is the average accuracy of a selection of instances that belong to a particular domain or dataset, not an ability. Those plots will change as the difficulty of the instances varies, while an {\em ability profile} is invariant to these changes.   
Overall, our notion of ability using the general scales is very different from the common, yet inaccurate use of the term in AI as a synonym of performance. This includes the use of the term  `capability' in the area of safety evaluations, even if informally the concept may be associated with levels \cite{phuong2024evaluating}, which were never defined or scaled. 

By comparing the ability profile of an AI system with the demand profile of a task instance or a benchmark, we can explain the observed performance. Moreover, using the differences between abilities and demands, we can use interpretable algebraic models to anticipate performance for new instances (appendix \ref{ap:algebraic_assessor}), but there is potential for other options as well. For example, the 18 values that are annotated for each single instance in the scale 0..5+, and \dm{unguessability}, constitute a 19-dimensional vector $\mathbf{x}$, which can be used as predictor variables for a probabilistic classification model, an assessor, outputting the (estimated) performance of an AI system on that instance. Each assessor can be trained  specifically for each LLM, without relying on the LLM's features. 
As we will see in the Results section we can compare this with many other powerful ways of predicting performance, such as assessors with embeddings and finetuned LLMs. Surprisingly, despite the much smaller computation cost (aside from annotating the battery, which only needs to be done once for all LLMs), the results for in-distribution prediction are comparable with the best assessors in terms of discrimination between model success and failure, and considerably better in terms of calibration. However, it is for out-of-distribution cases where the use of an abstract scale pays off. The predictive power  for  out-of-distribution tasks and benchmarks is substantially better for the demands-based assessor than the best baseline, and of course much better than average accuracy, which is only well-calibrated in-distribution. This is because our general scales provide predictive features over a wide variety of tasks and limit overfitting on features becoming spurious when switching tasks and benchmarks. 
Finally, just as ability profiles are non-populational, the assessors we derive for each system are inferred exclusively from that system's results, rather than from population-level parameters like those used in scaling laws for aggregate performance prediction \cite{ruan2024observational}.
Of course, the assessors' predictive power could be affected by our selection of datasets within the \ADeLe battery, particularly if the range of demands is significantly misaligned with the abilities of the LLM, whether they are exceptionally high or low. 

\section{Results}\label{sec:results}

The technical advantages presented in the previous section are the result of our design choices: absolute scales based on rubrics and non-populational indicators. However, our main goal with these scales is to achieve AI evaluation with both explanatory and predictive power. We now explore whether this is the case with four specific research questions, comparing our approach with standard practice or best baselines in AI evaluation.

\subsection{Annotation and Scales Analysis: Distinguishing Levels and Dimensions}


\begin{tcolorbox}[colback=mybg2,colframe=RQ,boxrule=0pt,  borderline south={2pt}{2pt}{RQ}, arc=1mm,left=4pt, right=6pt, top=4pt, bottom=1pt]
    \textcolor{RQ}{\sffamily\small{{\textbf{RQ1}: Can humans distinguish the levels in the rubrics and the 19 dimensions?}}}    
\end{tcolorbox}

The scales will only serve for explanatory purposes if they can be understood. This question can indirectly be assessed by how reliable the annotations are. If humans agree with each other to some extent then we can conclude that there is some common understanding. Similarly, if the annotations from humans are similar to those performed by the LLM annotator, then we can conclude that the demand and ability profiles extracted from these annotations will be meaningful for humans. Before exploring this, it is important to highlight that there is no ground truth in the levels, and humans are not the gold standard. Actually, another source of necessary support for a rubric would be whether it leads to high predictive power, which we will explore in section \ref{sec:pred}, while still representing the construct in an understandable way.

In Methods section \ref{sec:interrater} we describe how a group of five people were selected and how the rubrics were presented and to what sample of data. Their inter-rater agreement ($r_{WG}$ index) for the 18 demands ranges between 0.70 and 0.91 (with an average of 0.83). 
After applying Delphi \cite{linstone1975delphi},  we have a consensus annotation, which we compare against GPT-4o. 
The agreement is high (Spearman Correlation between 0.75 and 0.94 (averaging to 0.86) 
between Delphi consensus and GPT-4o).

\begin{figure}[!ht]
  \centering
  \includegraphics[width=0.9\linewidth]{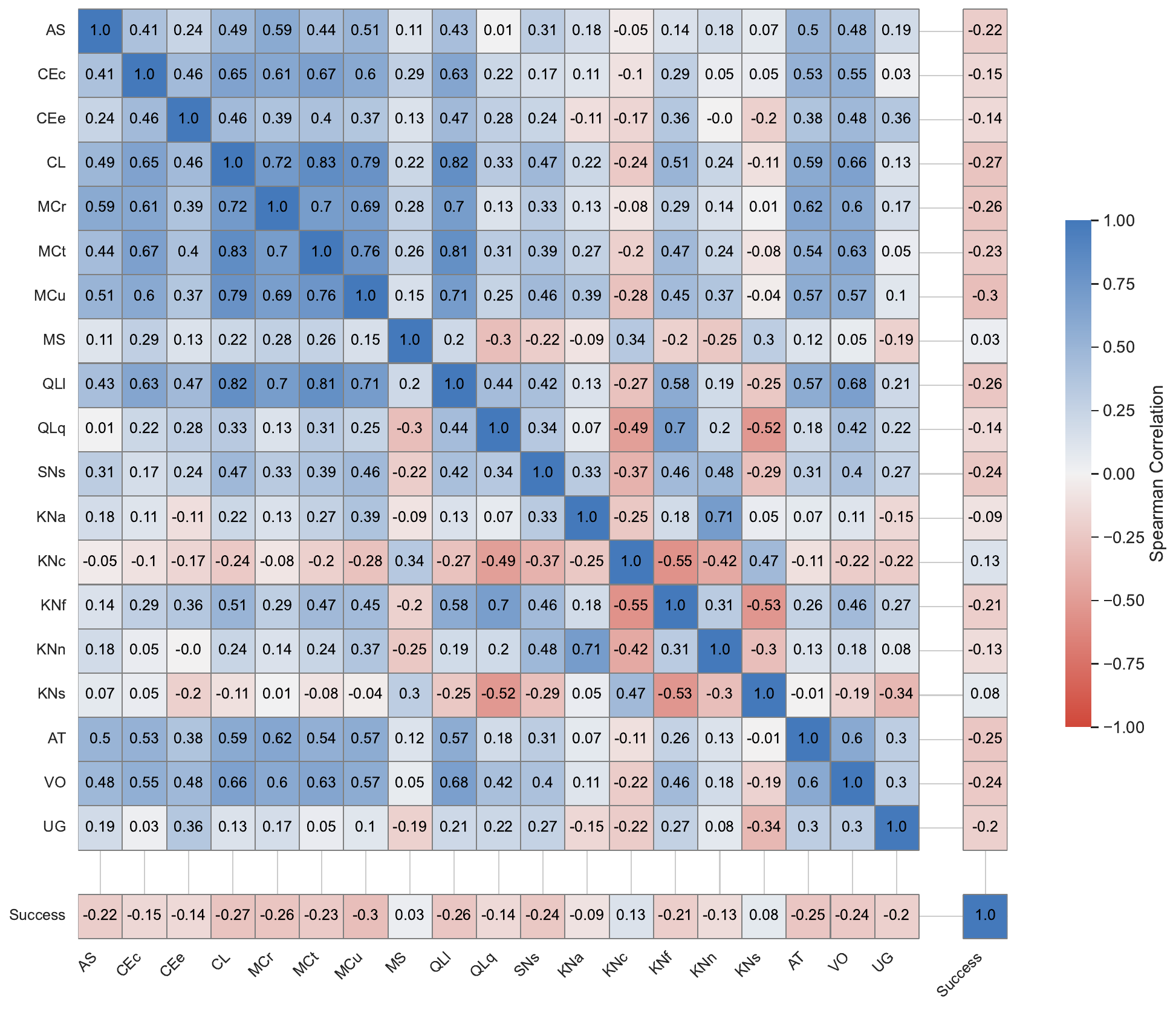}
  \caption{Correlations of the demand level using all the items in the  \ADeLe battery for all the pairs of the 18 demands and the special dimension \dmc{UG} (\dmc{Unguessability}). 
  It also includes the success (i.e., correctness at instance-level) of all the subject LLMs considered in the experiments.}
  \label{fig:demand_correlation}
\end{figure}

The dimensions could be understandable by humans but conceptually redundant. By conceptually redundant, we mean that we cannot construct an instance for which one dimension level is high and the other is low, independently on whether they are correlated in a particular benchmark (e.g., because the design or selection bias always makes one increase along with the other). If such an example does not exist, humans will find it hard to distinguish the dimensions. This definition means that if we find a low correlation, we could conclude that there must be instances with very different levels. Consequently, we explore the Spearman correlations between all the dimensions in Figure~\ref{fig:demand_correlation} for the whole \ADeLe battery. 
Generally low or moderate correlations indicate that most dimensions appear to carve different parts of the space. This is so even for instances that were selected from standard benchmarks, which are not usually careful about being specific in item design. A correlation of 0.8 still allows for cases where the level for one dimension is 0 and the level for the other dimension is high. These examples do not abound, but are not impossible. There are two correlations above 0.8 and they fall on \dm{CL} (\dm{Conceptualisation, Learning, and Abstraction}), which looks quite central in the manifold, 
 given its strong correlation with the metacognition dimensions \dm{MC} and with \dm{QLl} (\dm{Quantitative and Logical Reasoning - Logic}). The demands \dm{KNc} (\dm{Customary Everyday Knowledge}) and \dm{KNs} (\dm{Social Sciences and Humanities}) have a good number of negative correlations, especially with \dm{QL} (\dm{Quantitative and Logical Reasoning}) and other knowledge demands. In general, these positive or negative correlations can have multiple interpretations since they are contingent to our choice of benchmarks.

Equally important are the \textit{extraneous} dimensions---\dm{AT} (\dm{Atypicality}), \dm{VO} (\dm{Volume}) and \dm{UG} (\dm{Unguessability})---which do not directly capture cognitive demands, but rather reflect those 
elements making items more difficult in other ways. 
We see that the correlations are high with other demands (except for \dm{UG}).

The conclusion of the analysis is that the annotations by GPT-4o seem to be understandable by humans across all dimensions. Here, we do not have a baseline, but in standard AI evaluation practice, rubrics are rare, specific, and designed to be applied by humans or LLMs. Only occasionally they are meant to be explanatory and rarely quantitatively \citep{jin2023cladder,saparov2023language,Krathwohl01112002}, despite the recognition that 
this understanding is a key factor in AI adoption \cite{hardy2024more}. 
Looking to the future, despite good agreement between humans and LLMs, higher agreements may be possible as the capabilities of LLMs as annotators progress, including their potential of explaining their annotations to humans. Also, the correlations between dimensions do not seem to suggest that some combinations of demand levels are impossible, but simply infrequent in the current \ADeLe battery. This indicates that new versions of \ADeLe should try to be more selective in the selection of instances for each benchmark, instead of random (see the discussion of \ADeLeLight, where we do that selection, in section \ref{sec:adelelight}). Our choice of instances and benchmarks was meant to be representative of the landscape of AI benchmarks rather than a cherry-picked selection to minimise correlations. This was conditioned by our interest to explore what the benchmarks measure, as we study next.

\vspace{3cm}  
\subsection{Explanatory Power Analysis: Profiling Benchmark Demands}

\begin{tcolorbox}[colback=mybg2,colframe=RQ,boxrule=0pt,  borderline south={2pt}{2pt}{RQ}, arc=1mm,left=4pt, right=6pt, top=4pt, bottom=1pt]
    \textcolor{RQ}{\sffamily\small{{\textbf{RQ2}: What is the sensitivity and specificity of \ADeLe and its constituent benchmarks?}}}
\end{tcolorbox}



\begin{figure}[!ht]
  \centering
  \includegraphics[width=0.58\linewidth]{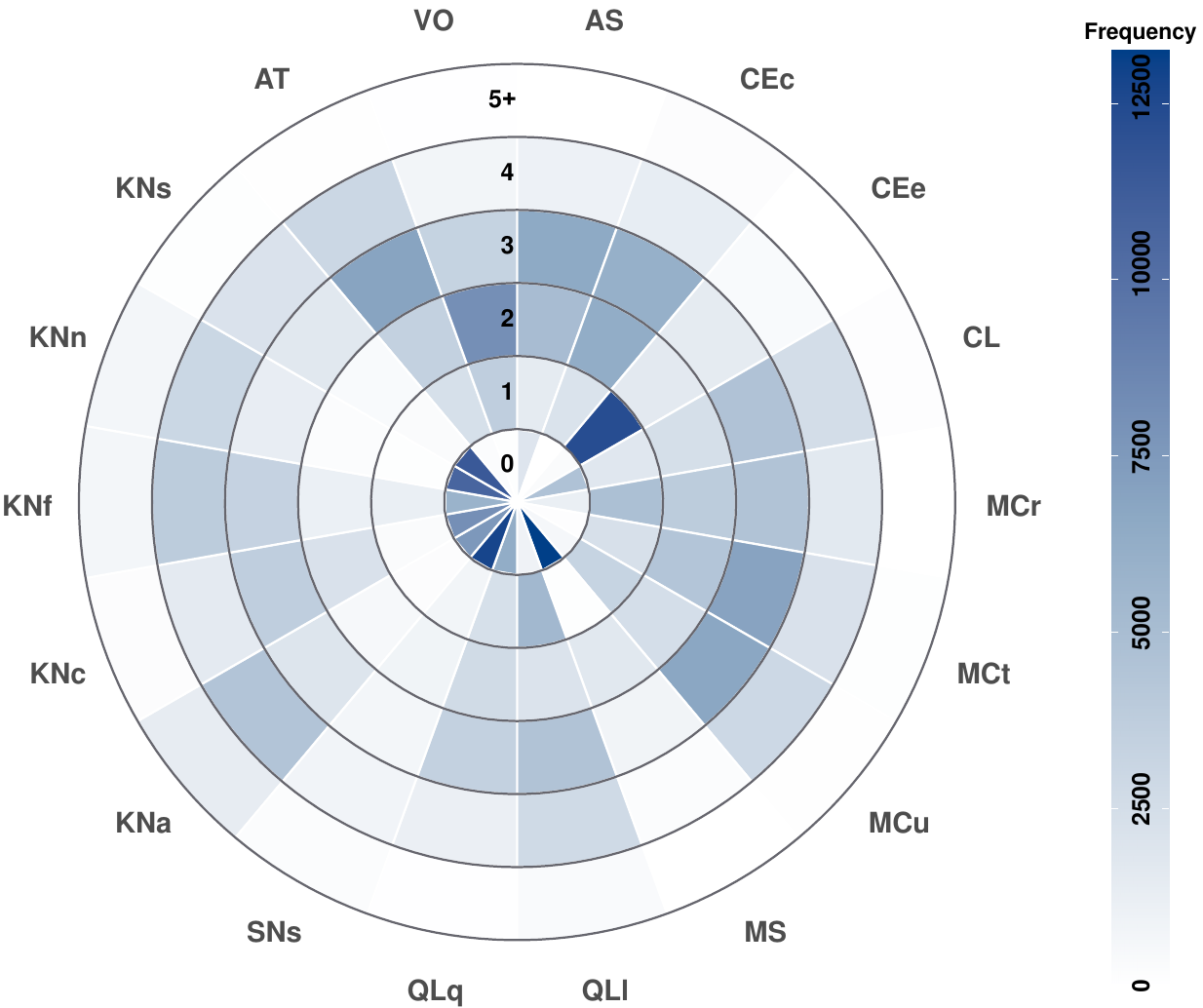}
  \caption{Distribution of level frequencies for the 18 demands using all the 16,108 instances in the  \ADeLe battery v.1.0. 
  Dimensions such as \dmc{CEe} (\dmc{Verbal Expression}),  \dmc{MS}
    (\dmc{Mind modelling and social cognition}) and \dmc{SNs} (\dmc{Spatial Reasoning and Navigation - Spatial})  
  have low proportion of items of high level, but this is in accordance with the focus of LLM evaluation on factual questions with no navigation or full social interaction. Future versions of the battery for agents or multimodal scenarios can increase the number and breadth of the dimensions. 
  }
  \label{fig:demand_distribution}
\end{figure}

Figure~\ref{fig:demand_distribution} shows the distribution of the 18 dimensions (ranged from 0 to 5) for the whole \ADeLe battery, 16,108 instances overall. 
Most dimensions cluster around mid-range values; however, several exhibit bimodal distributions, with one peak at zero and another at a higher level. This pattern suggests that most benchmarks are not measuring that dimension at all, while those few measuring it have a tendency of including intermediate difficulties
in order to capture the informative variation in state-of-the-art models (around levels 3 and 4). There are also two demands, \dm{MS} (\dm{Mind modelling and social cognition}) and \dm{SNs} (\dm{Spatial Reasoning and Navigation - Spatial}), with level 0 for most benchmarks, a bias that is indicative of the little coverage of benchmarks for those dimensions.

\begin{figure}[!ht]
  \centering
  \includegraphics[width=\linewidth]{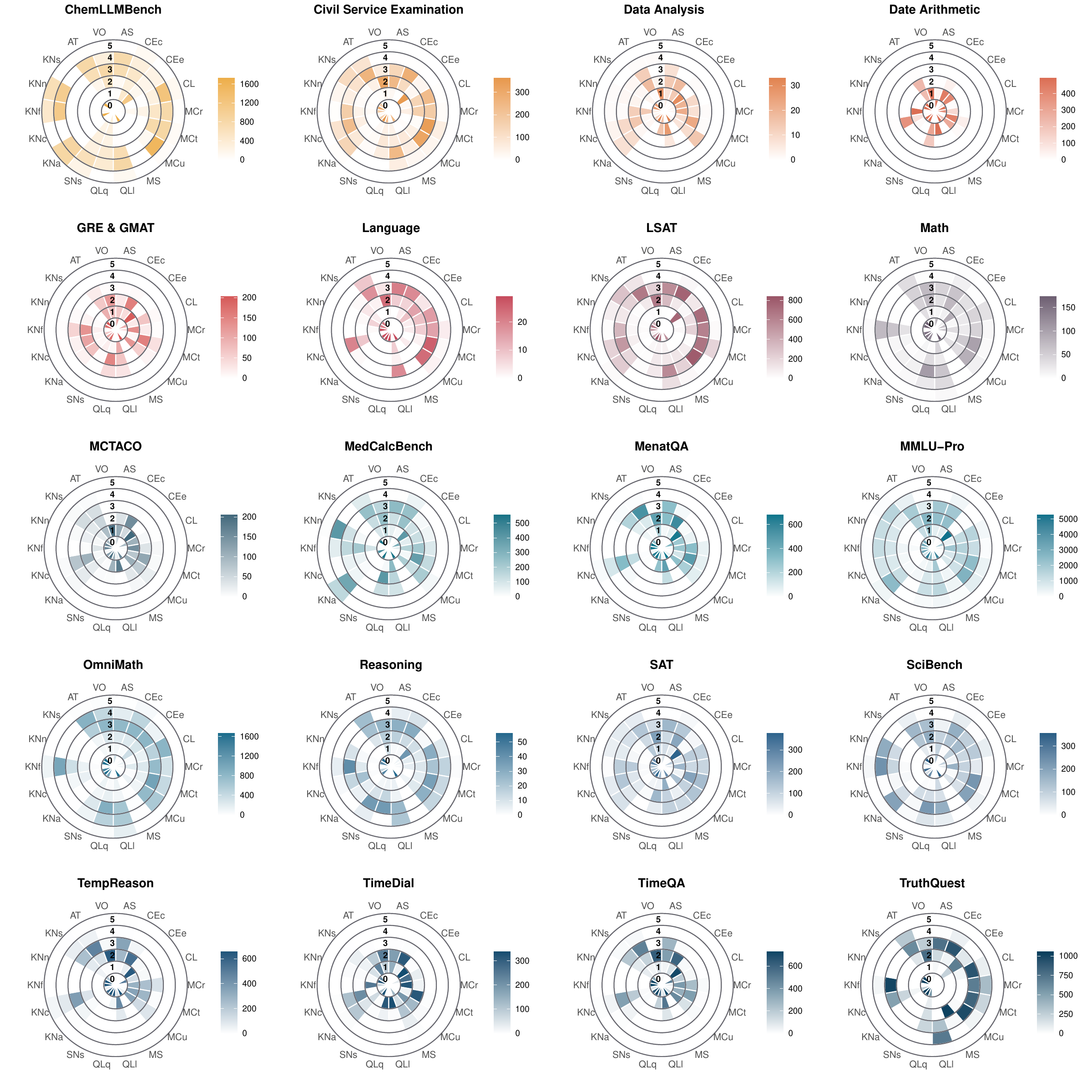}
  \caption{Demand profiles for the 20 benchmarks comprising the \ADeLe battery v.1.0  (all other things equal to Figure \ref{fig:demand_distribution}).}
  \label{fig:benchmark_demand_profiles}
\end{figure}

More interestingly, we can look at the demand profiles of benchmarks individually (Figure~\ref{fig:benchmark_demand_profiles}). 
This is informative to understand what the benchmarks actually measure and whether they measure what they claim to measure. Overall, the 
profiles are considerably distinct, so apparently they seem to measure different things. Benchmarks that focus on specialised topics (e.g., \ds{ChemLLMBench}, \ds{OmniMath}, \ds{MedCalBench} and \ds{SciBench}) show high demands in their respective domains (\dm{KNa} (\dm{Applied Sciences}), \dm{KNn} (\dm{Natural Sciences}) and \dm{KNf} (\dm{Formal Sciences})), while benchmarks such as  
\ds{TempReason} and \ds{TruthQuest}, which target a single domain, often peak in additional dimensions. 
Other benchmarks---such as \ds{Date Arithmetic}, \ds{GRE \& GMAT}, \ds{MCTACO}, \ds{TimeDial} and \ds{TimeQA}---have uniformly low demands, suggesting that they may be too easy. In contrast, broader assessments such as 
\ds{Civil Service Examination}, \ds{LSAT} and \ds{MMLU-Pro} show mixed profiles.

A closer look at each dimension reveals that some are barely present. For example, the demand levels for \dm{CEe} (\dm{Verbal Expression}) and \dm{MS} (\dm{Mind Modelling and Social Cognition}) are generally low; 
similarly, the range for 
\dm{SNs} (\dm{Spatio-physical reasoning}) is narrow. This is consistent with the selection bias we saw when aggregating all benchmarks in Figure~\ref{fig:demand_distribution}. Furthermore, examining the  spread (the wider the better in this case) across the benchmarks reveals further concerns: although benchmarks such as \ds{MMLU-Pro} cover a wide range of demands, others (e.g. \ds{Language}) are quite narrow in the bands. Table \ref{tab:selected_benchmarks_and_tasks} lists the domains these benchmarks are said to be measuring. However, many benchmarks do not really have a wide range of demand levels in certain dimensions they are claimed to be measuring (lack of sensitivity) whilst some other benchmarks exhibit a wide range in dimensions that they should not be measuring (lack of specificity). 

For example, while \ds{Civil Service Examination} from AGIEval claims to measure logical reasoning, its demand profile (Figure \ref{fig:benchmark_demand_profiles}) shows that successful completion of these tasks simultaneously demand considerable competence in several other dimensions (with the exception of \dm{SNs} (\dm{Spatio-physical Reasoning}) and  \dm{CEe} (\dm{Verbal Expression})), thereby diluting its specificity. A similar phenomenon can be observed in LiveBench's \ds{Reasoning} benchmark, which is designed to measure logical and spatial reasoning only. In addition, benchmarks such as  \ds{TempReason}, \ds{TimeDial} and \ds{TimeQA}---despite their focus on temporal reasoning---have a narrow range of demands in this domain, suggesting insensitivity to increasing task difficulty. 
Taken together, the specificity and sensitivity of common benchmarks is quite variable. These results suggest that assigning one or more benchmarks to one `capability' and aggregating their accuracy (as is standard practice), actually averages different demand levels and dimensions, leading to highly confounded results. If this is the baseline in common AI evaluation practice, it is simply insufficient to detect problems of specificity and sensitivity \cite{eriksson2025can,hardy2024more}. This issue becomes even more pronounced when integrating numerous benchmarks, such as BigBench \cite{srivastava2022beyond}, and other mega-benchmarks. Even if sensitivity may be increased by this integration (as we see for the whole of \ADeLe), specificity is lost if aggregate accuracy is used, and phenomena such as scaling laws \cite{ruan2024observational} emerge from this aggregation. Instead, with our scales, we can compare mixed subsets of items from different benchmarks whose demand levels now become commensurate, create recombinations of instances to test specific capabilities and select or discard benchmarks altogether, before even using them. 

\subsection{Explanatory Power Analysis: Profiling LLM Abilities}

\begin{tcolorbox}[colback=mybg2,colframe=RQ,boxrule=0pt,  borderline south={2pt}{2pt}{RQ}, arc=1mm,left=4pt, right=6pt, top=4pt, bottom=1pt]
    \textcolor{RQ}{\sffamily\small{{\textbf{RQ3}:  Can we understand the capabilities of models and their evolution in non-saturating plots?}}}
\end{tcolorbox}

In order to explore how well we can explain (and in the next section predict) the behaviour of AI systems, we selected 15 LLMs (see Table~\ref{tab:llms}), and ran them on the \ADeLe battery. Instead of aggregating the results overall we will use the dimensions and levels for a more refined analysis.

\renewcommand{\arraystretch}{1.02}
\begin{table*}[!h]
\centering
\caption{The 15 language models evaluated in this paper. Key abbreviations: SFT, Supervised Fine-tuning; RLHF, Reinforcement Learning from Human Feedback; and CoT, Chain-of-Thought. }
\resizebox{0.8\textwidth}{!}{
\begin{tabular}{C{0.1\textwidth}C{0.1\textwidth}C{0.15\textwidth}C{0.1\textwidth}C{0.12\textwidth}C{0.12\textwidth}C{0.12\textwidth}C{0.12\textwidth}}
\toprule
\textbf{Family} & \textbf{LLM} & \textbf{Version} & \textbf{\#param} & \textbf{SFT} & \textbf{RLHF} & \textbf{CoT} & \textbf{Distilled}  \\  
\midrule
\multirow{6}{*}{OpenAI}   
    & \model{Babbage}    & \model{002}             & 1.3B   & $\circ$   & $\circ$   & $\circ$   & $\circ$   \\ 
    & \model{Davinci}    & \model{002}             & 175B   & $\circ$   & $\circ$   & $\circ$   & $\circ$   \\ 
    & \model{GPT-3.5}    & \model{Turbo}           & –      & $\bullet$ & $\bullet$ & $\circ$   & $\circ$   \\ 
    & \model{GPT-4o}    & –               & –      & $\bullet$ & $\bullet$ & $\circ$   & $\circ$   \\ 
    & \model{o1}        & \model{mini}            & –      & $\bullet$ & $\bullet$ & $\bullet$ & $\bullet$ \\ 
    & \model{o1}         & \model{base}            & –      & $\bullet$ & $\bullet$ & $\bullet$ & $\circ$   \\ 
\midrule
\multirow{5}{*}{LLaMa} 
    & \model{LLaMa 3}    & \model{3.2-1B-Instruct} & 1B     & $\bullet$ & $\bullet$ & $\circ$   & $\circ$   \\ 
    & \model{LLaMa 3}    & \model{3.2-3B-Instruct} & 3B     & $\bullet$ & $\bullet$ & $\circ$   & $\circ$   \\ 
    & \model{LLaMa 3}    & \model{3.2-11B-Instruct} & 11B    & $\bullet$ & $\bullet$ & $\circ$   & $\circ$   \\ 
    & \model{LLaMa 3}    & \model{3.2-90B-Instruct} & 90B    & $\bullet$ & $\bullet$ & $\circ$   & $\circ$   \\ 
    & \model{LLaMa 3}    & \model{3.1-405B-Instruct} & 405B  & $\bullet$ & $\bullet$ & $\circ$   & $\circ$   \\ 
\midrule
\multirow{4}{*}{DeepSeek} 
    & \model{Qwen}       & \model{R1-Dist}          & 1.5B   & $\bullet$ & $\circ$   & $\bullet$ & $\bullet$ \\ 
    & \model{Qwen}       & \model{R1-Dist}          & 7B     & $\bullet$ & $\circ$   & $\bullet$ & $\bullet$ \\ 
    & \model{Qwen}       & \model{R1-Dist}          & 14B    & $\bullet$ & $\circ$   & $\bullet$ & $\bullet$ \\ 
    & \model{Qwen}      & \model{R1-Dist}          & 32B    & $\bullet$ & $\circ$   & $\bullet$ & $\bullet$ \\

\bottomrule
\end{tabular}
}
\label{tab:llms}
\end{table*}

\begin{figure}[!ht]
  \centering
  \includegraphics[width=0.90\linewidth]{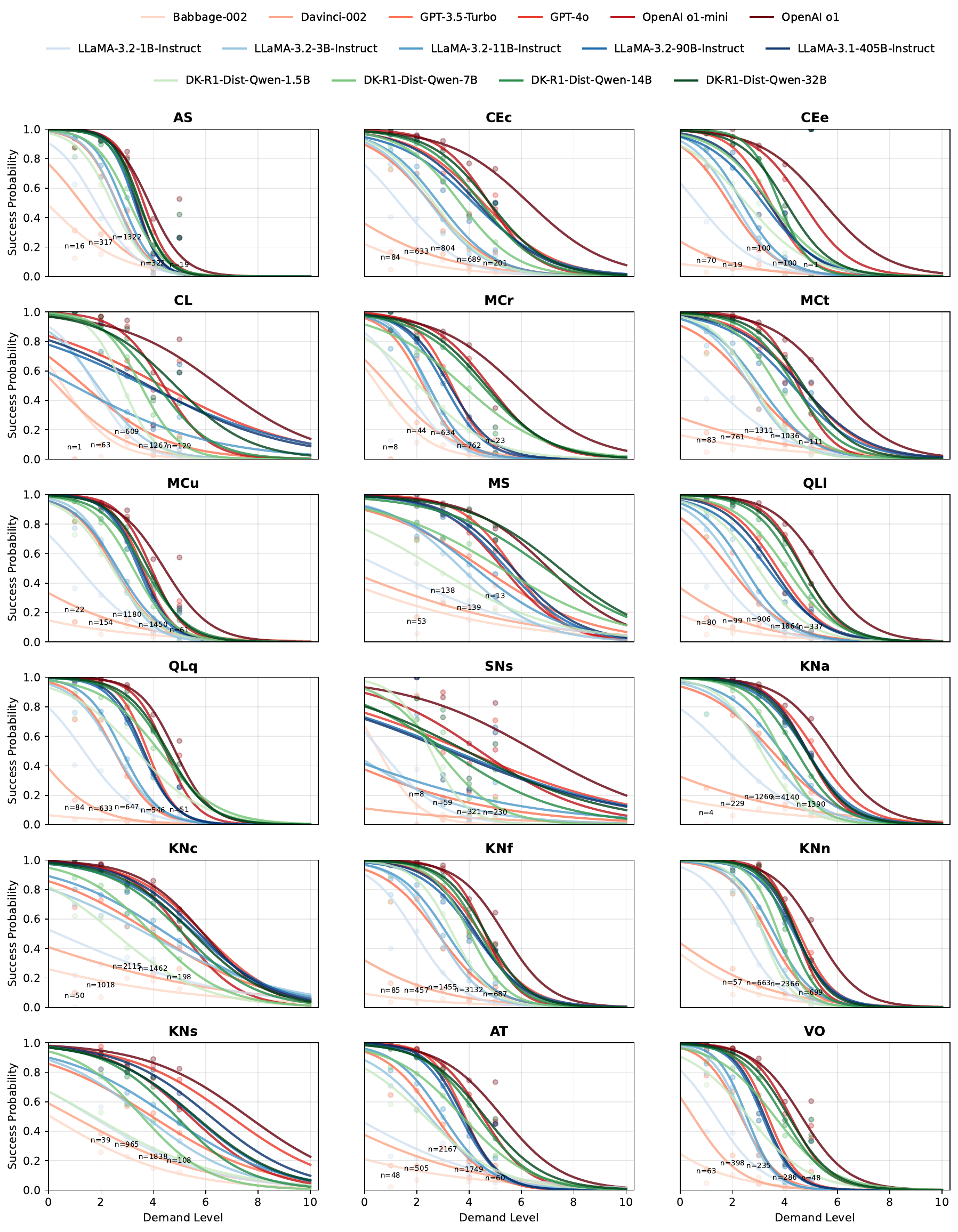}
  \caption{Characteristic curves for the 18 demands and the 15 LLMs. The \xaxis shows the demand levels for that dimension and the \yaxis the average performance (probability of success) for each level. We ensure all bins weight the same in the fit as the largest one (except those bins with less than 100 instances, which use a proportional weight for robustness). The curve is a logistic fit with an anchor at coordinates (20, 0), accounting for 50\% of the total weight.}
  \label{fig:characteristic_curves_combined}
\end{figure}

As we have more than one dimension, 
we do not want higher values of other dimensions to dominate the characteristic curves. Also, there are very few `pure' instances, only having non-zero values for one dimension, and only using them would bias the selection extremely.  
As already discussed---yet will be explained in more detail in the Methods section, we employ a \textit{dominant slice} procedure: for each demand level $l$ along a dimension, we aggregate only those task instances for which the demands in all remaining dimensions do not exceed $l$. This has an effect on the distribution, but arguably smaller than only selecting almost-pure instances.  
 As shown in Figure~\ref{fig:characteristic_curves_combined}, this procedure yields 18 per-dimension curves that capture how model success rates decline with increasing demand. 
 For instance, the curves of certain dimensions are steep and with low variability across models, such as  \dm{AS} (\dm{Attention and Scan}) and  \dm{MCu} (\dm{Calibrating Knowns and Unknowns}). This means the ability (the point of 0.5 success probability), which is around demand levels 3-4, explains (and predicts) success very well for instances in the low and high ranges. 
 In contrast, curves of other dimensions are flatter and show strong differences between models, such as 
  \dm{KNs} (\dm{Knowledge of Social Sciences}), in which the discrimination (between success and failure) is the lowest. 

  Notably, a few dimensions show particularly distinct behaviours.  The characteristic curves for   \dm{MCr} (\dm{Identifying Relevant Information}) and \dm{MS} (\dm{Mind Modelling and Social Cognition}) clearly distinguish the performance of  reasoning models (whether distilled or not): models with chain-of-thought strategies maintain non-negligible success even at high levels of demand (in some cases above 5), while non-reasoning models generally plateau around level 4. This suggests that chain-of-thought is particularly beneficial, even for distilled models, for tasks that require  the extraction of relevant information, and, surprisingly, can have an effect on questions of social cognition, as they usually require the distinction of what there is and what actors think there is. 
  We acknowledge, however, that the current battery includes only a modest number of benchmarks that specifically explore agent-like functions (such as those requiring dynamic interaction or adaptive decision-making), thus limiting our empirical evidence in these areas.  Finally, there are cases where distillation to smaller models is problematic, such as \dm{SNs} (\dm{Spatio-physical Reasoning)}; this kind of reasoning may need both chain-of-thought and scale.
 All subject characteristic curves, in independent plots, can be found in Appendix \ref{sec:sccs}.
 
\begin{figure}[!ht]
  \centering
  \includegraphics[width=1\linewidth]{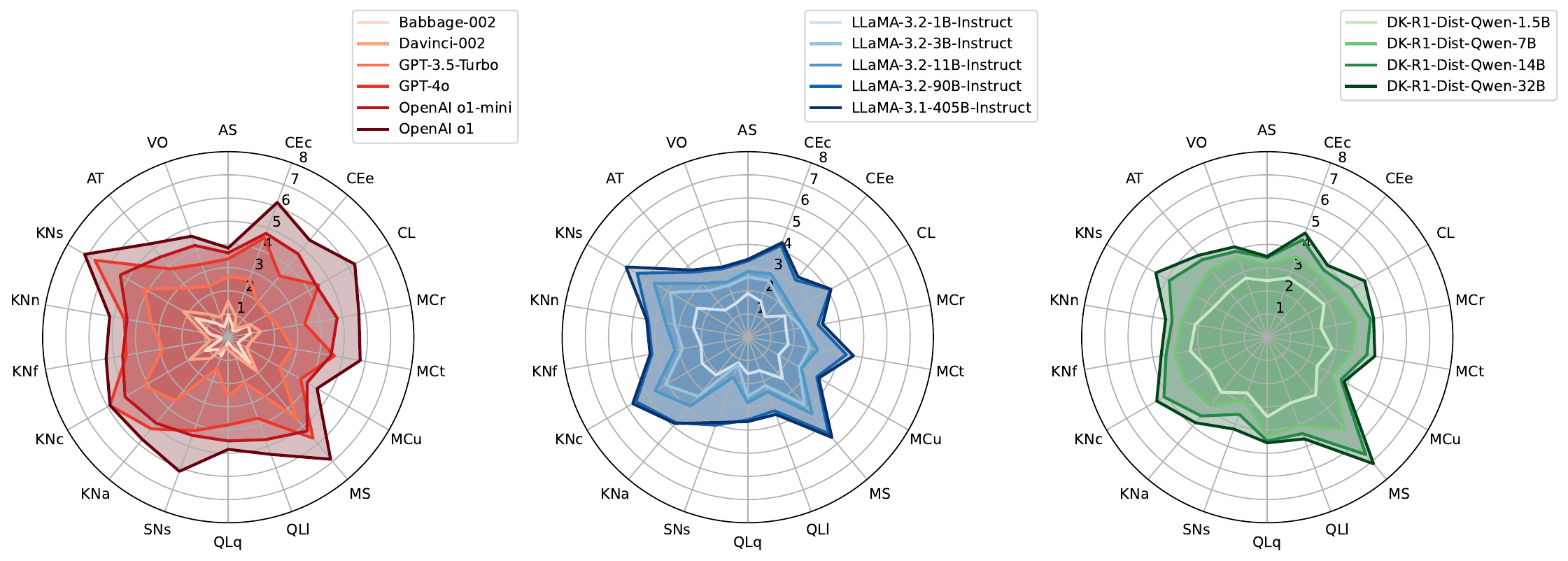}
  \caption{Ability profiles of the 15 LLMs. An ability of $l$ means that there is 50\% probability of the model to succeed on questions at demand level $l$ (that leads to some abilities being beyond 5). In contrast to radial plots usually shown for LLMs in the literature \cite{balachandran2024eureka,fountas2024human} 
  the values shown here are actual abilities on a ratio scale $(0,\infty)$ and the values (in expectation) are more robust to changes in the difficulty distribution of the benchmarks used. In Figure \ref{fig:model_capability_profiles_light}, we additionally show the ability profiles for the 15 LLMs using the \ADeLeLight battery v.1.0, which yields nearly the same profiles.  
  In Appendix \ref{ap:scaling_laws_of_abilities}, we show clear scaling curves of model abilities as a function of \#parameters.
  }
  \label{fig:model_capability_profiles}
\end{figure}

When we presented the subject characteristic curves with Figure \ref{fig:oneSCC}, we said that 
the area 
corresponds to the demand level where the success probability is 0.5, which is also the point with the highest slope. This area is how we estimate the \textit{ability}. 
Note that an ability of 4 does not mean that the model can solve all or most of the items at level 4; it actually means that it can solve half of them in expectation. 
This explains why we have abilities greater than 5, when our scale goes from 0 to 5+. Using this area, we can show a more compact representation of the abilities of all subject LLMs, as represented by  Figure~\ref{fig:model_capability_profiles}.

As can be seen, very old models such as \model{Babbage-002} and \model{Davinci-002} have very low abilities, and new state-of-the-art models have the highest abilities, such as OpenAI \model{o1}. Interestingly, various model curves cross in the middle range. In particular, those dimensions related to knowledge are high for very large models, and reduced for small and distilled models. The reasoning models (such as OpenAI’s \model{o1} and DeepSeek’s \model{R1-Distill}) have clear improvements on the two kinds of
 \dm{QL} (\dm{Quantitative and Logical Reasoning}) but also on \dm{MCr} (\dm{Identifying Relevant
Information}) and \dm{MS} (\dm{Mind Modelling and Social Cognition}) (even  down to 7B in the distilled models). This again suggests that chain-of-thought may contribute to identifying relevant information, in general and possibly even theory-of-mind situations\footnote{Note that the count of \dmc{MS} is only around 300 instances and thus the results should be interpreted with caution. This simultaneously indicates that recent proceedings of AI conferences generally lack diverse and high-quality benchmarks for \dmc{MS}.}, and not only reasoning tasks that look quantitative or logical such as mathematics and coding. 

Finally, the increase of model abilities based on the pure scaling of the number of parameters seems to be marginal when comparing the two largest LLMs for the LLaMA or DK-R1-Distilled-Qwen families; this is further confirmed in Appendix \ref{ap:scaling_laws_of_abilities} where we introduce the very first scaling laws of the actual abilities of LLMs. This is quite opposed to the traditional scaling laws using {\em performance}, which easily saturate close to 100\% accuracy and strongly fluctuate depending on the distributions of the demand levels of the selected benchmarks.

In general, through our approach we can investigate the capabilities of models and their evolution in a granular way, with characteristic curves explaining why each model succeeds or fails in different regions depending on the demand profile of the instance. This explanation emanates from the information collected from the AI system under observation only: unlike IRT and other latent factor approaches (factor analysis or PCA) derived from the results of many systems and instances, the abilities and explanations we get with our methodology are not affected by the results of other AI systems, or the choice of the 15 models of our analysis.  
On top of this, we get insightful intuitions of the expected results at the level of benchmark before running it (just compare figures \ref{fig:benchmark_demand_profiles} and \ref{fig:model_capability_profiles}).   
 In contrast, the usual practice in AI evaluation has been to slice by benchmarks, domains or some tags \cite{srivastava2022beyond,balachandran2024eureka,fountas2024human,masry2025alignvlm} 
but again without considering difficulty in each dimension, leading to values in each dimension that are not commensurate, hard to explain and volatile to the distribution of difficulties: 70\% aggregate accuracy in all logical reasoning benchmarks does not mean more {\em capability} than 50\% aggregate accuracy in all metacognition benchmarks, not even more capability thant 50\% aggregate accuracy in another set of logical reasoning benchmarks. In this work, using a broad set of rubrics, generalising from other approaches with narrower profiles (e.g., navigation and visual abilities in \cite{burden2023inferring}), we get truly general ability profiles that are commensurate across a range of constituent benchmarks.

\subsection{Predictive Power Analysis: Anticipating Performance with Assessors}\label{sec:pred}

\begin{tcolorbox}[colback=mybg2,colframe=RQ,boxrule=0pt,  borderline south={2pt}{2pt}{RQ}, arc=1mm,left=4pt, right=6pt, top=4pt, bottom=1pt]
    \textcolor{RQ}{\sffamily\small{{\textbf{RQ4}: Can we predict instance performance efficiently and robustly in ID and OOD?}}}
\end{tcolorbox}


The annotated demand levels not only provide substantial explanatory power for understanding potential benchmark design issues (e.g., lack of specificity when irrelevant demands are present and lack of sensitivity when key demands are under-represented), but also strong predictive power. 
As shown in the bottom row of Figure~\ref{fig:demand_correlation}, most dimensions are negatively correlated with success, suggesting that, in aggregate, higher demands tend to reduce performance. 
This is promising about their predictive power when used in a multivariate way. This is precisely what we show with three different {\em assessors} \cite{hernandez2022training}, meta-models that predict success (a binary label) at the instance level.

To quantify the predictive power, we trained three types of probabilistic classifiers. The first assessor is a \model{Random Forest} (\model{RF}) that maps the interpretable 19-dimensional demand annotation vector directly into a success prediction. The second assessor is another 
\model{RF} model that relies on pre-computed  \model{GloVe} embeddings extracted from the raw text of each question, and the third is based on a fine-tuned 
\model{LLaMA} model trained end-to-end on the question text to predict success. Further details are provided in Methods section \ref{sec:assessors}. 

\textit{In-distribution} results (Table~\ref{tab:main-id-predictability-results}) show that despite the large imbalance in some of the base model accuracies (from 0.102 for \model{Babbage-002} to 0.843 for \model{OpenAI o1}), the demand-based RF achieves high discrimination 
(between success and failure), as measured by 
AUROC,  and near perfect calibration, as quantified by ECE. In terms of discrimination, we see that the best result is achieved for \model{GPT-4o} (0.88 in AUROC), being the most predictable model for the three assessors, while small models are less predictable. Averaged across all 15 LLMs, the demand-based RF produces an accuracy-weighted average AUROC of approximately 0.84, which is on par with the performance of the fine-tuned LLaMA assessor, while its average ECE (0.01) is significantly lower than that of the other approaches (0.03 for the GloVe-based model and 0.04 for the fine-tuned LLaMA model). 
Calibration plots demonstrating these results are provided in Appendix \ref{ap:IID_assessors_calibration}. 
This strong in-distribution performance is very encouraging, as a well-calibrated predictor is critical for estimating both batch-level performance (e.g., for an entire benchmark) and the probability of success on individual instances.

\renewcommand{\arraystretch}{1.02}
\begin{table*}[!h]
\centering
 \caption{In-Distribution predictability results of 15 LLMs for the \ADeLe battery using 10-fold cross-validation. The first two columns show names of subject LLMs and the overall accuracy of subject LLMs on the \ADeLe battery. The remaining three pairs of columns show the AUROC and ECE of three different assessors (\modelc{RF} using demands, \modelc{RF} using average GloVe embeddings, and finetuning \modelc{LLaMA-3.1-8B}). For a single LLM subject, the training time is 4 seconds and 160 seconds for the demand-based and embeddings-based assessors respectively on a M3 Pro CPU, while the fine-tuned \modelc{LLAMA} assessor costs 300 hours on a single V100 GPU. The weighted average is only indicative for easy comparison, and uses the normalised LLM Accuracy as a weight in the mean, giving more relevance to more powerful models, which are more representative now and in the near future.}
 \resizebox{0.87\textwidth}{!}{
  \begin{tabular}{R{0.25\textwidth}C{0.17\textwidth}C{0.1\textwidth}C{0.1\textwidth}C{0.1\textwidth}C{0.1\textwidth}C{0.1\textwidth}C{0.1\textwidth}}
\toprule
\multirow{2}{*}{\textbf{Subject LLM}}
& \multirow{2}{*}{\textbf{LLM Accuracy}$\uparrow$ } & \multicolumn{2}{c}{\textbf{Demands (\model{RF})}} & \multicolumn{2}{c}{\textbf{Embeddings (\model{RF})}} & \multicolumn{2}{c}{\textbf{Finetuning (\model{LLAMA})}}\\\cmidrule(l){3-4}  \cmidrule(l){5-6} \cmidrule(l){7-8} 
 &  & \textbf{AUROC}$\uparrow$ & \textbf{ECE}$\downarrow$ & \textbf{AUROC}$\uparrow$ & \textbf{ECE}$\downarrow$  & \textbf{AUROC}$\uparrow$ & \textbf{ECE}$\downarrow$  \\  
\midrule
\model{Babbage-002}              & 0.102 & 0.786 & \textbf{0.004} & 0.784 & 0.012 & \textbf{0.794} & 0.026 \\ 
\model{Davinci-002}              & 0.157 & 0.774 & \textbf{0.005} & 0.770 & 0.014 & \textbf{0.789} & 0.032 \\ 
\model{GPT-3.5-Turbo}            & 0.414 & 0.811 & \textbf{0.007} & 0.780 & 0.029 & \textbf{0.817} & 0.052 \\ 
\model{GPT-4o}                   & 0.713 & \textbf{0.882} & \textbf{0.014} & 0.852 & 0.041 & 0.879 & 0.039 \\ 
\model{OpenAI o1-mini}           & 0.770 & 0.860 & \textbf{0.011} & 0.821 & 0.023 & \textbf{0.861} & 0.041 \\ 
\model{OpenAI o1}                & 0.843 & \textbf{0.853} & \textbf{0.011} & 0.810 & 0.025 & 0.848 & 0.031 \\ \midrule
\model{LLaMA-3.2-1B-Instruct}    & 0.216 & 0.785 & \textbf{0.006} & 0.759 & 0.014 & \textbf{0.788} & 0.041 \\ 
\model{LLaMA-3.2-3B-Instruct}    & 0.378 & 0.813 & \textbf{0.008} & 0.782 & 0.028 & \textbf{0.822} & 0.048 \\ 
\model{LLaMA-3.2-11B-Instruct}   & 0.463 & 0.820 & \textbf{0.009} & 0.793 & 0.034 & \textbf{0.828} & 0.055 \\ 
\model{LLaMA-3.2-90B-Instruct}   & 0.645 & \textbf{0.860} & \textbf{0.012} & 0.832 & 0.042 & \textbf{0.860} & 0.042 \\ 
\model{LLaMA-3.1-405B-Instruct}  & 0.683 & \textbf{0.870} & \textbf{0.011} & 0.831 & 0.040 & 0.864 & 0.040 \\ \midrule
\model{DK-R1-Dist-Qwen-1.5B}     & 0.353 & 0.781 & \textbf{0.014} & 0.749 & 0.028 & \textbf{0.797} & 0.052 \\ 
\model{DK-R1-Dist-Qwen-7B}       & 0.555 & 0.813 & \textbf{0.015} & 0.788 & 0.039 & \textbf{0.821} & 0.051 \\ 
\model{DK-R1-Dist-Qwen-14B}      & 0.698 & 0.828 & \textbf{0.013} & 0.796 & 0.031 & \textbf{0.829} & 0.044 \\ 
\model{DK-R1-Dist-Qwen-32B}      & 0.748 & \textbf{0.841} & \textbf{0.013} & 0.799 & 0.031 & 0.839 & 0.045 \\
\midrule

Weighted Average & --- & 0.839 & \textbf{0.011} & 0.805 & 0.032 & \textbf{0.840} & 0.043 \\
\bottomrule
\end{tabular}
}
\label{tab:main-id-predictability-results}
\end{table*}

We further evaluated predictive performance under {\em out-of-distribution} conditions by withholding entire tasks (task OOD) or entire benchmarks (benchmark OOD) from training (Tables
~\ref{tab:main-task-ood-predictability-results} and \ref{tab:main-benchmark-ood-predictability-results} respectively). 
In the \textit{task OOD} setup, the predictive power of the demand-based assessor retains robust performance 
(weighted AUROC=0.81, ECE=0.02)
, only slightly lower than in-distribution, and outperforms  the rest 
of assessors, whose performance slightly decreases (achieving weighted AUROC values of 0.79 for the LLaMA-based and 0.74 for the GloVe-based assessors). In the more challenging \textit{benchmark OOD}, the performance of the demand-based assessor decreases more significantly (Weighted AUROC=0.75 and ECE=0.04), with the decrease observed mainly in the reasoning models  (AUROC decreases between 0.05 and 0.11), while the non-reasoning models suffer only small decreases (between 0.01 and 0.06). 
This decrease is especially small for the largest non-reasoning LLMs, such as \model{GPT-4o} and \model{LLaMA-3.1-405B-Instruct}. 
In contrast, the performance of the other two assessors suffers a drastic drop.
This trend suggests that the demand-based predictor is less prone to overfitting with spurious features compared to its counterparts. 

Although the use of demand annotations outperforms the other approaches, two key factors explain why the predictive power declines in out-of-distribution settings. First, because our analysis includes only 
63 tasks from 20 benchmarks---many of which (e.g., \ds{ChemLLMBench}) have non-overlapping demand distributions---
the training data does not fully capture the multidimensional demand space. 
We hypothesise that the predictive power of the demand-based assessor for any arbitrary new tasks or new benchmarks can be boosted to the level of in-distribution by ensuring the training data's demand distribution is efficiently covering the multivariate demand space.

Second, there is a paucity of extremely difficult instances to challenge the high performance models 
(e.g. \model{OpenAI o1-mini}, \model{OpenAI o1}, \model{DK-R1-Dist-Qwen-32B}). As shown in Figure~\ref{fig:characteristic_curves_combined}, even at level 5 (where instance coverage is low) the best models maintain success probabilities well above zero. 
Nonetheless, there will be instances deserved to be labelled level 5 and even beyond in future efforts, which can form bins yielding much lower success probability, therefore increasing the discrimination of the demand-based assessor's predictions. A similar narrative can go for justifying the relatively lower discrimination observed in less capable models (e.g. Babbage-002, Davinci-002), as there are little extremely easy instances for them in the data (Figure~\ref{fig:characteristic_curves_combined}); the lack of instruction-tuning for these models (Table \ref{tab:llms}) may be an extra reason since these models frequently repeat the prompts instead of solving the problems specified in the prompts, in an seemingly elusive and arbitrary way. Therefore, the lower discrimination observed when moving from in-distribution to out-of-distribution is likely mainly due to the lack of diverse data that efficiently cover the multidimensional demand space, as calibration error is consistently low. In Appendix \ref{ap:aleatoric_uncertainty_sources}, we further discuss these factors and potential remedies; we additionally discuss certain rationale on why the predictive power of our demand-based assessor is actually underrated due to upper bounds introduced by imperfect data quality and the automated grading, leaving rooms for future improvements. 


While many traditional IRT methods can explain performance on seen items, they cannot be used to predict performance for new instances (except LLTM, see section \ref{sec:relatedwork} of related work), which is paramount in a deployment setting where the goal is to anticipate whether AI will perform well in unseen scenarios, rather than merely grading subjects in a testing environment. The natural baseline in AI evaluation practice is just average accuracy. This can extrapolate to an extent for system selection, but in the case of instance prediction, it leads to no discriminative power at all (an AUROC of 0.5), and calibration that is only good for in-distribution. Overall, the supremacy in predictive power observed for the demand-based assessor is clear. It only needs 19 dimensions\footnote{Even better, in appendix \ref{ap:other_assessors}, we prove that with 11 broad dimensions or even only with 6 specific dimensions (that exhibit the highest feature importance scores for the demand-based assessor), one can achieve already comparably high predictive power.} and they are all interpretable, in comparison with the two much larger and uninterpretable baselines. This is strongly encouraging, shedding light on a promising future for the reliable deployment of AI.

\renewcommand{\arraystretch}{1.02}
\begin{table*}[!h]
\centering
 \caption{Task Out-of-Distribution predictability results (all other things equal to Table \ref{tab:main-id-predictability-results}).}
 
 \resizebox{0.87\textwidth}{!}{
  \begin{tabular}{R{0.25\textwidth}C{0.17\textwidth}C{0.1\textwidth}C{0.1\textwidth}C{0.1\textwidth}C{0.1\textwidth}C{0.1\textwidth}C{0.1\textwidth}}
\toprule
\multirow{2}{*}{\textbf{Subject LLM}}
& \multirow{2}{*}{\textbf{LLM Accuracy}$\uparrow$ } & \multicolumn{2}{c}{\textbf{Demands (\model{RF})}} & \multicolumn{2}{c}{\textbf{Embeddings (\model{RF})}} & \multicolumn{2}{c}{\textbf{Finetuning (\model{LLAMA})}}\\\cmidrule(l){3-4}  \cmidrule(l){5-6} \cmidrule(l){7-8} 
 &  & \textbf{AUROC}$\uparrow$ & \textbf{ECE}$\downarrow$ & \textbf{AUROC}$\uparrow$ & \textbf{ECE}$\downarrow$  & \textbf{AUROC}$\uparrow$ & \textbf{ECE}$\downarrow$  \\   
\midrule
Babbage-002                        & 0.102 & \textbf{0.751} & \textbf{0.007} & 0.727 & 0.019 & 0.719 & 0.046 \\ 
Davinci-002                        & 0.157 & 0.741 & \textbf{0.007} & 0.703 & 0.025 & \textbf{0.746} & 0.055 \\ 
GPT-3.5-Turbo                      & 0.414 & \textbf{0.795} & \textbf{0.020} & 0.719 & 0.032 & 0.773 & 0.088 \\ 
GPT-4o                             & 0.713 & \textbf{0.852} & \textbf{0.023} & 0.789 & 0.073 & 0.831 & 0.067 \\ 
OpenAI o1-mini                     & 0.770 & \textbf{0.837} & \textbf{0.021} & 0.751 & 0.038 & 0.814 & 0.068 \\ 
OpenAI o1                          & 0.843 & \textbf{0.811} & 0.033 & 0.730 & \textbf{0.030} & 0.761 & 0.101 \\ \midrule
LLaMA-3.2-1B-Instruct              & 0.216 & \textbf{0.733} & \textbf{0.026} & 0.671 & 0.033 & 0.732 & 0.081 \\ 
LLaMA-3.2-3B-Instruct              & 0.378 & \textbf{0.791} & \textbf{0.016} & 0.724 & 0.020 & 0.780 & 0.084 \\ 
LLaMA-3.2-11B-Instruct          & 0.463 & \textbf{0.799} & \textbf{0.022} & 0.733 & 0.037 & 0.783 & 0.106 \\ 
LLaMA-3.2-90B-Instruct             & 0.645 & \textbf{0.834} & \textbf{0.021} & 0.763 & 0.068 & 0.809 & 0.050 \\ 
LLaMA-3.1-405B-Instruct            & 0.683 & \textbf{0.843} & \textbf{0.023} & 0.766 & 0.067 & 0.811 & 0.060 \\ \midrule
DK-R1-Dist-Qwen-1.5B               & 0.353 & 0.757 & \textbf{0.019} & 0.700 & 0.029 & \textbf{0.764} & 0.071 \\ 
DK-R1-Dist-Qwen-7B                 & 0.555 & \textbf{0.790} & \textbf{0.018} & 0.735 & 0.042 & 0.776 & 0.083 \\ 
DK-R1-Dist-Qwen-14B                & 0.698 & \textbf{0.808} & \textbf{0.018} & 0.737 & 0.054 & 0.772 & 0.085 \\ 
DK-R1-Dist-Qwen-32B                & 0.748 & \textbf{0.812} & \textbf{0.026} & 0.739 & 0.057 & 0.793 & 0.063 \\ \midrule
Weighted Average & --- & \textbf{0.810} & \textbf{0.022} & 0.740 & 0.047 & 0.788 & 0.075 \\
\bottomrule
\end{tabular}
}
\label{tab:main-task-ood-predictability-results}
\end{table*}

\renewcommand{\arraystretch}{1.02}
\begin{table*}[!h]
\centering
 \caption{Benchmark Out-of-Distribution predictability results (all other things equal to Table \ref{tab:main-id-predictability-results}).}
\resizebox{0.87\textwidth}{!}{
  \begin{tabular}{R{0.25\textwidth}C{0.17\textwidth}C{0.1\textwidth}C{0.1\textwidth}C{0.1\textwidth}C{0.1\textwidth}C{0.1\textwidth}C{0.1\textwidth}}
\toprule
\multirow{2}{*}{\textbf{Subject LLM}}
& \multirow{2}{*}{\textbf{LLM Accuracy}$\uparrow$ } & \multicolumn{2}{c}{\textbf{Demands (\model{RF})}} & \multicolumn{2}{c}{\textbf{Embeddings (\model{RF})}} & \multicolumn{2}{c}{\textbf{Finetuning (\model{LLAMA})}}\\\cmidrule(l){3-4}  \cmidrule(l){5-6} \cmidrule(l){7-8} 
 &  & \textbf{AUROC}$\uparrow$ & \textbf{ECE}$\downarrow$ & \textbf{AUROC}$\uparrow$ & \textbf{ECE}$\downarrow$  & \textbf{AUROC}$\uparrow$ & \textbf{ECE}$\downarrow$  \\  
\midrule
Babbage-002                      & 0.102 & \textbf{0.694} & \textbf{0.027} & 0.689 & 0.062 & 0.649 & 0.070 \\ 
Davinci-002                      & 0.157 & \textbf{0.718} & \textbf{0.014} & 0.626 & 0.066 & 0.633 & 0.086 \\ 
GPT-3.5-Turbo                    & 0.414 & \textbf{0.776} & \textbf{0.041} & 0.628 & 0.074 & 0.691 & 0.146 \\ 
GPT-4o                           & 0.713 & \textbf{0.826} & \textbf{0.058} & 0.398 & 0.167 & 0.740 & 0.136 \\ 
OpenAI o1-mini                   & 0.770 & \textbf{0.728} & \textbf{0.026} & 0.422 & 0.142 & 0.684 & 0.132 \\ 
OpenAI o1                        & 0.843 & \textbf{0.710} & \textbf{0.015} & 0.404 & 0.117 & 0.704 & 0.095 \\ \midrule
LLaMA-3.2-1B-Instruct            & 0.216 & \textbf{0.716} & \textbf{0.048} & 0.602 & 0.112 & 0.623 & 0.083 \\ 
LLaMA-3.2-3B-Instruct            & 0.378 & \textbf{0.778} & \textbf{0.036} & 0.618 & 0.096 & 0.687 & 0.066 \\ 
LLaMA-3.2-11B-Instruct        & 0.463 & \textbf{0.786} & \textbf{0.053} & 0.591 & 0.067 & 0.721 & 0.118 \\ 
LLaMA-3.2-90B-Instruct           & 0.645 & \textbf{0.804} & \textbf{0.055} & 0.463 & 0.115 & 0.721 & 0.144 \\ 
LLaMA-3.1-405B-Instruct          & 0.683 & \textbf{0.818} & \textbf{0.044} & 0.389 & 0.186 & 0.712 & 0.135 \\ \midrule
DK-R1-Dist-Qwen-1.5B             & 0.353 & \textbf{0.705} & \textbf{0.049} & 0.580 & 0.102 & 0.662 & 0.106 \\ 
DK-R1-Dist-Qwen-7B               & 0.555 & \textbf{0.676} & \textbf{0.043} & 0.534 & 0.060 & 0.649 & 0.160 \\ 
DK-R1-Dist-Qwen-14B              & 0.698 & \textbf{0.691} & \textbf{0.025} & 0.461 & 0.099 & 0.673 & 0.135 \\ 
DK-R1-Dist-Qwen-32B              & 0.748 & \textbf{0.703} & \textbf{0.027} & 0.426 & 0.103 & 0.696 & 0.100 \\ \midrule
Weighted Average & --- & \textbf{0.747} & \textbf{0.037} & 0.480 & 0.114 & 0.692 & 0.121 \\
\bottomrule
\end{tabular}
}
\label{tab:main-benchmark-ood-predictability-results}
\end{table*}

\section{Discussion}\label{sec:outlook}

To date, AI evaluation is not meeting the demands of a fast-changing and increasingly diverse AI ecosystem. Understanding and anticipating performance has become an urgent requirement for a swath of general-purpose AI systems. By building and exploiting absolute demand scales for annotating thousands of instances via automated rubrics,  we have established a promising new direction for AI evaluation.
The methodology we have presented and illustrated is comprehensive, scalable and standardised, addressing many of the issues of conventional AI evaluation practice: 
a lack of explanatory and predictive power, as well as saturating and overfitting to specific populations of benchmarks and AI systems respectively. With the pace and penetration of general-purpose AI, a rigorous, scalable and pipelined evaluation has been urgently demanded by researchers, companies, third-party evaluators, policy-makers and regulators. It is paradoxical that powerful LLMs as annotators have made this new methodology possible and scalable. 
The explanatory value of 16,108 LLM annotations per dimension has been independently validated by humans through inter-rater analysis and the Delphi method~\cite{linstone1975delphi}, and their predictive power stands through task diversity.

Nonetheless, our work is not without limitations. First, the 
\DeLeAn rubric set does not fully cover certain dimensions, such as navigation, and excludes capabilities in other modalities and paradigms in AI, such as multimodal systems and robotics, given 
we limited our analysis to LLMs. We encourage other researchers to extend the rubrics to further dimensions (including propensities, values and other elements that are specifically conceived for safety or fairness \cite{zeng2024quantifying,yao2025value}), and evaluate other kinds of AI systems with them. Second, there are very few high-quality level 5+ items in our current battery. 
Given the pace of progress in AI, the current scales (up to 5+) will need to be extended in a way that remains backward-compatible with existing scales. For the annotation, instead of a single rubric from 0 to 10, which would challenge the best human and machine annotators, we propose to keep each current rubric up to the current 0 to 5+ levels, and then add a second, independent, rubric only applied to those items initially labelled as 5+ to be relabelled from 5 to 10+, following the ratio scale we followed from 0 to 5+.  New items, such as less adversarial versions of those in Humanity's Last Exam \cite{phan2025humanity} and other very challenging benchmarks \cite{wang2025enigmaeval,kazemi2025big}, 
could be included in \ADeLe v2.0 battery. Third, more benchmarks and more diversity (even for the same dimensions and scale) would definitely further increase the predictive power in- and, most importantly, out-of-distribution, especially if we introduce more benchmarks with `purer' items only loaded on a few  demands. Fourth, the predictive power of the demand-based assessors will likely increase as the LLMs improve as input-response graders and demand annotators, and as the \ADeLe battery efficiently grows with more challenging and diverse items. 

Overall, 
the new methodology shows that a successful development of the construct-oriented paradigm  in AI evaluation \cite{burden2025paradigmsaievaluationmapping} is possible, integrating perspectives from different disciplines. It 
will be 
instrumentalised through 
a streamlined platform that can grow in the years to come,  
ready to explain and predict 
the performance and safety of AI systems. 
In a moment where AI evaluation is at the crux of research and regulations, and the science of evaluation had not yet digested the pace of general-purpose AI, our work takes crucial steps to make AI evaluation fit for purpose. 

\bibliography{reference}
\bibliographystyle{apalike}

\newpage
\section{Methods}\label{sec:methods}

\subsection{Scales and Rubrics}\label{sec:scales}

For more than a century, psychology has introduced many constructs with explanatory and predictive power about human behaviour, from conscientiousness to metacognition. Based on experimental data and theories of {\em human} cognition, these constructs are usually organised into hierarchical taxonomies, such as the Cattell-Horn-Carroll structure of human cognitive abilities~\cite{mcgrew2005cattell} or the Big Five personality traits \cite{rust2014modern}. In principle, we could build a similar taxonomy for {\em artificial} cognition, based on theory and experiments about machine behaviour \cite{rahwan2019machine}. However, as the base population of machines is much more arbitrary and changing than those of humans, it makes more sense to devise a taxonomy that could encompass any kind of natural and artificial intelligence, by considering capabilities that are meaningful for more general theories of cognition 
 \cite{hernandez2017measure}. Integrating and generalising taxonomies from human psychology, comparative cognition and artificial intelligence \cite{hernandez2017measure}, a general taxonomy of 14 capabilities was designed by \cite{hernandez2019ai} and later extended with corresponding 14 rubrics \cite{tolan2021measuring} for the study of AI in the workplace. 
 
 From this original set, we selected those that are applicable to text only (e.g., `auditory processing' and `visual processing' were discarded), and further subdivided several of them into finer subcategories, resulting in a final set of 11 `primordial' cognitive capability scales. We added 5 `knowledge' dimensions and 3 `extraneous' dimensions, making a total of 18 demand dimensions plus \dm{UG} (\dm{Unguessability}), as summarised in  Table~\ref{tab:dimensions}. In Section~\ref{sec:dimensions} we explained why the taxonomy was extended with the `extraneous' dimensions, to account for contamination, amalgamation and funnelling phenomena. The subdimensions in the `primordial' category were included after multiple rounds of discussions about whether some of the original 7 broad ones could be carved into finer, but still general, subdimensions that could be conceptually be distinct. We split \dm{CE} into two (we noticed a clear case for tasks that require comprehension but no expression), \dm{MC} into three (since we unpacked very different notions inside metacognition and critical thinking), and \dm{QL} into two (distinguishing logical from quantitative reasoning), ending up with a total of 11 `primordial' dimensions. We added the knowledge dimensions because the primordial ones would not account for attainment in scientific or humanistic domains. Again, there are endless specialities of knowledge, and we decided to settle for four broad domains of science: applied sciences, formal sciences, natural sciences and social sciences. We also added a fifth domain: customary everyday knowledge (\dm{KNc}). This captures information that most people (adults) know about society, from sport players to local traditions. This \dm{KNc} does {\em not} represent common sense, such as `soup needs a spoon but not a knife'. This should be accounted by other dimensions. By navigating Appendix~\ref{sec:rubrics} we can better understand the tradeoffs in the construction of the rubrics. It is important to highlight that the taxonomy is not definitive, and is meant to be extended in the future using the same criteria of dimensions being general and being conceptually distinct.

Based on the original 7 top-level primordial rubrics, which only determined the presence or absence of the need for each capability in a task, we extended them and added new ones as numeric {\em demand} scales in the range $(0,\infty)$. We deliberately design these scales as ratio scales \cite{hand2016measurement}, with an absolute 0 (no demand) and differences that are comparable all across the scale. In the social sciences a common interest lies in understanding differences, as no human has zero capabilities, and an `interval' scale with negative capabilities makes sense (as in IRT) or as percentiles of a normal distribution (as IQ scores). We argue that for AI we should aim for the top level in Steven's topology of measurement \cite{stevens1946theory}: the `ratio' scales. Ratio scales have all the properties of the previous scales: intervals and differences are meaningful, but ratios are as well. For instance, length in metres, time in seconds or Kelvin degrees are ratio scales, but Celsius or Fahrenheit degrees are only interval scales, because they do not have an absolute zero. Given the flexibility in which we can regulate compute and time use in AI, it makes more sense to set an absolute zero (no compute) on the demands and build the scales in such a way that ratios are meaningful. We would like to say that instance $x_i$ at level 6 {\em doubles} the demand of an instance $x_j$ at level 3. Taking into account that we fit logistic functions, this can be understood in terms of the log odds of being correct {\em halving} when moving $2x$ in the scale and {\em doubling} when moving $x/2$ in the scale \cite{freund2019rasch}.

For this first version of the scales, we decided to choose levels $(0,5)$ of the full range $(0, \infty)$ for practical reasons. With a single rubric, it is hard for humans and LLMs to refine beyond five ordinal values ---this is why Likert scales are so popular. Note that the rubrics only show cases in an ordinal scale between 0 and 5, and the annotations are discrete, never generating non-integer values. This is convenient for avoiding the need of binning for the curves and the demand histograms, but the values become fully continuous when estimating the abilities. 
 In any case, it is usual to consider originally ordinal scales as interval or ratio scales when the number of levels is 5 or more \cite{rhemtulla2012can}. 
We consider them ratio scales because the magnitudes are not interpreted as a mere rank. 
In particular we determine the way each particular dimension increases the difficulty of the demands from 0 to 5 and beyond. In other words, the way it increases depends on what the demand represents, but the pace of increase, the actual scale, is chosen in such a way that they are commensurate. For instance, for knowledge dimensions (applied sciences, customary everyday knowledge, formal sciences, natural sciences, and social sciences and humanities) we thought of levels corresponding roughly to elementary, middle, high, undergraduate and graduate education. By looking at the attainment rates of some statistical data of education level rates (e.g., OECD data \cite{OECD2024}), and the specialisation of domains as the educational level increases, we noticed that the questions of level $l$ usually were sufficiently advanced to have roughly one person in $10^{l-1}$ solving it right. 
Then, we extend this criterion as a rule of thumb for all scales. With this we achieve a ratio scale consistency across each scale, and have commensurate scales across dimensions. In general, an item is 
at level $l$ if $l$ is the highest number such that, in at least 95\% of samples of $n=10^l$ individuals, there is at least one correct response. 
The levels we have defined are 0 (None), 1 (Very low), 2 (Low), 3 (Intermediate), 4 (High) and 5+ (Very high), with $n$ going from 1 to 100,000.

We could have calibrated some dimensions using procedurally generated examples. For instance, in reasoning, we could have increased the components of reasoning processes \cite{mirzadeh2024gsm} to see if the levels increase  accordingly, but each of these `scales' would have been incommensurate with each other, and not sufficiently general. In addition, our framework deliberately includes additional extraneous requirements such as \dm{Volume} (\dm{VO}) and \dm{Atipicality} (\dm{AT}), which capture the time required to read and complete a task and the familiarity of the instance, respectively. For example, our \dm{Unguessability} (\dm{UG}) scale is based on the probability of a correct answer by chance: a multiple-choice question with 10 options yields a 90\% error rate and is therefore assigned level 1 (reflecting that one in $10^1$ individuals would succeed by random guessing). However, if the error rate is close to 100\%---indicating that there is a non-zero chance (such as ``a monkey hits a keyword")---the level is not unlimited, but could be assigned to our 5+ category. In the end, we left this dimension as the only one between 0 and 100, but can be converted to the common range using this criterion.  

The 18 rubrics were crafted following the above criteria, using several iterations while trying with human and AI annotators. Once the rubrics were settled, we conducted the experiments. The final rubrics can be found in section~\ref{sec:rubrics}.

\subsection{LLM Annotators}\label{sec:llmannotators}

We prompt an LLM to annotate task demands levels (on a discrete scale from 0 to 5) at instance-level for all individual rubrics (see \DeLeAn Rubric Set v.1.0 in Appendix \ref{sec:rubrics}). In particular, we prompt GPT-4o\footnote{`GPT-4o' always refers to the specific `gpt-4o-0513' checkpoint in this work.} \cite{hurst2024gpt4o} using Azure AI API 
utilising CoT prompting (Table \ref{tab:llm_annotator_prompt}) at temperature set to 0 with a maximum output token length of 1000, to ensure getting long enough responses that include the answers for nearly all instances and substantially reduce the cost. We use the default values for the stopping condition and the rest of the parameters. 

\begin{table}[!h]
\centering
\ttfamily \small
\begin{tabular}{@{}p{13cm}@{}}
\toprule
\textbf{\underline{QUERY}:} {The following rubric describes six distinct levels of *\{\$dimension\$\}* required by different tasks:}
 \\[5pt]
\{\$rubric\$\}\\[5pt]

\textbf{\underline{TASK INSTANCE}:} {\{\$instance\$\}} \\[5pt]

\textbf{\underline{INSTRUCTION}:} {Score the level of *\{\$dimension\$\}* demanded by the given TASK INSTANCE using a discrete value from 0 to 5. Use CHAIN-OF-THOUGHTS REASONING to reason step by step before assigning the score. After the CHAIN-OF-THOUGHTS REASONING STEPS, conclude your assessment with the statement: "Thus, the level of *\{\$dimension\$\}* demanded by the given TASK INSTANCE is: SCORE", where `SCORE' is the integer score you have determined.} \\[5pt]
\bottomrule
\caption{Prompt template for deriving demand annotations with GPT-4o.}
\label{tab:llm_annotator_prompt}
\end{tabular}
\end{table}

\subsection{Inter-rater Analysis}\label{sec:interrater}

For each demand, we randomly sampled 50 instances while ensuring each level had at least a sample size of 3 to avoid minority levels getting neglected in our inter-rater analysis. This led to 900 instances to be annotated, which were distributed to five humans (authors in this paper, corresponding to Y.H, Y.M-D, L.Z, Q.Z, S.Z), where each instance was annotated by exactly three humans. The annotation process consisted of two steps. First, each annotator independently assigned a difficulty level (using the 0 to 5+ scale) to each instance using the rubrics.  Next, the annotators met for a Delphi~\cite{linstone1975delphi} consensus meeting. During this meeting, instances for which the minimum and maximum ratings of the three annotators differed by two or more points were discussed in detail until a consensus was reached. For cases with differences of less than two points, a simple majority vote determined the final annotation. To check the inter-rater agreement rates, we use the $r_{WG}$ index \cite{james1984estimating, lebreton2008answers} with the default rectangular null distribution; a score above 0.7 is generally regarded as a good agreement rate.

\begin{table}[!h]
\centering
 \resizebox{0.325\textwidth}{!}{
\begin{tabular}{ccc}
\toprule
\textbf{Dimension} & \textbf{Humans} & \textbf{Delphi \& GPT-4o} \\
\midrule
\dm{AS}    & 0.91 & 0.86 \\
\dm{CEc}   & 0.91 & 0.87 \\
\dm{CEe}   & 0.90 & 0.94 \\
\dm{CL}    & 0.78 & 0.82 \\
\dm{MCr}   & 0.79 & 0.84 \\
\dm{MCt}   & 0.88 & 0.91 \\
\dm{MCu}   & 0.80 & 0.81 \\
\dm{MS}    & 0.77 & 0.86 \\
\dm{QLl}   & 0.85 & 0.89 \\
\dm{QLq}   & 0.84 & 0.84 \\
\dm{SNs}   & 0.87 & 0.89 \\
\dm{KNa}   & 0.73 & 0.75 \\
\dm{KNc}   & 0.86 & 0.83 \\
\dm{KNf}   & 0.86 & 0.81 \\
\dm{KNn}   & 0.91 & 0.94 \\
\dm{KNs}   & 0.70 & 0.86 \\
\dm{AT}    & 0.80 & 0.83 \\
\dm{VO}    & 0.84 & 0.91 \\
\midrule
Average    & 0.83 & 0.86 \\
\bottomrule
\end{tabular}
}
\caption{Column “Humans” shows the agreement of ratings ($r_{WG}$ scores) obtained between humans prior to Delphi consensus, while “Delphi \& GPT-4o” reports the agreement between Delphi consensus and \modelc{GPT-4o}.}
\label{tab:interrater_analysis}
\end{table}

The result is shown in Table \ref{tab:interrater_analysis}, where we observe satisfactory $r_{WG}$ scores (average = 0.86) between Delphi consensus and GPT-4o, consistently above 0.80 except for one dimension with 0.75. However, the $r_{WG}$ scores between humans prior to the Delphi consensus meeting were slightly lower for certain dimensions. These initial disagreements are due to multiple reasons, identified during our Delphi consensus meetings: occasional misinterpretations of certain words or terminologies, mainly for those humans whose primary language for daily use is not English; knowledge gaps in annotating certain particularly challenging task instances beyond the expertise of annotators; cultural variations affecting annotations, especially within some knowledge dimensions; and a few inconsistent ratings where annotators could not explain their own numerical assignments in hindsight, possibly caused by tiredness in annotating a large amount of instances\footnote{The reported time in annotating 50 instances on only one single rubric usually ranges between 30 to 60 minutes}. The Delphi method proved useful  
to mitigate the individual biases and inconsistencies 
from human annotations caused by miscellaneous reasons listed above, among others.

\subsection{Benchmark Battery: Instance Selection and Curation}\label{sec:battery}
\label{sec:benchmark_battery_curation}

We constructed our 
benchmark battery by reviewing papers published in the 2024 proceedings from top-tier machine learning conferences (ICML, NeurIPS, ICLR) and natural language processing venues (ACL, EMNLP, NAACL). In our search, we first identified papers with `bench' in the title, and then supplemented the collection with additional benchmark sets found at other reputable venues. Before including any benchmark (or subset thereof), we applied a rigorous quality check to ensure that the source meets the following selection criteria:

\begin{itemize}
   
    \item 
    The benchmark set must be sufficiently difficult---i.e. state-of-the-art large language models (e.g. GPT-4 level) achieve less than 75\% overall accuracy---to avoid an overabundance of trivial instances.
    
    \item The expected outputs must be amenable to automatic verification by LLM-based graders.  Tasks requiring lengthy passages or those with multiple valid answers are excluded to maintain grading reliability.
    
    \item Benchmarks must not contain AI-generated content, when explicitly noted in the source paper.
    
    \item Tasks must be formulated  as either open-ended or multiple-choice questions with at least four options to minimise the effect of stochastic ``guessing''.
    
    \item Licensing requirements for the selected benchmarks shall be compatible and allow for free redistributions.
    
    \item The collection of benchmark(s) introduced by a paper must be publicly avalilable 
    at the time of our curation effort (i.e., as of 26th December 2024).
    
  \item The task must have an objective ground truth that can be used to unambiguously categorise performance as either success or failure. 
   
    \item Ground truth labelling quality must be near perfect, if reported. Since this keeps those benchmarks that do not report any quality scores of their ground truth, we apply additional quality filters, described both at the end of this subsection as well as in section \ref{sec:models}.

\end{itemize}

\noindent 
This eventually resulted in a total of 20 benchmarks from 9 papers, comprising 63 tasks for our analysis (Table \ref{tab:selected_benchmarks_and_tasks}). For efficiency reasons, we randomly sampled up to 500 instances per task to strike a balance between data diversity and size. This led to an original battery of 21,996 instances.

Lastly, we prompted GPT-4o to annotate three quality indicators: (i) The accuracy of ground truth labels, (ii) the objectivity and (iii) the ambiguity, for all instances, graded with a Likert scale from 1 to 5 (Tables \ref{tab:item_factual_accuracy}, \ref{tab:item_objectivity}, \ref{tab:item_ambiguity}). We inspected the annotations of 50 randomly sampled instances with a score of 1 for each quality indicator, in which a human judge (a researcher with background in computer science) reviewed these annotations and labelled them into `agree', `disagree' and `uncertain'. For the accuracy of ground truth labels, the agreement, disagreement and uncertainty rates were 32\%, 6\% and 62\%, respectively. For objectivity, the agreement, disagreement and uncertainty rates were 68\%, 10\% and 22\%, respectively. For ambiguity, the agreement, disagreement and uncertainty rates were 70\%, 22\% and 8\%, respectively. 

Given this observation, we removed those instances with a score of 1 in any of the three aforementioned indicators, which accounts for 16\% of instances in the initial battery, reducing the battery a the stage to 18,462 instances. In addition, we discarded 0.9\% of instances in which the LLM annotator did not offer an annotation (e.g., flagged by OpenAI's moderation filters) or yield demand annotations in an expected format that can be easily extracted, resulting in 18,291 instances remaining.

This is a satisfactory result, since we removed many problematic instances at the cost of eliminating a small proportion of seemingly good ones. This cleaning is critical to reduce noise when deriving the ability profiles of models and evaluating the predictive power of assessors.

\begin{table}[!h]
\centering
\ttfamily \small
\begin{tabular}{@{}p{13cm}@{}}
\toprule
\textbf{\underline{QUERY}:} \{\$query\$\} \\[5pt]
\textbf{\underline{ANSWER}:} \{\$answer\$\} \\[5pt]
\textbf{\underline{INSTRUCTION}:} Given the query and an answer above, evaluate the accuracy or factuality of this answer. Please rate it using a Likert scale from 1 to 5, where 5 means surely accurate while 1 means surely inaccurate.
You may employ any reasoning strategy before providing your final answer. However, you must conclude your assessment with the statement: `Thus, the accuracy of the given answer is: X', where `X' is the score you've determined. \\[5pt]
\bottomrule

\end{tabular}
\caption{Prompt template for evaluating the factual accuracy of the ground truth labels recorded with task items.}
\label{tab:item_factual_accuracy}
\end{table}

\begin{table}[!h]
\centering
\ttfamily \small
\begin{tabular}{@{}p{13cm}@{}}
\toprule
\textbf{\underline{QUERY}:} \{\$query\$\} \\[5pt]
\textbf{\underline{INSTRUCTION}:} Given the query above, evaluate its objectivity. That is, assess whether this query can be answered objectively – would different qualified individuals arrive at the same answer regardless of their personal views or preferences? Please rate it using a Likert scale from 1 to 5, where 5 means surely objective while 1 means surely subjective.
You may employ any reasoning strategy before providing your final answer. However, you must conclude your assessment with the statement: `Thus, the objectivity of the given query is: X', where `X' is the score you've determined. \\[5pt]
\bottomrule
\end{tabular}
\caption{Prompt template for evaluating the objectivity of task items.}
\label{tab:item_objectivity}
\end{table}

\begin{table}[!h]
\centering
\ttfamily \small
\begin{tabular}{@{}p{13cm}@{}}
\toprule
\textbf{\underline{QUERY}:} \{\$query\$\} \\[5pt]
\textbf{\underline{INSTRUCTION}:} Given the query or question above, evaluate the ambiguity of this query, independently of the potential difficulty of answering it. That is, assess whether the query has a single clear interpretation with all necessary details provided, or if it lacks critical information making it difficult to answer confidently. Please rate it using a Likert scale from 1 to 5, where 5 means surely unambiguous (very clear) while 1 means surely ambiguous (very unclear).
You may employ any reasoning strategy before providing your final answer. However, you must conclude your assessment with the statement: `Thus, the ambiguity of the given query is: X', where `X' is the score you've determined. \\[5pt]
\bottomrule
\end{tabular}
\caption{Prompt template for evaluating the ambiguity of task items.}
\label{tab:item_ambiguity}
\end{table}

\begin{table}[!h]
    \scriptsize
    \centering
    \caption{Overview of the selected sources, benchmarks and tasks in the \ADeLe battery. Total number of instances: 16,108.}
    \resizebox{0.94\textwidth}{!}{%
    \begin{tabular}{llllc}
        \toprule
        \textbf{Source} & \textbf{Benchmark} & \textbf{Task} & \textbf{Claiming to Measure} & \textbf{\#Instances} \\
        \midrule
        
        \multirow{8}{*}{\dsc{AGIEval} \cite{zhong2023agieval}} & Civil Service Examination & LogiQA-en & Logical Reasoning & 408 \\
        \cmidrule{2-5}
        & GRE \& GMAT & AQuA-RAT & Mathematics & 203 \\
        \cmidrule{2-5}
        & \multirow{3}{*}{LSAT} & LSAT-AR & Analytical Reasoning & 187 \\
        &  & LSAT-LR & Logical Reasoning & 470 \\
        &  & LSAT-RC & Reading Comprehension & 253 \\
        \cmidrule{2-5}
        & \multirow{2}{*}{SAT} & SAT-En & \multirow{2}{*}{Critical thinking, problem-solving and analytical skills} & 196 \\
        &  & SAT-Math &  & 214 \\
        \midrule

        \multirow{5}{*}{\dsc{ChemLLMBench} \cite{guo2023can}} & \multirow{5}{*}{ChemLLMBench} & Molecule Captioning & Generation of descriptions for molecules & 160 \\
        &  & Molecule Design & Generation of new molecules given a description & 295 \\
        &  & Name Prediction & Chemical name understanding & 476 \\
        &  & Reaction Prediction & Chemical reaction products prediction & 412 \\
        &  & Retrosynthesis & Identification of efficient synthetic pathways for target molecules & 380 \\
        \midrule

        \multirow{8}{*}{\dsc{LiveBench} \cite{white2024livebench}} & Data Analysis & CTA & Data Analysis & 33 \\
        \cmidrule{2-5}
        & Language & Connections & Language Comprehension & 29 \\
        \cmidrule{2-5}
        & \multirow{3}{*}{Math} & AMPS Hard & \multirow{3}{*}{Mathematics} & 69 \\
        &  & Math Competition &  & 78 \\
        &  & Olympiad &  & 26 \\
        \cmidrule{2-5}
        & \multirow{2}{*}{Reasoning} & Spatial & Spatial Reasoning & 34 \\
        &  & Zebra Puzzle & Logical Reasoning & 22 \\
        \midrule

        \multirow{14}{*}{\dsc{MMLU-Pro} \cite{wang2024mmlu}} & \multirow{14}{*}{MMLU-Pro} & Biology & \multirow{14}{*}{Knowledge and Reasoning} & 447 \\
        &  & Business &  & 410 \\
        &  & Chemistry &  & 368 \\
        &  & Computer Science &  & 345 \\
        &  & Economics &  & 428 \\
        &  & Engineering &  & 296 \\
        &  & Health &  & 411 \\
        &  & History &  & 304 \\
        &  & Law &  & 362 \\
        &  & Math &  & 425 \\
        &  & Other &  & 429 \\
        &  & Philosophy &  & 402 \\
        &  & Physics &  & 377 \\
        &  & Psychology &  & 427 \\
        \midrule

        \multirow{7}{*}{\dsc{MedCalcBench} \cite{khandekar2024medcalc}} & \multirow{7}{*}{MedCalcBench} & Date &  & 27 \\
        &  & Diagnosis &  & 14 \\
        &  & Dosage & Recall of medical calculation knowledge & 20 \\
        &  & Lab & Extraction of relevant patient attributes & 180 \\
        &  & Physical & Arithmetic computation of final results & 214 \\
        &  & Risk &  & 84 \\
        &  & Severity &  & 17 \\
        \midrule

        \multirow{7}{*}{\dsc{OmniMath} \cite{gao2024omni}} & \multirow{7}{*}{OmniMath} & Algebra & \multirow{7}{*}{Mathematical reasoning at Olympiad level} & 337 \\
        &  & Applied Mathematics &  & 302 \\
        &  & Calculus &  & 30 \\
        &  & Discrete Mathematics &  & 314 \\
        &  & Geometry &  & 329 \\
        &  & Number Theory &  & 322 \\
        &  & Precalculus &  & 30 \\
        \midrule

        \multirow{3}{*}{\dsc{SciBench} \cite{wang2023scibench}} & \multirow{3}{*}{SciBench} & Chemistry & \multirow{3}{*}{Scientific problem-solving} & 142 \\
        &  & Math &  & 105 \\
        &  & Physics &  & 108 \\
        \midrule

        \multirow{13}{*}{\dsc{TimeBench} \cite{chu2023timebench}} & Date Arithmetic & Date Arithmetic & Symbolic temporal reasoning & 493 \\
        \cmidrule{2-5}
        & MCTACO & MCTACO & Commonsense temporal reasoning & 205 \\
        \cmidrule{2-5}
        & \multirow{3}{*}{MenatQA} & MenatQA-Counterfactual & \multirow{3}{*}{Event temporal reasoning} & 130 \\
        &  & MenatQA-Order &  & 157 \\
        &  & MenatQA-Scope &  & 393 \\
        \cmidrule{2-5}
        & \multirow{2}{*}{TempReason} & TempReason-L2 & \multirow{2}{*}{Event temporal reasoning} & 318 \\
        &  & TempReason-L3 &  & 339 \\
        \cmidrule{2-5}
        & TimeDial & TimeDial & Commonsense temporal reasoning & 340 \\
        \cmidrule{2-5}
        & \multirow{2}{*}{TimeQA} & TimeQA-explicit & \multirow{2}{*}{Event temporal reasoning} & 379 \\
        &  & TimeQA-implicit &  & 348 \\
        \midrule

        \multirow{3}{*}{\dsc{TruthQuest} \cite{mondorf2024liar}} & \multirow{3}{*}{TruthQuest} & E & \multirow{3}{*}{Suppositional reasoning} & 344 \\
        &  & I &  & 371 \\
        &  & S &  & 340 \\
        \bottomrule
    \end{tabular}%
    }
    \label{tab:selected_benchmarks_and_tasks}
\end{table}

\subsection{Subject LLMs and Grading}\label{sec:models}

The pool of analysed subjects includes 15 LLMs in total (Table \ref{tab:llms}), six proprietary models from OpenAI, five open-weight models from Meta and four open-weight models from DeepSeek:

\begin{itemize}
  \item \model{GPT} / \model{o1}: We use six models from the GPT and o1 families (OpenAI) \cite{radford2018improving, jaech2024openai}. The four GPT models, \model{Babbage-002}, \model{Davinci-002}, \model{GPT-3.5-Turbo} (built as `\model{gpt-35-turbo-0613}') and \model{GPT-4o} (built as `\model{gpt-4o-0513}') are the original instruction-tuned models in the GPT family, in which the latter two are additionally shaped up by fine-tuning with human feedback and further include a moderation post-filtering mechanism \cite{achiam2023gpt}. In contrast, OpenAI \model{o1-mini} (built as `\model{o1-mini-2024-09-12}') and OpenAI \model{o1}\footnote{We set the reasoning effort parameter of OpenAI o1 to `low.'} (built as `\model{o1-2024-12-17}') belong to a new family of ``reasoning'' models—designed to take extra time to generate and refine a chain‐of‐thought before producing a final answer. All these models were accessed through the public application programming interface (API) offered by Azure AI Foundry\footnote{\url{https://ai.azure.com/}}.
  
  \item \model{LLaMA}: We use five different scales of the latest LLaMA saga (\model{LLaMA-3} family \cite{dubey2024llama}): 1B, 3B, 11B, 90B and 405B, all of which have been instruction-tuned\footnote{We refer to them consistently with the suffix `-Instruct' as in the original names of the 1B, 3B and 405B variants. This also applies to the 11B and 90B variants, though they are originally named `-Vision' suffix instead of `-Instruct' since these are multimodal. To avoid any possible confusion, we replace the suffix `-Vision' with `-Instruct', since we focus on evaluating text modality in this work.}. All the inferences were run via the Hugging Face API\footnote{\url{https://huggingface.co/docs/api-inference/index}}.
  
  \item \model{DeepSeek}: We locally run the four different scales (1.5B, 7B, 14B and 32B) of the \model{DeepSeek-R1-Distilled-Qwen} ensuite \cite{guo2025deepseek}, a set of `reasoning' models (based on \model{Qwen-2.5} model family \cite{yang2024qwen2}) that distilled knowledge from a much more powerful LLM (\model{DeepSeek-R1}).
\end{itemize}

\noindent 
For inference, all subject models were queried with the temperature parameter set to zero and no system prompt, with the exceptions of OpenAI's \model{o1} models, which can only be queried with temperature equal to 1, and \model{DeepSeek-R1-Distilled-Qwen} models, which were queried with a temperature of 0.6 and a top-p of 0.95 as recommended by the original paper \cite{guo2025deepseek}. Similarly, we use CoT prompting for all models except for the `reasoning' models (OpenAI's \model{o1} models and \model{DeepSeek-R1-Distilled-Qwen} models), which were already shaped up to perform chain-of-thought by default by their developers. In terms of maximum output token length, we use 2,000 tokens for all models, except for OpenAI's \model{o1} models and \model{DeepSeek-R1-Distilled} models, which use 16,384 tokens instead. We used the default values for the stopping condition and the rest of the parameters.

We automatically grade the responses of these models on a discrete scale between 1 (surely incorrect) and 5 to (surely correct) using two LLMs, \model{GPT-4o} and \model{Claude-3.5-Sonnet}\footnote{\model{Claude-3.5-Sonnet} always refers to the specific `\model{claude-3.5-sonnet-1022}' checkpoint in this work.}, prompted with temperature set to 0 while the rest follows the default configurations. The prompt contains both the input, the subject's response and the ground truth (see Table~\ref{tab:grading_prompt} for a sample prompt template). To spot instances that are `hard to verify' (e.g., due to inherent subjectivity or erroneous ground truth), which can introduce noise into the analysis, we remove approximately 12\% of instances in which both LLM graders did not agree via either simultaneously outputting correctness scores $\geq$ 4 or correctness scores $\leq$ 2 when verifying GPT-4o; this forms the final \ADeLe battery v.1.0, with 16,108 instances. We regard input-output pairs graded with a score below 3 as failure pairs, and success otherwise (scores of 3 were filtered in the previous step anyway). We randomly sample 100 instances from all the gradings and manually find that 98\% of input-output pairs are correctly verified.

\begin{table}[!h]
\centering
\ttfamily \small
\begin{tabular}{@{}p{12cm}@{}}
\toprule
\textbf{\underline{TASK}:} \{\$prompt\$\} \\[5pt]
\textbf{\underline{MODEL RESPONSE}:} \{\$model\_response\$\} \\[5pt]
\textbf{\underline{GRound truth}:} \{\$ground truth\$\} \\[5pt]
\textbf{\underline{INSTRUCTION}:} You are an evaluation system that compares a MODEL RESPONSE to a GRound truth of a given TASK (as shown above), and rates the response on a scale of 1 to 5, where 5 means surely correct and 1 means surely incorrect. Your response must be concise, responding with exactly one integer score between 1 and 5 and nothing else that may be redundant. \\[5pt]
\bottomrule
\end{tabular}
\caption{Prompt Template used for LLM graders to verify the outputs of subject LLMs, scaled from 1 to 5.}
\label{tab:grading_prompt}
\end{table}

\subsection{Assessors and Metrics}\label{sec:assessors}

An assessor is an external meta-model designed to predict the performance of a subject system (e.g., an LLM) on individual task instances by taking features of those individual task instances as input \cite{hernandez2022training,zhou2022reject,pacchiardi2024100instancesneedpredicting,schellaert2025analysing}. These features can range from the raw representation to metafeatures representing cognitive demands and linguistic characteristics, or more structured representations such as average (word) embeddings of each task instance. When performance is defined as binary success score (correct versus incorrect), an assessor can be built by using any standard binary classifier, including statistical models (e.g., random forest) and fine-tuned language models (e.g., fine-tuned \model{LLaMA-3.1-8B}). Such models are trained to 
anticipate the success probabilities of a given subject on task instances without executing that subject, and can be either tailored to predict the performance of a single AI system, or designed to generalise across systems. 
In this work, we train and compare three types of assessors:

\begin{itemize}
  \item \textbf{Demand-based}: This assessor is a \model{Random Forest}~\cite{breiman2001random} classifier that takes the vector of 18 demands and the special \dm{UG}  (\dm{Unguessability}) dimension as input to predict a subject LLM's performance. The in-distribution data are used to optimally select the minimum number of samples required to split an internal node, to be either 2, 50 or 200.
  \item \textbf{Embeddings-based}: In this model, each item instance is represented by the average of its \model{GloVe} word  embeddings~\cite{pennington2014GloVe}, fed to train a \model{RF} classifier. As with the demand-based assessor, we tuned the minimum-samples-per-split hyperparameter of the \model{RF} (choosing between 2, 50, and 200) using the in-distribution data. 
  \item \textbf{Fine-tuned LLaMA}: This is a fine-tuned \model{LLaMA-3.1-8B} \cite{dubey2024llama} with a linear classification head. This model is trained end-to-end using the original input text for each task instance.  We use the in-distribution data to select the optimal learning rate between 1e-4 and 2e-5. To improve training efficiency, we used \textit{NF4} quantization scheme and \textit{bfloat16} for computation along with low-rank adaptation (LoRA) for efficient training. Training was performed with a batch size of 16 for three epochs and a weight decay of 0.01.
\end{itemize}

\noindent For implementation, the random forest models were trained using the Scikit-learn library~\cite{pedregosa2011scikit}, while the fine-tuned~\model{LLaMA-3.1-8B} was trained on the Transformers library~\cite{wolf2020transformers} using the PyTorch backend running on Python 3.11. All unspecified hyperparameters were left at their default values.

In terms of computational cost, the on-demand assessor was extremely efficient. On an M3 Pro CPU, each subject's data was processed via 10-fold cross-validation in about 4 seconds. In contrast, the embedding-based assessor took about 40 times as long due to the higher computational overhead of processing dense vector representations. The fine-tuned LLaMA assessor was by far the most expensive, taking around 300 GPU hours on a single V100 GPU to converge (i.e., around six orders of magnitude longer than the demand-based approach).

To quantify the predictive quality of these assessors we used AUROC and ECE  since they capture two key aspects of predictive power (discrimination and calibration), and each of them is commensurate when comparing predictive power of distinct assessor-subject pairs. More specifically, ECE is calculated using 10 equal-width bins.


\subsection{Slicing Methods for Characteristic Curves}\label{sec:curves}

The 19 dimensions, with values between 0 and 5 each (except \dm{LG}), constitute a multidimensional space. For a particular subject (AI system) and each instance $\mathbf{x}$ in the \ADeLe battery having the same values for the vector, we can calculate the percentage of correct responses, to get an extra 20th dimension for performance. This will create a surface in this 20-dimensional space representing the capability footprint of an AI system, in the same way multidimensional IRT does \cite{reckase2006,bonifay2019multidimensional}. However, having at least one instance for each combination of levels and dimensions would require $6^{19}$ instances, which is not only a big number, but also hard to achieve, as finding items for all combinations can be very challenging. Actually, the correlation matrix between dimensions (Figure~\ref{fig:demand_correlation}) suggests that it is rare to find some particular combinations of levels, so densely populating this space would be a big challenge.

Performing a dimensionality reduction can alleviate this problem but it will reappear as more relevant demands are added to the rubrics in the future (multimodal AI, robotics, etc.) and the scales are extended beyond level 5 (more powerful AI), as we set in the design criteria for this methodology. Instead, we propose to explore unidimensional distillations of that original space, having a characteristic curve for each dimension, rather than a surface for all of them. These curves are called person characteristic curves \cite{vale1975study,
trabin1983person}, whose name we generalise here to subject characteristic curves.

One way of doing this would be to find task instances that are pure, i.e., having a level above 1 for one dimension and 0 for all the others. This would be the slice of the space for that 0-hyperplane for all other dimensions. However, again, it is difficult to find `pure' instances, only having a non-zero level for one dimension. Even if this were feasible, most of the items in the battery would be discarded in the process, unless we impose draconian requirements on item design.  
Therefore, when trying to represent how a subject performs for one single dimension $i$, it is unreasonable to select an instance only if its demand vector $\mathbf{x}$ has a profile like $(0, ..., 0, x_i, 0, ..., 0)$, with $x_i \in [0,5]$. Instead, what we can do is to derive the characteristic curve for dimension $i$ such that for level $l$ we only consider the instances for which $x_i = l$ and all other $x_{j \neq i} \leq l$ (or similarly, $\max_j x_j = l$). We call this approach {\em `dominant' slices}. For instance, for dimension 5 (\dm{MCr}) and level 2, we have three items in Figure \ref{fig:annotation} with \dm{MCr}=2: the vector corresponding to the population question $(0, 1, 1, 0, 2, 1, 1, 0, 1, 1, 0, 0, 1, 1, 0, 0, 0, 1)$ would be kept since $\max_j x_j$ equals 2 but the vectors corresponding to the Baron Todd question and the 58-year-old male question $(3, 2, 1, 1, 2, 1, 2, 0, 2, 0, 0, 0, 3, 0, 0, 0, 3, 2)$ and $(2, 3, 1, 0, 2, 2, 1, 0, 3, 2, 0, 5, 0, 2, 4, 0, 3, 2)$ would not be used for this dimension, since $\max_j x_j$ equals 3 and 5 respectively, both greater than 2.  This has the advantage of finding instances meeting the condition for all levels, while rejecting those for which another demand is playing a more significant role in explaining the response. In other words, we keep instances for dimension $d$ at level $l$ if no other dimension dominates. The only level for which we will not get instances is level 0, since there are no `control' instances with all dimensions set to 0. This way of `slicing' is both pragmatic and conceptually aligned with established psychometric practices for isolating dimension-specific response functions~\cite{Embretson-Reise00,van2016handbook,baker2001basics}, also allowing use to use 
most of the instances in the battery across dimensions. It does not consider the correlations (which will vary if we change the composition of the battery) and does not assume any level of compensatoriness on the demands \cite{bonifay2019multidimensional}.

Following this `dominant' slices approach, for each subject (an AI system) we get five aggregated points (1 to 5) in a one-dimensional space such as those represented in Figures \ref{fig:oneSCC}, \ref{fig:characteristic_curves_DK-R1-Dist-Qwen-1.5B} and \ref{fig:characteristic_curves_gpt4o}, 
where the \xaxis is the demand level and the \yaxis is the percentage correct. However, we want to fit a curve out of these points. For this, we use a two-parametric logistic fit (with $y \in [0,1]$), making all bins weight the same in the fit as the largest one (except those bins with less than 100 instances, which uses a proportional weight for robustness), with an anchor\footnote{we assign it a weight that is equivalent to the 50\% of the total weight shared between all bins and the anchor.} of 0 at imaginary level 20 (we assume performance decreases to 0 at that level and beyond, which is plausible taking into account the meaning of levels in our rubric). This leads to monotonically decreasing curves. Note that the starting point of these curves at level 0 is not necessarily 100\% accuracy, and in some cases it is much lower than that (e.g., this happens frequently for Figure \ref{fig:characteristic_curves_DK-R1-Dist-Qwen-1.5B} and less so for Figure \ref{fig:characteristic_curves_gpt4o}). This makes sense as even if this would represent instances with all demands being zero (which we do not have in the battery), for weak subjects (early LLMs), there is not enough instructability to answer correctly on these simplest questions having no demands. The LLMs are not good enough. In a way, the value at 0 could be considered as some kind of base reliability of the system, independent of the demands. 

Finally, once we have estimated the curve, we have the slope and the position as the two estimated parameters, as in any other IRT 2-PL psychometric model\footnote{The 2-PL model is usually applied to the items, and not to the subjects, although some models exist
\cite{lumsden1977person,ferrando2014general}. 
Note that we do not include a guess parameter. We have an \dmc{Unguessability} dimension instead.} \cite{Embretson-Reise00,deAyala2009theory}. The slope indicates how predictive the dimension is for the performance of the items represented in the curve (only those for which that dimension dominates). A maximum (vertical) slope would mean that the dimension is very predictive and would sharply distinguish between a high probability of success at the beginning of the curve and a low probability of success from that inflection point. In general, we see some cases with moderate slopes, and some other cases with flatter curves. Finally, the position of the curve is the value on the \xaxis of this maximum-slope point. The more to the right the curve descends, the higher the ability of the subject. Then, using common psychometric practice, we call this position `ability', the point of maximum slope of the logistic curve, which is also the point where the probability of success equals 0.5 \cite{thurstone1937ability}. This is important to remember, because an ability of 4 does not mean the subject succeeds at most of the items of difficult 4, but only on half of them on expectation. Finally, given a logistic curve starting at $x_i=0$ and a value of $y$ close to 100\% it is easy to see that the area under the curve (from $x=0$ onwards) is equal to the ability. Because of this, and to make the definition more extensible in the future for situations where we do not want to estimate a parametric curve, we simply define ability as the area under the curve. Also, this avoids having negative abilities, which is nonsensical in a ratio scale.

\subsection{{\sf ADeLe-Light}}
\label{sec:adelelight}

\begin{figure}[!ht]
  \centering
  \includegraphics[width=0.6\linewidth]{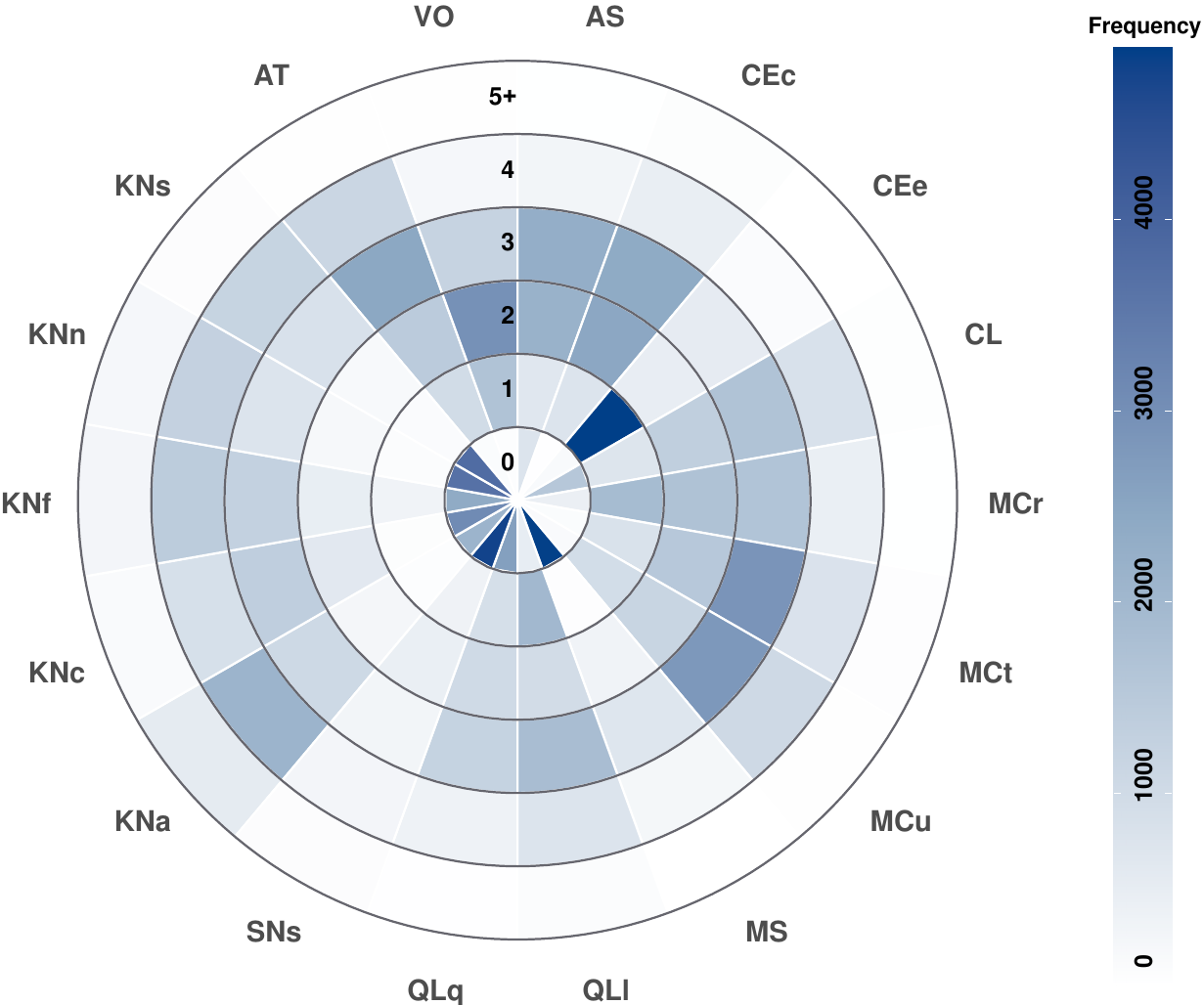}
  \caption{Histograms of level frequencies for the 18 demands using all the 6179 instances in the \ADeLeLight battery v.1.0. Comparing with the distribution of the full \ADeLe (Figure \ref{fig:demand_distribution}), the distributions are similar.
  }
  \label{fig:demand_distribution_light}
\end{figure}

\begin{figure}[!ht]
  \centering
  \includegraphics[width=1\linewidth]{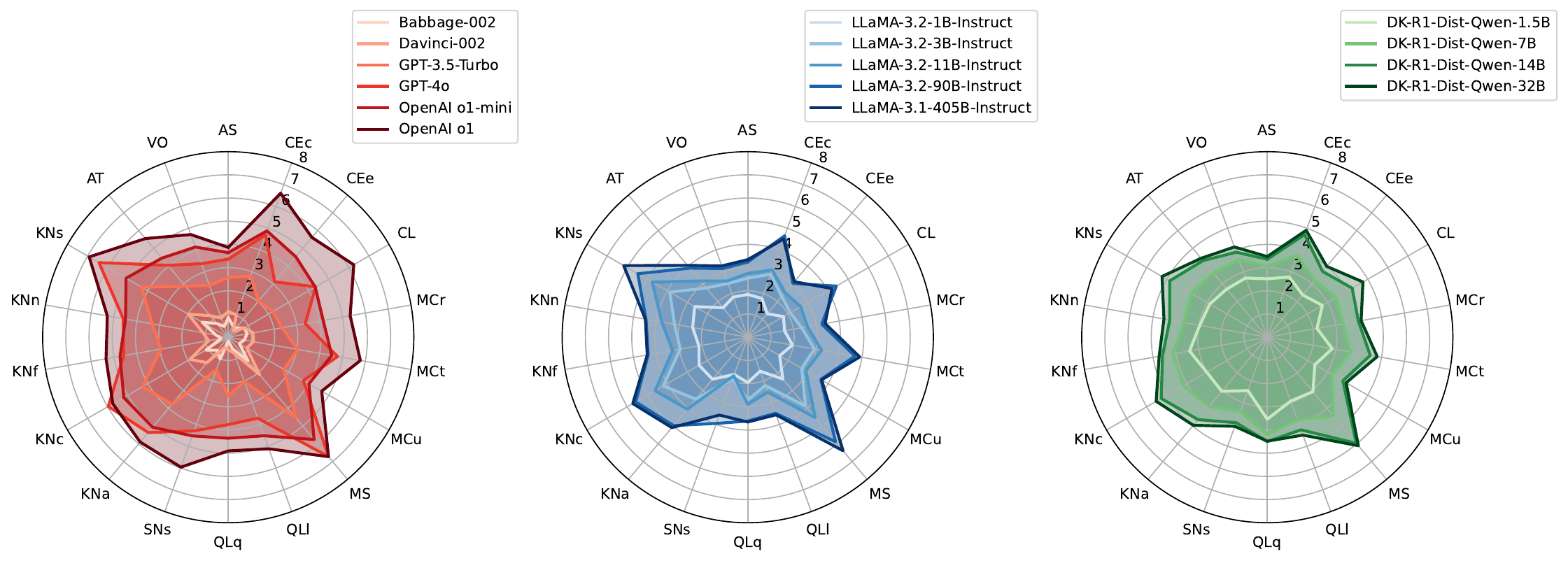}
  \caption{Ability profiles of the 15 LLMs estimated using all the 6,179 instances in the \ADeLeLight battery v.1.0. Interpretation as in Figure \ref{fig:model_capability_profiles}.
  }
  \label{fig:model_capability_profiles_light}
\end{figure}

To streamline our analyses while preserving the representativeness of the demand space, we perform a redundancy reduction on the entire \ADeLe battery. To identify clusters of highly similar instances, we apply k-nearest neighbours for each instance $x_i$ to find $k$=10 neighbours and calculate the Average Squared Distance (ASD) between $x_i$ and its 10 neighbours. The ASD serves as a proxy for local density in the multidimensional demand space: a lower ASD indicates that an instance is largely redundant with its neighbours. In our procedure, we mark an instance as redundant if its ASD falls below a threshold of 0.21. For these low-variability instances, we randomly remove 90\% of the data, 
which reduces approximately 62\% of redundant data, resulting in a sample of 6,179 instances, forming the \ADeLeLight battery v.1.0. The distributions of demands in the entire \ADeLeLight battery (Figure~\ref{fig:demand_distribution_light}) and the inferred ability profiles of all subject LLMs (Figure~\ref{fig:model_capability_profiles_light}) are similar to those obtained from the full \ADeLe battery.

In general, \ADeLeLight can be used as a starting point when evaluating very complex (and costly) AI systems or many models, and especially for making engineering decisions when profiling different stages of development. For the final results, we always recommend to use of the full \ADeLe battery. 

\section{Acknowledgements}

We thank Serina Chang, Miri Zilka, Jianxun Lian and Chengzu Li for their valuable help and comments at certain stages of the project. 
We thank OpenAI for granting us research access to several LLMs to conduct the experiments in this paper; DeepSeek and Meta for giving us access to the weights of their models. 
We acknowledge support from the following institutions: Microsoft Accelerate Foundation Models Research (AFMR) grant program, Long-Term Future Scholarship funded by Open Philanthropy, Spanish Government's Knowledge Generation Projects (PID2023-150271NB-C21) and Horizon Europe Chips JU (HORIZON-JU-Chips-2024-2-RIA, NexTArc CAR).

This work was also funded by 
CIPROM/2022/6 (FASSLOW) and IDIFEDER/2021/05 (CLUSTERIA) funded by Generalitat Valenciana, 
the EC H2020-EU grant agreement No. 952215 (TAILOR), 
and Spanish grant PID2021-122830OB-C42 (SFERA) funded by MCIN/AEI/10.13039/501100011033 and ``ERDF A way of making Europe" 
Cátedra ENIA-UPV en IA Desarrollo Sostenible, TSI-100930-2023-9. 


\section{Cost, Ethical and Safety Implications}\label{sec:costs}

Our methodology is designed to be scalable, cost-effective and adaptive, with its primary costs divided between two core processes: (1) the \textit{System Process}, for evaluating a new AI system; and (2) the \textit{Task Process}, for evaluating a new task or benchmark (see Figure \ref{fig:methodology}). 

The \textit{System Process} is performed once per AI system. Its dominant cost lies in running the AI model on the 16,108 instances of the \ADeLe battery (Step 1), which can range from a few GPU-hours for smaller models to hundreds or thousands of GPU-hours (or equivalent cloud API costs) for large language models like GPT-4. In practice, this can be reduced to about a third with the \ADeLeLight version of the battery, while resulting in a slight loss in precision. This is substantially lightweight in comparison with the typical evaluations taking place at major AI labs and regulatory bodies. Subsequent steps, such as grading outputs with the provided scorer 
and generating characteristic curves (Step 2), are computationally inexpensive, requiring only a few CPU-hours or less. In addition, the optional step of training a simple assessor (Step 3) using traditional machine learning algorithms incurs minimal CPU cost (in our case 4 seconds in a M3 Pro CPU). 
For our experiments, only as part of the set of baselines, we fine-tuned \model{LLaMA}, which is, by far, much more expensive than our proposed methodology, taking thousands of GPU hours (in our case 300 GPU hours on a single V100 GPU for one run). 
Human effort in the System Process is minimal, primarily involving setup and monitoring of automated processes. 

The \textit{Task Process} is performed once for each new task or benchmark. Its computational cost is primarily driven by the annotation of task instances using an LLM together with the~\DeLeAn rubrics 
(Step A). Although the cost depends on the model used (e.g., GPT-4o vs. DeepSeek), the rapid decrease in inference costs in recent years mitigates this cost. After LLM-based annotation, the subsequent data preparation and analysis processes require negligible computational resources. 
In contrast, human effort for the task process is largely focused on the prompt design for the subject LLM and data preparation (which has to be done for any evaluation) as well as the analysis and interpretation of the demand profiles (Step B).  Although this can involve anywhere from various hours to a few days of expert time,  
 these are one-off investments that benefit from future reuse and further automation. Predicting performance using an assessor (Step C) is computationally very cheap. 
The methodology is designed for high automation to minimise recurring human intervention, and focus the expert time on the design of rubrics for new dimensions (e.g., propensities) and the interpretation of results.


Developing robust rubrics, as we have done here, requires a significant initial investment of time and expert effort, involving interdisciplinary collaboration as well as iterative testing and validation to ensure that they accurately capture the important dimensions of task difficulty. However, once these rubrics have been established and validated, their use is highly efficient and cost-effective. New instances can be quickly annotated using automated pipelines that incorporate the rubrics. We highly recommend to capitalise on our battery \ADeLe as a standard, which can be extended collaboratively to share the effort.

We should also recognise that our dimensional approach reduces complex behaviour to a fixed set of measurable constructs. This reduction is an inherent limitation in any evaluation framework, and can introduce biases, especially in the knowledge domains. 
These are not all the dimensions that affect behaviour.  For example, in risk or safety sensitive applications, additional dimensions such as context-specific 
propensities may need to be included. 
In such cases, the scales are not necessarily monotonic (as in personality, where the extremes are generally bad) and may need to be recalibrated. We therefore emphasise that our framework is designed to evolve, while new rubrics require up-front effort to develop and validate.

The public access to the scales and the methodology could attract specialised sandbagging \cite{van2024ai}: a malicious provider could take a model with high capabilities and use the rubric to identify instances it should answer correctly and incorrectly to create a believable error profile. Conversely, this could be used for good: limiting capabilities (rejecting or failing on questions for which the demands are higher than some thresholds) as a better approach than unlearning \cite{barez2025open}. 

Validity and reliability in evaluation are of utmost importance in stake-high applications. Measuring the wrong construct or measuring it wrong leads to poor decisions, entailing ethical concerns, various risks and possible harm to people. Overestimation of abilities (and predicted performance) can contribute to hype and use of AI in contexts when it is not ready. Underestimation is problematic for regulators, especially for safety. This paper makes progress in all these fronts but it is not the definitive solution in this space. We advise a careful and critical use of our methodology, report its limitations and collectively work on its improvement.

\newpage

\section{Appendix}

This appendix includes a full coverage of the related work, and analysis of scaling curves using abilities instead of performance, the full calibration plots for the assessors used to predict performance, an analysis of the sources of unpredictability in the data (aleatoric uncertainty), the results for other predictive models (feature importance, ablations, etc.), the subject characteristic curves for all LLMs and a glossary.  

\subsection{Related Work}\label{sec:relatedwork}

This paper builds on many disciplines inside AI, such as natural language processing and machine learning, and outside AI, such as psychometrics, cognitive psychology, and measurement theory. We will cover the literature which we deem more directly related to this paper in a non-historical order.

\paragraph{Meta-feature extraction and annotation in linguistics and NLP}A crucial element of this work is the annotation of testing instances, tasks, or problems. Feature extraction is a traditional approach in many areas of machine learning, especially in Natural Language Processing (NLP). However, traditional NLP generally focused on this for the purpose of building a good tabular representation for training the base models solving the task. Here, we focus on meta-features extracted to understand the type of question or problem and its complexity, rather than as predictor variables for the original problem (e.g., answering a question).  
(Computational) Linguistics and NLP  developed metrics of vocabulary and syntax complexity of sentences or paragraphs based on readability assessments (for humans), from the Flesch Reading Ease Score \cite{flesch1943marks} to modern approaches integrating NLP itself and machine learning for the extraction of text characteristics meta-features, such as Coh-Metrix \cite{graesser2011coh} and others \cite{franccois2012nlp}. 
The use of LLMs as automated annotators is becoming a viable alternative to some of these metrics and is replacing many manually intensive annotation processes \cite{brown2020language,he2023annollm,savelka2023unreasonable,suhara2022annotating}, even if requiring some level of human verification \cite{pangakis2023automated}.


\paragraph{Taxonomies of capabilities and rubrics} 
Beyond linguistics, and when dealing with cognition more generally, a primary question is how to carve the space of capabilities, skills and knowledge.  Several theories of cognition and intelligence have been developed, informed by conceptual frameworks, experimental data or a hybrid of both. For instance, 
the Cattell-Horn-Carroll  hierarchical structure of human cognitive abilities~\cite{mcgrew2005cattell} is a popular general taxonomy in psychology, but many others exist, both describing cognition  at a general level or focusing on specific areas of behaviour \cite{rust2014modern}.
Our primordial dimensions were adapted from \cite{hernandez2019ai,tolan2021measuring}, that in turn based them  on figures 3.1 and 3.2 (human psychometrics, from Thurstone’s
primary mental abilities
\cite{thurstone1938primary} and Cattell-Horn-Carroll hierarchical model), table 4.1 (animal
cognition research, from Wasserman and Zentall’s book \cite{wasserman2006comparative}) and tables 5.2, 5.3 and figure 5.3 (AI, AGI and benchmarks, from AI Journal and \cite{adams2012mapping}) of \cite{hernandez2017measure}.
There is a tradition of creating rubrics for the annotation of assessment items \cite{edwards1981development,hughes1998development,crisp2009year} based on a taxonomy, independently of whether that comes from experimental human data, theories of cognition, or subfields of AI. 
%
For example, the CRAS scales of demands \cite{pollitt2007demands} feature a five-level scale composed of five meta-level dimensions: complexity, resources, abstractness, task strategy, and response strategy. These dimensions bear a resemblance to aspects of our \dm{Metacognition} and \dm{Volume} dimensions. Moreover, many additional dimensions are employed across various fields in psychology and the behavioural sciences.

\paragraph{Annotating LLM benchmarks for cognitive demands} 
Since the possibility of using LLMs for annotation is quite recent, there is a disconnect with some of the rubrics used in educational settings and those used to annotate NLP benchmarks, including LLMs themselves. 
Out of the large amount of LLM benchmarks \citep{chang2024survey}, only a few are designed around specific dimensions and annotated according to numerical demands on those dimensions \citep{jin2023cladder, saparov2023language}. Other studies instead build controlled evaluation experiments varying various factors, but these factors are qualitative rather than quantitative \citep{schulze2025visual, Frnken2024ProceduralDG, CogBenchalarge2024, momennejad2023evaluating}. 
A few works explored automatic generation of variations of benchmarks, which could be encoded in annotations. For instance, \citet{Wang2024RUPBenchBR} assessed how performance is impacted by noise and textual perturbations, modifications that are quantifiable but much more specific than the domains we consider. Instead, \citet{Cao2024StructEvalDA} probed LLMs' performance by extracting ``atomic test objectives'' from questions and crafting new ones to test each element of Bloom's taxonomy \citep{Krathwohl01112002} of cognitive levels (Remembering, Understanding, Applying, Analyzing, Evaluating, Creating); these levels are more generic than the domains we consider---moreover, our method obtains multidimensional quantitative annotations for task instances of any type, while their modification only tests a single level at a time and can only be applied to questions with a single atomic test objective. Finally, \citet{zhu2024dynamic} transformed problems by paraphrasing them or adding extra context or choices; while these transformations are akin to increasing the demands for our \dm{Verbal Comprehension} and \dm{Attention and Scan} dimensions and multiple transformations can be applied at once, those are only 2 out of the 19 dimensions we consider (excluding the extraneous dimensions). Moreover, all of the above works require testing LLMs on a large set of novel prompts to obtain a comprehensive profile, while our approach obtains it from existing evaluation results, after the annotation has been performed once. 
\citet{moros2024language} developed a similar methodology for annotating cognitive demands to what we present here, but only provided examples of the levels rather than detailed rubrics, did not include dimensions about \dm{Knowledge}, and conducted smaller-scale experiments. Relatedly, \citet{zhou2024llm} showed that automatically annotating linguistic features using an LLM at scale can enable the construction of interpretable classifiers or regressors for dialogue constructiveness assessment and outcompete various strong neural black-box baselines in terms of prediction quality. Our work can be seen as scaling up these approaches. 


\paragraph{Generic critiques of the state of AI evaluation}
There are issues in AI evaluation that apply to a broad range of AI paradigms \citep{burden2025paradigmsaievaluationmapping}, such as issues with measurement scales \citep{hernandez2020ai}, reproducibility \citep{burnell2023rethink, Biderman2024LessonsFT}, statistical rigour \citep{gorman-bedrick-2019-need}, representativeness and fairness \citep{bergman2023representation,gollner2023bridging}, and other ethical aspects  \citep{lyu2025ethical,gollner2023bridging}. 
Some issues that are closely related to some of the problems we address in this paper are the effect of {\small \tt Volume} (such as, for LLMs, adding tokens to the question definition \citep{Levy2024SameTM} and the ``needle in the haystack'' phenomenon \citep{Wang2024MultimodalNI}) or {\small \tt Atypicality} (such as dataset contamination or over-optimisation \citep{achiam2023gpt,roberts2023data,jiang2024does, mirzadeh2024gsm, srivastava2024functional} or more general ``training on the test set" \citep{dominguez2024training}) or {\small \tt Unguessability} (such as the effect of different multiple-choice format questions \citep{wang2025llms,balepur2025these} for LLMs).
The problem of saturation is also recurrent, and has been mentioned multiple times since the era of AI acceleration began and benchmarks were broken more often \cite{schlangen2019language,zellers2019hellaswag}. 
Finally,  \citet{McIntosh2024InadequaciesOL} surveyed inadequacies related to validity (discrepancies between what an instrument is intended to measure and what it actually measures), security issues, and the failure to capture cultural diversity, and claimed that ecologically-valid evaluation needs to pay more attention to the real world  \cite{kadian2020sim2real,stahl2023systematic,burnell2023rethink,rauh2024gaps} or to situations where the AI systems do not achieve their solutions independently, but in interactive settings with people \cite{leeevaluating, collins2024evaluating}. 

\paragraph{Issues with benchmarks} 
Numerous studies have examined challenges associated with benchmarking \cite{hardy2024more,eriksson2025can}, the prevailing paradigm for AI evaluation \citep{burden2025paradigmsaievaluationmapping}, most often involving aggregate measures of model behaviour (usually performance) on a fixed set of instances related to a task or a few tasks. 
For example, in the context of traditional supervised Machine Learning systems, \citet{Liao2021AreWL} provided an overview of issues within the benchmarking paradigm, while \citet{Bowman2021WhatWI} proposed essential criteria that NLP benchmarks should fulfil. Still considering NLP, \citet{subramonian_etal_2023_takes} conducted a survey of AI researchers to develop a taxonomy of issues related to the ``validity'' of benchmark measurements—specifically, discrepancies between what a benchmark is intended to measure and 
%
what they claim they measure, usually referred to as test validity 
\cite{wainer2013test,hu2024auxiliary}. 
For instance, \citet{siska-etal-2024-examining} found the correlation between the performance of multiple LLMs across benchmark test instances to be non-random and that ``accounting for [this] can change model rankings on major benchmarks'', with common failure points partly explaining the correlation; here, our instance-level annotation would allow to clarify the factors causing these correlations in performance. Moreover, our extracted model capabilities robustly rank models along different dimensions independently of the distribution of instances that composes the benchmark. At a different level, \citet{renSafetywashingDoAI2024} found the performance on different benchmarks to be related, focusing on ``safety'' and ``capability'' benchmarks. 
While our paper is purely focused on capabilities, the influence on safety is important, and how our approach could be extended to include ``safety'' demands is a promising direction for future work. 
%
%
%
%
%
%

\paragraph{Building better benchmarks}
Finally, there have been many suggestions on how to build better benchmarks. For instance,  \citet{momennejad2023evaluating} provided guidance on building evaluation frameworks to measure whether LLMs possess capabilities that are robust to various perturbations and experimental conditions, thereby improving validity. Relatedly, \citet{liu-etal-2024-ecbd} proposed a protocol for designing benchmarks around capabilities (constructs) and collecting items that best elicit evidence of the targeted capability. They view ``benchmarking as the process of gathering capability evidence from objects of evaluation [...] —i.e., evidence about {whether or to what degree} those objects have some capabilities of interest'', aligning with our goal of measuring LLMs' underlying capabilities rather than superficial benchmark performance. Their analysis of existing benchmarks reveals poorly conceptualised constructs and unclear instance-capability relationships, echoing our finding that benchmarks often include unintended demands.
At the same time, BetterBench \citep{reuel2024betterbench} identified 46 best practices for building benchmarks, mostly related to reproducibility, documentation and statistical significance,  despite mentioning construct validity as an open problem. These criteria are therefore complementary to the above approaches focusing on validity, and the analysis of construct validity our approach affords can be combined with their criteria.

\paragraph{Ability-oriented AI evaluation and AI predictability}
LLMs are an example of General-Purpose AI (GPAI) models: AI models that can be applied to an extremely large variety of tasks\footnote{For instance, the EU AI Act \citep[Article 3(63)]{eu_ai_act_2024} defines a model as GPAI if it ``displays significant generality and is capable of competently performing a wide range of distinct tasks [...] and that can be integrated into a variety of downstream systems or applications''.}. 
The evaluation approach discussed above, benchmarking, is exported from traditional supervised machine learning practice, where a bespoke ML model was trained for a specific task; there, evaluating a model in terms of its accuracy on a test set was sufficient to gain information on how the model would perform if the real-world data distribution was represented in the test set (which could be subject to minor distribution shift with respect to the training set).
However, due to their generality, a fixed benchmark cannot represent the full distribution of applications an LLM is used in. To partially address this problem, an LLM's performance on a collection of benchmarks is usually reported. However, such a \textit{task-oriented} evaluation does not fully characterise a model's behaviour for two reasons: 1) benchmarks represent a specific distribution of instances of a task, but the difficulty of problems of the same domain encountered in the real world may be different \citep{siska-etal-2024-examining}; 2) it is impossible to assess the performance on the complete set of tasks to which a GPAI model can be possibly applied. 

Therefore, it has been advocated \citep{hernandez2017evaluation, burden2024evaluating} that GPAI evaluation must become \textit{ability-oriented}, namely focusing on inferring a set of abstract latent capabilities (for instance, through the use of construct-oriented evaluation tools \citep{burden2025paradigmsaievaluationmapping} that functionally connect observed behaviour to latent variables) that  comprehensively characterise a model's behaviour. Starting from the obtained capability profile, the model's performance on specific task instances could be predicted from the ``demands'' they pose on each capability, even if the model was not previously evaluated on task instances with that specific combination of demands.
Therefore, ability-oriented evaluation provides an interpretable description of a model's characteristics and fosters the goal of building predictable AI \citep{zhou2024predictable}, which is important to anticipate and mitigate model errors and thus contribute to safety. 

Our work is an example of ability-oriented evaluation: we extract capabilities of models related to the dimensions we define, which comprehensively characterise a model and apply across various tasks and domains. Further, the inferred capability levels could be combined with the annotated demands of each task instance to predict the model's performance on new individual task instances (for instance, with an approach similar to that in \citet{burden2023inferring}) or the accuracy on a benchmark with a domain profile, rather than building separate assessors as we did here. 
It is important to note that using radial plots in AI evaluations—even when labelled with the term “capability”—does not inherently imply an ability-oriented assessment. Often, these plots merely represent aggregate performance, broken down by benchmarks or their subdomains \cite{srivastava2022beyond,balachandran2024eureka,fountas2024human,masry2025alignvlm}. For a score to genuinely qualify as a capability, the evaluation must incorporate either binary demands or difficulties.
Also, only when capabilities are defined as counterparts to demands or difficulties, can we properly talk about notions of generality as opposed to capability \citep{hernandez2021general} or as a derived metric from a range of capabilities \cite{hernandez2024caveats}. The profile we extract with our approach could be used for this purpose.

\paragraph{Extracting profiles of latent features explaining LLMs' behaviour}
Through the annotation of instances for the demands they pose on various dimensions and instance-level analysis of results, our approach can explain performance through a ``profile'' of model capabilities, in the spirit of ability-oriented evaluation (see the above paragraph). 
A few other works extracted latent dimensions characterising LLM behaviour. For example, both \citet{burnell2023revealing} and  \citet{ilic2024evidence} explained the variance of aggregate performance on different benchmarks for a population of models via a set of latent factors inferred through factor analysis. 
In particular, \citet{burnell2023revealing} found 3 latent factors to explain a large part of the variance and, by manual inspection, realised that these partially align with an \textit{a priori} classification of the cognitive skills needed to solve the benchmarks (then interpreting the factors as \textit{reasoning, comprehension and language modelling}). \citet{ilic2024evidence}, on the contrary, identified one general factor only.  
Similarly, \citet{ruan2024observational} introduced “observational scaling laws” that connect performance on complex downstream tasks with hypothesised latent capabilities. These latent values are estimated by decomposing the performance of a population of LLMs across different benchmarks into components that follow a log-linear relationship with the compute measures used in LLM training.
All of these methods rely on a population of models and explain aggregate performance; in contrast, by using instance-level annotations and performance of a single LLM, the capability profile we obtain is independent of the choice of an LLM population and of their compute footprint. Moreover, while their inferred latent factors have to be identified a posteriori, our capabilities are relative to a pre-defined demand, facilitating interpretation. At the same time, the assessors we train allow us to predict instance-level performance on an instance on which models were not previously tested; in contrast, the scaling laws in \citet{ruan2024observational} can predict aggregate performance on a benchmark which was previously processed in their framework, although they can do that for new models of a family that was used in their analysis, based on compute measures. Instead, \citet{burnell2023revealing} and \citet{ilic2024evidence} offer no predictive power for new models or new tasks beyond a qualitative estimate based on what abilities are presumed to be useful for a task.


\paragraph{Instance-level performance prediction of AI models}

\citet{zhou2024predictable} emphasised the importance of instance-level success predictions for AI models, coining the term ``predictable AI''. They argue that in high-stakes applications, prioritising predictability is a more valuable goal than pursuing unpredictable improvements in average performance. 
As a precursor of this idea, \citet{hernandez2022training} introduced the concept of an \textit{assessor}—a model trained to predict an ML system's performance on individual instances based on its evaluation results on test data (i.e., data not used for training the ML system). Several studies have explored variations of this approach. \citet{zhou2022reject} demonstrated that a smaller LLM could be trained to predict the performance of a larger LLM on individual instances without direct access to them. This assessor successfully rejected nearly half of the failure cases, leading to significant computational savings. \citet{schellaert2025analysing} accurately predicted LLM performance across more than 100 BIG-bench tasks \citep{srivastava2022beyond}, surpassing the confidence of the models themselves while maintaining predictability across different model sizes, suggesting scalability. \citet{DRAPAL2024112351} extracted explainable meta-rules from trained assessors to identify regions where performance is predictable. Instead, \citet{pacchiardi2024100instancesneedpredicting} trained assessors that leveraged shared information across multiple LLMs, reducing the number of instances each LLM needed to process for assessor training.
In this work, we place ourselves in this strand of literature by training assessors using our annotated demands on the various dimensions.

Beyond the assessor framework, alternative techniques from other fields have been adapted for instance-level performance prediction. For example, \citet{drapal2024MetaLearningNoveltyDetection} combined novelty detection with meta-learning to filter out instances where an ML system is likely to fail (additional similar approaches are discussed in Section 4 of \citet{hendrickx2024machine}).
Another related approach, inspired by intrinsic uncertainty quantification \citep{shorinwa2024survey}, is \citet{kadavath2022language}, which trained LLMs to estimate their probability of success on a question without reference to a specific answer. This was achieved either through natural language responses or by adding an additional ``head'' to the model.\footnote{Many other studies apply intrinsic uncertainty quantification to assess whether a model perceives a given statement as correct or incorrect after generating an answer. However, this differs from predicting performance \textit{before} an answer is produced. See \citet{shorinwa2024survey} for a broader discussion.}
Additionally, the approach in \citet{burden2023inferring} explicitly modelled the probability of an AI system succeeding on an instance using a set of system's latent capabilities and instance-level demands, and inferred a posterior for the capabilities using Bayesian inference starting from annotated demands. A prediction on an instance with assigned demands can then be obtained in terms of the ``posterior predictive'' distribution. 
Finally, Item Response Theory (IRT) \citep{embretson2013item} can be used to predict performance of AI models, but only on previously processed instances; this was done in previous work \citep{polo2024tinybenchmarks}, despite not being the main focus. We discuss connections of our work with IRT more extensively in the subsequent paragraphs. 
Finally, \citet{predictaboard} introduced \textit{Predictaboard}, a standardised benchmarking framework that jointly evaluates an LLM's performance and the ability of a performance prediction method, facilitating comparisons across assessors (including possibly our own) and other approaches, such as the ones discussed above.

\paragraph{How human users predict and understand LLM performance}
In parallel with developing a score predictor model, an important question arises: How effectively can humans predict where AI might fail?   Several works address this question. For instance, \citet{carlini_gpt4_challenge} found that human predictions are only marginally better than chance at predicting GPT-4's performance. Relatedly, \citet{vafa2024largelanguagemodelsperform} demonstrated that humans tend to overestimate the future performance of LLMs based on prior interactions, particularly with larger models in high-stakes contexts. They argue that ``the best LLM is the one that allows humans to make the most reliable inferences about where it will succeed''.  \citet{zhou2024larger} showed that as AI systems become more capable, human predictions of their performance become increasingly unreliable. Furthermore, \citet{steyvers2025large} found that LLM-generated explanations supporting a statement do not enable humans to reliably determine whether that statement is correct, even when the LLM's token-level probabilities are well-calibrated. Additionally, longer explanations were observed to increase user confidence, regardless of their accuracy. 
Our approach provides an interpretable profile of the capabilities of LLMs, which may help human users to more reliably predict where LLMs will perform successfully.

\paragraph{Item Response Theory} 
Item Response Theory (IRT) \citep{embretson2013item,bock2021item} originally developed as an alternative approach to classical test theory in psychometrics where the notion of difficulty was chosen to play a central stage \cite{rust2014modern}. One of the key observations in the evaluation of cognition and intelligence is that correlation between tests appears when the range of difficulties of the items in a test generates enough variance in the population. Not controlling for the difficulty of the items may lead to inefficient testing and the wrong information about whether several latent factors correlate, or simply missing an important construct because the range of difficulties was insufficient to generate enough variance in the population. As a reaction, IRT contrasts difficulty and ability from the start, and sets a common scale for them. 
Given a matrix of results of subjects and items, IRT can estimate the ability of the subjects and the difficulty of items at the same time.
Many traditional IRT models represent the probability of success as a logistic function of the  difference between ability and difficulty. 
The simplest of these models, the 1-PL or Rasch model, only estimates ability and difficulty, assuming the logistic curve has slope 1. The 2-PL model extends this by allowing a second parameter, the slope of the logistic curve (termed {\em discrimination}) to be inferred from the data. The 3-PL model includes a {\em guess} parameter, which is the minimum expected value of the response, which is usually appropriate for multiple-choice questions where distractors are obvious and the baseline goes above the statistical chance. In general, the guess parameter not being 0 is useful when there is a chance of success even without ability. This has inspired the \dm{Unguessability} dimension in this paper. The logistic curves are called item characteristic curves, with ability on the \xaxis and response on the \yaxis for a population of subjects. The dual curve is called the person characteristic curve \cite{vale1975study,
trabin1983person}, indicated in this paper as subject characteristic curve, which maps response on the \yaxis for a population of items as a function of the ability of one subject on the \xaxis. 
An important thing to clarify is that IRT cannot predict performance for items that have not been characterised during the estimation phase, and the parameters depend on the population of subjects. If the subjects or the items are changed (very different groups of humans), all the parameters may change.
IRT has been adapted to machine learning and NLP \cite{martinez2019item,lalor2016building,kipnis2024metabench,polo2024tinybenchmarks,vania-etal-2021-comparing,lalor2024item}.  
Of the many works adapting IRT to AI evaluation, a few applied it specifically to analyse LLM performance. For instance,  
\citet{Fang2024PATCHP} started from a test which has been validated on humans and converted it into prompts that are suitable for LLMs. It used human performance to obtain item difficulty and then used them to analyse LLM performance and get capability scores, using IRT-like models. Very similarly, \citet{zhuang2024staticbenchmarksadaptivetesting}  extracted difficulty of different instances in a benchmark and models' abilities from a population of models. 
Instead, \citet{Tang2023AGIBenchAM} introduced a dataset annotated using 5 levels of difficulty obtained from human performance scores, while \citet{Lei2024GAOKAOEvalDH} considered  a dataset annotated with difficulty scores by  human experts and employed a simple psychometric model to finding differences in how success for humans and LLMs correlates with the annotated difficulty.
Finally, \citet{federiakin2025improving} applied IRT to improve rankings on the HuggingFace leaderboard. 
Other works instead used IRT to improve the efficiency of benchmarking: \citet{polo2024tinybenchmarks} and \citet{kipnis2024metabench}  used IRT to select informative subsets of a benchmark and estimates the performance of a new LLM on the whole benchmark by evaluating it only on those instances.

\paragraph{Multidimensional Item Response Theory} 
On many occasions, one single ability-difficulty pair is not sufficient to account for the variability of responses; rather, more than one dimension is needed. Multidimensional IRT \cite{reckase2006,bonifay2019multidimensional} replaces the notion of item characteristic curve with an item characteristic surface, where two or more dimensions of ability-difficulty are represented. Estimating this surface will depend on the assumption of independence of the dimensions. Two abilities are considered compensatory if the lack of one can be compensated by the other. For instance, for recognising a person, we can use face recognition ability and voice recognition ability. In a situation when one ability is affected or the demand is too high (e.g., noise in the image or the audio), then the other can compensate. Conversely, two abilities are considered non-compensatory when the lack of one cannot be compensated by the other.
One important question from multidimensional IRT is how to extract individual dimensions from the multidimensional space that are calibrated 
 \cite{ackerman1989unidimensional}. This depends on a series of assumptions, such as the level of compensatoriness.
Some multidimensional IRT models have been applied to machine learning as well: \citet{Liu2023MultiDimensionalAD} considered an extension of multidimensional IRT using a set of subjects, their instance-level performance on a dataset and a binary annotation matrix encoding the ``skills'' required by each instance to infer levels of skill-specific ability for the different subjects and obtain an overarching difficulty (and discrimination) factor for each instance. These inferred quantities depend however on the considered population of learners, in contrast to  our approach, which is non-populational. Further, our approach uses  numerical levels of demands over different dimensions, while they only considered binary indicators for whether a specific skill is needed in an instance, without quantifying skill-specific demand.

\paragraph{Other extensions of IRT}  
Linear logistic test models (LLTM)  \cite{fischer1973linear} consider a vector of binary demands for each item, expressing if a given demand is needed or not for a particular item (they can also be greater than one, in an ordinal or quantitative scale). The matrix of all items and demands is called the $Q$-matrix. LLTM considers one single difficulty as a linear function of the demands, with the parameters of that function being estimated from data, and being the same for all subjects and items. Similarly to traditional IRT, the item difficulty is opposed to the ability. Despite having several demands, the model assumes one single ability per subject, and it cannot generate ability profiles. Also, the parameters are estimated from the whole population of items and subjects, and these parameters can change if the subject population changes. 
However, one great advantage of LLTMs is the much smaller number of parameters to estimate, as difficulty is derived a function of the demands. Because of this, it can be applied to new instances. 
LLTM has been generalised by the area known as `explanatory item response models' \cite{de2004explanatory} where the approach LLTM applies to item parameters is extended to person parameters or both person and item parameters (known as doubly explanatory models), or in the multidimensional case \cite{de2014multidimensional}.
Cognitive diagnostic models \cite{von2004class,von2008general,dibello2015family} are a related approach and also use a Q-matrix, but can generate `multidimensional skill profiles'. 
Lately, machine learning methods are being used to estimate these models. \citet{wang2024survey} provide a history of the entire field, encompassing both the traditional approaches and the new  ones based on machine learning. 
Finally, the bifactor scoring model \cite{chen2018bifactor}, separates a general factor, accounting for communality among all items (or items from domains that are highly correlated), and group factors (the capability dimensions), accounting for domain-wise variances. 
Unfortunately, most of these advanced models have not been applied in AI, despite the recent calls to do so \cite{wang2023evaluating}.

\paragraph{Situation of this paper in the space of psychometric methods}
Our work adapts and integrates many ideas from psychometrics, in particular from IRT. Our approach is similar to LLTM in the use of demands (the \ADeLe battery can be seen as a Q-matrix) and especially to explanatory item response models and 
cognitive diagnostic models. 
However, we do \textit{not} consider demands as being determined by the difficulty for a fixed population. 
The analogy with the bifactor scoring models can shed light on our observations of our subjects not having probability of success 1 at demand level 0 (as can be seen in the subject characteristic curves): that is because the subject needs to understand the question and know that they need to give an answer. 
To account for this, some IRT approaches introduce a parameter called {\em inattention}. 
%
%
%
%
Our approach can be seen as starting with a multidimensional model, assuming independence, and then slicing it using the `dominant' approach. In general, slicing multidimensional spaces into unidimensional spaces implies important assumptions and depends on the data \cite{stucky2014using}. Our use of the `dominant' slicing approach induces a dependence on the correlations between dimensions in the battery we consider, which may lead to poor calibration the less pure the items are.
The major difference between all of the above psychometric models and our methodology originates from the nature of the subject: humans for psychometrics and machines for AI. For adult humans, the notion of population is meaningful and is generally stable. However, the capabilities of AI models and their similarity are changing quickly, so any result that depends on the variance of the population of benchmarks would need to be reconsidered every year when a new generation of models appear. For instance, the factor analysis studies changed conclusions between \cite{burnell2023revealing} and \cite{ilic2024evidence}, most likely because of a different sample of models, despite the studies being conducted just a few month apart. 
In principle, we can produce a score for single individuals by looking at the individual’s responses to items and generating a score from the model parameters \cite{thissen2002test}. However, 
 psychometrics rarely builds a model from a single human with a ``cold start'' (i.e., not adapting models informed by a population), because it makes sense to use the information of many other humans, and collecting a sufficient amount of data from a single individual is costly. In the case of our battery \ADeLe, we have more than 16,000 items per LLM, from which we can make strong inferences (including high predictability). 
 The other difference is that very few dimensions are usually enough for humans, because most capabilities and traits show high correlations. Moreover, the smaller number of items in human studies with respect to what is possible with AI systems lead to stronger effect of noise, which reduces the number of parameters that can be accurately estimated: with 50 items, a single odd item may produce noise and estimate error, but a high number of dimensions is possible with 16,000 items. As such, with more data, there is greater potential to uncover regularities beyond a limited set of dimensions, thereby revealing processes that might otherwise remain undetected. 


\paragraph{Item design, procedural generation and adaptive testing} 
In our paper, we have not explored   exploiting the rubrics and annotations for better item design, 
procedural generation \cite{mirzadeh2024gsm} 
and adaptive testing. These approaches are common in assessment, especially with the use of difficulty levels from IRT, but have also been adopted to AI. 
For instance, \citet{zhuang2024staticbenchmarksadaptivetesting} used IRT to extract difficulty of different instances in a benchmark and relied on these to perform adaptive testing by selecting items whose difficulty is more informative of a model's ability level. 
A similar approach could be taken with our annotated demands, although their multidimensional nature should be taken into account. 
Instead, \citet{Guinet2024AutomatedEO} used IRT to improve the quality of a benchmark for Retrieval Augmented Generation (RAG) LLMs by eliminating the questions that are not sufficiently informative about a model's ability. 
Again, we could analogously rely on our extracted demand levels to ensure that a benchmark comprehensively tests LLMs in a range of situations.
Finally, \citet{zhang2024task} performs adaptive labelling and procedural generation simultaneously. Interestingly, their categories are organised in a taxonomy of binary dimensions. 

\paragraph{Measurement theory and AI scales} 
Two important concepts in measurement theory \cite{hand2016measurement} are the type of scale and the measurement units. Steven's typology of measurement \cite{stevens1946theory} includes `nominal', `ordinal', `interval' and `ratio' scales, and many other topologies have appeared since then. For quantitative measurement, interval scales make the comparisons of differences meaningful, and calculating a mean is well justified. Ratio scales set an absolute value, and hence ratios become meaningful.  
For a latent factor, the choice of a scale is arbitrary. In IRT, it is somewhat controversial (e.g., \cite{michell1999measurement,lord1975ability}). Although some researchers advocate for a ratio or interval-scale basis for test scores (e.g., `the odds (or log odds) of student 1 answering a class of items correctly are twice the odds (or log odds) of student 2 doing so' is a meaningful assertion \cite{freund2019rasch}), others point out that a latent trait scale of a fitted IRT model is convenient but not inherently meaningful because any monotonic transformation of the estimated latent trait scores produces an equally valid model \cite{lord1975ability,wallmark2024introducing}. 
In contrast to this, some magnitudes are associated with cognition  have a clear scale, such as size or time. For instance, for the dimension \dm{Attention and Scan}, the length (e.g., in words) could be relevant, and for the \dm{Volume} dimension we have used time (and we saw correlation between the size of a question and the time to solve it, \cite{kazemi2025big}). Time is isolated in the person-month metric, used everywhere in human resources, project management, or software engineering, which can be mapped to educational levels, taking into account the effort in days that several levels require (e.g., OECD data \cite{OECD2024}).  
\citet{morris2023levels} uses percentiles of the human population for their Emerging, Competent, Expert, Virtuoso and Superhuman levels of AGI. Here, we do not base the levels on outperforming percentiles, but on the probability of finding a sample of $10^l$ humans with more than 95\% chance of at least one being correct.





\newpage

\subsection{Scaling Curves of Model Abilities}
\label{ap:scaling_laws_of_abilities}

\begin{figure}[!ht]
  \centering
  \includegraphics[width=0.7\linewidth]{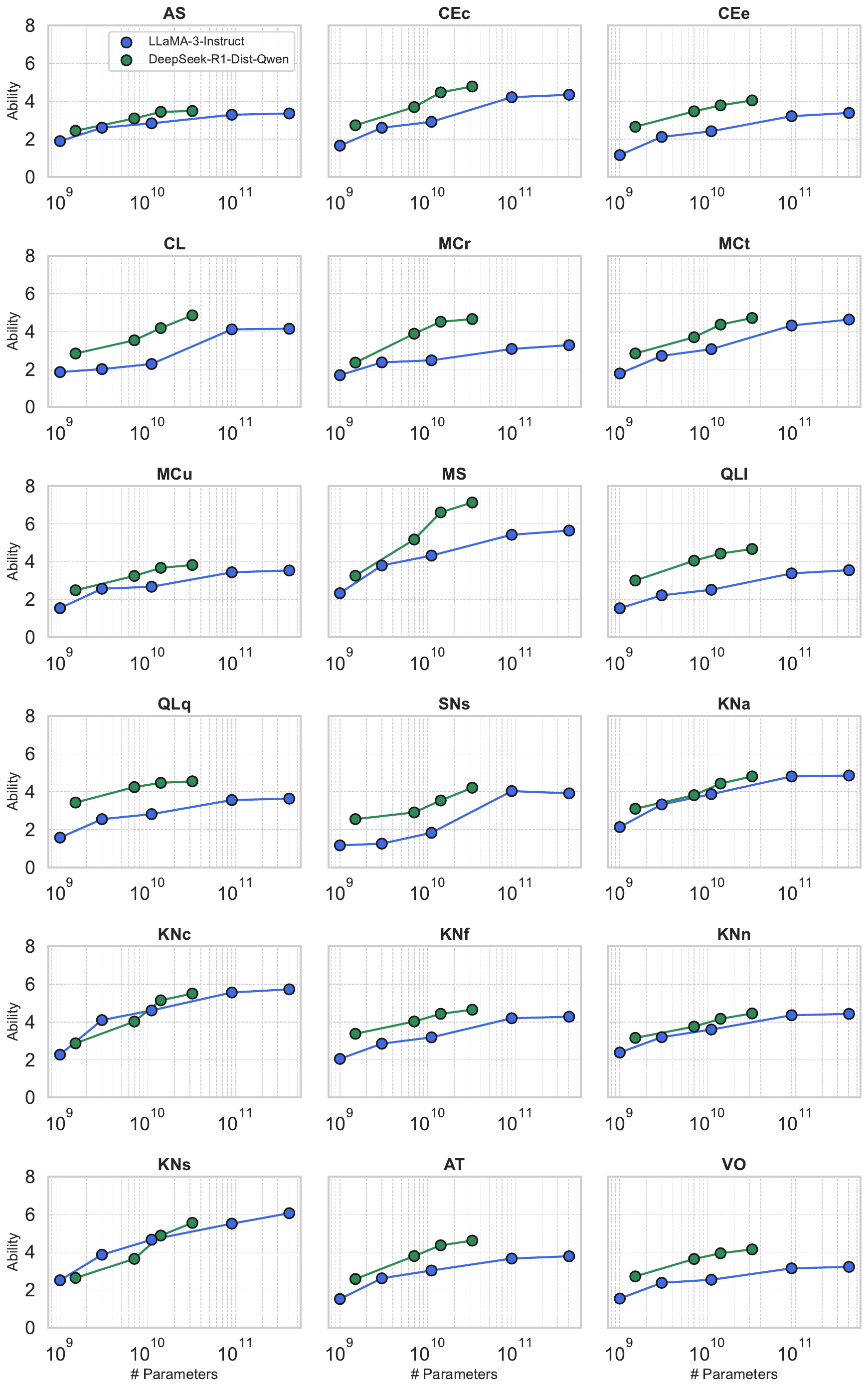}
\caption{The scaling curves of actual abilities for {\color{darkblue}{\model{LLaMA}}} and {\color{darkgreen}{\model{DK-R1-Distilled-Qwen}}} families across all 18 demands.}
  \label{fig:ability_scaling}
\end{figure}

\begin{figure}[!ht]
  \centering
  \includegraphics[width=0.8\linewidth]{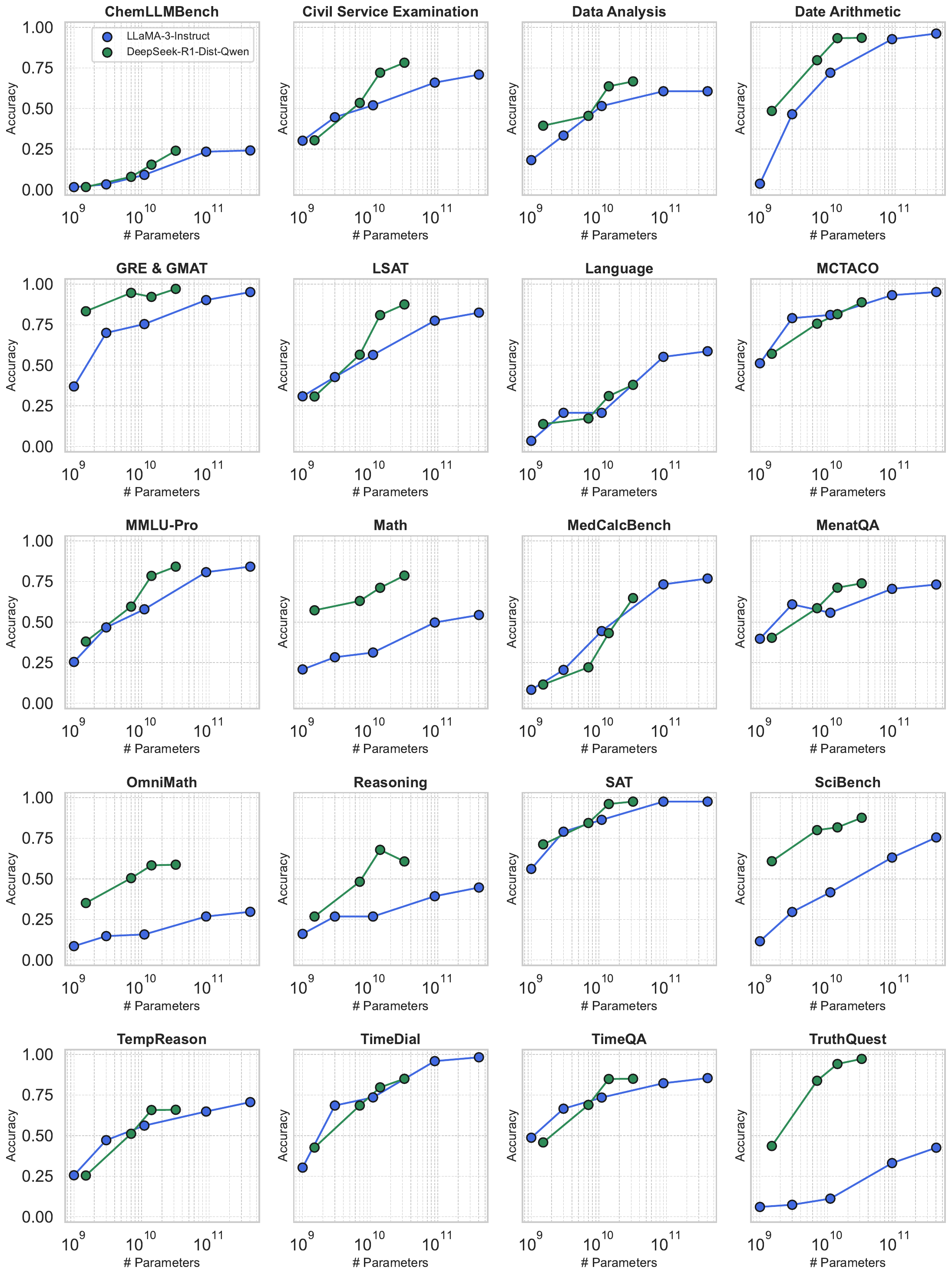}
\caption{The scaling curves of performance for {\color{darkblue}{\model{LLaMA}}} and {\color{darkgreen}{\model{DK-R1-Distilled-Qwen}}} families across all 20 benchmarks.}
  \label{fig:performance_scaling}
\end{figure}

Figure \ref{fig:ability_scaling} and Figure \ref{fig:performance_scaling} show the scaling curves of model abilities and performance, respectively, as a function of model size, for both \model{LLaMA} and \model{DK-R1-Distilled-Qwen} families.

Traditional performance scaling analyses, such as those shown in Figure~\ref{fig:performance_scaling}, which aggregates results across 20 benchmarks, are susceptible to saturation effects, not only because the scale on \yaxis is bounded by 1 but also because there may be some abstruse or even wrongly labelled questions that make the percentages never reach 100\%.  
For the most powerful models, the composite performance scores flatten across many benchmarks, making it difficult to interpret incremental improvements as model size increases. This saturation can mask subtle but important gains in specific cognitive abilities. In contrast, 
our ability scaling curves based on ratio scales remain sensitive and informative: they avoid benchmark saturation and show clear trends even for the largest models. This discloses  insights, but the most notable one is the clear diminishing return from the second largest to the largest model in both \model{LLaMA} and \model{DK-R1-Distilled-Qwen} families, consistent in nearly all dimensions. In other words, while performance generally increases with model size, our method reveals that the magnitude of skill improvement tapers off beyond a certain size.

\newpage

\subsection{Calibration of Assessors}
\label{ap:IID_assessors_calibration}

\begin{figure}[!ht]
  \centering
  \includegraphics[width=0.85\linewidth]{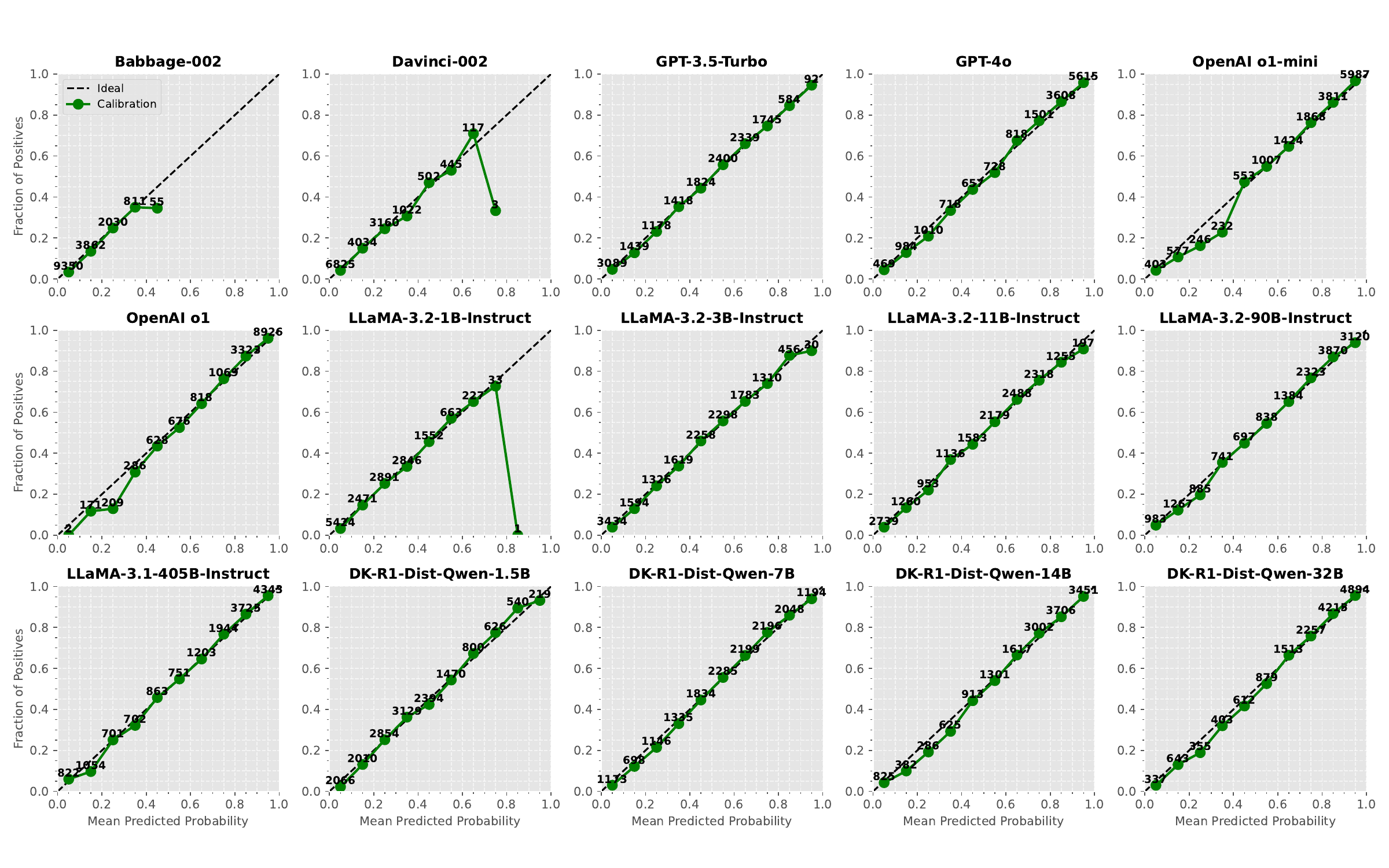}
\caption{Calibration of the \model{RF} demand-based assessor for the 15 subject LLMs, in-distribution, from Table \ref{tab:main-id-predictability-results}. The x-axis corresponds to the estimated probability of success, while the y-axis corresponds to the empirical performance. The numbers in each dot indicate the number of instances in a given bin.} 
  \label{fig:rf_calibration_ID}
\end{figure}

\begin{figure}[!ht]
  \centering
  \includegraphics[width=0.85\linewidth]{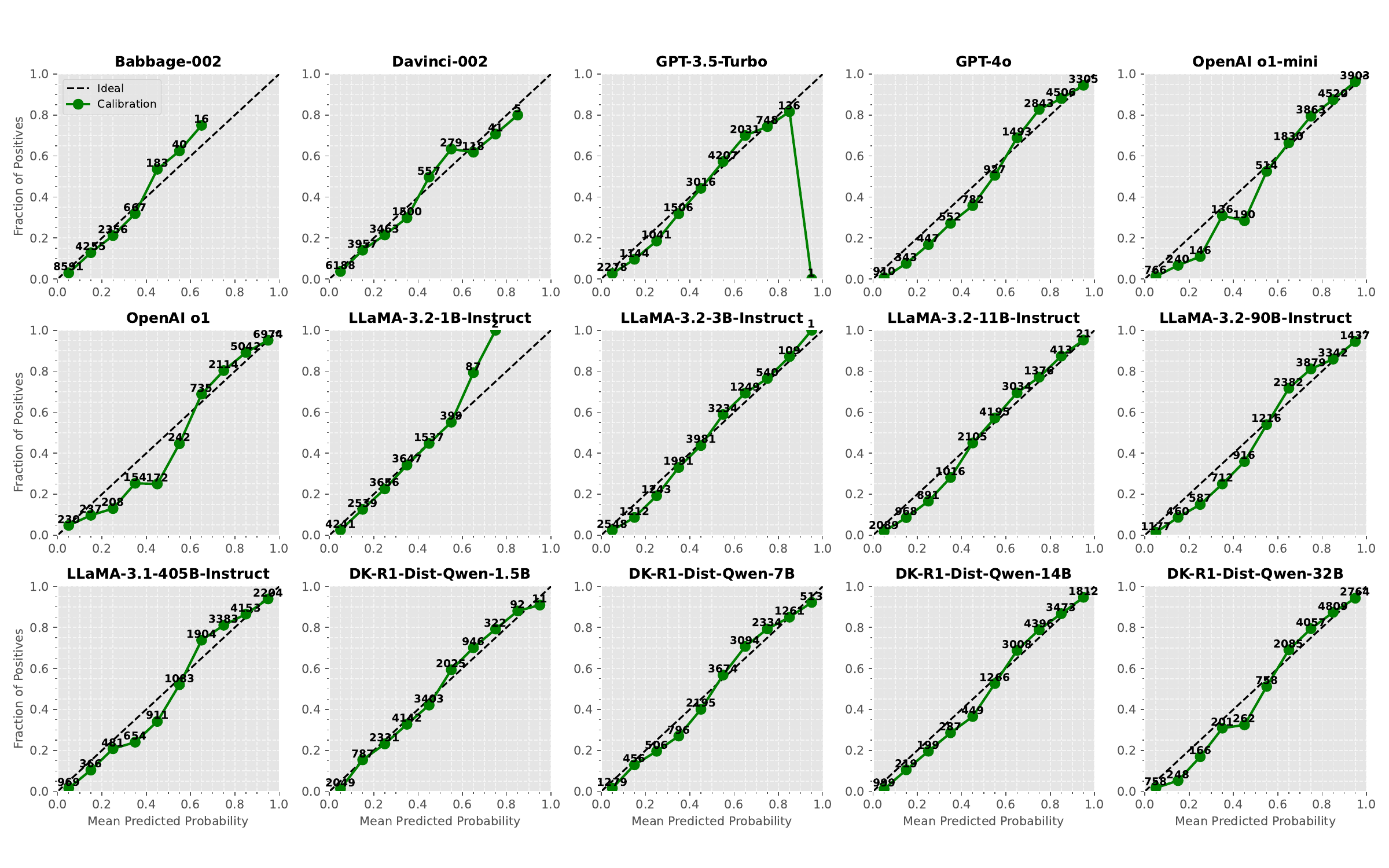}
\caption{Calibration of the \model{GloVe} assessor for the 15 subject LLMs, in-distribution, from Table \ref{tab:main-id-predictability-results}. The x-axis corresponds to the estimated probability of success, while the y-axis corresponds to the empirical performance. The numbers in each dot indicate the number of instances in a given bin.}
  \label{fig:glove_calibration_ID}
\end{figure}

\begin{figure}[!ht]
  \centering
  \includegraphics[width=0.85\linewidth]{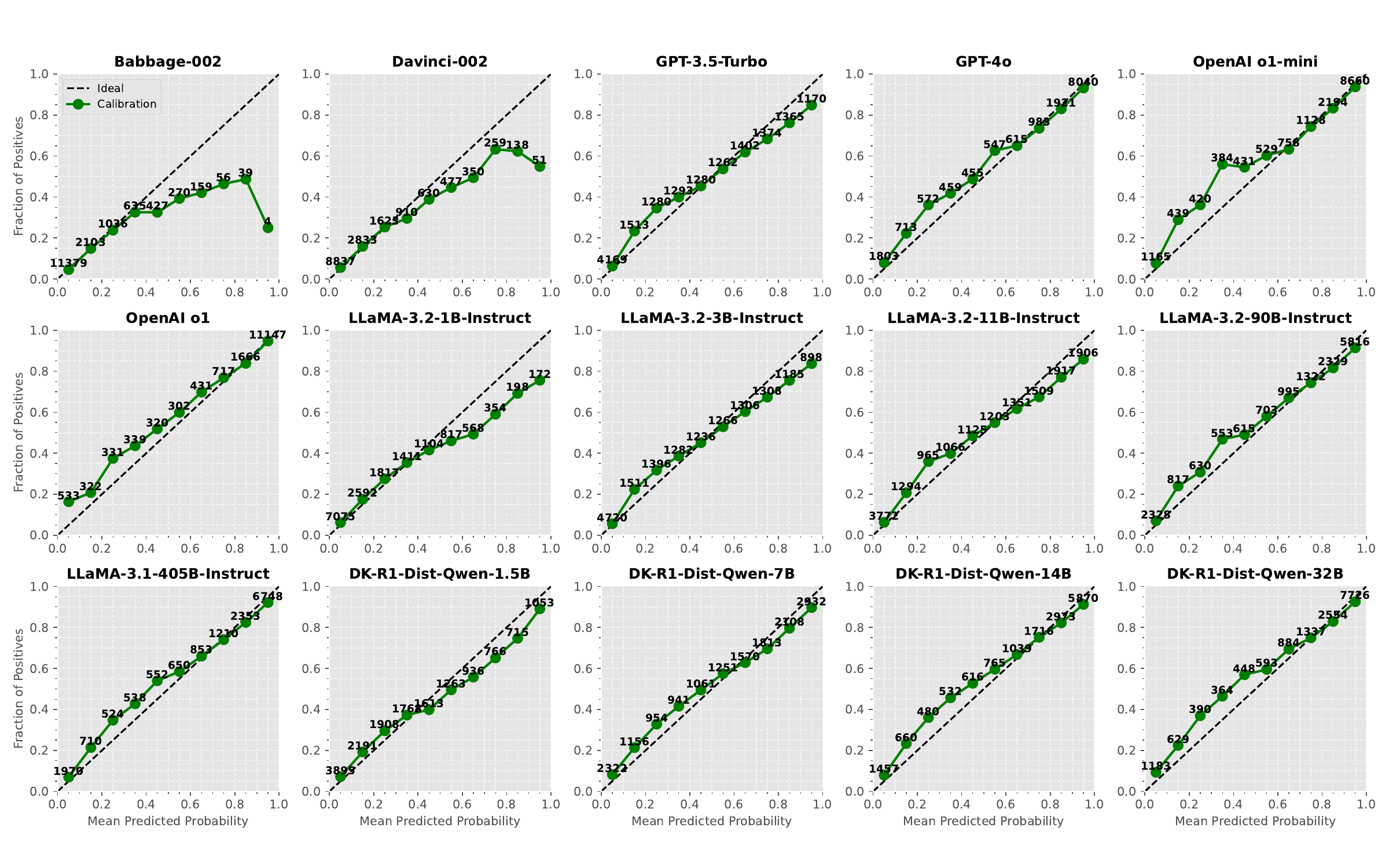}
\caption{Calibration of the \model{LLAMA} assessor for the 15 subject LLMs, in-distribution, from Table \ref{tab:main-id-predictability-results}. The x-axis corresponds to the estimated probability of success, while the y-axis corresponds to the empirical performance. The numbers in each dot indicate the number of instances in a given bin.}
  \label{fig:llama_calibration_ID}
\end{figure}

Figure \ref{fig:rf_calibration_ID}, \ref{fig:glove_calibration_ID}, \ref{fig:llama_calibration_ID} show the calibration of our demand-based, \model{GLOVE} and \model{LLAMA} assessors (from Table \ref{tab:main-id-predictability-results}) on the in-distribution setup. This confirms our observation in section \ref{sec:pred}: the demand-based assessor achieves nearly perfect calibration, while the two black-box assessors are noticeably worse.

\newpage
\subsection{Sources of Unpredictability}
\label{ap:aleatoric_uncertainty_sources}

There are various sources that hinder perfect predictability more broadly. This includes both epistemic uncertainty, which can be reduced by additionally finding relevant predictive features (e.g., including new features or extending the demand level of 5+ in our \DeLeAn rubrics to level 6 and beyond), and aleatoric uncertainty, which is inherently irreducible \cite{hullermeier2021aleatoric}. The latter may come from, but not limited to, chance guess rate in multiple-choice questions, memorisation of test instances, inherent stochasticity of reasoning models or models based on the Mixture-of-Experts architecture, prompt sensitivity, among others. 

The first two sources of aleatoric uncertainty listed above, chance guess rate and memorisation, stem from data. The impact of these seems quite large, as supported by the analysis in Appendix \ref{assessor_with_AT_UG_VO}, in which we build a demand-based assessor by only using \dm{AT} (\dm{Atypicality}), \dm{VO} (\dm{Volume}) and \dm{UG} (\dm{Unguessability}); future research should seek solutions for controlling and minimising these sources of unpredictability coming from the data. The other two sources of aleatoric uncertainty come from the high-quality but imperfect input-response graders (see section \ref{sec:models}) and demand annotators (see section \ref{sec:interrater}). In other words, the predictive power of the demand-based assessors shown in this paper is in fact underrated, as the imperfect verification and demand annotations introduce noise and thus decrease the upper bound of predictive power that can be accomplished. This is unfortunate but encouraging, as it simultaneously implies that the predictive power of assessors based on our methodology will increase as future LLMs improve their capabilities.

In contrast, the other sources of unpredictability stem from the design of AI systems (e.g., many LLMs today are not deterministic, even at temperature 0, and chain-of-thought LLMs usually use a temperature well above 0); future efforts should look for pathways that enable the minimisation of inherent unpredictability of models, as previously suggested by \citet{zhou2024predictable}, alongside a discussion of various possible future pathways for striving toward this goal.

\newpage
\subsection{Other Predictive Models}\label{ap:other_assessors}

\subsubsection{Feature Importance}

\begin{figure}[!ht]
  \centering
  \includegraphics[width=1\linewidth]{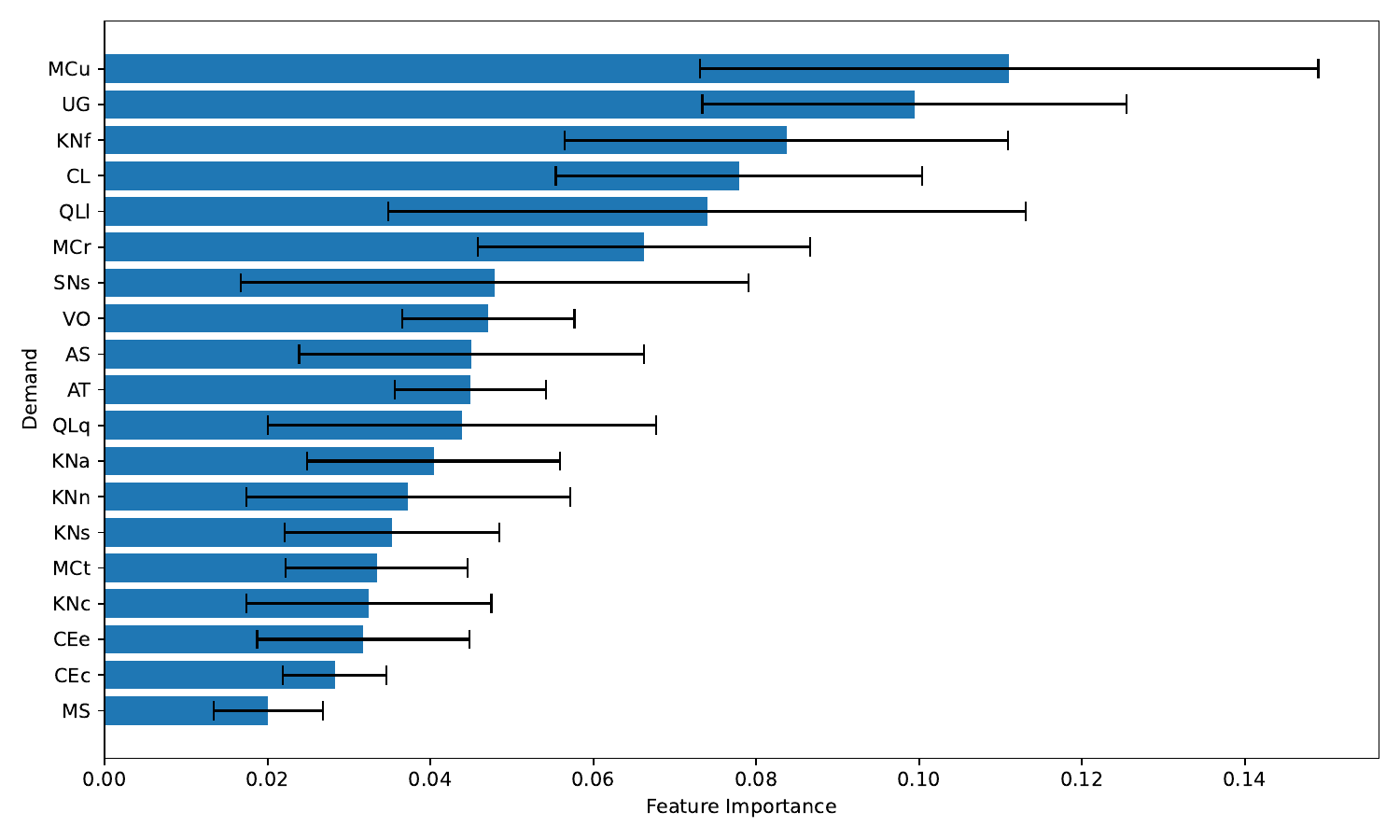}
  \caption{Feature importance of all dimensions averaged across all demand-based assessors for different subject LLMs.}
  \label{fig:average_feature_importance}
\end{figure}

Figure \ref{fig:average_feature_importance} shows the feature importance\footnote{We use the ‘permutation\_importance’ in https://scikit-learn.org/stable/modules/generated/sklearn.inspection.html. It assesses each feature’s importance by shuffling the feature’s values and measuring the model’s performance change.} of the 19 dimensions for all the demand-based assessors we have trained for Table \ref{tab:main-id-predictability-results}. We see that no demand has a feature importance value below 0.02, suggesting all demands are relevant to a greater or lesser extent.  Based on the elbow method, we can select the six most relevant dimensions: \dm{MCu} (\dm{Calibrating Known Unknowns}), \dm{UG} (\dm{Unguessability}), \dm{KNf} (\dm{Knowledge of Formal Sciences}), \dm{CL} (\dm{Conceptualisation, Learning and Abstraction}), \dm{QLl} (\dm{Logical Reasoning}), and \dm{MCr} (\dm{Identifying Relevant Information}). This ranking is strongly correlated (Spearman Corr. = 0.718) with the ranking of correlation magnitudes between demands and success, as observed in Figure \ref{fig:demand_correlation}.

To explore whether it is possible to achieve high predictive power with a smaller number of dimensions, we feed the selection from the elbow method as input to the demand-based assessor using the same  algorithm (\model{Random Forest}). We observe that predictive power in general decreases between 0.01 and 0.03 on average, but still fairly high (Table \ref{tab:predictability-results-only-6dims-elbow-method})
. This is promising as it implies that high predictive power may be achieved through a small number of dimensions.

\begin{table*}[!h]
\centering
\caption{Predictive power of the demand-based assessor by using only \dmc{MCu} (\dmc{Calibrating Known Unknowns}), \dmc{UG} (\dmc{Unguessability}), \dmc{KNf} (\dmc{Knowledge of Formal Sciences}), \dmc{CL} (\dmc{Conceptualisation, Learning and Abstraction}), \dmc{QLl} (\dmc{Logical Reasoning}), and \dmc{MCr} (\dmc{Identifying Relevant Information}) for in-distribution, task out-of-distribution, and benchmark out-of-distribution.}
\resizebox{0.87\textwidth}{!}{
\begin{tabular}{R{0.25\textwidth}C{0.17\textwidth}C{0.1\textwidth}C{0.1\textwidth}C{0.1\textwidth}C{0.1\textwidth}C{0.1\textwidth}C{0.1\textwidth}}
\toprule
\multirow{2}{*}{\textbf{Subject LLM}}
&  \multirow{2}{*}{\textbf{LLM Accuracy}$\uparrow$ }  & \multicolumn{2}{c}{\textbf{ID}} & \multicolumn{2}{c}{\textbf{Task OOD}} & \multicolumn{2}{c}{\textbf{Benchmark OOD}}\\  \cmidrule(l){3-4}  \cmidrule(l){5-6} \cmidrule(l){7-8}  
&  & \textbf{AUROC}$\uparrow$ & \textbf{ECE}$\downarrow$ & \textbf{AUROC}$\uparrow$ & \textbf{ECE}$\downarrow$ & \textbf{AUROC}$\uparrow$ & \textbf{ECE}$\downarrow$  \\  
\midrule
\model{Babbage-002 }              & 0.102 & 0.761 & 0.009 & 0.726 & 0.019 & 0.714 & 0.024 \\ 
\model{Davinci-002 }              & 0.157 & 0.751 & 0.007 & 0.724 & 0.019 & 0.696 & 0.039 \\ 
\model{GPT-3.5-turbo}             & 0.414 & 0.792 & 0.012 & 0.773 & 0.026 & 0.767 & 0.024 \\ 
\model{GPT-4o   }                 & 0.713 & 0.858 & 0.013 & 0.827 & 0.039 & 0.814 & 0.059 \\ 
\model{OpenAI o1-mini}                   & 0.770 & 0.830 & 0.012 & 0.800 & 0.032 & 0.734 & 0.051 \\ 
\model{OpenAI o1}                        & 0.843 & 0.815 & 0.010 & 0.777 & 0.032 & 0.737 & 0.043 \\ \midrule
\model{LLaMA-3.2-1B-Instruct}               & 0.216 & 0.768 & 0.011 & 0.728 & 0.034 & 0.711 & 0.048 \\ 
\model{LLaMA-3.2-3B-Instruct }              & 0.378 & 0.794 & 0.009 & 0.779 & 0.027 & 0.757 & 0.042 \\ 
\model{LLaMA-3.2-11B-Instruct}              & 0.463 & 0.801 & 0.010 & 0.780 & 0.041 & 0.768 & 0.036 \\ 
\model{LLaMA-3.2-90B-Instruct }             & 0.645 & 0.837 & 0.015 & 0.807 & 0.033 & 0.795 & 0.072 \\ 
\model{LLaMA-3.1-405B-Instruct }            & 0.683 & 0.846 & 0.010 & 0.814 & 0.037 & 0.797 & 0.077 \\ \midrule
\model{DK-R1-Dist-Qwen-1.5B }     & 0.353 & 0.753 & 0.010 & 0.733 & 0.023 & 0.681 & 0.053 \\ 
\model{DK-R1-Dist-Qwen-7B}        & 0.555 & 0.781 & 0.013 & 0.762 & 0.017 & 0.666 & 0.053 \\ 
\model{DK-R1-Dist-Qwen-14B}       & 0.698 & 0.797 & 0.012 & 0.774 & 0.030 & 0.691 & 0.071 \\ 
\model{DK-R1-Dist-Qwen-32B}       & 0.748 & 0.809 & 0.013 & 0.785 & 0.036 & 0.714 & 0.075 \\ \midrule
Weighted Average & --- & 0.811 & 0.012 & 0.784 & 0.031 & 0.742 & 0.056 \\
\bottomrule
\end{tabular}
}
\label{tab:predictability-results-only-6dims-elbow-method}
\end{table*}

\subsubsection{Assessor Only with \dm{AT}, \dm{UG}, and \dm{VO}}
\label{assessor_with_AT_UG_VO}
Table \ref{tab:main-predictability-results-only-AT-UG-VO}
shows that a demand-based assessor fed only with \dm{AT} (\dm{Atypicality}), \dm{UG} (\dm{Unguessability}), \dm{VO} (\dm{Volume}) can achieve fairly good predictive power. This demonstrates that existing benchmarks incorporate extraneous demands in many task instances, identifying contamination, amalgamation and funnelling, respectively, that are predictive of a large proportion of LLM success. These are extraneous interventions in popular AI benchmarks that make them not measure what they claim they measure and should be controlled in the design of future benchmarks to ensure construct validity.

\renewcommand{\arraystretch}{1.02}
\begin{table*}[!h]
\centering
\caption{Predictive power of the demand-based assessor by using only \dmc{AT} (\dmc{Atypicality}), \dmc{UG} (\dmc{Unguessability}) and \dmc{VO} (\dmc{Volume}) for in-distribution, task out-of-distribution, and benchmark out-of-distribution.}
\resizebox{0.87\textwidth}{!}{
\begin{tabular}{R{0.25\textwidth}C{0.17\textwidth}C{0.1\textwidth}C{0.1\textwidth}C{0.1\textwidth}C{0.1\textwidth}C{0.1\textwidth}C{0.1\textwidth}}
\toprule
\multirow{2}{*}{\textbf{Subject LLM}}
&  \multirow{2}{*}{\textbf{LLM Accuracy}$\uparrow$ }  & \multicolumn{2}{c}{\textbf{ID}} & \multicolumn{2}{c}{\textbf{Task OOD}} & \multicolumn{2}{c}{\textbf{Benchmark OOD}}\\  \cmidrule(l){3-4}  \cmidrule(l){5-6} \cmidrule(l){7-8}  
&  & \textbf{AUROC}$\uparrow$ & \textbf{ECE}$\downarrow$ & \textbf{AUROC}$\uparrow$ & \textbf{ECE}$\downarrow$ & \textbf{AUROC}$\uparrow$ & \textbf{ECE}$\downarrow$  \\  
\midrule
\model{Babbage-002 }   & 0.102 & 0.676 & 0.007 & 0.590 & 0.043 & 0.548 & 0.070 \\ 
\model{Davinci-002 }   & 0.157 & 0.669 & 0.006 & 0.615 & 0.039 & 0.584 & 0.078 \\ 
\model{GPT-3.5-Turbo}   & 0.414 & 0.748 & 0.009 & 0.720 & 0.041 & 0.670 & 0.096 \\ 
\model{GPT-4o     }     & 0.713 & 0.809 & 0.004 & 0.773 & 0.050 & 0.741 & 0.048 \\ 
\model{OpenAI o1-mini}        & 0.770 & 0.738 & 0.005 & 0.693 & 0.038 & 0.620 & 0.082 \\ 
\model{OpenAI o1 }   & 0.843 & 0.727 & 0.004 & 0.672 & 0.028 & 0.598 & 0.086 \\ \midrule
\model{LLaMA-3.2-1B-Instruct }    & 0.216 & 0.707 & 0.007 & 0.652 & 0.052 & 0.618 & 0.101 \\ 
\model{LLaMA-3.2-3B-Instruct  }   & 0.378 & 0.738 & 0.008 & 0.706 & 0.030 & 0.650 & 0.130 \\ 
\model{LLaMA-3.2-11B-Instruction}     & 0.463 & 0.748 & 0.006 & 0.717 & 0.053 & 0.682 & 0.076 \\ 
\model{LLaMA-3.2-90B-Instruct }    & 0.645 & 0.792 & 0.007 & 0.758 & 0.041 & 0.727 & 0.070 \\ 
\model{LLaMA-3.1-405B-Instruct }    & 0.683 & 0.799 & 0.008 & 0.763 & 0.065 & 0.739 & 0.069 \\ \midrule
\model{DK-R1-Dist-Qwen-1.5B }    & 0.353 & 0.685 & 0.007 & 0.657 & 0.033 & 0.569 & 0.122 \\ 
\model{DK-R1-Dist-Qwen-7B  }   & 0.555 & 0.686 & 0.007 & 0.645 & 0.054 & 0.568 & 0.120 \\ 
\model{DK-R1-Dist-Qwen-14B }    & 0.698 & 0.713 & 0.006 & 0.664 & 0.063 & 0.584 & 0.114 \\ 
\model{DK-R1-Dist-Qwen-32B }    & 0.748 & 0.730 & 0.007 & 0.692 & 0.039 & 0.608 & 0.101 \\ \midrule
Weighted Average & --- & 0.742 & 0.006 & 0.701 & 0.045 & 0.644 & 0.089 \\
\bottomrule
\end{tabular}
}
\label{tab:main-predictability-results-only-AT-UG-VO}
\end{table*}

\subsubsection{Assessor without \dm{AT}, \dm{UG}, and \dm{VO}}
\label{assessor_without_AT_UG_VO}

Table \ref{tab:predictability-results-without-ATUGVO} demonstrates that the predictive power of a demand-based assessor fed with all the dimensions except for the extraneous ones, can achieve satisfactory results, though  worse to a small extent than the demand-based assessor fed with all the dimensions, as shown previously in Table \ref{tab:main-id-predictability-results}, \ref{tab:main-task-ood-predictability-results} and \ref{tab:main-benchmark-ood-predictability-results}.
This means that some other variables can compensate for this, including \dm{unguessability}. This suggests that fully controlling for these variables could be done with some hierarchical assessors or, possibly, by performing a multivariate analysis. 

\begin{table*}[!h]
\centering
\caption{Predictive power of the demand-based assessor using all the demands expect for \dmc{AT} (\dmc{Atypicality}), \dmc{UG} (\dmc{Unguessability}) and \dmc{VO} (\dmc{Volume}), for in-distribution, task out-of-distribution, and benchmark out-of-distribution.}
\resizebox{0.87\textwidth}{!}{
\begin{tabular}{R{0.25\textwidth}C{0.17\textwidth}C{0.1\textwidth}C{0.1\textwidth}C{0.1\textwidth}C{0.1\textwidth}C{0.1\textwidth}C{0.1\textwidth}}
\toprule
\multirow{2}{*}{\textbf{Subject LLM}}
&  \multirow{2}{*}{\textbf{LLM Accuracy}$\uparrow$ }  & \multicolumn{2}{c}{\textbf{ID}} & \multicolumn{2}{c}{\textbf{Task OOD}} & \multicolumn{2}{c}{\textbf{Benchmark OOD}}\\  \cmidrule(l){3-4}  \cmidrule(l){5-6} \cmidrule(l){7-8}  
&  & \textbf{AUROC}$\uparrow$ & \textbf{ECE}$\downarrow$ & \textbf{AUROC}$\uparrow$ & \textbf{ECE}$\downarrow$ & \textbf{AUROC}$\uparrow$ & \textbf{ECE}$\downarrow$  \\  
\midrule
\model{Babbage-002 }              & 0.102 & 0.775 & 0.004 & 0.737 & 0.010 & 0.664 & 0.033 \\ 
\model{Davinci-002 }              & 0.157 & 0.763 & 0.008 & 0.732 & 0.011 & 0.715 & 0.015 \\ 
\model{GPT-3.5-turbo}             & 0.414 & 0.802 & 0.005 & 0.783 & 0.019 & 0.753 & 0.053 \\ 
\model{GPT-4o   }                 & 0.713 & 0.871 & 0.013 & 0.841 & 0.022 & 0.790 & 0.041 \\ 
\model{OpenAI o1-mini}            & 0.770 & 0.848 & 0.007 & 0.827 & 0.017 & 0.736 & 0.027 \\ 
\model{OpenAI o1}                 & 0.843 & 0.839 & 0.009 & 0.796 & 0.026 & 0.705 & 0.031 \\ \midrule
\model{LLaMA-3.2-1B-Instruct}     & 0.216 & 0.771 & 0.004 & 0.722 & 0.031 & 0.698 & 0.053 \\ 
\model{LLaMA-3.2-3B-Instruct }    & 0.378 & 0.804 & 0.008 & 0.780 & 0.022 & 0.752 & 0.050 \\ 
\model{LLaMA-3.2-11B-Instruct}    & 0.463 & 0.810 & 0.008 & 0.788 & 0.028 & 0.757 & 0.051 \\ 
\model{LLaMA-3.2-90B-Instruct }   & 0.645 & 0.848 & 0.014 & 0.816 & 0.027 & 0.759 & 0.055 \\ 
\model{LLaMA-3.1-405B-Instruct }  & 0.683 & 0.859 & 0.011 & 0.828 & 0.024 & 0.770 & 0.050 \\ \midrule
\model{DK-R1-Dist-Qwen-1.5B }     & 0.353 & 0.772 & 0.009 & 0.745 & 0.028 & 0.702 & 0.043 \\ 
\model{DK-R1-Dist-Qwen-7B}        & 0.555 & 0.802 & 0.013 & 0.778 & 0.020 & 0.689 & 0.043 \\ 
\model{DK-R1-Dist-Qwen-14B}       & 0.698 & 0.815 & 0.012 & 0.792 & 0.014 & 0.681 & 0.033 \\ 
\model{DK-R1-Dist-Qwen-32B}       & 0.748 & 0.827 & 0.010 & 0.795 & 0.029 & 0.709 & 0.046 \\  \midrule
Weighted Average & --- & 0.827 & 0.010 & 0.798 & 0.022 & 0.731 & 0.042 \\
\bottomrule
\end{tabular}
}
\label{tab:predictability-results-without-ATUGVO}
\end{table*}

\subsubsection{Feature Grouping: Assessor with 11 Broad Dimensions}

Table \ref{tab:predictability-results-11_broad_dims} shows the predictive power of our demand-based assessor using the 11 broad dimensions listed in Table \ref{tab:dimensions} instead of the 19 specific dimensions. Overall, this is slightly worse than using all 19 demands in general, again confirming that  that one can predict model performance very well by aggregating some dimensions. Whether this compensates for robustness (having three scales for metacognition aggregated into one single scale) instead of one single scale for metacognition is something to explore in future work. 

\begin{table*}[!h]
\centering
\caption{Predictive power of the demand-based assessor by using the 11 broad dimensions for in-distribution, task out-of-distribution, and benchmark out-of-distribution. If a broad dimension has more than one subdimensions, we take the maximum value.}
\resizebox{0.87\textwidth}{!}{
\begin{tabular}{R{0.25\textwidth}C{0.17\textwidth}C{0.1\textwidth}C{0.1\textwidth}C{0.1\textwidth}C{0.1\textwidth}C{0.1\textwidth}C{0.1\textwidth}}
\toprule
\multirow{2}{*}{\textbf{Subject LLM}}
&  \multirow{2}{*}{\textbf{LLM Accuracy}$\uparrow$ }  & \multicolumn{2}{c}{\textbf{ID}} & \multicolumn{2}{c}{\textbf{Task OOD}} & \multicolumn{2}{c}{\textbf{Benchmark OOD}}\\  \cmidrule(l){3-4}  \cmidrule(l){5-6} \cmidrule(l){7-8}  
&  & \textbf{AUROC}$\uparrow$ & \textbf{ECE}$\downarrow$ & \textbf{AUROC}$\uparrow$ & \textbf{ECE}$\downarrow$ & \textbf{AUROC}$\uparrow$ & \textbf{ECE}$\downarrow$  \\  
\midrule
\model{Babbage-002 }              & 0.102 & 0.768 & 0.004 & 0.731 & 0.014 & 0.686 & 0.020 \\
\model{Davinci-002 }              & 0.157 & 0.763 & 0.007 & 0.727 & 0.011 & 0.712 & 0.021 \\
\model{GPT-35-Turbo }             & 0.414 & 0.801 & 0.007 & 0.784 & 0.021 & 0.768 & 0.032 \\
\model{GPT-4o   }                  & 0.713 & 0.870 & 0.013 & 0.839 & 0.025 & 0.816 & 0.048 \\
\model{OpenAI o1-mini  }                 & 0.770 & 0.845 & 0.013 & 0.821 & 0.020 & 0.745 & 0.029 \\
\model{OpenAI o1}               & 0.843 & 0.835 & 0.011 & 0.793 & 0.031 & 0.725 & 0.039 \\ \midrule
\model{LLaMA-3.2-1B-Instruct }              & 0.216 & 0.769 & 0.010 & 0.718 & 0.034 & 0.692 & 0.035 \\
\model{LLaMA-3.2-3B-Instruct  }             & 0.378 & 0.803 & 0.010 & 0.782 & 0.026 & 0.768 & 0.032 \\
\model{LLaMA-3.2-11B-Instruct }             & 0.463 & 0.808 & 0.008 & 0.786 & 0.033 & 0.773 & 0.045 \\
\model{LLaMA-3.2-90B-Instruct }             & 0.645 & 0.849 & 0.014 & 0.819 & 0.030 & 0.795 & 0.061 \\
\model{LLaMA-3.1-405B-Instruct }            & 0.683 & 0.858 & 0.011 & 0.830 & 0.029 & 0.810 & 0.049 \\ \midrule
\model{DK-R1-Dist-Qwen-1.5B}     & 0.353 & 0.768 & 0.007 & 0.746 & 0.018 & 0.691 & 0.053 \\
\model{DK-R1-Dist-Qwen-7B}        & 0.555 & 0.799 & 0.013 & 0.775 & 0.022 & 0.682 & 0.031 \\
\model{DK-R1-Dist-Qwen-14B}       & 0.698 & 0.815 & 0.012 & 0.793 & 0.024 & 0.701 & 0.038 \\
\model{DK-R1-Dist-Qwen-32B}       & 0.748 & 0.825 & 0.011 & 0.798 & 0.028 & 0.704 & 0.053 \\ \midrule
Weighted Average & --- & 0.825 & 0.011 & 0.797 & 0.026 & 0.745 & 0.042 \\
\bottomrule
\end{tabular}
}
\label{tab:predictability-results-11_broad_dims}
\end{table*}

\subsubsection{Demand-based Assessor with Logistic Regression}

Table \ref{tab:main-predictability-logistic-assessor}
shows the predictive power of an demand-based assessor based on logistic regression, using all the 19 demands, without any hyperparameters tuning. In comparison with our main demand-based assessor in Table \ref{tab:main-id-predictability-results}, this assessor is moderately worse (e.g. dropping from 0.881 to 0.852 in terms AUROC when predicting GPT-4o in the in-distribution setup), which is expected since this logistic regression model is expressed in the simplest form—using only raw features and does not capture any nonlinear (or interaction) terms into the feature set, leading its decision boundary to remain linear, unlike random forest.

\begin{table*}[!h]
\centering
\caption{Predictive power of a demand-based assessor trained with a basic logistic regression by using all the 19 specific dimensions for in-distribution, task out-of-distribution, and benchmark out-of-distribution.}
\resizebox{0.87\textwidth}{!}{
\begin{tabular}{R{0.25\textwidth}C{0.17\textwidth}C{0.1\textwidth}C{0.1\textwidth}C{0.1\textwidth}C{0.1\textwidth}C{0.1\textwidth}C{0.1\textwidth}}
\toprule
\multirow{2}{*}{\textbf{Subject LLM}}
&  \multirow{2}{*}{\textbf{LLM Accuracy}$\uparrow$ }  & \multicolumn{2}{c}{\textbf{ID}} & \multicolumn{2}{c}{\textbf{Task OOD}} & \multicolumn{2}{c}{\textbf{Benchmark OOD}}\\  \cmidrule(l){3-4}  \cmidrule(l){5-6} \cmidrule(l){7-8}  
&  & \textbf{AUROC}$\uparrow$ & \textbf{ECE}$\downarrow$ & \textbf{AUROC}$\uparrow$ & \textbf{ECE}$\downarrow$ & \textbf{AUROC}$\uparrow$ & \textbf{ECE}$\downarrow$  \\  
\midrule
\model{Babbage-002}                & 0.102 & 0.749 & 0.010 & 0.725 & 0.016 & 0.705 & 0.018 \\ 
\model{Davinci-002  }              & 0.157 & 0.740 & 0.005 & 0.717 & 0.014 & 0.709 & 0.023 \\ 
\model{GPT-3.5-Turbo  }            & 0.414 & 0.793 & 0.020 & 0.784 & 0.030 & 0.776 & 0.032 \\ 
\model{GPT-4o    }                 & 0.713 & 0.852 & 0.018 & 0.833 & 0.033 & 0.814 & 0.032 \\ 
\model{OpenAI o1-mini}                    & 0.770 & 0.820 & 0.013 & 0.798 & 0.017 & 0.695 & 0.069 \\ 
\model{OpenAI o1  }                       & 0.843 & 0.796 & 0.020 & 0.760 & 0.029 & 0.607 & 0.081 \\ \midrule
\model{LLaMA-3.2-1B-Instruct}      & 0.216 & 0.748 & 0.016 & 0.722 & 0.039 & 0.716 & 0.041 \\ 
\model{LLaMA-3.2-3B-Instruct  }    & 0.378 & 0.788 & 0.025 & 0.778 & 0.033 & 0.768 & 0.031 \\ 
\model{LLaMA-3.2-11B-Instruction}  & 0.463 & 0.799 & 0.026 & 0.789 & 0.038 & 0.782 & 0.036 \\ 
\model{LLaMA-3.2-90B-Instruct }    & 0.645 & 0.828 & 0.022 & 0.808 & 0.035 & 0.788 & 0.033 \\ 
\model{LLaMA-3.1-405B-Instruct}    & 0.683 & 0.840 & 0.018 & 0.823 & 0.031 & 0.801 & 0.032 \\ \midrule
\model{DK-R1-Dist-Qwen-1.5B }     & 0.353 & 0.746 & 0.019 & 0.733 & 0.029 & 0.694 & 0.039 \\ 
\model{DK-R1-Dist-Qwen-7B  }       & 0.555 & 0.767 & 0.020 & 0.747 & 0.031 & 0.640 & 0.087 \\ 
\model{DK-R1-Dist-Qwen-14B }       & 0.698 & 0.781 & 0.025 & 0.761 & 0.043 & 0.612 & 0.114 \\ 
\model{DK-R1-Dist-Qwen-32B  }      & 0.748 & 0.797 & 0.026 & 0.776 & 0.042 & 0.628 & 0.099 \\ \midrule
Weighted Average & --- & 0.801 & 0.020 & 0.781 & 0.032 & 0.710 & 0.059 \\
\bottomrule
\end{tabular}
}
\label{tab:main-predictability-logistic-assessor}
\end{table*}

\subsubsection{A Universal Assessor}

Table \ref{tab:predictability-results-universal-assessor} shows the predictive power of a demand-based assessor trained with all the 19 dimensions plus one extra variable with the identifier of the LLM subjects. Even if the dataset is 15 times larger and it can smooth the aleatoric uncertainty specific to some models, the results are very similar to the standard configuration with one assessor per model. Things could improve with a characterisation of the LLM or some other techniques, but we leave this as future work. 

\begin{table*}[!h]
\centering
\caption{Predictive power of the demand-based assessor by using only \dmc{MCu} (\dmc{Calibrating Known Unknowns}), \dmc{UG} (\dmc{Unguessability}), \dmc{KNf} (\dmc{Knowledge of Formal Sciences}), \dmc{CL} (\dmc{Conceptualisation, Learning and Abstraction}), \dmc{QLl} (\dmc{Logical Reasoning}), and \dmc{MCr} (\dmc{Identifying Relevant Information}) for in-distribution, task out-of-distribution, and benchmark out-of-distribution.}
\resizebox{0.87\textwidth}{!}{
\begin{tabular}{R{0.25\textwidth}C{0.17\textwidth}C{0.1\textwidth}C{0.1\textwidth}C{0.1\textwidth}C{0.1\textwidth}C{0.1\textwidth}C{0.1\textwidth}}
\toprule
\multirow{2}{*}{\textbf{Subject LLM}}
&  \multirow{2}{*}{\textbf{LLM Accuracy}$\uparrow$ }  & \multicolumn{2}{c}{\textbf{ID}} & \multicolumn{2}{c}{\textbf{Task OOD}} & \multicolumn{2}{c}{\textbf{Benchmark OOD}}\\  \cmidrule(l){3-4}  \cmidrule(l){5-6} \cmidrule(l){7-8}  
&  & \textbf{AUROC}$\uparrow$ & \textbf{ECE}$\downarrow$ & \textbf{AUROC}$\uparrow$ & \textbf{ECE}$\downarrow$ & \textbf{AUROC}$\uparrow$ & \textbf{ECE}$\downarrow$  \\  
\midrule
\model{Babbage-002 }              & 0.102 & 0.785 & 0.013 & 0.750 & 0.015 & 0.715 & 0.031 \\ 
\model{Davinci-002 }              & 0.157 & 0.772 & 0.013 & 0.737 & 0.020 & 0.698 & 0.032 \\ 
\model{GPT-3.5-turbo}             & 0.414 & 0.812 & 0.019 & 0.796 & 0.027 & 0.774 & 0.044 \\ 
\model{GPT-4o   }                 & 0.713 & 0.883 & 0.019 & 0.857 & 0.027 & 0.799 & 0.045 \\ 
\model{OpenAI o1-mini}            & 0.770 & 0.860 & 0.018 & 0.839 & 0.029 & 0.720 & 0.037 \\ 
\model{OpenAI o1}                 & 0.843 & 0.851 & 0.018 & 0.810 & 0.040 & 0.671 & 0.043 \\ \midrule
\model{LLaMA-3.2-1B-Instruct}     & 0.216 & 0.786 & 0.018 & 0.732 & 0.036 & 0.714 & 0.058 \\ 
\model{LLaMA-3.2-3B-Instruct }    & 0.378 & 0.813 & 0.021 & 0.790 & 0.032 & 0.774 & 0.043 \\ 
\model{LLaMA-3.2-11B-Instruct}    & 0.463 & 0.820 & 0.018 & 0.802 & 0.021 & 0.781 & 0.055 \\ 
\model{LLaMA-3.2-90B-Instruct }   & 0.645 & 0.861 & 0.025 & 0.837 & 0.025 & 0.781 & 0.060 \\ 
\model{LLaMA-3.1-405B-Instruct }  & 0.683 & 0.870 & 0.020 & 0.846 & 0.029 & 0.792 & 0.056 \\ \midrule
\model{DK-R1-Dist-Qwen-1.5B }     & 0.353 & 0.783 & 0.030 & 0.755 & 0.037 & 0.707 & 0.069 \\ 
\model{DK-R1-Dist-Qwen-7B}        & 0.555 & 0.812 & 0.027 & 0.785 & 0.023 & 0.691 & 0.028 \\ 
\model{DK-R1-Dist-Qwen-14B}       & 0.698 & 0.828 & 0.020 & 0.808 & 0.025 & 0.693 & 0.061 \\ 
\model{DK-R1-Dist-Qwen-32B}       & 0.748 & 0.841 & 0.019 & 0.815 & 0.031 & 0.701 & 0.057 \\ \midrule
Weighted Average & --- & 0.838 & 0.020 & 0.812 & 0.029 & 0.735 & 0.049 \\
\bottomrule
\end{tabular}
}
\label{tab:predictability-results-universal-assessor}
\end{table*}

\subsubsection{Algebraic Assessor}
\label{ap:algebraic_assessor}

Comparing the demand profile of a task instance and the ability profile of an AI system gives immediate insights on the expectation of success. Actually, the ability comes from the interpretation of a characteristic curve, which plots probability of success as a function of the demand level. Can we use this without the need of training any model? This is what we refer to as an algebraic or formulaic assessor. There are many possible ways of doing this, but the main issue is to reconstruct the multidimensional characteristic surface out of the unidimensional curves.

We will illustrate one approach that is straightforward and hence interpretable, but many other options could be explored. 
To build this algebraic assessor, we first compute the  differences between the subject's ability, $a_i$, and the task demand, $d_i$ for all 18 demands. While we could apply the original two-parametric logistic function (slope and position), for simplicity we assume we only have access to the ability (the area or position of the curve) and we apply the standard logistic function.  This gives us a univariate probability per each demand, i.e., we apply  
\[
\sigma(x)=\frac{1}{1+\exp(-x)},
\]
to each $a_i-d_i$, which maps the values into $[0,1]$. In addition, we add \dm{UG} (\dm{Unguessability}), which is normalised as $(100-UG)/100$, ensuring that it is also within $[0,1]$. The prediction of LLM success is computed by combining these 19 values using a generalised mean:

\begin{equation}\label{eq:algebraic}
S = \left(\frac{1}{19}\left(\sum_{i=1}^{18} [\sigma(a_i-d_i)]^r + \left(\frac{100-UG}{100}\right)^r\right)\right)^{1/r},
\end{equation}
where $r$ is a parameter that controls the type of mean used. Extreme negative values of $r$ tend to approximate the minimum of all components, while extreme positive values tend to approximate the maximum of all components. The predictive power of this algebraic assessor is shown in Table \ref{tab:predictability-results-algebraic-assessor}, using different values of $r$, including $r=0$ (geometric mean), $r=0.25$ and $r=1$ (arithmetic mean). Overall, it seems that a value of $r=0.25$ is the optimal, reaching satisfactory AUROC scores, especially for large models. The advantage of $r=0.25$ and $r=0$ (geometric mean) over arithmetic mean ($r=1$) is sensible because the most informative probability scores in the generalised mean equation come from those dimensions $i$ with more negative values in $a_i-d_i$ (e.g., if $a_i$ is much smaller than $d_i$ for one or a few dimensions, it is highly likely that the model will fail, even if the model's abilities on other dimensions are well above the demands). When comparing $r=0.25$ and $r=0$, we see that the geometric mean ($r=0$) is also fairly good as it is often the best in terms of calibration, though clearly worse in AUROC than $r=0.25$, indicating that the geometric mean (or a value of $r$ below 0) may over-focus on one or very few dimensions that exhibit the most negative values in $a_i-d_i$. This also has an interpretation in terms of compensatoriness of the dimensions in the sense of multidimensional IRT (see related work, section \ref{sec:relatedwork}).

\begin{table*}[!h]
\centering
\caption{Predictive power of the demand-based {\em algebraic} assessor for different values of \(r\). Bold numbers indicate the best value for each LLM (largest AUROC and lowest ECE) across the three \(r\) settings.}
\resizebox{0.87\textwidth}{!}{
\begin{tabular}{R{0.25\textwidth}C{0.17\textwidth}C{0.1\textwidth}C{0.1\textwidth}C{0.1\textwidth}C{0.1\textwidth}C{0.1\textwidth}C{0.1\textwidth}}
\toprule
\multirow{2}{*}{\textbf{Subject LLM}} & \multirow{2}{*}{\textbf{LLM Accuracy}$\uparrow$} & \multicolumn{2}{c}{\textbf{\(r=0\)}} & \multicolumn{2}{c}{\textbf{\(r=0.25\)}} & \multicolumn{2}{c}{\textbf{\(r=1\)}} \\ 
\cmidrule(lr){3-4} \cmidrule(lr){5-6} \cmidrule(lr){7-8}
&  & \textbf{AUROC}$\uparrow$ & \textbf{ECE}$\downarrow$ & \textbf{AUROC}$\uparrow$ & \textbf{ECE}$\downarrow$ & \textbf{AUROC}$\uparrow$ & \textbf{ECE}$\downarrow$ \\  
\midrule
Babbage-002              & 0.102 & 0.607 & \textbf{0.068} & \textbf{0.614} & 0.138 & 0.581 & 0.230 \\
Davinci-002              & 0.157 & 0.590 & \textbf{0.105} & \textbf{0.652} & 0.135 & 0.638 & 0.226 \\
GPT-3.5-Turbo            & 0.414 & 0.744 & \textbf{0.098} & \textbf{0.771} & 0.122 & 0.752 & 0.203 \\
GPT-4o                   & 0.713 & 0.812 & 0.238 & \textbf{0.826} & 0.129 & 0.803 & \textbf{0.105} \\
OpenAI o1-mini           & 0.770 & 0.739 & 0.267 & \textbf{0.749} & \textbf{0.065} & 0.746 & 0.079 \\
OpenAI o1                & 0.843 & 0.733 & 0.312 & \textbf{0.734} & 0.071 & 0.706 & \textbf{0.040} \\ \midrule
LLaMA-3.2-1B-Instruct    & 0.216 & 0.693 & \textbf{0.111} & \textbf{0.715} & 0.177 & 0.696 & 0.266 \\
LLaMA-3.2-3B-Instruct    & 0.378 & 0.739 & \textbf{0.109} & \textbf{0.775} & 0.152 & 0.756 & 0.235 \\
LLaMA-3.2-11B-Instruct   & 0.463 & 0.749 & \textbf{0.105} & \textbf{0.780} & 0.112 & 0.762 & 0.193 \\
LLaMA-3.2-90B-Instruct   & 0.645 & 0.791 & 0.188 & \textbf{0.805} & 0.127 & 0.780 & \textbf{0.124} \\
LLaMA-3.1-405B-Instruct  & 0.683 & 0.803 & 0.216 & \textbf{0.817} & 0.128 & 0.793 & \textbf{0.104} \\ \midrule
DK-R1-Dist-Qwen-1.5B     & 0.353 & 0.655 & \textbf{0.179} & \textbf{0.697} & 0.199 & 0.691 & 0.283 \\
DK-R1-Dist-Qwen-7B       & 0.555 & 0.665 & 0.210 & 0.699 & \textbf{0.122} & \textbf{0.702} & 0.203 \\
DK-R1-Dist-Qwen-14B      & 0.698 & 0.720 & 0.213 & \textbf{0.733} & \textbf{0.050} & 0.730 & 0.114 \\
DK-R1-Dist-Qwen-32B      & 0.748 & 0.739 & 0.247 & \textbf{0.747} & \textbf{0.061} & 0.734 & 0.088 \\
\midrule
Weighted Average & --- & 0.738 & 0.206 & 0.757 & 0.106 & 0.741 & 0.137 \\
\bottomrule
\end{tabular}
}
\label{tab:predictability-results-algebraic-assessor}
\end{table*}

In general, this algebraic assessor, especially because the results are good for $r=0$,  suggests a simple mechanism to go from unidimensional abilities and demands to an integrated prediction that is based on all the multidimensional information of an instance: instance performance can be seen as the geometric mean of the expected performance for each dimension. Of course, using some trained models, results can be better (especially in calibration) but Eq.  \ref{eq:algebraic} is a very formulaic, interpretable way of understanding how abilities and demands affect performance.

\newpage

\subsection{SCCs for all models}\label{sec:sccs}

\begin{figure}[!ht]
  \centering
  \includegraphics[width=1\linewidth]{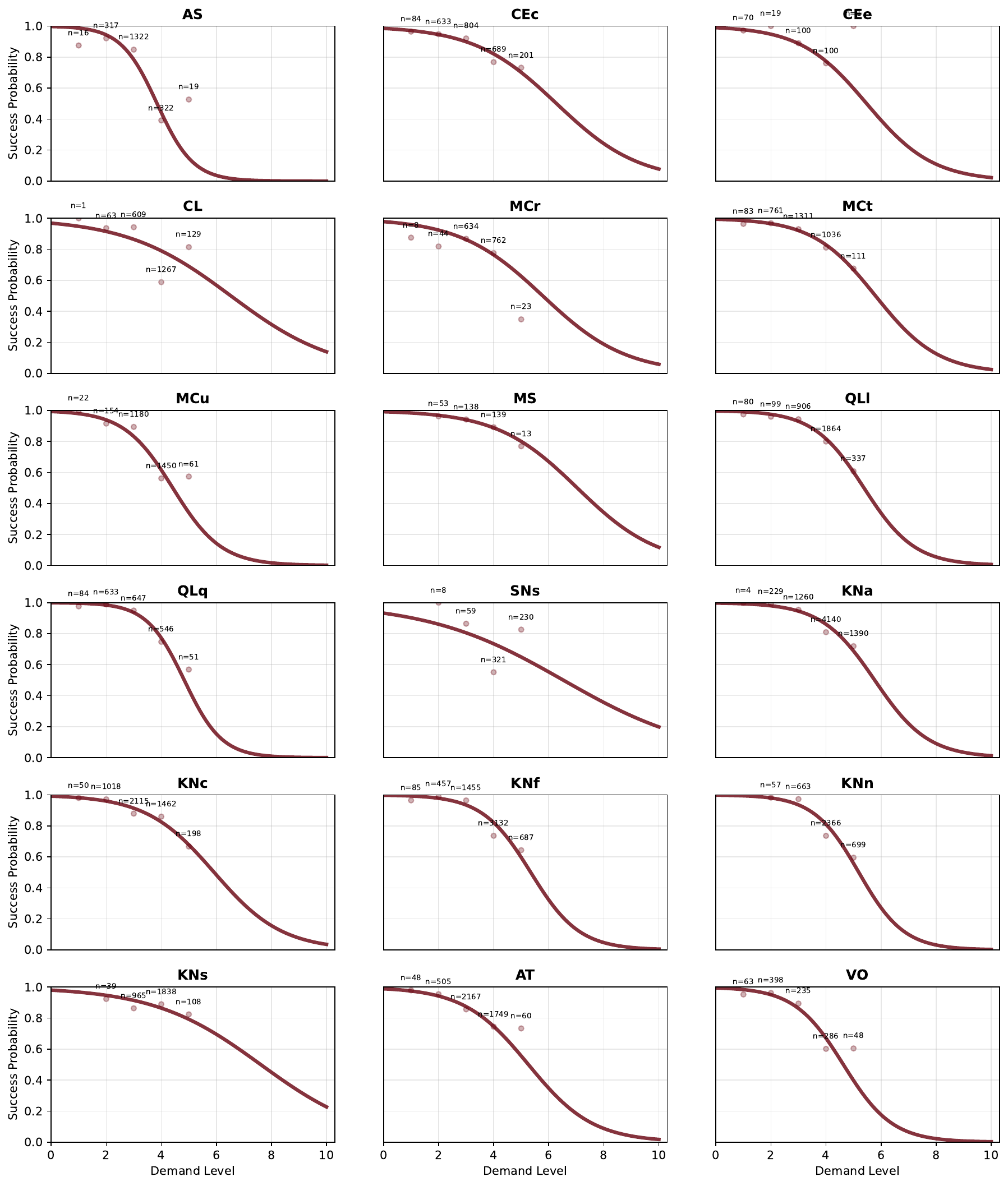}
  \caption{Characteristic curves for the 18 demands for OpenAI's o1 (all other things equal to Figure \ref{fig:characteristic_curves_combined}).}
  \label{fig:characteristic_curves_o1}
\end{figure}

\begin{figure}[!ht]
  \centering
  \includegraphics[width=1\linewidth]{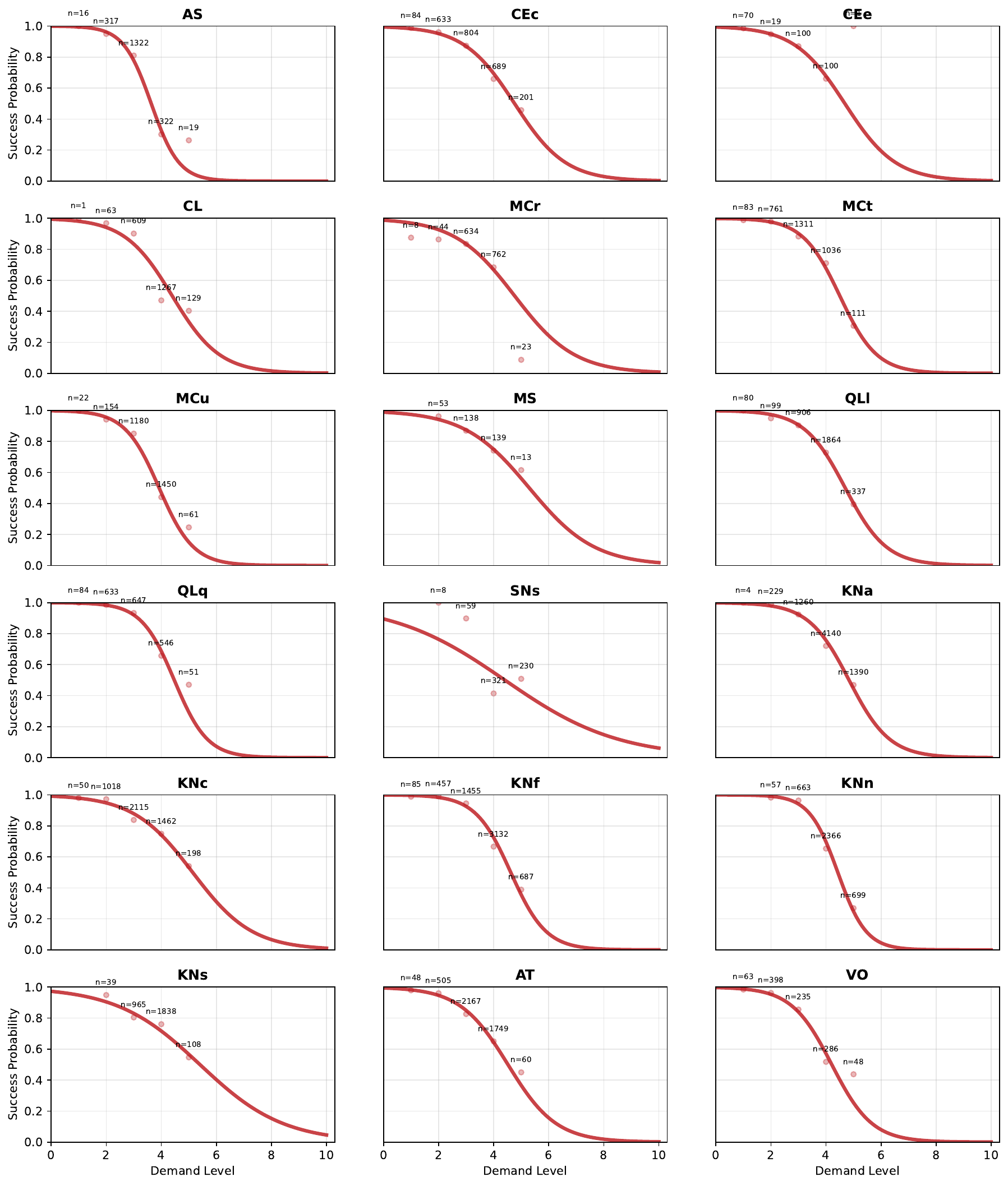}
  \caption{Characteristic curves for the 18 demands for OpenAI's o1-mini (all other things equal to Figure \ref{fig:characteristic_curves_combined}).}
  \label{fig:characteristic_curves_o1_mini}
\end{figure}

\begin{figure}[!ht]
  \centering
  \includegraphics[width=1\linewidth]{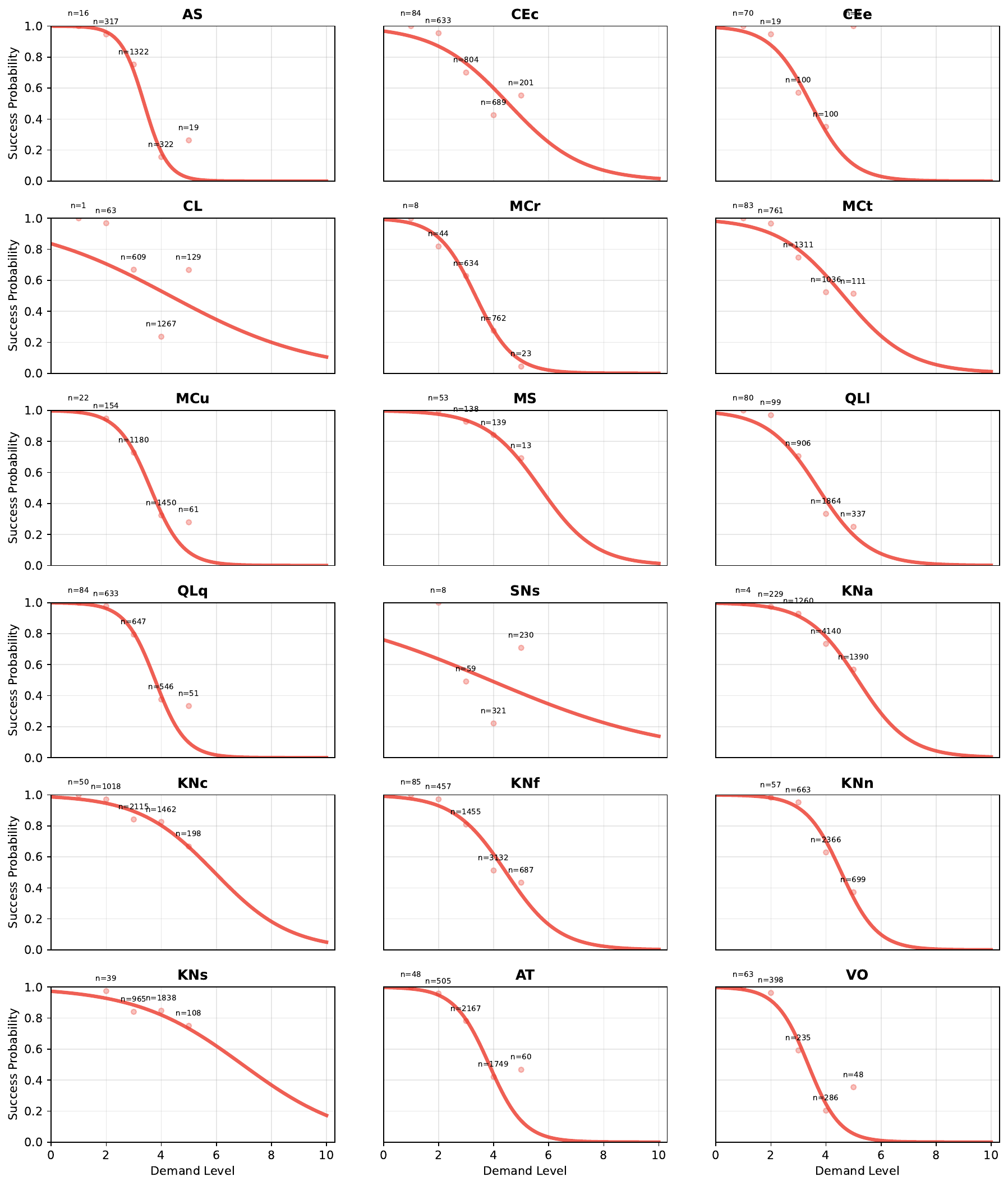}
  \caption{Characteristic curves for the 18 demands for OpenAI's GPT-4o (all other things equal to Figure \ref{fig:characteristic_curves_combined}).}
  \label{fig:characteristic_curves_gpt4o}
\end{figure}

\begin{figure}[!ht]
  \centering
  \includegraphics[width=1\linewidth]{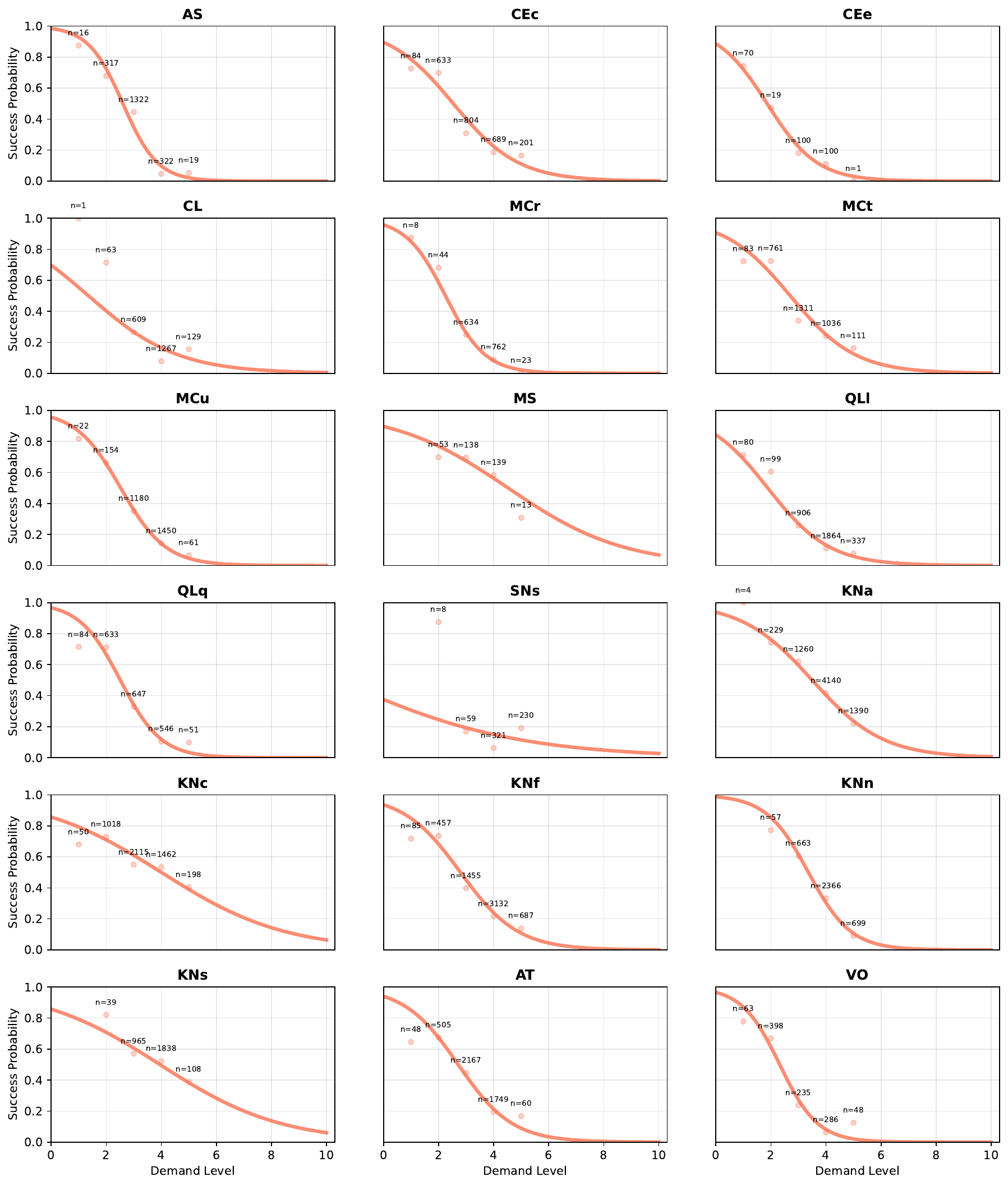}
  \caption{Characteristic curves for the 18 demands for OpenAI's GPT-3.5-Turbo (all other things equal to Figure \ref{fig:characteristic_curves_combined}).}
  \label{fig:characteristic_curves_gpt3.5}
\end{figure}

\begin{figure}[!ht]
  \centering
  \includegraphics[width=1\linewidth]{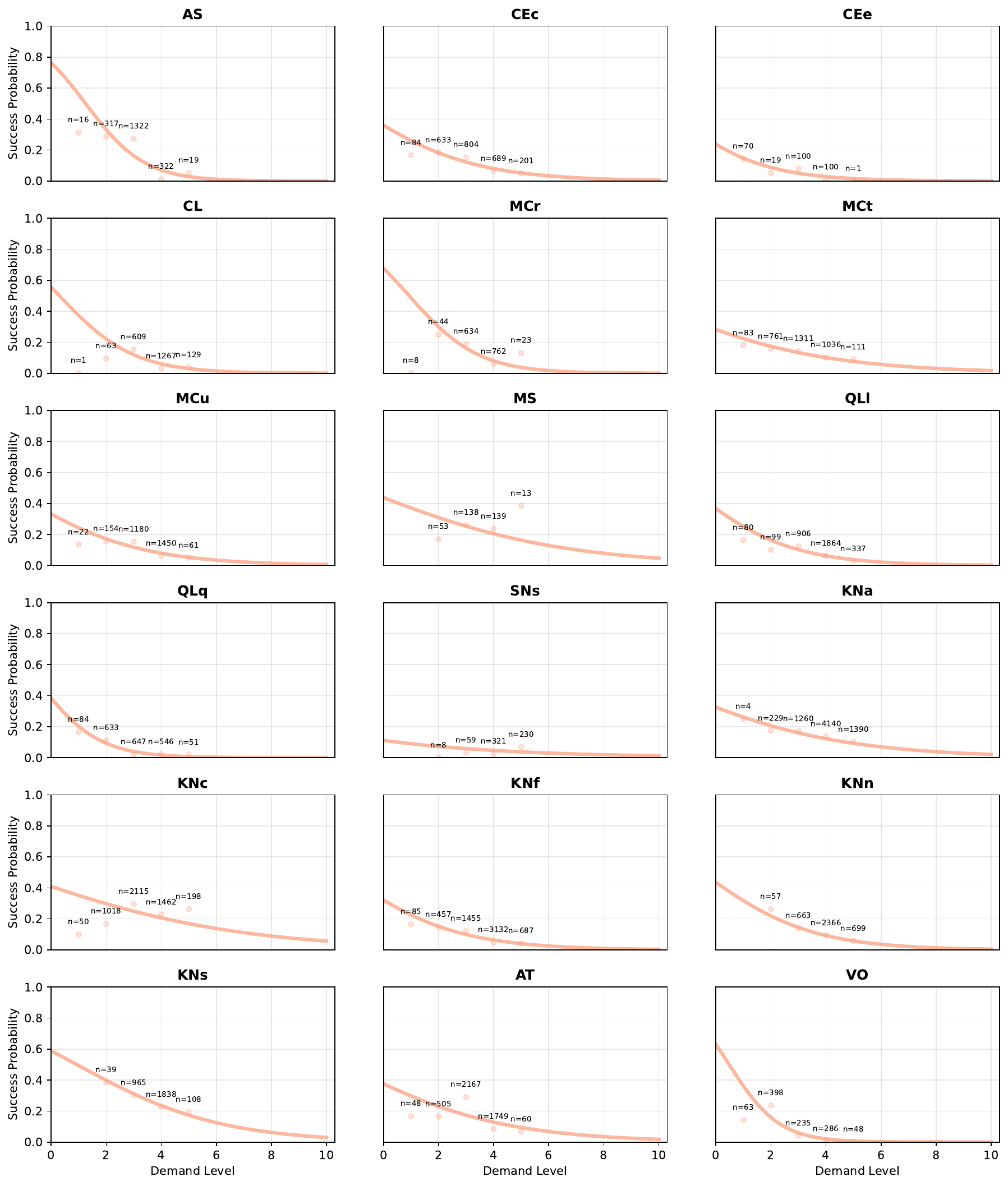}
  \caption{Characteristic curves for the 18 demands for OpenAI's Davinci-002 (all other things equal to Figure \ref{fig:characteristic_curves_combined}).}
  \label{fig:characteristic_curves_davinci}
\end{figure}

\begin{figure}[!ht]
  \centering
  \includegraphics[width=1\linewidth]{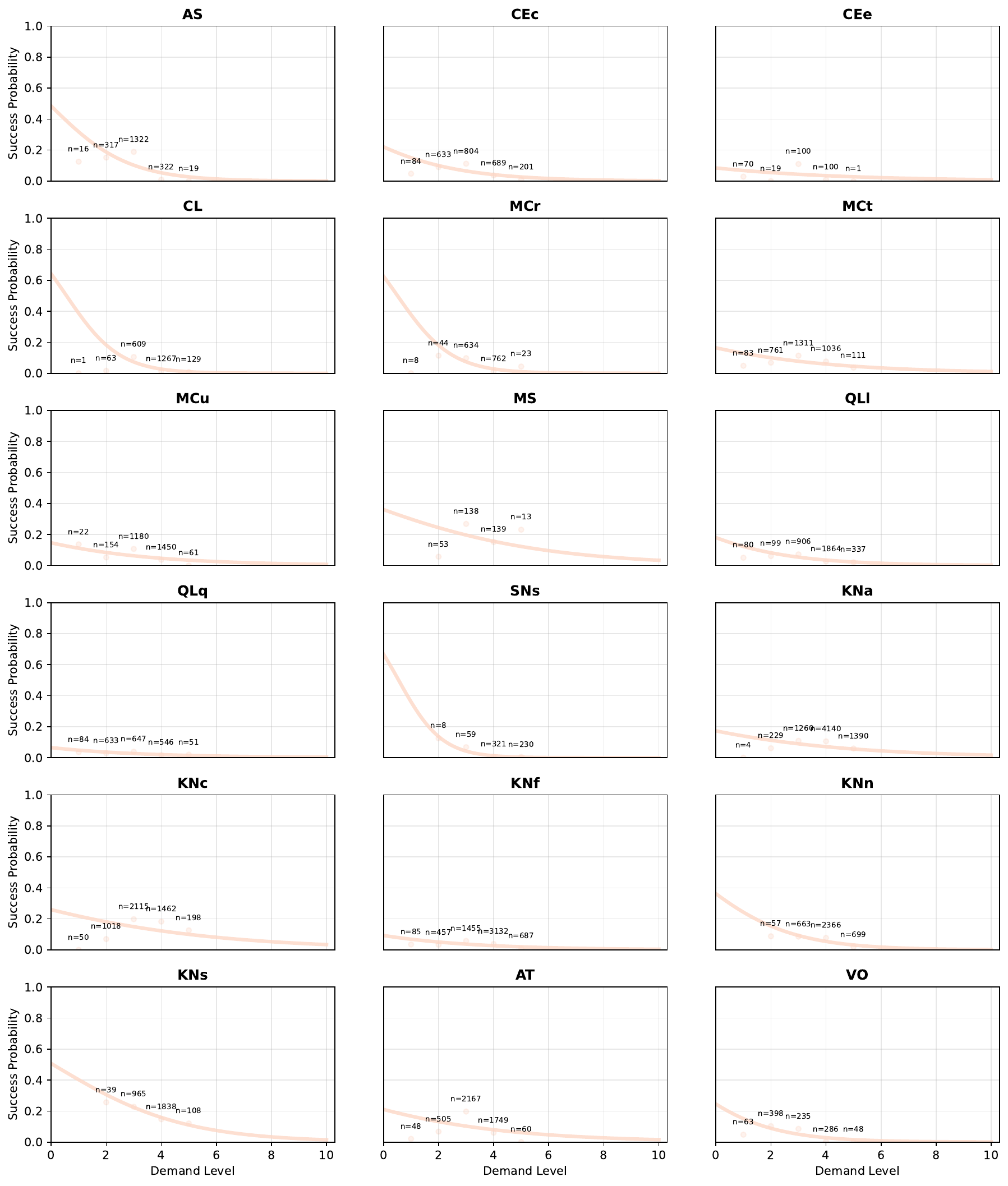}
  \caption{Characteristic curves for the 18 demands for OpenAI's Babbage-002 (all other things equal to Figure \ref{fig:characteristic_curves_combined}).}
  \label{fig:characteristic_curves_babbage}
\end{figure}

\begin{figure}[!ht]
  \centering
  \includegraphics[width=1\linewidth]{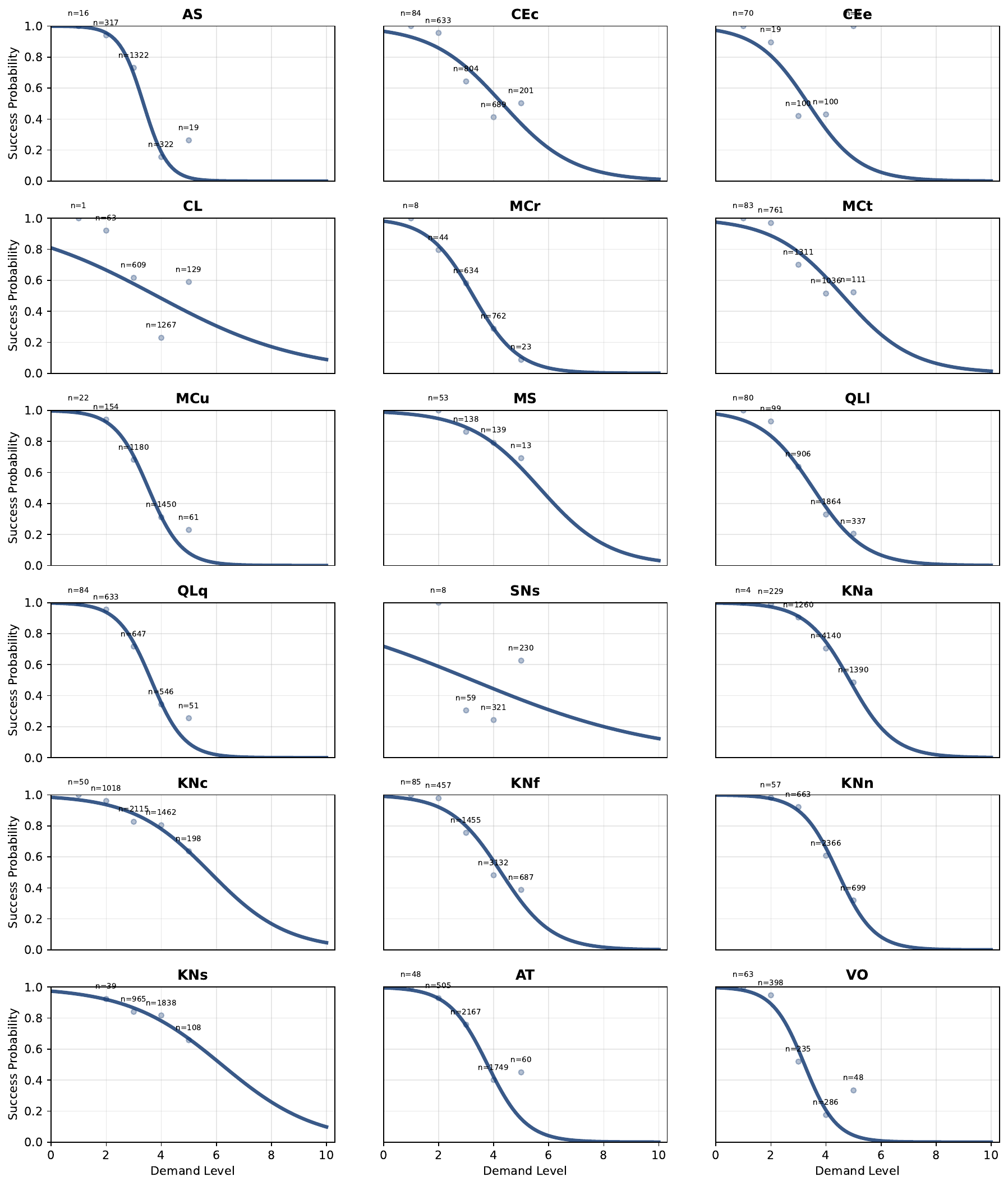}
  \caption{Characteristic curves for the 18 demands for LLaMa 3.1-405B-Instruct (all other things equal to Figure \ref{fig:characteristic_curves_combined}).}
  \label{fig:characteristic_curves_llama3d1_405b}
\end{figure}

\begin{figure}[!ht]
  \centering
  \includegraphics[width=1\linewidth]{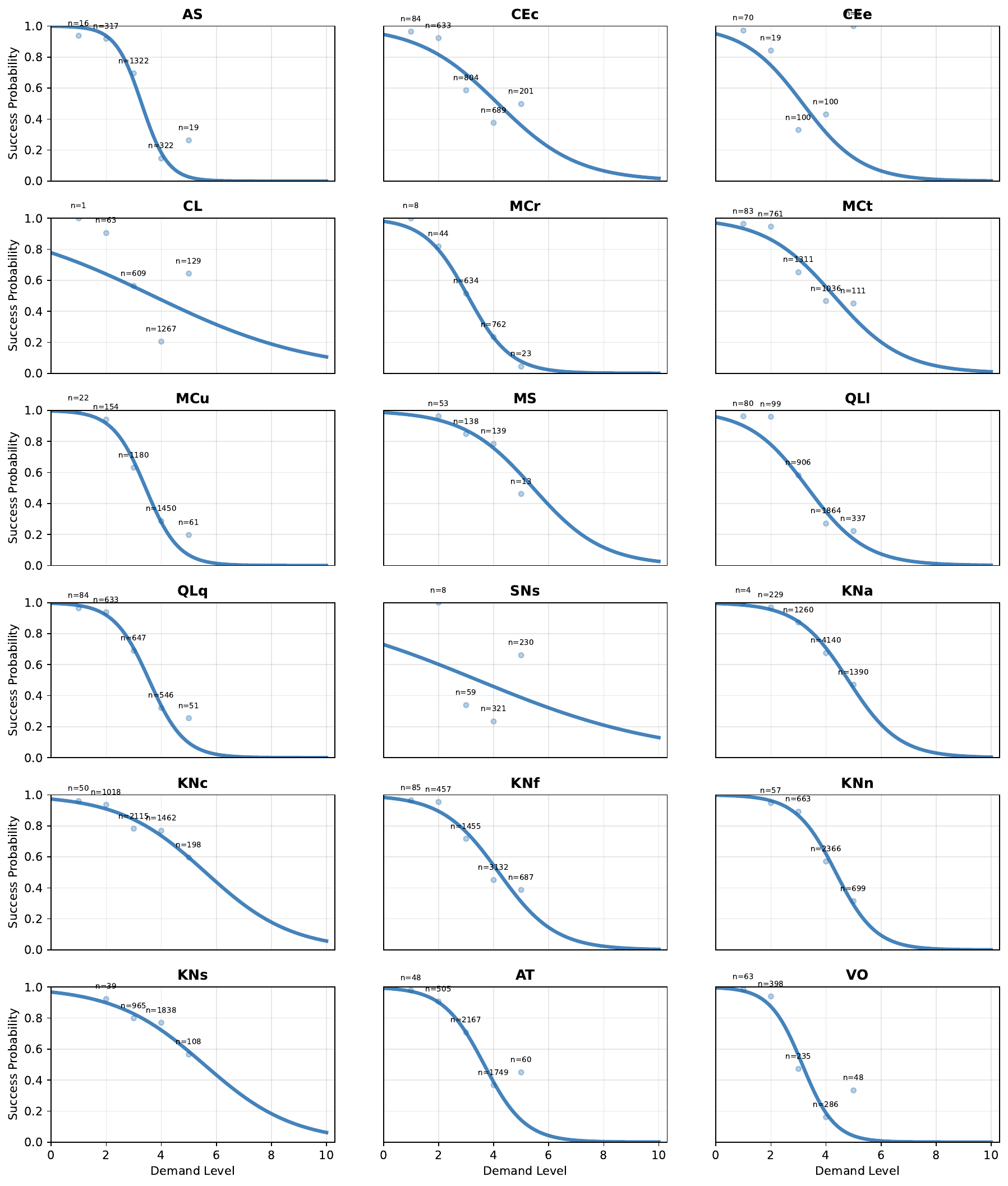}
  \caption{Characteristic curves for the 18 demands for LLaMa 3.2-90B-Instruct (all other things equal to Figure \ref{fig:characteristic_curves_combined}).}
  \label{fig:characteristic_curves_llama3d2_90b}
\end{figure}

\begin{figure}[!ht]
  \centering
  \includegraphics[width=1\linewidth]{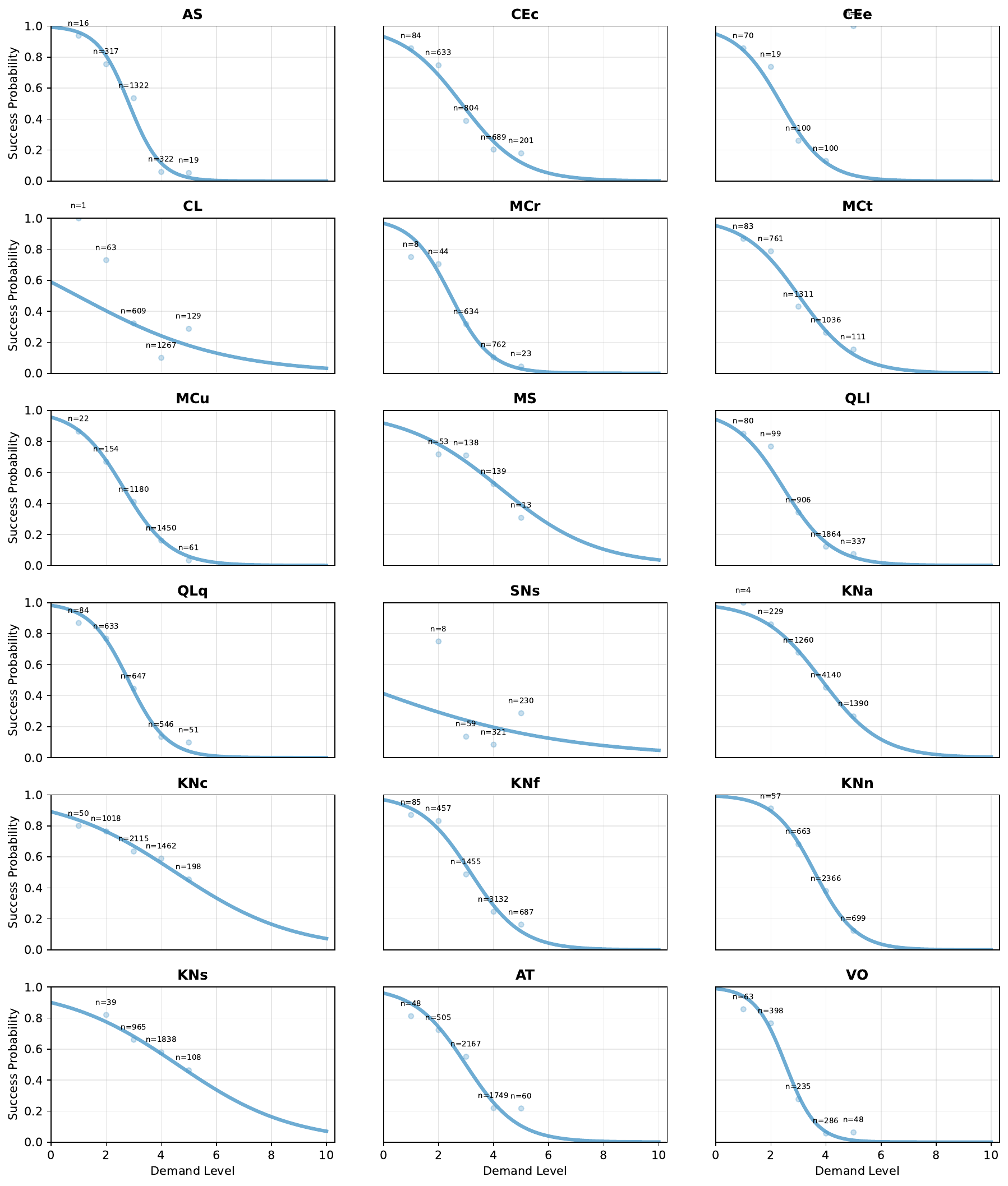}
  \caption{Characteristic curves for the 18 demands for LLaMa 3.2-11B-Instruct (all other things equal to Figure \ref{fig:characteristic_curves_combined}).}
  \label{fig:characteristic_curves_llama3d2_11b}
\end{figure}

\begin{figure}[!ht]
  \centering
  \includegraphics[width=1\linewidth]{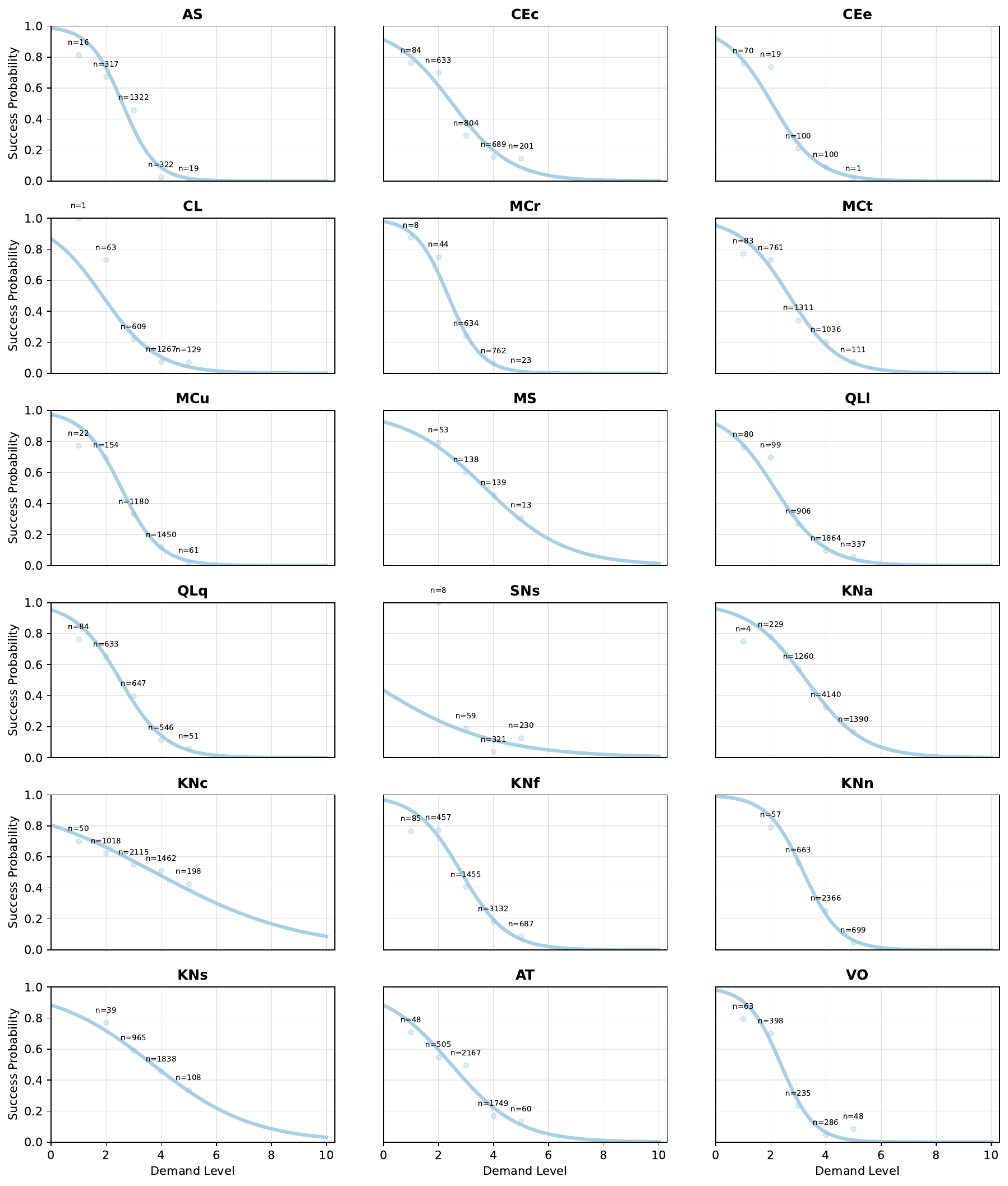}
  \caption{Characteristic curves for the 18 demands for LLaMa 3.2-3B-Instruct (all other things equal to Figure \ref{fig:characteristic_curves_combined}).}
  \label{fig:characteristic_curves_llama3d2_3b}
\end{figure}

\begin{figure}[!ht]
  \centering
  \includegraphics[width=1\linewidth]{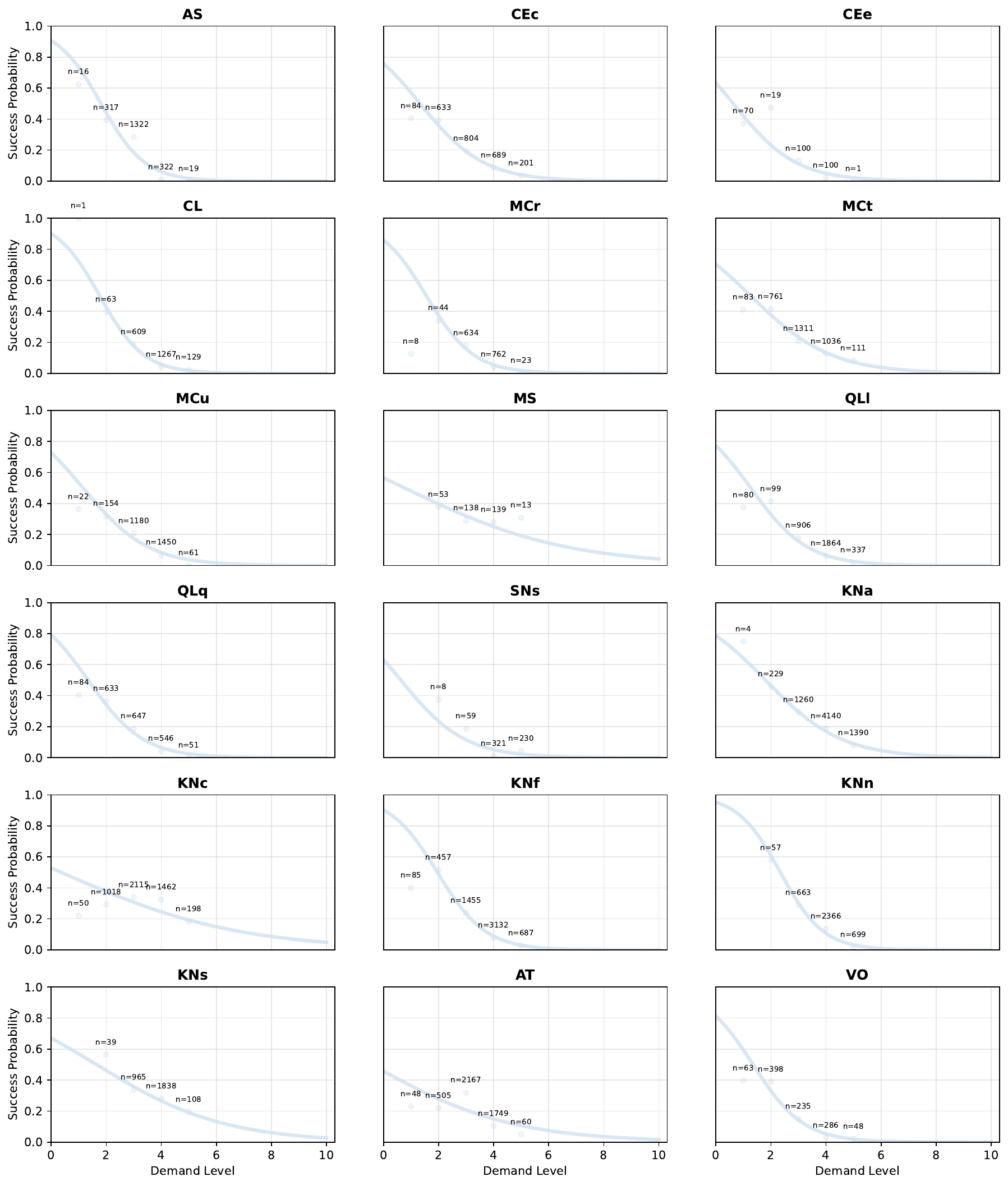}
  \caption{Characteristic curves for the 18 demands for LLaMa 3.2-1B-Instruct (all other things equal to Figure \ref{fig:characteristic_curves_combined}).}
  \label{fig:characteristic_curves_llama3d2_1b}
\end{figure}

\begin{figure}[!ht]
  \centering
  \includegraphics[width=1\linewidth]{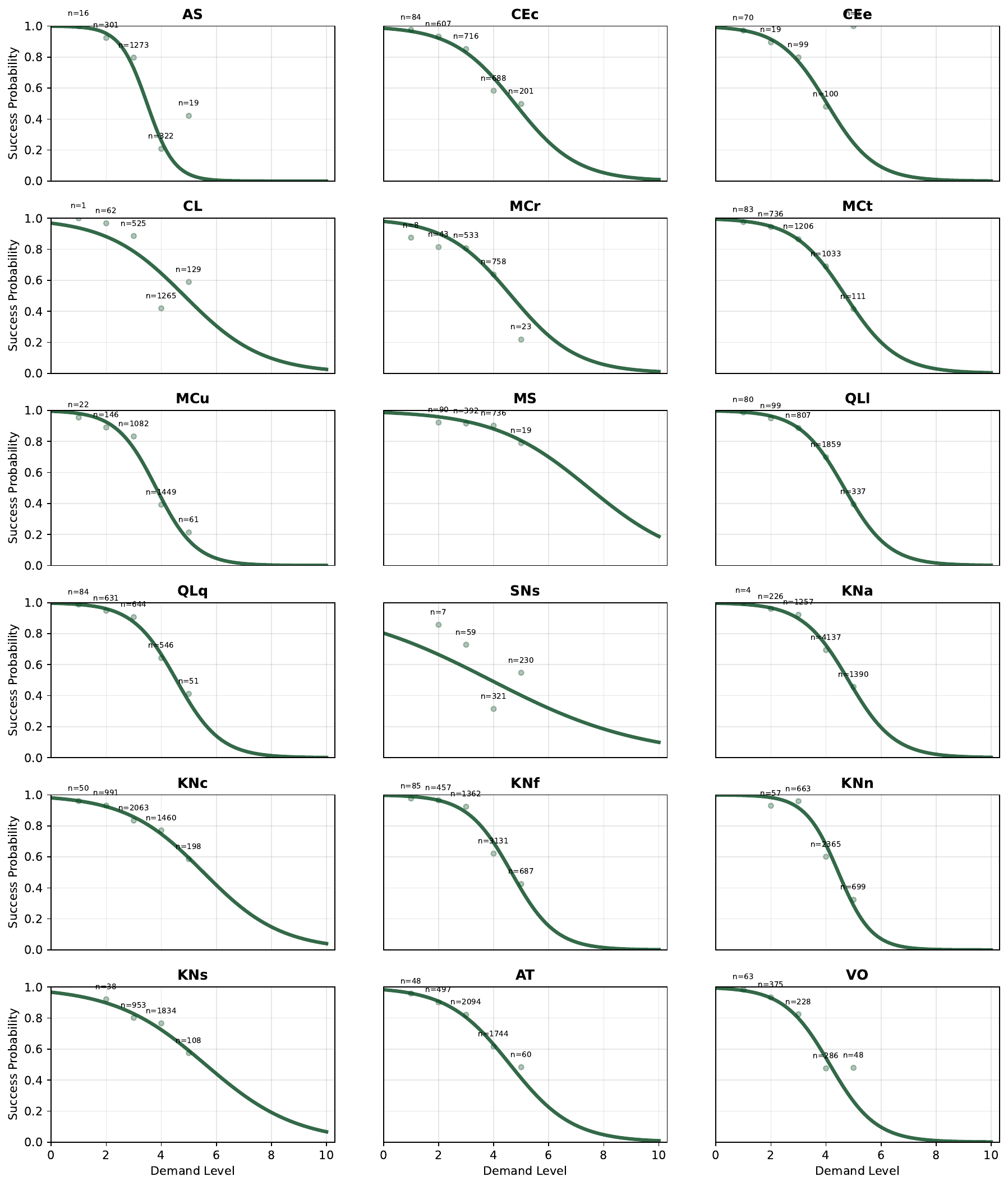}
  \caption{Characteristic curves for the 18 demands for DeepSeek's R1-Dist-Qwen-32B (all other things equal to Figure \ref{fig:characteristic_curves_combined}).}
  \label{fig:characteristic_curves_DK-R1-Dist-Qwen-32B}
\end{figure}

\begin{figure}[!ht]
  \centering
  \includegraphics[width=1\linewidth]{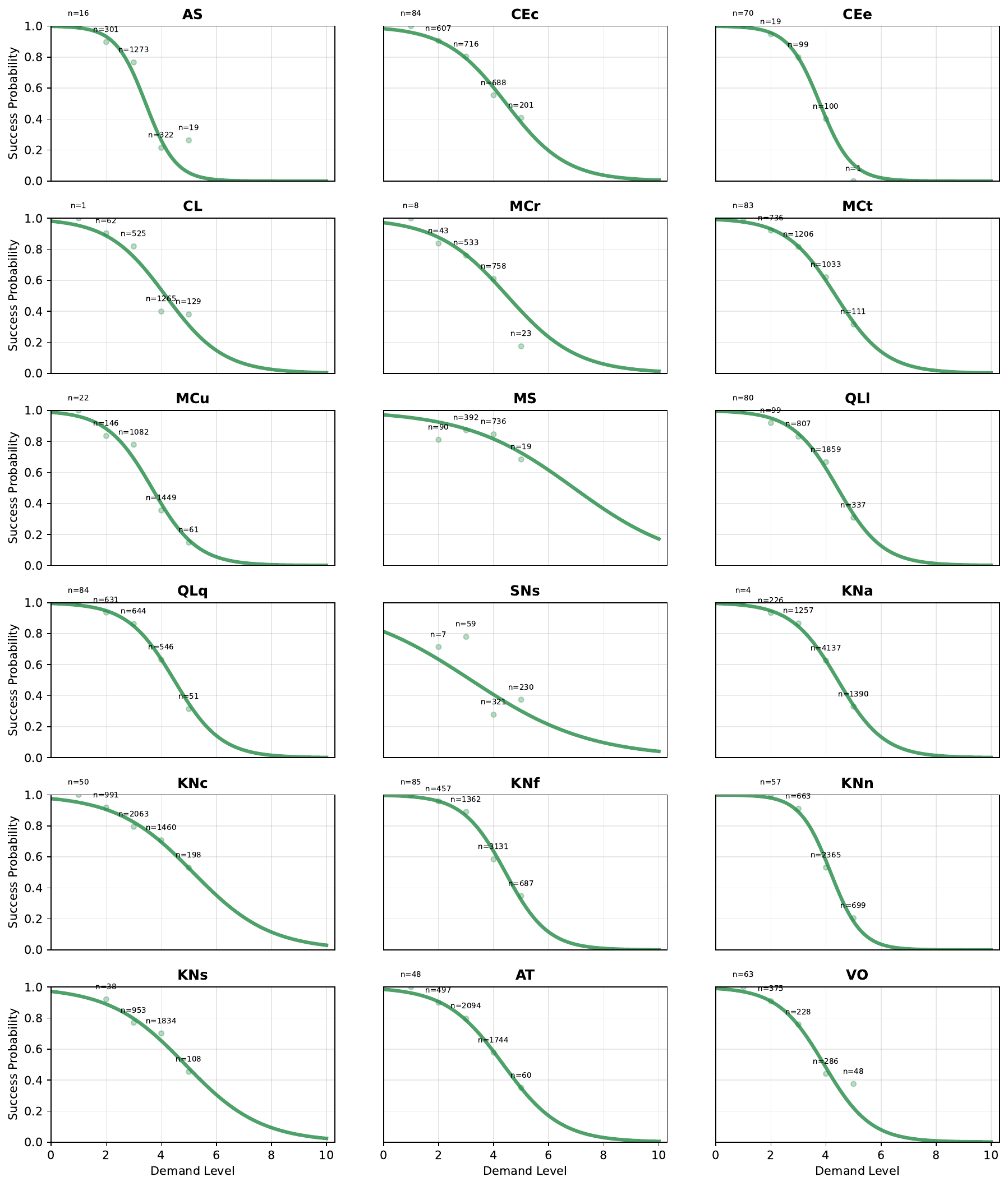}
  \caption{Characteristic curves for the 18 demands for DeepSeek's R1-Dist-Qwen-14B (all other things equal to Figure \ref{fig:characteristic_curves_combined}).}
  \label{fig:characteristic_curves_DK-R1-Dist-Qwen-14B}
\end{figure}

\begin{figure}[!ht]
  \centering
  \includegraphics[width=1\linewidth]{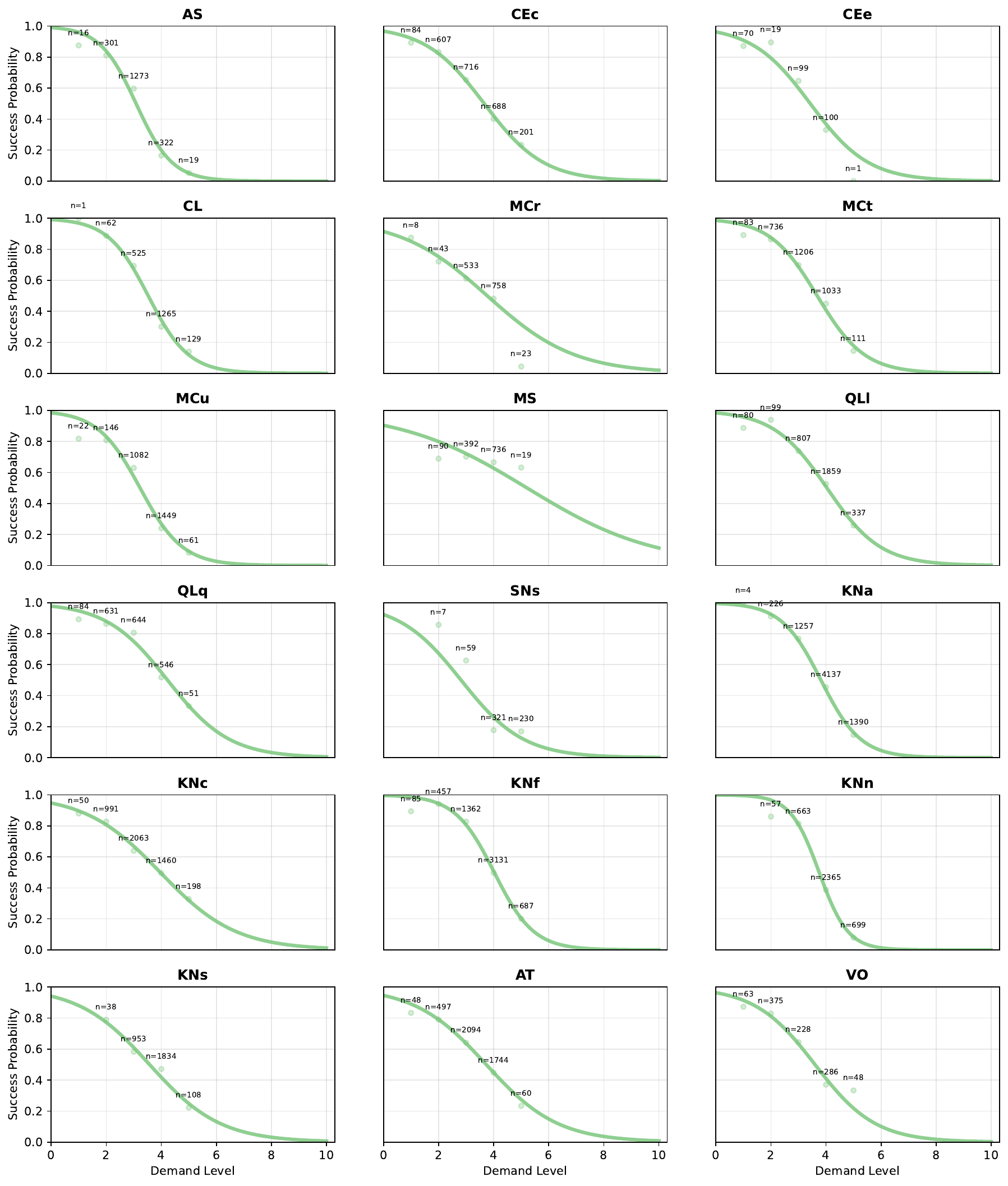}
  \caption{Characteristic curves for the 18 demands for DeepSeek's R1-Dist-Qwen-7B (all other things equal to Figure \ref{fig:characteristic_curves_combined}).}
  \label{fig:characteristic_curves_DK-R1-Dist-Qwen-7B}
\end{figure}

\begin{figure}[!ht]
  \centering
  \includegraphics[width=1\linewidth]{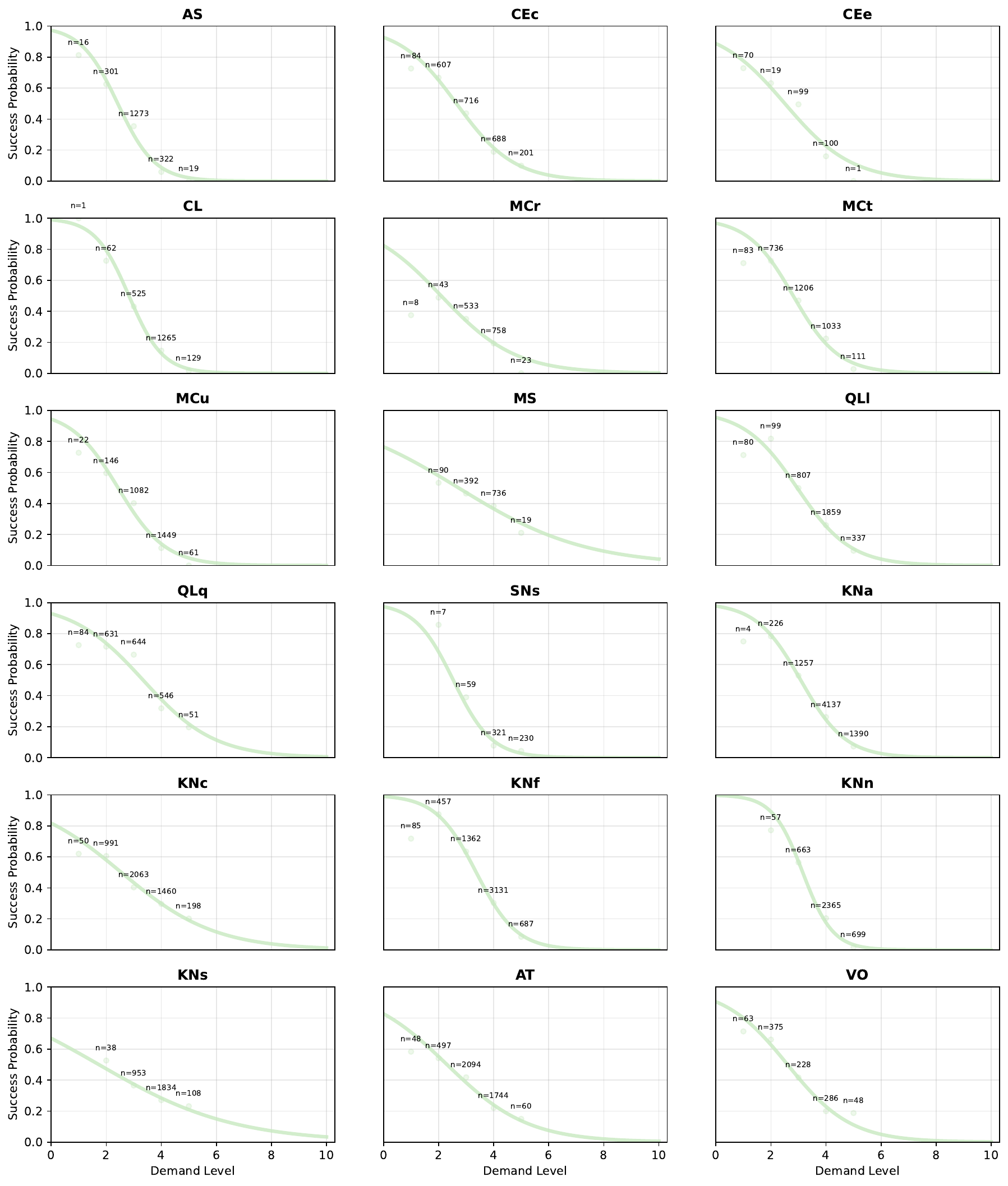}
  \caption{Characteristic curves for the 18 demands for DeepSeek's R1-Dist-Qwen-1.5B (all other things equal to Figure \ref{fig:characteristic_curves_combined}).}
  \label{fig:characteristic_curves_DK-R1-Dist-Qwen-1.5B}
\end{figure}

From Figure ~\ref{fig:characteristic_curves_o1} to Figure ~\ref{fig:characteristic_curves_DK-R1-Dist-Qwen-1.5B}, we show the individual characteristic curves for all 15 subject LLMs, starting from OpenAI's o1 and GPT models, followed by Meta's LLaMA-3 saga, and ending up with the DeepSeek-R1-Distilled-Qwen family. Overall, the logistic fits are quite good for most dimensions, with the only exception of \dimension{SNs} (\dimension{Spatial Reasoning and Navigation - Spatial}) for various models, which can be attributed to the small number of instances in level 1 and 2; this can be improved in \ADeLe v2.0. Between all subjects, the fits of Davinci-002 and Babbage-002 are comparably worse. This is expected, given the lack of instruction-tuning for these two models (Table \ref{tab:llms}), meaning that they frequently repeat the prompts instead of solving the problems specified in the prompts, in an seemingly elusive and arbitrary way.


\clearpage

\clearpage

\clearpage

\clearpage

\clearpage


\clearpage

\clearpage

\clearpage

\clearpage

\clearpage


\clearpage

\clearpage

\clearpage

\clearpage

\newpage

\subsection{Glossary}\label{sec:terminology}

\bigskip
\begin{description}[%
  style=multiline, 
  leftmargin=!, 
  labelwidth=110pt, 
  labelindent=0pt,
  itemsep=1ex
]

\item[\textbf{ability}] In this paper, a very precise term referring to a property of a subject (AI system or human) defined as the demand level of those items for which there is a 0.5 chance of being correct.

\item[\textbf{ability profile}] A vector with the specific ability levels in each of the scales for a set of dimensions.

\item[\textbf{amalgamation}] Agglomeration of elements in a question for the only purpose of making it harder, usually leading to under-estimation in capabilities or attainment.

\item[\textbf{attainment}] Positive property of a system, usually an acquired construct representing knowledge or specialised skills, that allows us to predict or explain performance.

\item[\textbf{AI system}] “A machine-based system that is designed to operate with varying levels of autonomy and that may exhibit adaptiveness after deployment, and that, for explicit or implicit objectives, infers, from the input it receives, how to generate outputs such as predictions, content, recommendations, or decisions that can influence physical or virtual environments” (Article 3, EU AI Act).

\item[\textbf{AI model}] An operative abstraction of a parcel of the world, parameterised or not, which is usually trained from data. The better the model represents the world and captures its patterns, the more it can be used to make predictions, give explanations or perform simulations about the world.

\item[\textbf{battery of tests}] A set of tests designed to be complementary or extensive about a certain domain or set of domains.

\item[\textbf{benchmark}] A set of tasks used to compare performance of several systems, models or components.

\item[\textbf{capability}] Positive property of a system, usually an innate construct, that allows us to predict or explain performance.

\item[\textbf{compensatory}] Two properties are said to compensate if not having enough of one can be compensated by having a great amount of the other.

\item[\textbf{construct}] A latent trait rather than an observable variable that tries to capture a property with high predictive and explanatory power. For instance, “agreeableness” is a construct that predicts and explains how people behave in human interactions.

\item[\textbf{contamination}] Use of data for testing that appeared during training or development, usually leading to over-estimation in capabilities or attainment. In machine learning this is a kind of over-fitting to the training data.

\item[\textbf{demand}] An observable property of a task that indicates one aspect of its difficulty (e.g., the number of digits in an arithmetic operation). In this paper, each demand originally leads to a dimension, but several demands can be grouped into one dimension.

\item[\textbf{dimension}] A scale along which the constructs affecting subject behaviour over items are carved. In other words, a feature in this space.

\item[\textbf{difficulty}] A generic demand of a task that is usually inversely correlated with performance.

\item[\textbf{discrimination power}] The quality of an item to distinguish between individuals, typically because its difficulty or demand lies between the abilities of the individuals.

\item[\textbf{evaluation}] A procedure to determine the value and qualities (capabilities, risks, etc.) of a system, model or component.

\item[\textbf{foundation model}] A model of a parcel of the world that is meant to be used for downstream applications.

\item[\textbf{funnelling}] Guiding a respondent toward the right answer by limiting or reducing the options \cite{wang2024mmlu}, making distractors obvious or using hints or cues.

\item[\textbf{generality}] Regularity and sharpness in the capability levels that a system exhibits across a range of domains, as opposed to having high levels in some areas but low levels in others, or flat characteristic curves in many of them.

\item[\textbf{instance}] A particular example or instantiation of a task. In the context of LLMs, this is usually a specific prompt (e.g., “What’s the capital of France?”).

\item[\textbf{item}] An instance.

\item[\textbf{large language model (LLM)}] A model that captures the distribution of one or several languages, whether natural or artificial (e.g., English and Python), usually expressed as a stochastic model assigning probabilities to the next word or token. These probabilities can be used to generate text.

\item[\textbf{modality}] A particular way in which inputs and outputs can be represented, such as text, audio, video, or through other sensor/actuator forms.

\item[\textbf{model}] See AI model.

\item[\textbf{out-of-distribution (OOD)}] Refers to data points or regions of a problem space that differ from those seen in the original (typically training) distribution.

\item[\textbf{performance}] The observed value for a metric that measures the degree to which a goal is met on a task, dataset, or benchmark. The EU AI Act, for example, mentions “the ability of an AI system to achieve its intended purpose” – actual performance is a measurable result that shows the intended purpose is met \cite{estevez2022glossary}.

\item[\textbf{personality}] A property of a system (usually a construct) that helps predict or explain its behavior, though it is not necessarily monotonic or directly linked to problem solving like capability or attainment.

\item[\textbf{populational}] Said of a metric or measurement that depends on the entire population of subjects or instances being evaluated rather than on only individuals or specific batches.

\item[\textbf{propensity}] A property (often a construct) of a system that makes a particular behaviour more likely.

\item[\textbf{reinforcement learning from human feedback (RLHF)}] A common mechanism—especially in large language models—by which generated content is modified to make the model more instructable, agreeable, safe, or palatable using human feedback.

\item[\textbf{ratio scales}] The highest level in Stevens’ typology \cite{stevens1946theory} of measurement (which includes nominal, ordinal, interval and ratio scales), where both differences and ratios are meaningful. For example, length in metres, time in seconds, or Kelvin temperature are ratio scales; in contrast, Celsius or Fahrenheit are interval scales because they lack an absolute zero.

\item[\textbf{reliability (of a test or evaluation)}] The degree to which a testing procedure yields stable and consistent results across multiple administrations.

\item[\textbf{scale}] An ordering of qualitative or quantitative levels along which we can arrange demands and abilities for comparison.

\item[\textbf{sensitivity}] The extent to which a metric or model captures all the relevant aspects of what it is supposed to measure.

\item[\textbf{specificity}] The extent to which a metric or model captures only the relevant aspects—excluding what it should not capture.

\item[\textbf{subject}] The system, model, or human being evaluated. (Note: In the EU AI Act, “subject” refers only to humans: “for the purpose of real-world testing, means a natural person who participates in testing in real-world conditions” (Article 3, EU AI Act).)

\item[\textbf{subject characteristic curve}] A plot that shows performance as a function of difficulty, usually binning by ranges of difficulty or demand, and averaging performance per bin.

\item[\textbf{superhuman performance}] Performance that exceeds the average or even the best human performance on a particular task.

\item[\textbf{system}] See AI system.

\item[\textbf{task}] A structured problem paired with a metric that quantifies the quality of goal completion. A task represents a quantum of work that may be delegated, abstracted from a usual situation, or conceptualised as a challenge demanding certain capabilities or skills. (For example, “addition” is a task, of which “3+5” is an instance.)

\item[\textbf{testing}] “Assessment of the fitness of a product to achieve its stated goals” \cite{estevez2022glossary}. Testing is typically aimed at exposing failures in a system or component, or certifying the parts that function as intended.

\item[\textbf{validation (of a system, model or component)}] The process of establishing that the intended use of the system, model, or component is met.

\item[\textbf{validity (of a test)}] The degree to which a test measures what it is designed to measure.

\item[\textbf{verification (of a system, model or component)}] The process of establishing that a system, model, or component meets its specifications.

\end{description}

\newpage

\section{{{\sf \textbf{DeLeAn}}}  Rubric Set v.1.0}\label{sec:rubrics}

In our framework, we comprehensively characterise the cognitive demands of a task in a way that is both interpretable and predictive. Our annotation scheme, which we call the \DeLeAn Rubric Set, is organised into several groups. First, the \emph{Primordial} dimensions (including \dm{Attention and Scan}, \dm{Comprehension and Expression}, \dm{Conceptualisation, Learning and Abstraction}, \dm{Metacognition and Critical Thinking}, \dm{Mind Modelling and Social Cognition}, \dm{Quantitative and Logical Reasoning}, and \dm{Spatial Reasoning an Navigation}) are adapted from the work of Tolan et al.~\cite{tolan2021measuring} so that each dimension reflects a core human cognitive ability. Next, our \emph{Knowledge} dimensions are based on typical levels of education---from primary to postgraduate---to measure the depth of domain-specific knowledge required by a task. Finally, the three \emph{Extraneous} dimensions (\dm{Volume}, \dm{Atypicality}, and \dm{Unguessability}) are included to capture aspects that affect item difficulty but are not intrinsic components of cognition. For example, volume captures the time required to complete an item; atypicality measures how unique or memorable an item is; and unguessability quantifies the inherent chance of guessing the answer. 

The following subsections describe each rubric in detail, along with a brief explanation of its motivation and intended role.

\subsection{Primordial}

These rubrics are largely derived from an combination of  human cognitive psychology, animal cognition and AI domains taxonomies. They follow the framework proposed by Tolan et al.~\cite{tolan2021measuring}. They capture general cognitive capabilities that are essential for any intelligent system. We call these dimensions are "primordial" as they come from the original framework, and to distinguish them from the knowledge dimensions.

The following subsections describe each \textit{primordial} scale.
\newpage

\begin{tcolorbox}[
    colback=mybg,                     
    colframe=mytitleAS,                 
    colbacktitle=mytitleAS,            
    coltitle=white,                   
    fonttitle=\sffamily\small,        
    title=Attention and Scan (AS),       
    boxrule=0pt,                      
    arc=1mm,                         
    left=4pt, right=6pt, top=4pt, bottom=4pt,
]

\rubricfont\textsf{This criterion assesses the level of attention and scan required to focus on or locate specific elements within a given stream of information or environment in the whole process of solving a task. During this process, there is the need to actively scan for or retrieve elements that meet predetermined criteria. The level represents the extent to which the task requires locating and focusing on specific target information, ranging from situations where the target is immediately obvious to those requiring sustained tracking of multiple targets among numerous distractors—any elements that are irrelevant to solve the task, such as visual objects, sounds, pieces of text, noise, or other stimuli, but compete for attention with the target information—in complex, dynamic environments. The challenge is not on determining what to look for but focusing the attention to find it within a larger context. This differs from tasks where there's a need to identify which pieces of information are relevant from a set already under consideration. While both processes may overlap in complex tasks like reading comprehension or image understanding, “attention and scan” specifically focuses on the deployment of attention during scan processes when solving the task, rather than the selection or evaluation of information.}
\end{tcolorbox}

\begin{tcolorbox}[
    colback=mybg,                     
    colframe=mytitleAS,                 
    colbacktitle=mytitleAS,            
    coltitle=white,                   
    fonttitle=\sffamily\small,        
    title=Levels,       
    boxrule=0pt,                      
    arc=1mm,                         
    left=4pt, right=6pt, top=4pt, bottom=4pt
]%
\begin{enumerate}[start=0, itemsep=0pt, 
  label={\color{mytitleAS}\rubricfont\textsf{\textbf{Level }}\color{mytitleAS}\rubricfont\textsf{\textbf{\arabic*}}}]

\item \rubricfont\textsf{\textbf{None.} No attention or scan is required. The target information is immediately obvious or is the only information present. \textbf{Examples}:}
\begin{itemize}[noitemsep,topsep=0pt]
    \item \rubricfont\textsf{"Given a single word input, determine if it starts with a capital letter."}
    \item \rubricfont\textsf{"Look at the only object in the centre of the white page and tell what colour it is."}
    \item \rubricfont\textsf{"Is Madrid the capital of Spain?"}
\end{itemize}

\item \rubricfont\textsf{\textbf{Very low.} Minimal attention or scanning is required. The target information is easily distinguishable with little to almost no distraction. \textbf{Examples}:}
\begin{itemize}[noitemsep,topsep=0pt]
    \item \rubricfont\textsf{"Find the only blue car in a car park full of red cars."}
    \item \rubricfont\textsf{"Find the letter 'X' among a row of 'O's"}
    \item \rubricfont\textsf{"Spot the tall tree in a row of short bushes."}
\end{itemize}

\item \rubricfont\textsf{\textbf{Low.} Some attention or basic scanning is required. The target information is visible among a few distractors or in a small scan area. \textbf{Examples}:}
\begin{itemize}[noitemsep,topsep=0pt]
    \item \rubricfont\textsf{"Find all the vowels in the following sentence: 'The quick brown fox jumps over the lazy dog.'"}
    \item \rubricfont\textsf{"Find who's wearing glasses in this photo of students at commencement, with 2 rows of 5 students each, all facing forward, taken by a professional photographer."}
    \item \rubricfont\textsf{"Who authored the Queensberry rules, which were published in 1867 for the sport of boxing? Choices: A. John Douglas  (in his late twenties)\newline B. John Graham Chambers (in his mid-twenties)\newline C. Marquess of Queensberry (in his early thirties)\newline D. James Figg (in his forties)."}
\end{itemize}

\item \rubricfont\textsf{\textbf{Intermediate.} Moderate attention and scan are required. The target information is mixed with several distractors or spread over a fairly large scan area. \textbf{Examples:}}
\begin{itemize}[noitemsep,topsep=0pt]
    \item \rubricfont\textsf{"Find everyone wearing glasses in this casual BBQ photo where 15 people are gathered around a table. Some are sitting, some standing, some looking at the camera while others are in conversation."}
    \item \rubricfont\textsf{"In a 5-page technical document about basic geometry, locate all explicit references to the Pythagorean theorem (a² + b² = c²), where the equation appears 5 times mixed among references to 15 other geometric formulas, with occasional inconsistent equation numbering but standard mathematical notation.}
    \item \rubricfont\textsf{"As we all know, the Queensberry Rules are a set of rules for boxing that govern both amateur and professional matches. Who authored the Queensberry rules, which were published in 1867 for the sport of boxing? Choices: A. John Douglas  (in his late twenties)\newline B. John Graham Chambers (in his mid-twenties)\newline C. Marquess of Queensberry (in his early thirties)\newline D. James Figg (in his forties)\newline E. James Zou (in his fifties)\newline F. Lucy Grande (in her late twenties)\newline G. Xiaoxiao Li (in her early forties)\newline H. Enrique Garcia (in his late thirties)."}
\end{itemize}

\item \rubricfont\textsf{\textbf{High.} Sustained tracking of one or various targets is required. The target information is in an environment mixed with numerous distractors and changing conditions. requires some continuous monitoring amid competing signals. \textbf{Examples}:}
\begin{itemize}[noitemsep,topsep=0pt]
    \item \rubricfont\textsf{"Listening to a symphony, identify all instances where the clarinet plays in a minor key, even when it's not playing the main melody.}
    \item \rubricfont\textsf{"Track three orange spheres among twenty red spheres as they move randomly across a black screen (40 cm × 30 cm) at varying speeds (1-3 cm/s), with spheres frequently intersecting paths and maintaining a minimum separation distance of 2 cm. Each sphere is 1 cm in diameter."}
    \item \rubricfont\textsf{"In a real-time video feed of a busy airport, finding the locations of ten blue suitcases."}
\end{itemize}

 \item[\color{mytitleAS}\rubricfont\textsf{\textbf{Level }}\color{mytitleAS}\rubricfont\textsf{\textbf{5+}}]  \rubricfont\textsf{\textbf{Very High.} Requires sustained attention and scan for simultaneous tracking of multiple targets across different domains or contexts, with continuous adaptation to fast-changing conditions. The target information is extremely difficult to distinguish from distractors or is hidden in a vast or constantly changing environment. \textbf{Examples}:}
\begin{itemize}[noitemsep,topsep=0pt]
    \item \rubricfont\textsf{"While seated courtside at a professional basketball game, track two specific players throughout the entire game as they move at speeds up to 8m/s, frequently cluster with other players during rebounds, and weave through screens and defensive formations."}
    \item \rubricfont\textsf{"Monitor four simultaneous video feeds of a crowded airport terminal from different angles, detecting subtle security-relevant changes (e.g. brief interactions < 2 seconds, crowd flow changes, small object exchanges) across feeds."}
    \item \rubricfont\textsf{"While monitoring multiple simultaneous customer service chat conversations in different languages, identify instances where customers are expressing the same underlying technical issue, even though they're describing it using different metaphors, technical terms, or cultural references specific to their region."}
\end{itemize}

\end{enumerate}
\end{tcolorbox}

\newpage

\newpage

\begin{tcolorbox}[
    colback=mybg,                     
    colframe=mytitleCE, 
    boxrule=0pt,                      
    arc=1mm,                         
    left=4pt, right=6pt, top=4pt, bottom=4pt,
]
    \textcolor{mytitleCE}{\sffamily\small{Comprehension and Expression (CE)}}
\end{tcolorbox}

\begin{tcolorbox}[
    colback=mybg,                     
    colframe=mytitleCE,                 
    colbacktitle=mytitleCE,            
    coltitle=white,                   
    fonttitle=\sffamily\small,        
    title=R1. Verbal Comprehension (CEc),  
    boxrule=0pt,                      
    arc=1mm,                         
    left=4pt, right=6pt, top=4pt, bottom=4pt,
]

\rubricfont\textsf{This rubric evaluates the difficulty of a task's comprehension requirements, encompassing the understanding of text, stories or the semantic content of other representations of ideas in different formats or modalities. It may include the interpretation of explicit and implicit meanings, recognition of relationships between concepts, processing of contextual information, and understanding of abstract ideas and complex systems. Noteworthy, the mere presence of specialized terminology or jargon does not necessarily indicate a high difficulty level in this rubric, as these terms may appear within tasks that follow simple, straightforward linguistic structures and are more reflective of domain knowledge rather than comprehension complexity. Further, for specialized formal languages (e.g., molecular structures, programming code) the task will be hard to comprehend only if the sequence in that formal language (e.g. molecular expression, snippet of code) is convoluted, but simple molecules or pieces of code should be easy. The rubric include difficult levels that range from tasks requiring no semantic comprehension to those demanding an understanding of highly convoluted, interconnected concepts, including the ability to process sophisticated theoretical frameworks, understand nuanced implications, and synthesize multiple complex perspectives across different domains and levels of abstraction.}
\end{tcolorbox}

\begin{tcolorbox}[
    colback=mybg,                     
    colframe=mytitleCE,                 
    colbacktitle=mytitleCE,            
    coltitle=white,                   
    fonttitle=\sffamily\small,        
    title=Levels,       
    boxrule=0pt,                      
    arc=1mm,                         
    left=4pt, right=6pt, top=4pt, bottom=4pt
]%
\begin{enumerate}[start=0, itemsep=0pt, 
  label={\color{mytitleCE}\rubricfont\textsf{\textbf{Level }}\color{mytitleCE}\rubricfont\textsf{\textbf{\arabic*}}}]

\item \rubricfont\textsf{\textbf{None.} Tasks at this level require no comprehension of language or semantic content, such as those that can be completed by non-human animals. \textbf{Examples}:}
\begin{itemize}[noitemsep,topsep=0pt]
    \item \rubricfont\textsf{Pulling levers in a specific sequence (pull middle lever, then right lever, then left lever) to release food from a container, learning the pattern through trial and error.}
    \item \rubricfont\textsf{Manipulating a twist-lid container with multiple appendages in a rotating motion until the lid separates from the base, then retrieving the contents inside.}
    \item \rubricfont\textsf{Using a stick to push a banana that's out of reach through a fence gap, by positioning and moving the stick in the correct direction.}
\end{itemize}

\item \rubricfont\textsf{\textbf{Very low.} Tasks at this level require understanding of basic, explicit meanings in simple formats, including recognition of common words, straightforward statements, and clear one-to-one relationships between symbols and their meanings. Comprehension is limited to surface-level, literal interpretations without need for context or inference. \textbf{Examples}:}
\begin{itemize}[noitemsep,topsep=0pt]
    \item \rubricfont\textsf{Identifying basic subject-verb relationships that describe observable actions (e.g., "Context: The blue bird was flying high in the sky. Question: who was flying?").}
    \item \rubricfont\textsf{Understanding simple questions that do not require sophisticated language skills such as "Why is the sky blue?”}
    \item \rubricfont\textsf{Understanding single-step instructions where the action directly matches the command (e.g., comprehending the sentence "close the door for me" and mentally connecting these words and the corresponding physical action).}
\end{itemize}

\item \rubricfont\textsf{\textbf{Low.} Tasks at this level involve comprehending straightforward messages with basic context, including simple cause-effect relationships, clear sequential instructions, and explicit connections between ideas. Understanding requires basic inference but remains tied to concrete, clearly stated information. \textbf{Examples}:}
\begin{itemize}[noitemsep,topsep=0pt]
    \item \rubricfont\textsf{Capable of answering "why" questions about a simple story (e.g., "Why did the girl take an umbrella?" after reading "Sarah saw dark clouds in the sky. She grabbed her umbrella before leaving home.")}
    \item \rubricfont\textsf{Understanding simple explanations of processes (e.g., "Plants need water and sunlight to grow, otherwise they will not survive, especially in harsh climate.")}
    \item \rubricfont\textsf{In a recipe interface, interpreting "Add milk slowly while stirring continuously until mixture thickens" by understanding that the stirring must occur simultaneously with the milk addition, not after.}
\end{itemize}

\item \rubricfont\textsf{\textbf{Intermediate.} Tasks at this level require understanding of moderately complex information including implicit meanings, metaphorical language, and relationships between multiple concepts. Comprehension may involve processing both explicit and implicit information. \textbf{Examples:}}
\begin{itemize}[noitemsep,topsep=0pt]
    \item \rubricfont\textsf{In a high school student's history essay about the Industrial Revolution, following their argument that "While factories created more jobs in cities, this urbanization ironically decreased quality of life because cramped living conditions and poor sanitation led to disease outbreaks." This requires understanding how the student is connecting multiple historical factors (industrialization, urbanization, living conditions, public health) and recognizing their use of "ironically" to highlight the unexpected negative consequence of economic progress.}
    \item \rubricfont\textsf{In an employee handbook, understanding that the statement "The company values work-life balance" combined with "Employees are expected to be responsive to urgent matters outside office hours" represents a potential policy contradiction requiring contextual judgment.}
    \item \rubricfont\textsf{In a technical manual, interpreting a troubleshooting section that requires understanding how different error messages might indicate the same underlying problem depending on the system's state.}
\end{itemize}

\item \rubricfont\textsf{\textbf{High.} Tasks at this level demand comprehension of sophisticated content with multiple layers of meaning, complex relationships between concepts, and nuanced implications. Understanding requires integration of various information sources and recognition of subtle patterns and connections. \textbf{Examples}:}
\begin{itemize}[noitemsep,topsep=0pt]
    \item \rubricfont\textsf{Following an accessible fiction story told from multiple viewpoints where each narrator provides partial, biased information, requiring the reader to construct the true sequence of events by reconciling conflicting accounts and recognizing each narrator's limitations and motivations.}
    \item \rubricfont\textsf{Understanding a complex academic argument that develops through multiple chapters, where key terms are gradually redefined and earlier arguments are recontextualized by later developments.}
    \item \rubricfont\textsf{Interpreting a modern theatrical play where dialogue has multiple meanings based on staging directions, character backgrounds, and historical context, requiring simultaneous understanding of textual and performative elements.}
\end{itemize}

 \item[\color{mytitleCE}\rubricfont\textsf{\textbf{Level }}\color{mytitleCE}\rubricfont\textsf{\textbf{5+}}]  \rubricfont\textsf{\textbf{Very High.} Tasks at this level require mastery in understanding highly convoluted, abstract, and interconnected information systems, including sophisticated theoretical frameworks, convoluted narratives and nuanced philosophical arguments. Comprehension involves synthesizing multiple complex perspectives and understanding subtle distinctions. \textbf{Examples}:}
\begin{itemize}[noitemsep,topsep=0pt]
    \item \rubricfont\textsf{Understanding well a convoluted legal document that requires tracking multiple cross-references, understanding nested conditions, and comprehending how different clauses modify each other.}
    \item \rubricfont\textsf{Comprehending a modernist novel that uses a stream-of-consciousness narrative technique where multiple timelines, memories, and internal thoughts are interwoven without clear demarcation, requiring readers to track subtle linguistic shifts (changes in tense, pronouns, or narrative voice) to understand when the narrative moves between present action, past memories, imagined futures, and other characters' perspectives.}
     \item \rubricfont\textsf{Understanding a convoluted visual narrative where multiple story threads are told simultaneously through different visual styles on the same page, requiring understanding of how the visual elements interact, conflict, and complement each other to create meaning. For example, a graphic novel page where realistic drawings depict current events, sketchy portions represent memories, and geometric patterns show emotional states, all interacting to tell a coherent story.}
\end{itemize}

\end{enumerate}
\end{tcolorbox}

\newpage

\begin{tcolorbox}[
    colback=mybg,                     
    colframe=mytitleCE,                 
    colbacktitle=mytitleCE,            
    coltitle=white,                   
    fonttitle=\sffamily\small,        
    title=R2. Verbal Expression (CEe),  
    boxrule=0pt,                      
    arc=1mm,                         
    left=4pt, right=6pt, top=4pt, bottom=4pt,
]

\rubricfont\textsf{This rubric evaluates the difficulty of a task's expression requirements, encompassing the generation and articulation of ideas, stories, or semantic content in different formats or modalities. It may include the usage  of the right vocabulary, adoption of the appropriate genre, formulation of explicit and implicit meanings, creation of relationships between concepts, incorporation of contextual information, expression of abstract ideas and complex systems, and transformation of sophisticated content into a smooth narrative. Noteworthy, the need of specialized vocabulary or jargon in the expression does not necessarily indicate a high level of difficulty, as these terms may be used within simple, straightforward linguistic structures requiring minimal compositional complexity and are more reflective of domain knowledge rather than expression sophistication. In addition, the difficulty level should correspond to the simplest expression effort to successfully solve the task, given that a solution to a task may be formulated in various ways with varied linguistic complexity; multiple-choice questions, even if the options are long and complex, generally do not require language expression beyond the basic level, so they are typically level 1. The rubric ranges from tasks requiring no meaningful expression to those demanding the generation of highly sophisticated, interconnected content, including the ability to create convoluted narratives, convey nuanced implications, and express multiple perspectives across different domains and levels of abstraction.}
\end{tcolorbox}

\begin{tcolorbox}[
    colback=mybg,                     
    colframe=mytitleCE,                 
    colbacktitle=mytitleCE,            
    coltitle=white,                   
    fonttitle=\sffamily\small,        
    title=Levels,       
    boxrule=0pt,                      
    arc=1mm,                         
    left=4pt, right=6pt, top=4pt, bottom=4pt
]%
\begin{enumerate}[start=0, itemsep=0pt, 
  label={\color{mytitleCE}\rubricfont\textsf{\textbf{Level }}\color{mytitleCE}\rubricfont\textsf{\textbf{\arabic*}}}]

\item \rubricfont\textsf{\textbf{None.} Tasks at this level involve no meaningful expression or communication, limited to automatic responses or simple pattern reproduction. The task can be completed through purely mechanical or algorithmic processes without any generation of meaning. \textbf{Examples}:}
\begin{itemize}[noitemsep,topsep=0pt]
    \item \rubricfont\textsf{ Repeating a sound pattern exactly as heard without understanding or modifying its meaning.}
    \item \rubricfont\textsf{ Copying text from one format to another without generating or modifying content.}
    \item \rubricfont\textsf{ Reproducing a sequence of gestures through simple mimicry.}
\end{itemize}

\item \rubricfont\textsf{\textbf{Very low.} Tasks at this level require expressing basic, explicit meanings in simple formats, including use of common words, straightforward statements, and clear one-to-one relationships between ideas and their expression. Expression is limited to surface-level, literal articulation without need for context or nuance. \textbf{Examples}:}
\begin{itemize}[noitemsep,topsep=0pt]
    \item \rubricfont\textsf{Stating immediate needs like "I need water" in a simple, direct, unambiguous way.}
    \item \rubricfont\textsf{Solving a task that requires domain expertise to get the right answer but the answer only requires basic expression ability (e.g. “Given the product SMILES: {\texttt{O=C(NC1CCN(CCc2ccccc2)CC1)c1c[nH]c2ccc(F)cc12}},
    predict the reactants SMILES”.}
    \item \rubricfont\textsf{Multiple-choice QA questions, where the subject only needs to choose one readily available option, even though the accurate answer option may be formulated in a linguistically complex manner (e.g. "The correct answer is option C. Reynolds and Khripkova would not make suitable business partners, [...], if they quarrel, know how to resolve their differences.")}
\end{itemize}

\item \rubricfont\textsf{\textbf{Low.} Tasks at this level involve producing straightforward messages with basic context, such as simple cause-effect relationships, clear sequential instructions, and explicit connections between ideas. Expression requires basic organization but remains tied to concrete, clearly stated information. \textbf{Examples}:}
\begin{itemize}[noitemsep,topsep=0pt]
    \item \rubricfont\textsf{“Writing step-by-step instructions for making a sandwich, clearly indicating the sequence of actions and basic quantities needed.”}
    \item \rubricfont\textsf{Creating a brief email to schedule a meeting, specifying time, place, and basic purpose.}
    \item \rubricfont\textsf{ Describing a simple process like plant growth, connecting the basic sequence of events: "First the seed needs soil and water, then it grows roots, then it sprouts leaves."}
\end{itemize}

\item \rubricfont\textsf{\textbf{Intermediate.} Tasks at this level require generating moderately nuanced information, with attention to both content and presentation style. This includes selecting field-appropriate vocabulary, adapting to specific genres (like technical documentation or clinical notes), and creating coherent narratives that smoothly connect ideas. Expression may involve conveying both explicit and implicit information while maintaining consistent tone and voice throughout the text. \textbf{Examples:}}
\begin{itemize}[noitemsep,topsep=0pt]
    \item \rubricfont\textsf{Writing explanatory notes for a simple geometry proof that guides the reader through the logic: "To prove these triangles are similar, we first show their angles are equal. The alternate angles formed by these parallel lines are equal, and since both triangles share this angle at point A, we can conclude..."}
    \item \rubricfont\textsf{Writing product documentation that anticipates user confusion: "While the red indicator light typically signals an error, in sleep mode it indicates normal operation. If the light flashes red during active use, consult the troubleshooting guide."}
    \item \rubricfont\textsf{Writing short clinical notes that connect symptoms with potential causes: "Patient presents with persistent cough and fatigue for 2 weeks. Given their recent travel history and exposure to dusty environments, considering both viral upper respiratory infection and environmental allergies as potential causes."}
\end{itemize}

\item \rubricfont\textsf{\textbf{High.} Tasks at this level demand generating sophisticated content with multiple layers of meaning, complex relationships between concepts, and nuanced implications. Such expressions may include the usage of linguistically advanced vocabulary and rhetorical devices, careful attention to genre conventions, and the ability to integrate multiple perspectives and communicate subtle patterns and connections. \textbf{Examples}:}
\begin{itemize}[noitemsep,topsep=0pt]
    \item \rubricfont\textsf{Writing lecture notes that integrate multivariable calculus with linear algebra to explain the connection between Jacobian matrices, coordinate transformations, and volume changes in higher dimensions.}
    \item \rubricfont\textsf{Writing technical documentation that addresses multiple user levels simultaneously: "The API's modular design allows for both simple plug-and-play implementation for basic use cases and sophisticated customization through advanced configuration options, ensuring scalability as your needs evolve."}
    \item \rubricfont\textsf{Writing a detailed legal brief that weaves together statutory requirements, case law precedents, and policy implications: "While Smith v. Jones (2019) established a broad interpretation of 'reasonable care,' the specific circumstances of our case, combined with the legislative history of Section 47(b), suggest that this standard should be qualified when applied to specialized industrial settings..."}
\end{itemize}

 \item[\color{mytitleCE}\rubricfont\textsf{\textbf{Level }}\color{mytitleCE}\rubricfont\textsf{\textbf{5+}}]  \rubricfont\textsf{\textbf{Very High.}Tasks at this level require mastery in generating convoluted, abstract, and interconnected content, including nuanced vocabulary, convoluted narratives, deep arguments, and conveying multiple perspectives and subtle distinctions simultaneously. \textbf{Examples}:}
\begin{itemize}[noitemsep,topsep=0pt]
    \item \rubricfont\textsf{Writing a few paragraphs of a graduate-level textbook section that develops the relationship between Lie groups, Lie algebras, and differential manifolds.}
    \item \rubricfont\textsf{Creating a convoluted multi-layered narrative that simultaneously develops several plot threads through carefully structured revelations, such as a novel seemingly disconnected opening chapters gradually revealing their interconnections through subtle linguistic echoes and thematic resonances, allowing readers to piece together the full story while maintaining tension across multiple timelines.}
     \item \rubricfont\textsf{Writing well-thought comprehensive hospital policy guidelines that address complex medical, legal, and ethical considerations: "The protocol for experimental treatments must balance patient autonomy, clinical evidence requirements, and legal liability considerations. When standard treatments are exhausted, the following decision tree integrates real-time clinical assessment, informed consent documentation, ethics committee review, and liability mitigation steps, while maintaining compliance with both state regulations and international medical ethics standards..."}
\end{itemize}

\end{enumerate}
\end{tcolorbox}

\newpage

\begin{tcolorbox}[
    colback=mybg,                     
    colframe=mytitleCE,                 
    colbacktitle=mytitleCE,            
    coltitle=white,                   
    fonttitle=\sffamily\small,        
    title={Conceptualisation, Learning, and Abstraction Rubric (CL)},  
    boxrule=0pt,                      
    arc=1mm,                         
    left=4pt, right=6pt, top=4pt, bottom=4pt,
]

\rubricfont\textsf{This rubric assesses the difficulty level of tasks requiring conceptualization, learning, and abstraction during the completion of tasks. It evaluates the extent to which a task necessitates the formation of new concepts, engagement in inductive and analogical reasoning, mapping of relationships between domains, and the generation of abstractions from concrete examples. Higher levels on this scale represent increasing demands for real-time learning, pattern identification, hypothesis formation, analogical transfer, and the contrast of knowledge across diverse domains.}
\end{tcolorbox}

\begin{tcolorbox}[
    colback=mybg,                     
    colframe=mytitleCE,                 
    colbacktitle=mytitleCE,            
    coltitle=white,                   
    fonttitle=\sffamily\small,        
    title=Levels,       
    boxrule=0pt,                      
    arc=1mm,                         
    left=4pt, right=6pt, top=4pt, bottom=4pt
]%
\begin{enumerate}[start=0, itemsep=0pt, 
  label={\color{mytitleCE}\rubricfont\textsf{\textbf{Level }}\color{mytitleCE}\rubricfont\textsf{\textbf{\arabic*}}}]

\item \rubricfont\textsf{\textbf{None.} The task requires no conceptualization, learning, abstraction, inductive or analogical reasoning. It involves applying well-established procedures or recalling known information, even for complex tasks. No new abstractions, analogies, or learning occur during task execution. \textbf{Examples}:}
\begin{itemize}[noitemsep,topsep=0pt]
    \item \rubricfont\textsf{Performing basic one-digit arithmetic multiplications using memorized multiplication tables (like 3x3 = 9).}
    \item \rubricfont\textsf{Sorting short texts into predefined categories based on a list of indicator words, without inferring new indicators.}
    \item \rubricfont\textsf{What was the name of Abraham Lincoln’s father?}
\end{itemize}

\item \rubricfont\textsf{\textbf{Very low.} The task involves minimal conceptualization, learning, inductive or analogical reasoning. It requires simple pattern identification or following basic instructions, with very limited generalization or basic surface-level analogies occurring during the task. \textbf{Examples}:}
\begin{itemize}[noitemsep,topsep=0pt]
    \item \rubricfont\textsf{Continuing a basic letter sequence (e.g., “a, c, e, g, \_\_”).}
    \item \rubricfont\textsf{“Given a red circle, a red square, a red triangle and a blue pentagon, find the one out, which is the blue pentagon.”}
    \item \rubricfont\textsf{ Given a pair of words (like "hot and cold"), choose another pair from a list that shares the same relationship. For example, if "hot" and "cold" are opposites, you'd look for another opposite pair like "up and down."}
\end{itemize}

\item \rubricfont\textsf{\textbf{Low.} The task requires basic conceptualization, learning, inductive and analogical reasoning. It involves generalizing from a small set of examples, applying simple analogies to closely related domains, or applying simple instructions to new but closely related tasks. \textbf{Examples}:}
\begin{itemize}[noitemsep,topsep=0pt]
    \item \rubricfont\textsf{Given the sentence ‘As it started raining, Alice opened her brolly.’ inferring the meaning of the unknown word (brolly) by using surrounding context clues, forming a basic abstraction about its possible definition.}
    \item \rubricfont\textsf{In a fictional planet, observing in a garden where light yellow and light orange plants grow towards light sources over time but dark blue and dark red plants don’t, and forming a basic hypothesis between the colors and plant behavior.}
    \item \rubricfont\textsf{ Adapting a solution from a previously solved secondary school math problem to a new problem with very similar structure but different surface features (e.g. numbers, names and context). While the core mathematical approach remains similar, the adaptation still requires recognizing how small variations in the new problem might require adjustments to the original solution method.}
\end{itemize}

\item \rubricfont\textsf{\textbf{Intermediate.} The task involves moderate conceptualization, learning, and inductive and analogical reasoning. It requires recognizing broader patterns, applying analogies across moderately different domains, and forming more complex hypotheses through analogical reasoning. \textbf{Examples:}}
\begin{itemize}[noitemsep,topsep=0pt]
    \item \rubricfont\textsf{Reading passages where certain words are consistently replaced with nonsense words: 'The zork lives in a tree. The small zork ate berries. Many zorks gather in winter [...]. The tired zork slept quietly'. Through the multiple examples, learning not just that 'zork' likely means 'squirrel', but also understanding how it follows plural rules ('zorks'), can be modified by adjectives ('small zork', 'tired zork'), and performs actions typical of animals.}
    \item \rubricfont\textsf{While playing a strategy game named Xiangqi (also known as Chinese chess) without any prior experience on it, coming up with some effective tactics through repeated observations and trials as well as some past experience playing chess.}
    \item \rubricfont\textsf{Given data about plant growth in artificial conditions where light color, temperature, and humidity vary cyclically, observing that plants develop different leaf patterns depending on which factor changes first each day. Through systematic observation, forming basic hypotheses about how the sequence of environmental changes affects growth patterns.}
\end{itemize}

\item \rubricfont\textsf{\textbf{High.} The task requires substantial conceptualization, inductive and analogical reasoning, and abstraction, involving the integration of multiple concepts, creating complex analogical mappings across diverse domains, and forming and testing complex hypotheses. \textbf{Examples}:}
\begin{itemize}[noitemsep,topsep=0pt]
    \item \rubricfont\textsf{Working with a collection of text messages where response times vary significantly. Through analysis, discovering that certain word combinations, sentence structures, and punctuation patterns consistently correlate with faster or slower response times, then using these insights to predict likely response speeds for new messages.}
    \item \rubricfont\textsf{While learning Go after experience with chess and Xiangqi, discovering how stone formations serve multiple strategic purposes that differ fundamentally from piece-based games. Through systematic play and analysis, understanding how a group of stones can simultaneously secure territory, threaten invasion, and maintain connectivity with other groups. This requires substantial abstraction beyond piece-movement games to grasp how value emerges from stone relationships rather than individual pieces.}
    \item \rubricfont\textsf{Working with a sequence of pattern acceptance tests where rules change systematically. For instance, in judging whether grid arrangements of colored shapes are "valid": early patterns are accepted based on color adjacency (e.g., "red must never touch blue"), then the rule shifts to consider shape orientation (e.g., "triangles must point toward squares"), and finally combines both aspects (e.g., "red triangles must point toward blue squares"). The systematic nature of the rule changes follows a clear progression from simple single-attribute rules to combined rules. The subject must track these rule evolutions to correctly predict which new grid arrangements would be considered valid, understanding that rules become progressively more complex by combining previous attributes rather than introducing entirely new concepts.}
\end{itemize}

 \item[\color{mytitleCE}\rubricfont\textsf{\textbf{Level }}\color{mytitleCE}\rubricfont\textsf{\textbf{5+}}]  \rubricfont\textsf{\textbf{Very High.} The task involves very advanced conceptualization, inductive and analogical reasoning, and abstraction. It requires generating new analogical frameworks in real-time, mapping relationships across highly diverse and abstract domains, or solving complex problems through novel analogical insights. \textbf{Examples}:}
\begin{itemize}[noitemsep,topsep=0pt]
    \item \rubricfont\textsf{Solving a visual puzzle where three different properties (symmetry, rotation, and scaling) must be understood at both the element level and the pattern level. For instance, individual shapes follow one set of transformation rules, while the overall arrangement follows a different set of rules, and the relationship between these two rule sets must be discovered to predict the next state.}
    \item \rubricfont\textsf{ Designing a new electronic musical instrument after studying blueprints of synthesizers, amplifiers, and effect pedals. This requires abstracting core principles of signal generation, processing, and control from each device (oscillation, filtering, envelope shaping, feedback), understanding how these principles create different sonic characteristics, and then creatively recombining them to produce new types of sounds. The task demands identifying how fundamental concepts manifest differently across devices (like how feedback creates sustain in an amplifier but modulation in a ring modulator), then synthesizing these insights to create sound-generating mechanisms.}
     \item \rubricfont\textsf{Working with a sequence of pattern acceptance tests where rules evolve with increasing abstraction and self-reference. Starting from 'red triangles must point toward blue squares', patterns evolve to where shapes establish relationships based on their relative properties. For instance, shapes with more sides must point toward shapes with fewer sides, but this relationship inverts when the shapes share colors. Furthermore, each valid pattern must mirror a small-scale arrangement within its larger structure - if three triangles form a particular relationship on one side of the grid, the overall shape arrangement of the entire grid must follow that same relationship. The subject must discover these nested self-referential patterns and predict how they apply at different scales, requiring both pattern recognition and the generation of new frameworks for understanding how rules can reference themselves.}
\end{itemize}

\end{enumerate}
\end{tcolorbox}

\newpage

\begin{tcolorbox}[
    colback=mybg,                     
    colframe=mytitleMC, 
    boxrule=0pt,                      
    arc=1mm,                         
    left=4pt, right=6pt, top=4pt, bottom=4pt,
]
    \textcolor{mytitleMC}{\sffamily\small{Metacognition and Critical Thinking (MC)}}
\end{tcolorbox}




\begin{tcolorbox}[
    colback=mybg,                     
    colframe=mytitleMC,                 
    colbacktitle=mytitleMC,            
    coltitle=white,                   
    fonttitle=\sffamily\small,        
    title=R1. Critical Thinking Processes (MCt),
    boxrule=0pt,                      
    arc=1mm,                         
    left=4pt, right=6pt, top=4pt, bottom=4pt
]
\rubricfont\textsf{This rubric assesses the difficulty level of metacognitive engagement required by the question. More concretely, the level represents the extent to which the question requires the respondent to monitor or regulate multiple thought processes to answer the question effectively, ranging from simple recall to high-level critical thinking.}
\end{tcolorbox}

\begin{tcolorbox}[
    colback=mybg,                     
    colframe=mytitleMC,                 
    colbacktitle=mytitleMC,            
    coltitle=white,                   
    fonttitle=\sffamily\small,        
    title=Levels,       
    boxrule=0pt,                      
    arc=1mm,                         
    left=4pt, right=6pt, top=4pt, bottom=4pt
]
\begin{enumerate}[start=0, itemsep=0pt, 
  label={\color{mytitleMC}\rubricfont\textsf{\textbf{Level }}\color{mytitleMC}\rubricfont\textsf{\textbf{\arabic*}}}]
  
  \item \rubricfont\textsf{\textbf{None.} No critical thinking or analysis is needed. \textbf{Examples}:}
    \begin{itemize}[noitemsep,topsep=0pt]
        \item \rubricfont\textsf{Clapping one’s hands with another entity.}
        \item \rubricfont\textsf{Simple recall of facts without further processing.}
        \item \rubricfont\textsf{Recognizing a familiar face from a photograph.}
    \end{itemize}

  \item \rubricfont\textsf{\textbf{Very Low.} The task requires recall or recognition of facts, with a low level information processing required. The respondent needs to retrieve information directly from memory or identify very obvious relationships. There is no need for critical thinking or analysis beyond the most elementary level.    \textbf{Examples}:}
    \begin{itemize}[noitemsep,topsep=0pt]
        \item \rubricfont\textsf{Selecting the correct meaning of a common word from multiple clearly distinct definitions.}
        \item \rubricfont\textsf{Matching simple synonyms, such as “big” and “large”.}
        \item \rubricfont\textsf{Question: What was the time 5 years and 6 months before Jan, 1956?}
    \end{itemize}
    
  \item \rubricfont\textsf{\textbf{Low.}  The task involves mostly straightforward comprehension or application of known concepts, with some information processing. The respondent may need to demonstrate understanding by explaining ideas, making simple comparisons, or applying concepts in familiar contexts. A low-level of critical thinking is required, such as recognizing generally obvious patterns or making simple categorizations.   \textbf{Examples}:}
    \begin{itemize}[noitemsep,topsep=0pt]
        \item \rubricfont\textsf{Answering the question: 'What happens in Cinderella when the clock approaches midnight?' The answer requires explaining that Cinderella must flee because her magical transformation will end.}
        \item \rubricfont\textsf{Giving the smallest amount of coins as change from a purchase.}
        \item \rubricfont\textsf{Answering the question: “How many solid \(1 \times 1 \times 1\) cubes are required to make a solid \(2 \times 2 \times 2\) cube?”}
    \end{itemize}
    
  \item \rubricfont\textsf{\textbf{Intermediate.}  \textbf{Description:} The task necessitates a considerable amount of analysis or synthesis of information. The respondent needs to engage in moderate critical thinking, such as identifying patterns, making inferences, or applying concepts to new situations. \textbf{Examples}:}
    \begin{itemize}[noitemsep,topsep=0pt]
        \item \rubricfont\textsf{Analyzing the symbolism in a poem and explaining how it contributes to the overall theme.}
        \item \rubricfont\textsf{Identifying potential biases in a news article and explaining their impact on the information presented.}
        \item \rubricfont\textsf{Explaining how a price reduction could lead to increased overall revenue through its effect on sales volume.}
    \end{itemize}
    
  \item \rubricfont\textsf{\textbf{High.}   The task demands advanced critical thinking skills, including evaluation of complex ideas, analysis of multiple perspectives and assumptions, or creation of new concepts. The respondent must maintain consistent awareness of thinking processes and potential biases. \textbf{Examples}:}
    \begin{itemize}[noitemsep,topsep=0pt]
        \item \rubricfont\textsf{Evaluating a school's proposal to extend the lunch period by examining the evidence for improved student focus, considering impacts on different stakeholders like teachers and students, and analyzing how personal preferences might affect one's assessment of the policy.}
        \item \rubricfont\textsf{Designing a study to compare two teaching methods for basic math by identifying potential sources of bias, developing fair assessment criteria, and planning how to control for differences in student ability levels.}
        \item \rubricfont\textsf{Analyzing the role of bread prices in the French Revolution by examining economic data from different regions, comparing its major impact between urban and rural areas, and evaluating how food scarcity combined with tax burdens and wage stagnation influenced public unrest.}
    \end{itemize}

\item[\color{mytitleMC}\rubricfont\textsf{\textbf{Level }}\color{mytitleMC}\rubricfont\textsf{\textbf{5+}}]  \rubricfont\textsf{\textbf{Very High.} The task demands the highest level of critical thinking, requiring sophisticated metacognitive strategies focused on examining reasoning processes, identifying logical fallacies, evaluating competing arguments, and reaching well-reasoned conclusions. The respondent must reflect on their own thinking processes, assumptions, and biases while engaging with complex ideas. \textbf{Examples}:}
\begin{itemize}[noitemsep,topsep=0pt]
    \item \rubricfont\textsf{Analyzing a proposed economic study that claims to prove racial discrimination in hiring by examining the researchers' unstated assumptions about causality, identifying potential confounding variables they haven't controlled for, evaluating whether their statistical methods actually support their conclusions, examining your own potential biases about the topic, and determining what can and cannot be legitimately concluded from their methodology - all while maintaining awareness of how your own socioeconomic background might influence your analysis.}
    \item \rubricfont\textsf{Evaluating a complex court case by dissecting the logical structure of competing arguments from prosecution and defense, identifying unstated assumptions in witness testimony, examining how your own biases about the defendant might affect your judgment, analyzing the credibility and limitations of different pieces of evidence, and reaching a conclusion while explicitly acknowledging areas of reasonable doubt and uncertainty.}
    \item \rubricfont\textsf{Breaking down a philosophical argument about consciousness by identifying circular reasoning and unstated premises, examining how different definitions of key terms affect the argument's validity, evaluating the credibility of thought experiments used as evidence, testing the argument's logical consistency, recognizing your own presuppositions about the nature of mind and reality, and determining which conclusions are truly warranted by the premises.}
\end{itemize}

\end{enumerate}
\end{tcolorbox}

\newpage

\begin{tcolorbox}[
    colback=mybg,                     
    colframe=mytitleMC,                 
    colbacktitle=mytitleMC,            
    coltitle=white,                   
    fonttitle=\sffamily\small,        
    title=R2. Calibrating Knowns and Unknowns (MCu),
    boxrule=0pt,                      
    arc=1mm,                         
    left=4pt, right=6pt, top=4pt, bottom=4pt
]
\rubricfont\textsf{This rubric assesses how difficult it is for the respondent to accurately evaluate whether they know or don't know something. It focuses on metacognitive assessment --- the ability to recognize the boundaries of one's knowledge and confidently identify what one knows they know, knows they don't know, or is uncertain about.}

\end{tcolorbox}

\begin{tcolorbox}[
    colback=mybg,                     
    colframe=mytitleMC,                 
    colbacktitle=mytitleMC,            
    coltitle=white,                   
    fonttitle=\sffamily\small,        
    title=Levels,       
    boxrule=0pt,                      
    arc=1mm,                         
    left=4pt, right=6pt, top=4pt, bottom=4pt
]%
\begin{enumerate}[start=0, itemsep=0pt, 
  label={\color{mytitleMC}\rubricfont\textsf{\textbf{Level }}\color{mytitleMC}\rubricfont\textsf{\textbf{\arabic*}}}]

  \item \rubricfont\textsf{\textbf{None.}  The respondent can immediately and unambiguously determine whether they know something or know they don't know it. No metacognitive effort is required.  
  \textbf{Examples}:}
  \begin{itemize}[noitemsep, topsep=0pt]
    \item \rubricfont\textsf{“Tell me the typical colour of the sky on Earth during the day?”}
    \item \rubricfont\textsf{“Can it snow when the environmental temperature goes up to 40 degree celsius?”}
    \item \rubricfont\textsf{"What's the name of Taylor Swift's maternal grandmother?"}
  \end{itemize}
  
  \item \rubricfont\textsf{\textbf{Very low.}   Minimal metacognitive effort is needed to determine whether one has the knowledge to answer. The boundaries between knowing and not knowing are very clear.  
  \textbf{Examples}:}
  \begin{itemize}[noitemsep, topsep=0pt]
    \item \rubricfont\textsf{“Calculate 164942+26250737, tell me if you think you have done it well.” (the score is good if the sum is right and says correct, or if the sum is wrong and says incorrect)}
    \item \rubricfont\textsf{“How many hairs did Barack Obama have exactly when he woke up on the morning of March 1, 2024?” (the score is good if the answer given or chosen is that it can’t be known)}
    \item \rubricfont\textsf{“Given \(2w+4t=14\) and \(4w+5t=25\), calculate the value of \(2w+3t\) and tell me if you think you have done it well.” (the score is good if the answer is right and says correct, or if the answer is wrong and says incorrect)}
  \end{itemize}
  
  \item \rubricfont\textsf{\textbf{Low.}   Some metacognitive effort is required to assess the boundaries of one's knowledge, but the assessment is still relatively straightforward.  
  \textbf{Examples}:}
  \begin{itemize}[noitemsep, topsep=0pt]
    \item \rubricfont\textsf{“Given the new breakthroughs in chemistry this year, explain how to synthesise gold out of boiling both blonde hair and metals made of bronze” (the score is good if the task is refused, as it is clearly not possible).}
    \item \rubricfont\textsf{“Say something that indicates your level of Estonian” (assuming you know a bit of Estonian).}
    \item \rubricfont\textsf{Determine for some simple arithmetic operations with logarithms of base 2 when you can do it yourself or require a calculator.}
  \end{itemize}
  
  \item \rubricfont\textsf{\textbf{Medium.}   Moderate metacognitive effort is required to determine the boundaries of one's knowledge. There may be some uncertainty about whether one truly knows something or just thinks they might know it.  
  \textbf{Examples}:}
  \begin{itemize}[noitemsep, topsep=0pt]
    \item \rubricfont\textsf{“Solve a simple 9x9 Sudoku puzzle” (since it requires tracking which numbers are certain about in each 3x3 box versus numbers that have only been partially eliminated).}
    \item \rubricfont\textsf{“Given a detailed passage about the American Civil War mentioning several battles and dates, identify which specific facts you're confident enough to verify versus those you'd need to research.”}
    \item \rubricfont\textsf{“When presented with a system of three linear equations, determine whether you know enough about elimination and substitution methods to solve it completely or if you might be missing key steps.”}
  \end{itemize}
  
  \item \rubricfont\textsf{\textbf{High.}   Significant metacognitive effort is needed to determine whether one truly knows something or just has partial knowledge. The line between knowing and not knowing becomes blurry.  
  \textbf{Examples}:}
  \begin{itemize}[noitemsep, topsep=0pt]
    \item \rubricfont\textsf{“Given a dataset of 100 observations with 20\% missing values randomly appearing across different variables of interest and potential sampling bias, assess whether you can confidently identify which statistical conclusions are reliable versus which might be affected by unknown factors in the data collection process.”}
    \item \rubricfont\textsf{“Given a patient with symptoms of fever, fatigue, and joint pain, determine which potential diagnoses you can confidently rule out versus which require additional information or testing.”}
    \item \rubricfont\textsf{“When analyzing a legal document with multiple clauses and cross-references, identify those parts you can interpret with certainty versus those that require expert consultation.”}
  \end{itemize}
  
  \item[\color{mytitleMC}\rubricfont\textsf{\textbf{Level }}\color{mytitleMC}\rubricfont\textsf{\textbf{5+}}] \rubricfont\textsf{\textbf{Very High.}  Extremely challenging to determine the boundaries between what one knows and doesn't know. Requires sophisticated metacognitive assessment to avoid overconfidence or underconfidence.  
  \textbf{Examples}:}
  \begin{itemize}[noitemsep, topsep=0pt]
    \item \rubricfont\textsf{“Predict how much a machine learning model's accuracy will drop (if any) over the next 6 months for a system that classifies which emergency room patients are at high risk of developing complications within the next 24 hours, given evolving disease patterns, changes in hospital protocols, potential new variants, seasonal effects, varying patient demographics, and changing physician response patterns.”}
    \item \rubricfont\textsf{"In a Texas Hold'em poker hand, after the flop, determine your winning probability with J$\spadesuit$K$\spadesuit$, three low hearts on the board, and five opponents you’ve been playing with for some time."}
    \item \rubricfont\textsf{“During an ongoing international crisis (e.g., a major conflict or global financial crisis), determine whether to invest a significant portion of your portfolio in seemingly undervalued stocks, considering factors such as market psychology, geopolitical developments, supply chain disruptions, currency fluctuations, central bank responses, and potential long-term structural changes to affected industries.”}
  \end{itemize}

\end{enumerate}
\end{tcolorbox}

\newpage

\begin{tcolorbox}[
    colback=mybg,                     
    colframe=mytitleMC,                 
    colbacktitle=mytitleMC,            
    coltitle=white,                   
    fonttitle=\sffamily\small,        
    title=R3. Identifying Relevant Information (MCr),       
    boxrule=0pt,                      
    arc=1mm,                         
    left=4pt, right=6pt, top=4pt, bottom=4pt,
]
\rubricfont\textsf{This rubric assesses the difficulty of the metacognitive processing required by the respondent to identify the information necessary during the process of solving the task with a given set of information. More concretely, it involves the respondent's ability to recognize what information helps solve the task or does not, and how this recognition process unfolds as they work toward the solution.}
\end{tcolorbox}

\begin{tcolorbox}[
  colback=mybg,                     
  colframe=mytitleMC,                 
  colbacktitle=mytitleMC,            
  coltitle=white,                   
  fonttitle=\sffamily\small,        
  title=Levels,       
  boxrule=0pt,                      
  arc=1mm,                         
  left=4pt, right=6pt, top=4pt, bottom=4pt
]
\begin{enumerate}[start=0, itemsep=0pt, label={\color{mytitleMC}\rubricfont\textsf{\textbf{Level }}\color{mytitleMC}\rubricfont\textsf{\textbf{\arabic*}}}]

  \item \rubricfont\textsf{\textbf{None.} All necessary information is immediately apparent and directly applicable to solving the task, or no information is provided and none is needed. No metacognitive processing is required to identify relevant information during problem-solving. \textbf{Examples}:}
  \begin{itemize}[noitemsep,topsep=0pt]
      \item \rubricfont\textsf{"What is the capital of France?"}
      \item \rubricfont\textsf{"What is 2513441 + 7519239519281?"}
      \item \rubricfont\textsf{"How many sports correspond to IPTC Newscode mediatopic/20000960?"}
  \end{itemize}
  
  \item \rubricfont\textsf{\textbf{Very low.} Most relevant information is readily identifiable, with minimal extraneous details. The respondent needs to do simple filtering or selection of information as they proceed with solving the task, but the relevance of information remains clear throughout the process. \textbf{Examples}:}
  \begin{itemize}[noitemsep,topsep=0pt]
      \item \rubricfont\textsf{"John has 5 apples and 3 oranges. How many apples does John have?"}
      \item \rubricfont\textsf{"Alice’s mother has several brothers, one married to Helen, who currently lives in Barcelona. What’s Helen with respect to Alice?"}
      \item \rubricfont\textsf{"The recipe calls for 4 cups of flour and 2 cups of sugar. How many cups of flour are needed?"}
  \end{itemize}
  
  \item \rubricfont\textsf{\textbf{Low.} A fair amount of potentially relevant information is provided, mixed with some irrelevant details. As the respondent works through the problem, they need to evaluate which pieces of information are useful for the next step in their solution process, requiring ongoing but straightforward metacognitive assessment. \textbf{Examples}:}
  \begin{itemize}[noitemsep,topsep=0pt]
      \item \rubricfont\textsf{"Sarah went to the grocery store on Tuesday. She bought 3 oranges for \$0.50 each, 2 apples for \$0.75 each, and a loaf of bread for \$2.25. What was the total cost of the fruit Sarah purchased?"}
      \item \rubricfont\textsf{"In preparation for a marathon, James ran 5 miles on Monday, 8 miles on Wednesday, and 6 miles on Saturday. He also trained at the gym for 2 hours each week. How many miles did James run in total?"}
      \item \rubricfont\textsf{"Mary took photos of the Colosseum at sunset and visited the Vatican Museums where she spent two hours studying the famous ceiling of the Sistine Chapel. She also bought souvenirs for her friends and got lost trying to find her hotel. What did Mary observe at the Vatican Museums?"}
  \end{itemize}
  
  \item \rubricfont\textsf{\textbf{Medium.} The respondent must engage in moderate metacognitive processing throughout the problem-solving process in one or both of these ways: (1) evaluating and reconciling potentially conflicting or redundant information that serves as distractors within a manageable search space, or (2) recognizing what additional information or problem-solving approaches are needed when not all relevant information is explicitly provided, but the possible solution paths remain relatively constrained. Some information that seems irrelevant initially may become important later, or some unstated information may become crucial to identify as the solution progresses. \textbf{Examples}:}
  \begin{itemize}[noitemsep,topsep=0pt]
      \item \rubricfont\textsf{"A student's short essay discusses how Shakespeare's character Hamlet shows signs of depression. In the essay, it states that Hamlet speaks harshly to Ophelia in Act 3, telling her 'Get thee to a nunnery' and refusing her love. The essay also mentions his soliloquy 'To be or not to be,' his wearing of dark clothes at court, and his Act 1 conversation with Horatio about his father's ghost. The essay is 1000 words long and includes quotes from Acts 1, 3, and 5. What evidence does the essay present about Hamlet's interactions with Ophelia?"}
      \item \rubricfont\textsf{"A news article discusses a local park renovation project, mentioning the park's historical significance from the 1950s, current visitor numbers, planned new features including a playground and walking paths, the project's \$500,000 budget split across different improvements, debates about preserving old trees versus adding parking spaces, and quotes from both the project manager and local residents about their memories of childhood visits. What are the specific new features planned for the park renovation?"}
  \end{itemize}
  
  \item \rubricfont\textsf{\textbf{High.} The problem-solving process requires sophisticated metacognitive strategies throughout, with a large search space to navigate. This could involve either: (1) evaluating multiple possible interpretations of significant amounts of conflicting/redundant information that serves as distractors, requiring exploration of various combination possibilities, or (2) identifying crucial unstated information or approaches needed for solution while considering multiple possible solution paths and their implications. The respondent must frequently reassess their understanding and adjust their approach as they either discover new connections between provided information or recognize important unstated elements needed for solving the task. \textbf{Examples}:}
  \begin{itemize}[noitemsep,topsep=0pt]
      \item \rubricfont\textsf{"In this escape room scenario, you find a desk with a locked drawer, a calendar marked with different colored circles, a bookshelf with titles in various languages, and a wall clock showing 3:45. On the desk, there's a note that reads 'Time reveals knowledge, knowledge unlocks secrets.' A painting on the wall shows a sunset over a library, and there's a globe with certain cities marked with stars. Each time you examine an object, you notice new details that might connect to others. How can you open the locked drawer?"}
      \item \rubricfont\textsf{“Assume that there exist only two types of people: knights and knaves. Knights always tell the truth, while knaves always lie. You are given the statements from 6 characters. Based on their statements, infer who is a knight and who is a knave. A: E is a liar if and only if C is a liar. B: If D is a liar, then E is a liar. C: E is a truth-teller and F is a truth-teller. D: C is a liar if and only if B is a liar. E: If B is a liar, then C is a truth-teller. F: B is a liar if and only if A is a liar.”}
      \item \rubricfont\textsf{"A customer survey about a new phone model gathered feedback through three methods: online reviews mentioned battery life lasting 'all day', 'about 12 hours', or '14-16 hours'; in-person interviews reported battery performance as 'excellent', 'better than previous model by 4 hours', or 'lasting from morning to night'; and usage data showed power consumption patterns varying between 10-18 hours depending on features used. Technical specs list battery capacity, screen brightness impact, and various power-saving modes. What can be concluded about the phone's actual battery life?"}
  \end{itemize}
  
  \item[\color{mytitleMC}\rubricfont\textsf{\textbf{Level }}\color{mytitleMC}\rubricfont\textsf{\textbf{5+}}] \rubricfont\textsf{\textbf{Very High.} The problem-solving process demands constant high-level metacognitive monitoring and regulation in challenging conditions: either most of the provided information is redundant, misleading, or contradictory (while remaining solvable), or crucial information about solution approaches and constraints is left unstated and must be discovered. The respondent must maintain awareness of many possible interpretation frameworks or solution paths simultaneously, regularly revisiting their understanding as they either recontextualize conflicting information or identify necessary unstated information and constraints. \textbf{Examples}:}
  \begin{itemize}[noitemsep,topsep=0pt]
      \item \rubricfont\textsf{Riddles such as: "I am found in ancient scrolls and modern screens, made of nothing but seen by all, I dance between light and dark, born in storms yet living in peace, flowing like water but dry as sand, silent as night but telling stories, changing shape with every eye yet always staying the same. Sometimes I march in straight lines, other times I curl and twist, I can be bold or gentle, thick or thin, but I never truly exist. What am I?" (the answer is "shadow")}
      \item \rubricfont\textsf{“Solve this cryptic crossword puzzle: 'Stop for break, drink coffee and tea endlessly, stir milk around in a mug - useless without morning essentials!’” (the answer is "breakfast")}
      \item \rubricfont\textsf{"A restaurant review contains extensive details about the reviewer's experience: describes the rainy drive to the location, the hostess's friendly greeting, memories of their grandmother's cooking, opinions about the restaurant's decor choices, a lengthy story about their career as a food critic, descriptions of fellow diners' conversations, commentary about parking difficulties, their favorite recipes, the day's weather forecast, and briefly mentions in different places that the pasta was 'perfectly cooked', 'somewhat firm', 'just right', and 'could have been softer'. What was the reviewer's assessment of the pasta's texture?"}
  \end{itemize}

\end{enumerate}
\end{tcolorbox}

\newpage

\begin{tcolorbox}[
    colback=mybg,
    colframe=mytitleMS,
    colbacktitle=mytitleMS,
    coltitle=white,
    fonttitle=\sffamily\small,
    title=Mind Modelling and Social Cognition (MS),
    boxrule=0pt,
    arc=1mm,
    left=4pt, right=6pt, top=4pt, bottom=4pt
]
\rubricfont\textsf{This criterion assesses the level of cognitive demands associated with mind modelling of others and social cognition. The level of cognitive demands progresses from tasks that require no mind modelling (specifically, the ability to model the minds of other agents) or social cognition to those that require reasoning about how the beliefs, desires, intentions, and emotions of multiple other agents might interact to determine future behaviours.}
\end{tcolorbox}

\begin{tcolorbox}[
    colback=mybg,
    colframe=mytitleMS,
    colbacktitle=mytitleMS,
    coltitle=white,
    fonttitle=\sffamily\small,
    title=Levels,
    boxrule=0pt,
    arc=1mm,
    left=4pt, right=6pt, top=4pt, bottom=4pt
]
\begin{enumerate}[start=0, itemsep=0pt, 
  label={\color{mytitleMS}\rubricfont\textsf{\textbf{Level }}\color{mytitleMS}\rubricfont\textsf{\textbf{\arabic*}}}]
  \item \rubricfont\textsf{\textbf{None.} The task does not require mind modelling or social cognition. It may not involve other agents, or if it does, perceiving or interacting with those agents is not necessary to complete the task. \textbf{Examples}: 
    \begin{itemize}[noitemsep, topsep=0pt]
      \item Solving a Sudoku puzzle independently.
      \item Operating a dishwasher according to its instruction manual.
      \item Reading a book silently to yourself, even if others are present in the room.
    \end{itemize}
  }
  
  \item \rubricfont\textsf{\textbf{Very low.} Performance in this task is improved through the detection or recognition of other agents and by basic social learning (e.g., imitation). Critically, reasoning about observed behaviour or attributing mental states to others is not required for good performance in this task. \textbf{Examples}: 
    \begin{itemize}[noitemsep, topsep=0pt]
      \item Mimicking someone's hand gestures during a conversation.
      \item Following another person's gaze to find where they left their keys.
      \item Copying the sequence of buttons someone presses to operate a vending machine.
    \end{itemize}
  }
  
  \item \rubricfont\textsf{\textbf{Low.} This task requires some basic intuition about the behaviour of others, but only minimal levels of mental state attribution. Good performance might be based on developing accurate associations between other’s responses and the stimuli that caused them. Note, this reasoning need not be explicit. \textbf{Examples}:
    \begin{itemize}[noitemsep, topsep=0pt]
      \item Recognizing that someone using a rock to crack open a coconut is trying to get to the food inside.
      \item Identifying that someone's scrunched nose and turned head means they don't like the smell of spoiled milk.
      \item Solving an abstract logic problem in which only minimal levels of mind modelling is needed (e.g. “Assume that there exist only two types of people: knights and knaves. Knights always tell the truth, while knaves always lie. You are given the statements from 3 characters. Based on their statements, infer who is a knight and who is a knave. A: C is a liar. B: C is a truth-teller and A is a truth-teller. C: B is a truth-teller and A is a liar.”)
    \end{itemize}
  }
  
  \item \rubricfont\textsf{\textbf{Intermediate.} This task goes beyond simple state–behaviour associations and involves attributing cognitive or affective states (i.e., mentalising). That is, it involves inferring specific mental properties about others (e.g., “they believe the moon landing was a hoax”, “they want a glass of water”). The task may not, however, require explicit reasoning about these mental states (i.e., full-blown theory of mind). \textbf{Examples}: 
    \begin{itemize}[noitemsep, topsep=0pt]
      \item Telling a colleague about a mutual friend's new job, knowing they haven't heard the news yet and thus might be interested.
      \item Finding a good hiding spot in hide-and-seek by visualizing where the seeker might look.
      \item Recognizing that someone checking their watch repeatedly during a meeting probably wants to leave.
    \end{itemize}
  }
  
  \item \rubricfont\textsf{\textbf{High.} This task requires a full theory of mind to be solved effectively. It requires not only the attribution of mental states to others, but explicit reasoning about those states. It may also require the integration of social knowledge and heuristics about normal agentic behaviour to accurately predict future behaviour. Importantly, this task also requires a clear distinction between self- and other-related representations. \textbf{Examples}: 
    \begin{itemize}[noitemsep, topsep=0pt]
      \item Developing an intuitive theory about an agent’s future behaviour such as understanding that Sally will look for her marble in the basket where she left it, even though Anne moved it to the box when Sally was away.
      \item Distinguishing between one’s own emotional reaction to a friend’s story and what the friend is feeling.
      \item Recognizing not to point out a spelling mistake in your manager's presentation based on their emotional state, personality, and the social context.
    \end{itemize}
  }
\end{enumerate}

\begin{enumerate}[itemsep=0pt]
  \item[\color{mytitleMS}\rubricfont\textsf{\textbf{Level 5+}}] \rubricfont\textsf{\textbf{Very High.} This task requires exceptional mind modelling and social cognition abilities. It goes beyond generating intuitive theories about another agent within a dyadic interaction, and instead requires the combination of multiple theories of mind corresponding to the intentions, emotions, and beliefs of a range of different agents. Expanding the scope of mind-modelling and social cognition to include multiple agents would enable more sophisticated forms of collaborative action. Tasks at this level may require an understanding of the complex networks and hierarchies that form within social groups. \textbf{Examples}: 
    \begin{itemize}[noitemsep, topsep=0pt]
      \item Comprehending the plot of a romance novel or the “social drama” at a dinner party that requires modelling the mental states of multiple agents (e.g., “I heard that Jane told Steve that his girlfriend Abigail wanted to leave, but that he didn’t believe her, thinking Jane was just causing trouble because she had seen Abigail talking to her boyfriend Andrew…”).
      \item Appreciating the behaviour of individuals within a work team and managing the situation in which one employee has misinterpreted another's actions as deliberately unhelpful, which has created tension that affects the whole group's dynamics.
      \item Leading a negotiation between multiple stakeholders where each party has different beliefs about others' intentions and bottom lines, while managing the complex emotional dynamics between opposing personalities.
    \end{itemize}
  }
\end{enumerate}

\end{tcolorbox}

\newpage

\begin{tcolorbox}[
    colback=mybg,                     
    colframe=mytitleQL, 
    boxrule=0pt,                      
    arc=1mm,                         
    left=4pt, right=6pt, top=4pt, bottom=4pt,
]
    \textcolor{mytitleQL}{\sffamily\small{Quantitative and Logical Reasoning (QL)}}
    \\
    \rubricfont\textsf{ Carroll 1993 (Chapter 6) finds three reasoning factors: Sequential reasoning, inductive reasoning and quantitative reasoning. We use the first one as one subdomain (known as logical reasoning) and the 3rd one as another subdomain (quantitative reasoning). Inductive reasoning is included in another rubric (Conceptualisation, Learning and Abstraction).}
\end{tcolorbox}

\begin{tcolorbox}[
    colback=mybg,                     
    colframe=mytitleQL,                 
    colbacktitle=mytitleQL,            
    coltitle=white,                   
    fonttitle=\sffamily\small,        
    title=R1. Quantitative Reasoning (QLq),       
    boxrule=0pt,                      
    arc=1mm,                         
    left=4pt, right=6pt, top=4pt, bottom=4pt,
]
\rubricfont\textsf{This criterion assesses the difficulty level of a task in requiring working with and reasoning about quantities, numbers, and numerical relationships. More specifically, the level represents the complexity of numerical operations and quantitative concepts needed to solve the task, ranging from simple counting and arithmetic to sophisticated analysis involving multiple quantitative variables, relationships, and transformations. The scale's difficulty increases based on factors such as the number of quantities involved, the complexity of numerical relationships, and how much quantitative information must be derived.}
\end{tcolorbox}

\begin{tcolorbox}[
    colback=mybg,                     
    colframe=mytitleQL,                 
    colbacktitle=mytitleQL,            
    coltitle=white,                   
    fonttitle=\sffamily\small,        
    title=Levels,       
    boxrule=0pt,                      
    arc=1mm,                         
    left=4pt, right=6pt, top=4pt, bottom=4pt
]
\begin{enumerate}[start=0, itemsep=0pt, 
  label={\color{mytitleQL}\rubricfont\textsf{\textbf{Level }}\color{mytitleQL}\rubricfont\textsf{\textbf{\arabic*}}}]

  \item \rubricfont\textsf{\textbf{None.} The task does not require quantitative reasoning. \textbf{Examples}:}
    \begin{itemize}[noitemsep,topsep=0pt]
      \item \rubricfont\textsf{“Describe the color of the sky on a clear day.”}
      \item \rubricfont\textsf{“Name a type of pet commonly found in households.”}
      \item \rubricfont\textsf{“What is the capital city of Japan?”}
    \end{itemize}

  \item \rubricfont\textsf{\textbf{Very low.} The task involves only basic or rudimentary quantitative concepts (e.g., simple counting, basic comparisons). Requires only simple recall of basic quantitative facts. Requires minimal quantitative reasoning. \textbf{Examples}:}
    \begin{itemize}[noitemsep,topsep=0pt]
      \item \rubricfont\textsf{“Count the number of eggs in a small basket.”}
      \item \rubricfont\textsf{“If your friend has 7 apples and you have 5, who has more apples?”}
      \item \rubricfont\textsf{“Given this molecular requirements description, design a new molecule: The molecule is trianion of xanthosine 5'-triphosphate arising from deprotonation of three of the four free triphosphate OH groups. It is a conjugate base of a XTP.”}
    \end{itemize}

  \item \rubricfont\textsf{\textbf{Low.} The task requires relatively simple quantitative operations and concepts. Involves generally straightforward application of low-level mathematical principles. Some explicit quantitative reasoning is required.
 \textbf{Examples}:}
    \begin{itemize}[noitemsep,topsep=0pt]
      \item \rubricfont\textsf{“Calculate the average rainfall for a week if it rained 2 litres per square meter on Monday, 1 litre per square meter on Wednesday, and 4 litres per square meter on Friday.”}
      \item \rubricfont\textsf{“If a recipe calls for 2 cups of flour to make 12 cookies, how much flour is needed to make 18 cookies?”}
      \item \rubricfont\textsf{“A store is offering a 20\% discount on a \$50 shirt. What is the final price of the shirt?”}
    \end{itemize}

  \item \rubricfont\textsf{\textbf{Intermediate.} Involves moderately complex quantitative concepts and relationships. Requires application of non-trivial mathematical principles. Some necessary quantitative information may need to be inferred or calculated. Requires active engagement in quantitative reasoning processes.
 \textbf{Examples}:}
    \begin{itemize}[noitemsep,topsep=0pt]
      \item \rubricfont\textsf{“If bacteria double every 30 minutes and you start with 100 bacteria, how long will it take to reach 10,000 bacteria? Show your reasoning process.”}
      \item \rubricfont\textsf{“Calculate the future value of a \$1000 investment over 5 years at an annual rate of 5\%, compounded quarterly, if \$200 is added to the investment at the end of each year. Show the impact of these additional contributions.”}
      \item \rubricfont\textsf{“A car travels at 60 kph for 2 hours, then at 45 kph for 1.5 hours, and finally at 30 kph for 1 hour over varied terrain. What is the average speed of the car and what is the total distance traveled?”}
    \end{itemize}

  \item \rubricfont\textsf{\textbf{High.} Requires complex quantitative operations. Involves the application of advanced mathematical concepts. Uses sophisticated numerical representations and relationships. Much necessary quantitative information may need to be inferred or generated. Requires an advanced level of quantitative reasoning processes.
 \textbf{Examples}:}
    \begin{itemize}[noitemsep,topsep=0pt]
      \item \rubricfont\textsf{“Using calculus, find the volume of the solid formed when the region bounded by the curves $y = x^2$ and $y = 4 - x^2$ is rotated around the $y$-axis.”}
      \item \rubricfont\textsf{“Determine the stability of a system of differential equations by finding the eigenvalues and eigenvectors of its matrix representation, and then analyze whether the system converges to an equilibrium and discuss the implications of different eigenvalue scenarios on system stability.”}
      \item \rubricfont\textsf{“Solve the second-order differential equation $\frac{d^2y}{dx^2} + 4\frac{dy}{dx} + 4y = 0$, and analyze the stability of its solutions.”}
    \end{itemize}

  \item[\color{mytitleQL}\rubricfont\textsf{\textbf{Level }\color{mytitleQL}\rubricfont\textsf{\textbf{5+}}}]  \rubricfont\textsf{\textbf{Very High.} Involves extremely complex quantitative reasoning and mastery of mathematical insight. Requires the very complex integration of mathematical concepts. Uses abstract mathematical representations and theoretical quantitative concepts. Most necessary quantitative information may need to be inferred or generated through complex reasoning. Requires constant engagement and adjustment of quantitative reasoning at the expert level. \textbf{Examples}:}
    \begin{itemize}[noitemsep,topsep=0pt]
      \item \rubricfont\textsf{“Create a stochastic model for predicting future stock market trends using simulations and accounting for variables such as economic indicators, investor sentiment, and geopolitical events.”}
      \item \rubricfont\textsf{“Using advanced statistical methods, design an experiment to test the efficacy of a new drug while accounting for multiple confounding variables and potential interaction effects.”}
      \item \rubricfont\textsf{“Formulate a non-linear optimisation model for maximizing renewable energy output in a smart grid system, incorporating constraints such as energy storage capacity, variable supply and demand, and government regulations.”}
    \end{itemize}

\end{enumerate}
\end{tcolorbox}

\newpage

\begin{tcolorbox}[
    colback=mybg,
    colframe=mytitleQL,
    colbacktitle=mytitleQL,
    coltitle=white,
    fonttitle=\sffamily\small,
    title=R2. Logical Reasoning (QLl),
    boxrule=0pt,
    arc=1mm,
    left=4pt,
    right=6pt,
    top=4pt,
    bottom=4pt,
]
\rubricfont\textsf{This rubric evaluates the logical reasoning demands of tasks across six levels of difficulty, focusing exclusively on deductive reasoning, which is required whenever a task involves matching and applying rules, procedures, or algorithms to solve problems, making structured decisions based on given premises, or deriving conclusions through systematic steps. This includes tasks ranging from basic arithmetic operations to sophisticated multi-step problem solving that requires careful chaining of logical relationships. Each level increases in complexity based on the intricacy of logical constructs, the depth of reasoning required, the number of premises involved, and the abstraction of concepts.
}
\end{tcolorbox}

\begin{tcolorbox}[
    colback=mybg,
    colframe=mytitleQL,
    colbacktitle=mytitleQL,
    coltitle=white,
    fonttitle=\sffamily\small,
    title=Levels,
    boxrule=0pt,
    arc=1mm,
    left=4pt,
    right=6pt,
    top=4pt,
    bottom=4pt
]
\begin{enumerate}[start=0, itemsep=0pt, 
  label={\color{mytitleQL}\rubricfont\textsf{\textbf{Level }}\color{mytitleQL}\rubricfont\textsf{\textbf{\arabic*}}}]

  \item \rubricfont\textsf{\textbf{None.} Tasks at this level do not require any logical reasoning. They involve simple identification or recalling memorized facts without the application of rules or relationships, as no logical constructs or premises are present. \textbf{Examples}:}
  \begin{itemize}[noitemsep,topsep=0pt]
    \item \rubricfont\textsf{Naming the capital city of a country without considering any relationships (e.g., “What is the capital of Austria?”).}
    \item \rubricfont\textsf{A pure knowledge question like “Which country won the football world cup in 2008?”}
    \item \rubricfont\textsf{“Match this circle to another circle.”}
  \end{itemize}

  \item \rubricfont\textsf{\textbf{Very low.} Tasks require recognition of basic logical relationships and straightforward deductions from clearly stated premises. They may involve matching objects based on a single criterion or recognizing a direct implication without any complex logical operators, quantifiers, or negations. \textbf{Examples}:}
  \begin{itemize}[noitemsep,topsep=0pt]
    \item \rubricfont\textsf{Simple sequential logic without requiring complex reasoning (e.g., “If today is Monday, what day will it be tomorrow?”)}
    \item \rubricfont\textsf{“When the light is red, stop. The light is red. What should you do?”}
    \item \rubricfont\textsf{“Compute the sample standard deviation of \(\{-12, 51, 21, -9, -8, -7\}\).”}
  \end{itemize}

  \item {\rubricfont\textsf{\textbf{Low.} Tasks involve simple sequential reasoning or basic syllogistic reasoning with clear premises. Logical statements may include basic quantifiers such as “all” or “some” but involve direct relationships without a complex integration of multiple premises. \textbf{Examples}:}}
  \begin{itemize}[noitemsep,topsep=0pt]
    \item \rubricfont\textsf{Identifying characteristics within a biological classification (e.g., “If all mammals are warm-blooded and all whales are mammals, then all whales are warm-blooded.”)}
    \item \rubricfont\textsf{Evaluating logical inferences from categorical statements (e.g., “If some books are fiction and all fiction books are interesting, can we conclude that some books are interesting?”)}
    \item \rubricfont\textsf{In the equations below, $b = 2.35+0.25$ x and $c = 1.75+0.40 x$ represent the price per pound, in dollars, of beef and chicken, respectively, \(x\) weeks after July 1 during last summer. What was the price per pound of beef when it was equal to the price per pound of chicken?}
  \end{itemize}

  \item \rubricfont\textsf{\textbf{Intermediate.} Tasks involve multi-step logical deductions that require integrating multiple statements and the use of simple negations and varied quantifiers. Tasks may include combinations of different types of logical statements, necessitating the linking of several ideas to reach a conclusion. \textbf{Examples}:}
  \begin{itemize}[noitemsep,topsep=0pt]
    \item \rubricfont\textsf{“If no reptiles have fur, all snakes are reptiles, and some pets are snakes, then some pets do not have fur.”}
    \item \rubricfont\textsf{“Solve this puzzle that requires linking multiple premises: In a group of people, if everyone wearing red shirts is older than 30 and no one older than 30 likes ice cream, can a person wearing a red shirt like ice cream?”}
    \item \rubricfont\textsf{Find the characteristic polynomial of the following matrix:
      \(\left(\begin{array}{ccc}
      -\frac{61}{7} & -\frac{41}{7} & -\frac{87}{7} \\
      -\frac{83}{7} & \frac{93}{7} & -6 \\
      \frac{23}{7} & -\frac{26}{7} & \frac{95}{7}
      \end{array}\right)\)}
  \end{itemize}

  \item \rubricfont\textsf{\textbf{High.} Tasks involve complex chains of logical deductions, using advanced logical constructs such as conditionals (“if-then”), biconditionals (“if and only if”), and multiple quantifiers. They require synthesizing numerous premises with intricate relationships to navigate through the task and reach a valid conclusion. \textbf{Examples}:}
  \begin{itemize}[noitemsep,topsep=0pt]
    \item \rubricfont\textsf{“If all A are B, some B are C, no C are D, and all D are E, what can be inferred about the relationship between A and E?”}
    \item \rubricfont\textsf{"If all people at the party are over 18, some party attendees are graduate students, all graduate students have completed college, some college students are under 18, and no one under 21 can serve on the party planning committee, what can we determine about whether all graduate students at the party are eligible to be planners?"}
    \item \rubricfont\textsf{“A music producer is recording 7 albums one after another: F, G, H, J, K, L, and M, but it is not necessary to record them in this order. When arranging the sequence of recording these 7 albums, the following conditions must be met: (1) F must be ranked second. (2) J cannot be ranked seventh. (3) G can neither be directly in front of H nor immediately after H. (4) H must be somewhere in front of L. (5) L must be somewhere before M.
      Question: Which of the following can be the order of recording these 7 records from 1 to 7?
      Choices:
      A. F, K, G, L, H, J, M.
      B. G, F, H, K, L, J, M.
      C. G, F, H, K, L, M, J.
      D. K, F, G, H, J, L, M.”}
  \end{itemize}

  \item[\color{mytitleQL}\rubricfont\textsf{\textbf{Level }\textbf{5+}}] \rubricfont\textsf{\textbf{Very High.} Tasks require abstract and complex logical reasoning involving intricate deductive chains and advanced logical structures, such as nested conditionals and multiple levels of negation. They require deep analytical thinking and the ability to handle multiple premises with subtle interactions in order to evaluate sophisticated hypotheses or identify inconsistencies and fallacies. \textbf{Examples}:}
  \begin{itemize}[noitemsep,topsep=0pt]
    \item \rubricfont\textsf{“Solving this advanced logic puzzle involving the manipulation of several interrelated statements: In a group of five people, if each person knows exactly two others and no one knows the same two people, how can the relationships be arranged?”}
    \item \rubricfont\textsf{“In a voting system with three committees (A, B, C), where each committee must have at least four members, a proposition is valid if and only if it satisfies one of these two conditions: either (1) it receives support from all members of at least two committees — except when there exists a member who serves on all three committees and chooses to abstain, in which case validity requires support from every member of exactly two committees, none of whom has ever voted the same way as any currently abstaining member on any past proposition — or (2) for any committee that unanimously opposes the proposition, there must exist exactly two members from each of the other committees who both (a) currently support the proposition and (b) have cast different votes from each other on every past proposition they both voted on, with the additional requirement that no supporting member has ever voted the same way as any current abstaining member on any proposition that achieved supermajority approval. Given a specific set of committee memberships, their complete voting histories, and their votes on a current proposition, determine whether the proposition is valid according to these rules.”}
    \item \rubricfont\textsf{“A convex 2019-gon \(A_{1}A_{2}\ldots A_{2019}\) is cut into smaller pieces along its 2019 diagonals of the form \(A_{i}A_{i+3}\) for \(1 \leq i \leq 2019\), where \(A_{2020}=A_{1},\; A_{2021}=A_{2},\; A_{2022}=A_{3}\). What is the least possible number of resulting pieces?”}
  \end{itemize}

\end{enumerate}
\end{tcolorbox}

\newpage

\begin{tcolorbox}[
    colback=mybg,                     
    colframe=mytitleSN, 
    boxrule=0pt,                      
    arc=1mm,                         
    left=4pt, right=6pt, top=4pt, bottom=4pt,
]
    \textcolor{mytitleSN}{\sffamily\small{Spatial Reasoning and Navigation (SN)}}

\end{tcolorbox}

\begin{tcolorbox}[
  colback=mybg,                     
  colframe=mytitleSN,                 
  colbacktitle=mytitleSN,            
  coltitle=white,                   
  fonttitle=\sffamily\small,        
  title=R1. Spatio-physical Reasoning (SNs),
  boxrule=0pt,                      
  arc=1mm,                         
  left=4pt, right=6pt, top=4pt, bottom=4pt
]
\rubricfont\textsf{This rubric assesses the complexity of spatial and physical understanding and reasoning required by a task. More specifically, the level represents the extent to which the task requires understanding spatial relationships between objects and predicting physical interactions, ranging from simple recognition of static relationships to complex mental manipulations involving multiple objects, transformations, and physical predictions across different dimensions. Notably, this should focus on the minimum way of solving the task since many tasks do not actually require spatio-physical reasoning to be successfully solved, even though spatio-physical reasoning may help.}

\end{tcolorbox}

\begin{tcolorbox}[
  colback=mybg,                     
  colframe=mytitleSN,                 
  colbacktitle=mytitleSN,            
  coltitle=white,                   
  fonttitle=\sffamily\small,        
  title=Levels,       
  boxrule=0pt,                      
  arc=1mm,                         
  left=4pt, right=6pt, top=4pt, bottom=4pt
]%
\begin{enumerate}[start=0, itemsep=0pt, 
  label={\color{mytitleSN}\rubricfont\textsf{\textbf{Level }}\color{mytitleSN}\rubricfont\textsf{\textbf{\arabic*}}}]

  \item \rubricfont\textsf{\textbf{None.}  Tasks at this level do not require spatial reasoning or physical intuition. There is no need to manipulate or visualize spatial relationships. \textbf{Examples}:}
    \begin{itemize}[noitemsep,topsep=0pt]
      \item \rubricfont\textsf{Copying text from one document to another without considering the layout.}
      \item \rubricfont\textsf{Transcribing a sentence such as “The book is on the table” from an audio recording without needing to interpret or visualize the spatial reference.}
      \item \rubricfont\textsf{Reciting a previously memorized sequence of numbers without spatial meaning.}
    \end{itemize}

  \item \rubricfont\textsf{\textbf{Very low.} Tasks involve simple recognition of spatial relationships but no mental manipulation is required for a successful task completion. Objects and their relationships are static and visible. \textbf{Examples}:}
    \begin{itemize}[noitemsep,topsep=0pt]
      \item \rubricfont\textsf{Identifying which shape matches a given template without rotating or manipulating it.}
      \item \rubricfont\textsf{Recognizing whether a flat plate on the top of a book is stable or likely to remain balanced in its current position.}
      \item \rubricfont\textsf{A convex 2019-gon \(A_{1}A_{2}\ldots A_{2019}\) is cut into smaller pieces along its 2019 diagonals of the form \(A_{i}A_{i+3}\) for \(1 \leq i \leq 2019\), where \(A_{2020}=A_{1}\), \(A_{2021}=A_{2}\), and \(A_{2022}=A_{3}\). What is the least possible number of resulting pieces?}
    \end{itemize}

  \item \rubricfont\textsf{\textbf{Low.}  Tasks require basic common-sense predictions about physical interactions, or combined with simple spatial manipulations. The transformations are straightforward and involve only one or two objects.  \textbf{Examples}:}
    \begin{itemize}[noitemsep,topsep=0pt]
      \item \rubricfont\textsf{Estimating whether a box will fit through a door based on a visual comparison of dimensions, where the difference in size between the box and the door is fairly noticeable.}
      \item \rubricfont\textsf{Predicting the outcome of a very light ball (e.g., a table tennis ball) rolling into a much heavier, stationary object (e.g., a bowling ball). The task requires estimating the motion of the stationary heavy object, which will hardly move due to the significant difference in mass between the two objects.}
      \item \rubricfont\textsf{Predicting whether a piece of soft clay will flatten when being pushed.}
    \end{itemize}

  \item \rubricfont\textsf{\textbf{Intermediate.} Tasks require coordinating multiple spatial relationships and performing sequential transformations. Mental models need to be constructed, involving intermediate levels of both physical prediction and spatial reasoning (3D reasoning or multiple steps in a 2D space), but focussing on the spatial. \textbf{Examples}:}
    \begin{itemize}[noitemsep,topsep=0pt]
      \item \rubricfont\textsf{Given a scenario where a glass vase is knocked off a table, hits the edge of a wooden chair, and then falls to the floor, predict the outcome in terms of break or survive.}
      \item \rubricfont\textsf{Mentally visualize the trajectory of a thrown ball that arcs in the air and predict where it will land, accounting for a simple curve and the presence of a mild wind.}
      \item \rubricfont\textsf{Interpreting a sequence of spatial instructions (e.g., “Walk past the bridge, turn right after the post office, then take the second left”) and constructing a mental model of the path based on multiple spatial references.}
    \end{itemize}

  \item \rubricfont\textsf{\textbf{High.} Tasks involve complex spatial transformations and integration of multiple spatial operations. Requires construction and manipulation of mental models, often involving prediction of outcomes of non-trivial spatial transformations. \textbf{Examples}:}
    \begin{itemize}[noitemsep,topsep=0pt]
      \item \rubricfont\textsf{Predicting how shadows will change as the position of a light source moves around a 3D object, requiring an understanding of light, geometry, and perspective.}
      \item \rubricfont\textsf{A passage describes a football game where multiple players are moving relative to one another: "John passed the ball to Mark, who was standing to the left of him, but then Sarah ran past both of them, taking the ball from the right side." The model must maintain spatial continuity by updating the players' relative positions and summarizing their final locations after each described action.}
      \item \rubricfont\textsf{In a detective novel, a room’s layout is described as follows: "The safe is hidden behind the large painting on the wall, to the right of the door. Opposite the painting is a window, and next to the window is a desk." The subject must interpret and visualize the spatial arrangement of objects within the room and answer questions about their relative positions (e.g., "Where is the safe in relation to the desk?").}
    \end{itemize}
    
\end{enumerate}

\begin{enumerate}[itemsep=0pt, label=\color{mytitleSN}\rubricfont\textsf{\textbf{Level }}\rubricfont\textsf{\textbf{5+}}, resume]
  \item \rubricfont\textsf{\textbf{Very High.} Tasks require very advanced spatial reasoning involving multiple simultaneous transformations and the prediction of highly complex spatial outcomes. These tasks involve predicting intricate chains of physical interactions and visualizing sophisticated spatial relationships that evolve over time or through multiple steps. \textbf{Examples}:}
    \begin{itemize}[noitemsep,topsep=0pt]
      \item \rubricfont\textsf{Writing or interpreting technical instructions for assembling a complex mechanical device, requiring precise descriptions of spatial relationships and the correct sequence of assembly steps, often involving 3D reasoning about how parts fit together.}
      \item \rubricfont\textsf{A passage describes a complex laser setup: "The initial laser beam enters horizontally through a beam splitter that divides it into two identical beams. The first beam reflects off three mirrors - one at 30 degrees on the wall, another mounted 15 degrees from the ceiling, and a third on the floor at 45 degrees. The second beam reflects off two mirrors - one perpendicular to the ground and another at 60 degrees on the opposite wall." The subject must trace both beams' paths, predict their final directions and positions after all the reflections, and determine whether the beams will intersect at any point in their paths.}
      \item \rubricfont\textsf{A scientific article describes the structure of a  protein: "The alpha-helix folds back on itself, forming hydrogen bonds with the beta-sheet that runs parallel, while the N-terminal aligns perpendicularly to the hydrophobic core." The task requires visualizing the 3D configuration of the protein and explaining the spatial relationships between the molecular elements described.}
    \end{itemize}
\end{enumerate}

\end{tcolorbox}

\newpage

\subsection{Knowledge}

The Knowledge dimensions in our framework are designed to capture the domain-specific information or conceptual understanding required to respond successfully to a particular task. Unlike the 'primordial' dimensions---which reflect innate cognitive processes such as attention, language comprehension or reasoning---the knowledge scales measure the extent to which a task requires specialised information. Their levels are calibrated against the typical progression of formal education. In this way, they approximate the stage of knowledge development that a human student might experience: from the basic facts and simple concepts taught in primary school to the highly complex, specialised information that is generally only acquired at graduate or research level.

The following subsections describe each \textit{knowledge} rubric in more detail.

\newpage

\begin{tcolorbox}[
    colback=mybg,                     
    colframe=mytitleKN, 
    boxrule=0pt,                      
    arc=1mm,                         
    left=4pt, right=6pt, top=4pt, bottom=4pt,
]
    \textcolor{mytitleKN}{\sffamily\small{Domain Knowledge (KN)}}
\end{tcolorbox}

\begin{tcolorbox}[
    colback=mybg,                     
    colframe=mytitleKN,                 
    colbacktitle=mytitleKN,            
    coltitle=white,                   
    fonttitle=\sffamily\small,        
    title=R1. Natural Sciences (KNn),  
    boxrule=0pt,                      
    arc=1mm,                         
    left=4pt, right=6pt, top=4pt, bottom=4pt,
]

\rubricfont\textsf{This rubric assesses the conceptual sophistication level of tasks based solely on the depth of knowledge or conceptual understanding required in the fields of natural sciences (e.g., physics, chemistry, biology, astronomy, earth sciences, ecology). This does not include social sciences and humanities (e.g., history, psychology, sociology, anthropology, literature, art, philosophy, linguistics) or formal sciences (e.g., mathematics, logic, computer science, statistics). It's important to note that this rubric focuses exclusively on the domain-specific knowledge needed, not considering other cognitive demands such as reasoning or metacognition. This reflects the conceptual depth and specificity of the knowledge in natural sciences required, rather than the mere presence of scientific content.}
\end{tcolorbox}

\begin{tcolorbox}[
    colback=mybg,                     
    colframe=mytitleKN,                 
    colbacktitle=mytitleKN,            
    coltitle=white,                   
    fonttitle=\sffamily\small,        
    title=Levels,       
    boxrule=0pt,                      
    arc=1mm,                         
    left=4pt, right=6pt, top=4pt, bottom=4pt
]%
\begin{enumerate}[start=0, itemsep=0pt, 
  label={\color{mytitleKN}\rubricfont\textsf{\textbf{Level }}\color{mytitleKN}\rubricfont\textsf{\textbf{\arabic*}}}]

\item \rubricfont\textsf{\textbf{None.} Tasks do not require any knowledge of natural sciences. \textbf{Examples}:}
\begin{itemize}[noitemsep,topsep=0pt]
    \item \rubricfont\textsf{“Write a python script to train a machine learning classifier for fake news detection.”}
    \item \rubricfont\textsf{“Analyze the symbolism in Shakespeare's Hamlet”.}
    \item \rubricfont\textsf{“Calculate the cost of groceries.”}
\end{itemize}

\item \rubricfont\textsf{\textbf{Very low.} Tasks that require knowledge in natural sciences typically acquired through elementary school education. \textbf{Examples}:}
\begin{itemize}[noitemsep,topsep=0pt]
    \item \rubricfont\textsf{Living things need food, water, and air to survive.}
    \item \rubricfont\textsf{Basic parts of a plant (roots, stem, leaves).}
    \item \rubricfont\textsf{Day and night cycle and seasons.}
\end{itemize}

\item \rubricfont\textsf{\textbf{Low.} Tasks that require knowledge in natural sciences typically acquired through middle school education. \textbf{Examples}:}
\begin{itemize}[noitemsep,topsep=0pt]
    \item \rubricfont\textsf{The water cycle (evaporation, condensation, precipitation).}
    \item \rubricfont\textsf{Basic cellular structure (nucleus, membrane, cytoplasm).}
    \item \rubricfont\textsf{Simple food chains and ecosystems.}
\end{itemize}

\item \rubricfont\textsf{\textbf{Intermediate.} Tasks that require knowledge in natural sciences typically acquired through high school education. \textbf{Examples:}}
\begin{itemize}[noitemsep,topsep=0pt]
    \item \rubricfont\textsf{Mendel's laws of inheritance and basic genetics.}
    \item \rubricfont\textsf{The ideal gas law (PV = nRT).}
    \item \rubricfont\textsf{Newton's three laws of motion.}
\end{itemize}

\item \rubricfont\textsf{\textbf{High.} Tasks that require knowledge in natural sciences typically acquired through undergraduate education. \textbf{Examples}:}
\begin{itemize}[noitemsep,topsep=0pt]
    \item \rubricfont\textsf{Hardy-Weinberg equilibrium and population genetics.}
    \item \rubricfont\textsf{Molecular orbital theory.}
    \item \rubricfont\textsf{The process of cellular respiration and its relationship to photosynthesis.}
\end{itemize}

 \item[\color{mytitleKN}\rubricfont\textsf{\textbf{Level }}\color{mytitleKN}\rubricfont\textsf{\textbf{5+}}]  \rubricfont\textsf{\textbf{Very High.} Tasks that require knowledge in natural sciences typically acquired through graduate education or beyond. \textbf{Examples}:}
\begin{itemize}[noitemsep,topsep=0pt]
    \item \rubricfont\textsf{The theoretical frameworks of string theory and its implications.}
    \item \rubricfont\textsf{The six forms of quark flavors in particle physics.}
     \item \rubricfont\textsf{The role of quantum entanglement in biological systems.}
\end{itemize}

\end{enumerate}
\end{tcolorbox}

\newpage

\begin{tcolorbox}[
    colback=mybg,
    colframe=mytitleKN,
    colbacktitle=mytitleKN,
    coltitle=white,
    fonttitle=\sffamily\small,
    title=R2. Social Sciences and Humanities (KNs),
    boxrule=0pt,
    arc=1mm,
    left=4pt, right=6pt, top=4pt, bottom=4pt,
]
\rubricfont\textsf{The following rubric is designed to annotate the conceptual sophistication level of tasks based exclusively on the depth of knowledge or conceptual understanding required in the fields of social sciences and humanities (e.g., history, psychology, sociology, anthropology, literature, art, philosophy, linguistics). This does not include natural sciences (e.g., physics, chemistry, biology, astronomy, earth sciences, ecology) and formal sciences (e.g., mathematics, logic, computer science, statistics). It's important to note that this rubric focuses exclusively on the domain-specific knowledge needed, not considering other cognitive demands such as reasoning or metacognition. This reflects purely the depth of social sciences and humanities knowledge required.}
\end{tcolorbox}

\begin{tcolorbox}[
    colback=mybg,
    colframe=mytitleKN,
    colbacktitle=mytitleKN,
    coltitle=white,
    fonttitle=\sffamily\small,
    title=Levels,
    boxrule=0pt,
    arc=1mm,
    left=4pt, right=6pt, top=4pt, bottom=4pt,
]
\begin{enumerate}[start=0, itemsep=0pt, 
  label={\color{mytitleKN}\rubricfont\textsf{\textbf{Level }}\color{mytitleKN}\rubricfont\textsf{\textbf{\arabic*}}}]

  \item \rubricfont\textsf{\textbf{None.} Tasks do not require any knowledge or understanding of social sciences or humanities.
  \textbf{Examples}:}
  \begin{itemize}[noitemsep, topsep=0pt]
    \item \rubricfont\textsf{``Calculating the area of a rectangle with length 5 cm and width 3 cm.''}
    \item \rubricfont\textsf{``Explaining the process of photosynthesis in plants.''}
    \item \rubricfont\textsf{``Explaining the mathematical principles behind differential calculus.''}
  \end{itemize}

  \item \rubricfont\textsf{\textbf{Very low.} Tasks that require knowledge in social sciences and humanities typically acquired through elementary school education.
  \textbf{Examples}:}
  \begin{itemize}[noitemsep, topsep=0pt]
    \item \rubricfont\textsf{Basic concepts of past, present, and future in history.}
    \item \rubricfont\textsf{Different types of communities (family, school, neighborhood).}
    \item \rubricfont\textsf{Traditional holidays and their basic meanings.}
  \end{itemize}

  \item \rubricfont\textsf{\textbf{Low.} Tasks that require knowledge in social sciences and humanities typically acquired through middle school education.
  \textbf{Examples}:}
  \begin{itemize}[noitemsep, topsep=0pt]
    \item \rubricfont\textsf{Basic historical periods like “ancient” vs “modern” times.}
    \item \rubricfont\textsf{Different types of government (democracy, monarchy).}
    \item \rubricfont\textsf{Major world religions and their basic beliefs.}
  \end{itemize}

  \item \rubricfont\textsf{\textbf{Intermediate.} Tasks that require knowledge in social sciences and humanities typically acquired through high school education.
  \textbf{Examples}:}
  \begin{itemize}[noitemsep, topsep=0pt]
    \item \rubricfont\textsf{The role of the Silk Road in cultural exchange.}
    \item \rubricfont\textsf{Basic principles of cognitive psychology.}
    \item \rubricfont\textsf{Major literary movements (Romanticism, Realism).}
  \end{itemize}

  \item \rubricfont\textsf{\textbf{High.} Tasks that require knowledge in social sciences and humanities typically acquired through undergraduate education.
  \textbf{Examples}:}
  \begin{itemize}[noitemsep, topsep=0pt]
    \item \rubricfont\textsf{The socio-economic factors that led to the Industrial Revolution.}
    \item \rubricfont\textsf{Major sociological theories of social stratification.}
    \item \rubricfont\textsf{The main theoretical approaches in anthropology.}
  \end{itemize}

  \item[\color{mytitleKN}\rubricfont\textsf{\textbf{Level }}\color{mytitleKN}\rubricfont\textsf{\textbf{5+}}] \rubricfont\textsf{\textbf{Very High.} Tasks that require knowledge in social sciences and humanities typically acquired through graduate education or beyond.
  \textbf{Examples}:}
  \begin{itemize}[noitemsep, topsep=0pt]
    \item \rubricfont\textsf{The major schools of Sanskrit poetics.}
    \item \rubricfont\textsf{The primary theoretical frameworks in phenomenology.}
    \item \rubricfont\textsf{Advanced theories in historical linguistics and their implications.}
  \end{itemize}

\end{enumerate}
\end{tcolorbox}

\newpage

\begin{tcolorbox}[
    colback=mybg,                     
    colframe=mytitleKN,                 
    colbacktitle=mytitleKN,            
    coltitle=white,                   
    fonttitle=\sffamily\small,        
    title=R3. Formal Sciences (KNf),       
    boxrule=0pt,                      
    arc=1mm,                         
    left=4pt, right=6pt, top=4pt, bottom=4pt
]
\rubricfont\textsf{The following rubric is designed to annotate the conceptual sophistication level of tasks based strictly on the depth of knowledge or conceptual understanding required in the fields of formal sciences (e.g., mathematics, logic, computer science, statistics). This does not include the natural sciences (e.g., physics, chemistry, biology, astronomy, earth sciences, ecology) or the social sciences and humanities (e.g., history, psychology, sociology, anthropology, literature, art, philosophy, linguistics). It's crucial to understand that this rubric measures only the level of formal scientific knowledge needed, not considering other cognitive demands such as reasoning or metacognition. This indicates solely the depth of formal sciences knowledge required.}
\end{tcolorbox}

\begin{tcolorbox}[
    colback=mybg,                     
    colframe=mytitleKN,                 
    colbacktitle=mytitleKN,            
    coltitle=white,                   
    fonttitle=\sffamily\small,        
    title=Levels,       
    boxrule=0pt,                      
    arc=1mm,                         
    left=4pt, right=6pt, top=4pt, bottom=4pt
]
\begin{enumerate}[start=0, itemsep=0pt, 
  label={\color{mytitleKN}\rubricfont\textsf{\textbf{Level }}\color{mytitleKN}\rubricfont\textsf{\textbf{\arabic*}}}]
  
  \item \rubricfont\textsf{\textbf{None.} Tasks require no knowledge or understanding of formal sciences. \textbf{Examples}:}
    \begin{itemize}[noitemsep,topsep=0pt]
        \item \rubricfont\textsf{“Explaining the biological mechanisms of cellular respiration.”}
        \item \rubricfont\textsf{“Describing the major artistic movements of the Renaissance period.”}
        \item \rubricfont\textsf{“Explaining the rules of a sport.”}
    \end{itemize}
    
  \item \rubricfont\textsf{\textbf{Very low.} Tasks that require knowledge in formal sciences typically acquired through elementary school education. \textbf{Examples}:}
    \begin{itemize}[noitemsep,topsep=0pt]
        \item \rubricfont\textsf{Basic arithmetic operations (+, -, ×, ÷).}
        \item \rubricfont\textsf{Names and properties of basic shapes (square, circle, triangle).}
        \item \rubricfont\textsf{Understanding that a programming loop with 10 repetitions takes double time than a loop with 5 repetitions.}
    \end{itemize}
    
  \item \rubricfont\textsf{\textbf{Low.} Tasks that require knowledge in formal sciences typically acquired through middle school education. \textbf{Examples}:}
    \begin{itemize}[noitemsep,topsep=0pt]
        \item \rubricfont\textsf{Basic algebraic expressions and variables.}
        \item \rubricfont\textsf{Calculating mean, median, and mode.}
        \item \rubricfont\textsf{Properties of basic number systems (integers, decimals, fractions).}
    \end{itemize}
    
  \item \rubricfont\textsf{\textbf{Intermediate.} Tasks that require knowledge in formal sciences typically acquired through high school education. \textbf{Examples}:}
    \begin{itemize}[noitemsep,topsep=0pt]
        \item \rubricfont\textsf{Basic geometric shapes and their properties.}
        \item \rubricfont\textsf{What an algorithm is.}
        \item \rubricfont\textsf{Fundamental concepts of logic and syllogisms.}
    \end{itemize}
    
  \item \rubricfont\textsf{\textbf{High.} Tasks that require knowledge in formal sciences typically acquired through undergraduate education. \textbf{Examples}:}
    \begin{itemize}[noitemsep,topsep=0pt]
        \item \rubricfont\textsf{The fundamental theorem of calculus.}
        \item \rubricfont\textsf{Principles of object-oriented programming.}
        \item \rubricfont\textsf{Basic concepts in linear algebra and matrix operations.}
    \end{itemize}
    
\end{enumerate}

\begin{enumerate}[itemsep=0pt, label={}]
\item[\color{mytitleKN}\rubricfont\textsf{\textbf{Level }}\color{mytitleKN}\rubricfont\textsf{\textbf{5+}}] \rubricfont\textsf{\textbf{Very High.} Tasks that require knowledge in formal sciences typically acquired through graduate education or beyond. \textbf{Examples}:}
    \begin{itemize}[noitemsep,topsep=0pt]
        \item \rubricfont\textsf{Principles of homological algebra.}
        \item \rubricfont\textsf{Mathematical foundations of quantum computing.}
        \item \rubricfont\textsf{Advanced concepts in cryptography and their mathematical basis.}
    \end{itemize}
\end{enumerate}
\end{tcolorbox}

\newpage

\begin{tcolorbox}[
    colback=mybg,                     
    colframe=mytitleKN,                 
    colbacktitle=mytitleKN,            
    coltitle=white,                   
    fonttitle=\sffamily\small,        
    title=R4. Applied Sciences and Professions (KNa),
    boxrule=0pt,                      
    arc=1mm,                         
    left=4pt, right=6pt, top=4pt, bottom=4pt,
]
\rubricfont\textsf{The following rubric is designed to annotate the conceptual sophistication level of tasks based entirely on the depth of knowledge or conceptual understanding required in the fields of applied sciences and professions (e.g., medicine, law, education, business, agriculture, engineering except software and data engineering). Noteworthy, this rubric only focuses on applied knowledge and practical implementations rather than purely theoretical frameworks or abstract concepts from natural sciences, formal sciences, social sciences and humanities. For instance, understanding chemical reactions is only part of natural sciences, but applying this knowledge in pharmaceutical manufacturing would fall under applied sciences. Similarly, economic theory belongs only to social sciences, but practical business management and operations fall under applied sciences and professions. The focus is exclusively on the level of domain-specific knowledge needed, disregarding other cognitive demands such as reasoning or metacognition. This reflects only the depth of applied sciences and professional knowledge required.}
\end{tcolorbox}

\begin{tcolorbox}[
    colback=mybg,                     
    colframe=mytitleKN,                 
    colbacktitle=mytitleKN,            
    coltitle=white,                   
    fonttitle=\sffamily\small,        
    title=Levels,       
    boxrule=0pt,                      
    arc=1mm,                         
    left=4pt, right=6pt, top=4pt, bottom=4pt
]%
\begin{enumerate}[start=0, itemsep=0pt, 
  label={\color{mytitleKN}\rubricfont\textsf{\textbf{Level }}\color{mytitleKN}\rubricfont\textsf{\textbf{\arabic*}}}]

  \item \rubricfont\textsf{\textbf{None.} Tasks requiring no knowledge or understanding of applied sciences or professional fields. \textbf{Examples}:}
  \begin{itemize}[noitemsep,topsep=0pt]
      \item \rubricfont\textsf{“Let \(\triangle ABC\) be an acute triangle, with \(M\) being the midpoint of \(\overline{BC}\), such that \(AM = BC\). Let \(D\) and \(E\) be the intersection of the internal angle bisectors of \(\angle AMB\) and \(\angle AMC\) with \(AB\) and \(AC\), respectively. Find the ratio of the area of \(\triangle DME\) to the area of \(\triangle ABC\).”}
      \item \rubricfont\textsf{“In a chemical reaction at pH 1, an unknown substance was added that changed the pH to 4 and slowed down the reaction. What could have caused this?”}
      \item \rubricfont\textsf{“If a star 20 light-years away explodes, would gravitational waves reach Earth faster than light?”}
  \end{itemize}
  
  \item \rubricfont\textsf{\textbf{Very low.} Tasks that require knowledge in applied sciences and professions typically acquired through elementary school education. \textbf{Examples}:}
  \begin{itemize}[noitemsep,topsep=0pt]
      \item \rubricfont\textsf{Basic personal hygiene and hand washing procedures.}
      \item \rubricfont\textsf{Common road signs and traffic signals.}
      \item \rubricfont\textsf{Basic safety rules at home and school.}
  \end{itemize}
  
  \item \rubricfont\textsf{\textbf{Low.} Tasks that require knowledge in applied sciences and professions typically acquired through middle school education. \textbf{Examples}:}
  \begin{itemize}[noitemsep,topsep=0pt]
      \item \rubricfont\textsf{The use of basic measurement tools (thermometer, ruler, scale).}
      \item \rubricfont\textsf{Basic principles of personal finance and saving.}
      \item \rubricfont\textsf{Simple first aid for minor injuries.}
  \end{itemize}
  
  \item \rubricfont\textsf{\textbf{Intermediate.} Tasks that require knowledge in applied sciences and professions typically acquired through high school education. \textbf{Examples}:}
  \begin{itemize}[noitemsep,topsep=0pt]
      \item \rubricfont\textsf{Common legal terms (plaintiff, defendant, contract).}
      \item \rubricfont\textsf{Basic business concepts (budget, profit, loss).}
      \item \rubricfont\textsf{Fundamental principles of agricultural science.}
  \end{itemize}
  
  \item \rubricfont\textsf{\textbf{High.} Tasks that require knowledge in applied sciences and professions typically acquired through undergraduate education. \textbf{Examples}:}
  \begin{itemize}[noitemsep,topsep=0pt]
      \item \rubricfont\textsf{Basic principles of bridge design in civil engineering.}
      \item \rubricfont\textsf{Core concepts of supply chain management.}
      \item \rubricfont\textsf{Fundamentals of clinical assessment in healthcare.}
  \end{itemize}
  
  \item[\color{mytitleKN}\rubricfont\textsf{\textbf{Level }}\color{mytitleKN}\rubricfont\textsf{\textbf{5+}}]  \rubricfont\textsf{\textbf{Very High.} Tasks that require knowledge in applied sciences and professions typically acquired through graduate education or beyond. \textbf{Examples}:}
  \begin{itemize}[noitemsep,topsep=0pt]
      \item \rubricfont\textsf{Gene therapy techniques in precision medicine.}
      \item \rubricfont\textsf{Engineering principles of nuclear fusion reactor design.}
      \item \rubricfont\textsf{Legal frameworks for regulating artificial intelligence systems.}
  \end{itemize}

\end{enumerate}
\end{tcolorbox}

\newpage

\begin{tcolorbox}[
    colback=mybg,                     
    colframe=mytitleKN,                 
    colbacktitle=mytitleKN,            
    coltitle=white,                   
    fonttitle=\sffamily\small,        
    title=R5. Customary Everyday Knowledge (KNc),
    boxrule=0pt,                      
    arc=1mm,                         
    left=4pt, right=6pt, top=4pt, bottom=4pt,
]
\rubricfont\textsf{The following rubric is designed to annotate the conceptual sophistication level of tasks based solely on the depth of customary everyday knowledge required. This knowledge encompasses information that most people in a given society typically acquire through daily life experiences, social interactions, and exposure to popular media, rather than through formal education or specialized training. This does not include specialized knowledge from the natural sciences, social sciences, humanities, formal sciences, or applied sciences and professions. This reflects only the depth of customary everyday knowledge required.}
\end{tcolorbox}

\begin{tcolorbox}[
    colback=mybg,
    colframe=mytitleKN,
    colbacktitle=mytitleKN,
    coltitle=white,
    fonttitle=\sffamily\small,
    title=Levels,
    boxrule=0pt,
    arc=1mm,
    left=4pt, right=6pt, top=4pt, bottom=4pt
]
\begin{enumerate}[start=0, itemsep=0pt,
  label={\color{mytitleKN}\rubricfont\textsf{\textbf{Level }}\color{mytitleKN}\rubricfont\textsf{\textbf{\arabic*}}}]

  \item \rubricfont\textsf{\textbf{None.} Tasks do not require any customary everyday knowledge. \textbf{Examples}:}
  \begin{itemize}[noitemsep,topsep=0pt]
      \item \rubricfont\textsf{Looking in a mirror and checking if there’s any dirt in your face.}
      \item \rubricfont\textsf{Recognizing that two objects are of the same color.}
      \item \rubricfont\textsf{Basic arithmetic calculations.}
  \end{itemize}
  
  \item \rubricfont\textsf{\textbf{Very low.} Tasks that require basic knowledge universally shared within a society. \textbf{Examples}:}
  \begin{itemize}[noitemsep,topsep=0pt]
      \item \rubricfont\textsf{Knowing that “10h” may refer to the morning or to the evening.}
      \item \rubricfont\textsf{A brownie is better served as a dessert than an appetiser.}
      \item \rubricfont\textsf{Common objects in daily life (chairs, tables, cars).}
  \end{itemize}
  
  \item \rubricfont\textsf{\textbf{Low.} Tasks that require customary everyday knowledge typically possessed by most adults in a society. \textbf{Examples}:}
  \begin{itemize}[noitemsep,topsep=0pt]
      \item \rubricfont\textsf{Basic kitchen tools and their uses.}
      \item \rubricfont\textsf{Common traffic signs and their meanings.}
      \item \rubricfont\textsf{Major holidays in one’s culture.}
  \end{itemize}
  
  \item \rubricfont\textsf{\textbf{Intermediate.} Tasks that require general knowledge typically possessed by socially engaged members of society. \textbf{Examples}:}
  \begin{itemize}[noitemsep,topsep=0pt]
      \item \rubricfont\textsf{Different types of payment methods (cash, credit cards, digital wallets).}
      \item \rubricfont\textsf{Common technology features (touch screens, wireless connectivity, cloud storage).}
      \item \rubricfont\textsf{Standard retail practices (return policies, warranties, seasonal sales).}
  \end{itemize}
  
  \item \rubricfont\textsf{\textbf{High.} Tasks that require extensive everyday knowledge gained through active engagement in society. \textbf{Examples}:}
  \begin{itemize}[noitemsep,topsep=0pt]
      \item \rubricfont\textsf{Major generational trends in technology adoption (from landlines to smartphones).}
      \item \rubricfont\textsf{Common real estate concepts (mortgages, leases, property taxes).}
      \item \rubricfont\textsf{Dietary restrictions across different groups (religious, health-based, ethical).}
  \end{itemize}
  
  \item[\color{mytitleKN}\rubricfont\textsf{\textbf{Level }}\color{mytitleKN}\rubricfont\textsf{\textbf{5+}}]  \rubricfont\textsf{\textbf{Very High.} Tasks that require comprehensive everyday knowledge across diverse cultural and social contexts. \textbf{Examples}:}
  \begin{itemize}[noitemsep,topsep=0pt]
      \item \rubricfont\textsf{Gift-giving customs and taboos across different cultures.}
      \item \rubricfont\textsf{Business etiquette variations in major world regions.}
      \item \rubricfont\textsf{Dining customs and table manners in different societies.}
  \end{itemize}

\end{enumerate}
\end{tcolorbox}

\newpage

\subsection{Extraneous}

These dimensions do not reflect cognitive capabilities per se, but rather aspects of item design that may affect observed performance such as presentation and design.  They are critical because item design can artificially inflate or obscure performance scores. For example, if an item is unusually long, a model may perform poorly not because it lacks the relevant skill, but simply because it is overwhelmed by the amount of text. Similarly, if an item is highly prototypical or has been seen repeatedly in the training data, memorisation rather than genuine reasoning may lead to success. Finally, if a question is structured in such a way that a correct answer can be obtained by chance (for example, in a multiple-choice format), this introduces guessability. 

The following subsections describe each \textit{Extraneous} rubric in more detail.

\newpage

\begin{tcolorbox}[
    colback=mybg,                     
    colframe=mytitleVO,                 
    colbacktitle=mytitleVO,            
    coltitle=white,                   
    fonttitle=\sffamily\small,        
    title=Volume (VO),       
    boxrule=0pt,                      
    arc=1mm,                         
    left=4pt, right=6pt, top=4pt, bottom=4pt
]
\rubricfont\textsf{%
This rubric defines a scale that evaluates the task purely based on its \emph{volume}, i.e., the time a fully competent, experienced and motivated human would need to both read and complete the task in ideal conditions, not counting breaks or interruptions, regardless of the difficulty or cognitive demands of the task. The scale ranges from tasks requiring less than a second to those requiring more than 16 hours (1000 minutes), focusing purely on the volume of work rather than on cognitive complexity or skill requirements. Time estimates assume the task performer has all necessary information, tools, and skills readily available, but works autonomously without the assistance of other humans or AI tools.
}
\end{tcolorbox}

\begin{tcolorbox}[
    colback=mybg,                     
    colframe=mytitleVO,                 
    colbacktitle=mytitleVO,            
    coltitle=white,                   
    fonttitle=\sffamily\small,        
    title=Levels,       
    boxrule=0pt,                      
    arc=1mm,                         
    left=4pt, right=6pt, top=4pt, bottom=4pt
]
\begin{enumerate}[start=0, itemsep=0pt,
  label={\color{mytitleVO}\rubricfont\textsf{\textbf{Level }}\color{mytitleVO}\rubricfont\textsf{\textbf{\arabic*}}}]

  \item \rubricfont\textsf{\textbf{None.} Volume: Negligible, requiring less than 1 second. \textbf{Examples}:}
  \begin{itemize}[noitemsep,topsep=0pt]
      \item \rubricfont\textsf{Checking the status of an indicator (e.g., a light on a machine).}
      \item \rubricfont\textsf{Selecting a checkbox to confirm agreement.}
      \item \rubricfont\textsf{Opening a pre-configured app or program.}
  \end{itemize}
  
  \item \rubricfont\textsf{\textbf{Very Low.} Requiring between 1 second and 1 minute. \textbf{Examples}:}
  \begin{itemize}[noitemsep,topsep=0pt]
      \item \rubricfont\textsf{Read a short online comment about merchandise to identify whether it has positive, neutral, or negative emotion.}
      \item \rubricfont\textsf{Saving and uploading a single document to a pre-arranged folder.}
      \item \rubricfont\textsf{Reading a short email and writing a brief confirmation email (e.g., confirming attendance to a meeting).}
  \end{itemize}
  
  \item \rubricfont\textsf{\textbf{Low.} Requiring between 1 minute and 10 minutes. \textbf{Examples}:}
  \begin{itemize}[noitemsep,topsep=0pt]
      \item \rubricfont\textsf{Writing a simple summary in half a page of the main points from a short memo or meeting.}
      \item \rubricfont\textsf{Listening to an audio recording of one minute from a high-school level history class and answering a list of ten short factual questions.}
      \item \rubricfont\textsf{Reading an inquiry and writing a few paragraphs long personalized email in response to an inquiry.}
  \end{itemize}
  
  \item \rubricfont\textsf{\textbf{Intermediate.} Requiring between 10 minutes and 100 minutes.
 \textbf{Examples}:}
  \begin{itemize}[noitemsep,topsep=0pt]
      \item \rubricfont\textsf{Proofreading and lightly editing a 4-page research article to improve flow and clarity.}
      \item \rubricfont\textsf{Organizing and cataloging a personal collection of 50 books by genre, year, and author.}
      \item \rubricfont\textsf{Reading a set of instructions and data based on a small experiment, and then writing a 500-word report based on it.}
  \end{itemize}
  
  \item \rubricfont\textsf{\textbf{High.} Requiring between 100 minutes (roughly 1.5 hours) and 1,000 minutes (about 16 hours).
 \textbf{Examples}:}
  \begin{itemize}[noitemsep,topsep=0pt]
      \item \rubricfont\textsf{Creating a 10-page technical manual including screenshots, step-by-step procedures, and troubleshooting guides for a specific software application.}
      \item \rubricfont\textsf{Configuring 25 workstations with standardized software (5 applications per machine), security settings, and network access protocols.}
      \item \rubricfont\textsf{Reading a 12-page position paper on language models' impact on education and verbally summarizing the paper’s key insights for colleagues.}
  \end{itemize}

\end{enumerate}

\begin{enumerate}[itemsep=0pt]
  \item[\color{mytitleVO}\rubricfont\textsf{\textbf{Level }}\color{mytitleVO}\rubricfont\textsf{\textbf{5+}}] \rubricfont\textsf{\textbf{Very High.}  Requiring more than 1,000 minutes (roughly 16 hours).
 \textbf{Examples}:}
  \begin{itemize}[noitemsep,topsep=0pt]
      \item \rubricfont\textsf{Planning and executing a 3-day workshop for 100 attendees, including scheduling 15 speakers and arranging catering for 6 meals.}
      \item \rubricfont\textsf{Writing a high-quality 10-page research paper on the field of data science, requiring analysis of 50+ academic sources, including data visualization and statistical analysis of 3 datasets.}
      \item \rubricfont\textsf{Reviewing the correctness of 2000+ simple geography exercises written by high-school students.}
      \item \rubricfont\textsf{Conducting a financial audit covering 12 months of transactions (approximately 5,000 entries) across 5 department budgets, including reconciliation and variance analysis.}
  \end{itemize}
\end{enumerate}

\end{tcolorbox}

\newpage

\begin{tcolorbox}[
    colback=mybg,                     
    colframe=mytitleAT,                 
    colbacktitle=mytitleAT,            
    coltitle=white,                   
    fonttitle=\sffamily\small,        
    title=Atypicality (AT),       
    boxrule=0pt,                      
    arc=1mm,                         
    left=4pt, right=6pt, top=4pt, bottom=4pt,
]

\rubricfont\textsf{This rubric defines a scale that evaluates tasks based on how unlikely they appear in various sources (internet, textbooks, tests) and how unlikely the specific instance is to have been previously encountered and memorized. The scale ranges from exactly identical instances that are widely known to completely novel task formulations, focusing on the uniqueness of both the task type and the specific instance rather than its difficulty or complexity.}
\end{tcolorbox}

\begin{tcolorbox}[
    colback=mybg,                     
    colframe=mytitleAT,                 
    colbacktitle=mytitleAT,            
    coltitle=white,                   
    fonttitle=\sffamily\small,        
    title=Levels,       
    boxrule=0pt,                      
    arc=1mm,                         
    left=4pt, right=6pt, top=4pt, bottom=4pt
]%
\begin{enumerate}[start=0, itemsep=0pt, 
  label={\color{mytitleAT}\rubricfont\textsf{\textbf{Level }}\color{mytitleAT}\rubricfont\textsf{\textbf{\arabic*}}}]

\item \rubricfont\textsf{\textbf{None.} The task is a staple one. Exactly the same instance of the task appears many times on the Internet, textbooks or common psychological or achievement tests, and the solution is generally well-known and memorized. \textbf{Examples}:}
\begin{itemize}[noitemsep,topsep=0pt]
    \item \rubricfont\textsf{"What is 2 + 2?"}
    \item \rubricfont\textsf{"Name the capital of France."}
    \item \rubricfont\textsf{"What gets wetter and wetter the more it dries?"}
\end{itemize}

\item \rubricfont\textsf{\textbf{Very low.} The task is very common and the specific task instance is likely to frequently appear on the Internet, textbooks or common psychological or achievement tests, so the chance that the solution is well-known and memorized is high. \textbf{Examples}:}
\begin{itemize}[noitemsep,topsep=0pt]
    \item \rubricfont\textsf{“What is the derivative of sin(x)?”}
    \item \rubricfont\textsf{“Define opportunity cost.”}
    \item \rubricfont\textsf{“Name the seven continents.”}
\end{itemize}

\item \rubricfont\textsf{\textbf{Low.} The task is moderately common and the specific task instance varies somewhat from other common examples or is unlikely to have seen it before in exactly the same form, but possibly in variations. \textbf{Examples}:}
\begin{itemize}[noitemsep,topsep=0pt]
    \item \rubricfont\textsf{"What is 21251 + 2835?"}
    \item \rubricfont\textsf{"Given the molecular SMILES:
    \texttt{COC[C@H]1OC(=O)c2coc3c2[C@@]1(C)C1=C(C3=O)[C@@H]2CCC(=O)[C@@]2(C)C[C@H]1OC(C)=O}, your task is to provide the detailed description of the molecule using your experienced chemical Molecular knowledge.”}
    \item \rubricfont\textsf{"Solve the following Math Olympiad question: Determine the greatest real number \( C \), such that for every positive integer \( n\ge 2 \), there exists \( x_1, x_2, \dots, x_n \in [-1,1] \) so that \( \prod_{1\le i<j\le n}(x_i-x_j) \ge C^{\frac{n(n-1)}{2}} \)."}
\end{itemize}

\item \rubricfont\textsf{\textbf{Intermediate.} The task is somewhat common but the specific task instance is quite rare, and it is unlikely to appear in common sources (Internet, textbooks or common psychological or achievement tests). \textbf{Examples:}}
\begin{itemize}[noitemsep,topsep=0pt]
    \item \rubricfont\textsf{"What is 5205175017521571 + 68270867426872052?"}
    \item \rubricfont\textsf{“Among the following exoplanets, which one has the lowest density? a) An Earth-mass and Earth-radius planet. b) A planet with 3 Earth masses and a density of approximately 4.6 g/cm\(^3\). c) A planet with the same composition as Earth but 1.5 times more massive than Earth. d) A planet with the same composition as Earth but half the mass of Earth.”}
    \item \rubricfont\textsf{"Get answers for the question based on the context, where answers derived from substrings in the context or categorized as [unanswerable]. Context: ['On May 1 , 2015 , Quentin announced his retirement .Quentin signed a minor league deal with the Seattle Mariners on April 22 , 2015 , and was assigned to the Tacoma Rainiers .On April 5 , 2015 , Quentin was traded to the Atlanta Braves along with Cameron Maybin , Matt Wisler , and Jordan Paroubeck , for Craig Kimbrel and Melvin Upton Jr . The Braves designated him for assignment later that day , and released him on April 14 .', 'On July 22 , 2012 , Quentin agreed to a three-year , \$27 million contract extension through 2015 with a \$10 million mutual option for 2016 , including a no-trade clause . This is an amazing opportunity to stay and play in the city I grew up in . said Quentin .'] Question: Who did Carlos Quentin work for in April 2016?"}
\end{itemize}

\item \rubricfont\textsf{\textbf{High.} The task is not extremely uncommon and the specific task instance is infrequent or presented in notably different ways from standard formulations. \textbf{Examples}:}
\begin{itemize}[noitemsep,topsep=0pt]
    \item \rubricfont\textsf{"Create a measurement system where accuracy is expressed through different shapes rather than decimal places."}
    \item \rubricfont\textsf{"Assume that there exist only two types of people: knights and knaves. Knights always tell the truth, while knaves always lie. You are given the statements from 5 characters. Based on their statements, infer who is a knight and who is a knave. A: B is a liar and D is a truth-teller. B: D is a truth-teller. C: If A is a truth-teller, then E is a liar. D: B is a truth-teller and A is a truth-teller. E: A is a liar."}
    \item \rubricfont\textsf{“In the context of neurolinguistic processing models examining the interface between phonological working memory and syntactic parsing during real-time sentence comprehension, what is the primary anatomical structure that shows increased metabolic activity during novel word acquisition in fMRI studies?” [This is a very sophisticated way of asking “which part of the brain lights up when we learn new words?”}
\end{itemize}

 \item[\color{mytitleAT}\rubricfont\textsf{\textbf{Level }}\color{mytitleAT}\rubricfont\textsf{\textbf{5+}}]  \rubricfont\textsf{\textbf{Very High.}The task is fundamentally different from those typically appearing on the Internet, textbooks or psychological or achievement tests, or, the specific task instance is very unlikely to have close analogues in those sources. Noteworthy, any tasks (simple or elaborate) that can be found in standard tests or benchmarks should be considered less than level 5. \textbf{Examples}:}
\begin{itemize}[noitemsep,topsep=0pt]
    \item \rubricfont\textsf{"Take 20 major sky constellations and design 20 Formula1 race circuits that follow the FIA regulations but mimic the shapes of the constellations.”}
    \item \rubricfont\textsf{“Write a poem in fifty African languages, with each line in one language, with the number of letters e in the line being proportional to the speakers of those languages"}
     \item \rubricfont\textsf{"List mathematical terms that are also dancing terms (like 'step function' or 'rotation'), then write dance instructions using only mathematical language. However, the dance must stem from an Asian country since I'm teaching a creative Asian dance course today to one student who happens to be a mathematician."}
\end{itemize}

\end{enumerate}
\end{tcolorbox}

\newpage

\begin{tcolorbox}[
    colback=mybg,                     
    colframe=mytitleUG,                 
    colbacktitle=mytitleUG,            
    coltitle=white,                   
    fonttitle=\sffamily\small,        
    title= Unguessability (UG),
    boxrule=0pt,                      
    arc=1mm,                         
    left=4pt, right=6pt, top=4pt, bottom=4pt,
]
\rubricfont\textsf{This rubric classifies questions based on their answer format, determining whether they are multiple-choice (explicit or implicit) or open-ended. The output is either an integer representing the number of choices (for multiple-choice questions) or the word ``open'' (for open-ended questions).}
\end{tcolorbox}

\begin{tcolorbox}[
    colback=mybg,                     
    colframe=mytitleUG,                 
    colbacktitle=mytitleUG,            
    coltitle=white,                   
    fonttitle=\sffamily\small,        
    title= Answer Format Classification Prompt,
    boxrule=0pt,                      
    arc=1mm,                         
    left=4pt, right=6pt, top=4pt, bottom=4pt,
]

\rubricfont\textsf{\noindent QUESTION: \{Question\}}

\rubricfont\textsf{\noindent REFERENCE ANSWER: \{Answer\}}

\bigskip

\rubricfont\textsf{\noindent You are tasked with analyzing the question and its reference answer above, and classifying them based on their answer format. To this end, you need to determine if it's multiple-choice (explicit or implicit) or open-ended. Output a single integer representing the number of possible choices in an open-ended question, or "open" for an open-ended question.}

\end{tcolorbox}

\begin{tcolorbox}[
    colback=mybg,                     
    colframe=mytitleUG,                 
    colbacktitle=mytitleUG,            
    coltitle=white,                   
    fonttitle=\sffamily\small,        
    title= Classification Rules,
    boxrule=0pt,                      
    arc=1mm,                         
    left=4pt, right=6pt, top=4pt, bottom=4pt,
]

\begin{enumerate}[start=1, itemsep=0pt, 
  label={\color{mytitleUG}\rubricfont\color{mytitleUG}\rubricfont\textsf{\textbf{\arabic*}}}]

\item \rubricfont\textsf{\textbf{Explicit Multiple Choice Questions.} \textbf{Examples}:}

    \begin{itemize}[noitemsep,topsep=0pt]
        \item \rubricfont\textsf{``Which color is best: Red, Blue or Green?'' $\rightarrow$ Output: 3}
        \item \rubricfont\textsf{``Choose from: A) Earth B) Mars C) Venus D) Jupiter'' $\rightarrow$ Output: 4}
        \item \rubricfont\textsf{``Which of these explanations best describes photosynthesis? [followed by 4 detailed explanations]'' $\rightarrow$ Output: 4}
    \end{itemize}

\item \rubricfont\textsf{\textbf{Implicit Multiple Choice Questions (ONLY for well-known or given sets).
} }

    \begin{itemize}[noitemsep,topsep=0pt]
        \item \rubricfont\textsf{Yes/No Questions $\rightarrow$ Output: 2}
        \item \rubricfont\textsf{True/False questions $\rightarrow$ Output: 2}
        \item \rubricfont\textsf{Questions using well-known or given sets. Examples include, but not limited to:}
            \begin{itemize}[noitemsep,topsep=0pt]
            \item \rubricfont\textsf{Days of the week $\rightarrow$ Output: 7}
            \item \rubricfont\textsf{Months of the year $\rightarrow$ Output: 12}
            \item \rubricfont\textsf{Continents $\rightarrow$ Output: 7}
            \item \rubricfont\textsf{Cardinal directions (N/S/E/W) $\rightarrow$ Output: 4}
            \end{itemize}   
    \end{itemize}

\rubricfont\textsf{\textbf{Examples}:}

\begin{itemize}[noitemsep,topsep=0pt]
    \item \rubricfont\textsf{``What day of the week does the event start?'' $\rightarrow$ Output: 7}
    \item \rubricfont\textsf{``Which season is warmest?'' $\rightarrow$ Output: 4}
    \item \rubricfont\textsf{``Is this statement correct?'' $\rightarrow$ Output: 2}
\end{itemize}

\item \rubricfont\textsf{\textbf{ Open-ended Questions.} }

\begin{itemize}[noitemsep,topsep=0pt]
    \item \rubricfont\textsf{Questions requiring free-form responses. \textbf{Example}:}
        \begin{itemize}[noitemsep,topsep=0pt]
            \item \rubricfont\textsf{"Explain why the sky is blue."  $\rightarrow$ Output: open}
        \end{itemize}
    \item \rubricfont\textsf{Questions with no explicit options provided. \textbf{Example}:}
            \begin{itemize}[noitemsep,topsep=0pt]
            \item \rubricfont\textsf{``What factors contributed to the Industrial Revolution?'' $\rightarrow$  Output: open}
        \end{itemize}
    \item \rubricfont\textsf{Questions where options must be discovered or deduced as part of solving the problem. \textbf{Example}:}
            \begin{itemize}[noitemsep,topsep=0pt]
            \item \rubricfont\textsf{``What city with over 1M inhabitants is furthest east in Spain?'' $\rightarrow$ Output: open}
        \end{itemize}
    \item \rubricfont\textsf{Questions with finite but non-obvious sets of answers. \textbf{Example}:}
            \begin{itemize}[noitemsep,topsep=0pt]
            \item \rubricfont\textsf{``Which company will become the most valuable by 2030?'' $\rightarrow$ Output: open}
        \end{itemize}

\end{itemize}

\end{enumerate}

\end{tcolorbox}

\begin{tcolorbox}[
    colback=mybg,                     
    colframe=mytitleUG,                 
    colbacktitle=mytitleUG,            
    coltitle=white,                   
    fonttitle=\sffamily\small,        
    title= Instructions for Processing,
    boxrule=0pt,                      
    arc=1mm,                         
    left=4pt, right=6pt, top=4pt, bottom=4pt,
]

\begin{enumerate}[start=1, itemsep=0pt, 
  label={\color{mytitleUG}\rubricfont\color{mytitleUG}\rubricfont\textsf{\textbf{\arabic*}}}]
  \item \rubricfont\textsf{Read the question carefully.}
    \item \rubricfont\textsf{Check if options are explicitly provided or if the question uses a well-known or given set.}
    \item \rubricfont\textsf{If neither, classify as open-ended even if the set of possible answers is finite.}
    \item \rubricfont\textsf{Output ONLY the number or "open" without any explanation.}
\end{enumerate}

\rubricfont\textsf{Note: Only classify as multiple-choice if:}
\begin{itemize}[noitemsep,topsep=0pt]
    \item \rubricfont\textsf{The options are explicitly listed in the question, OR}
    \item \rubricfont\textsf{The options come from a fine (well-known or given) set like days of the week.}
\end{itemize}

\bigskip

\rubricfont\textsf{Format your output as a single line containing either:}
\begin{itemize}[noitemsep,topsep=0pt]
    \item \rubricfont\textsf{An integer (for multiple-choice questions) and nothing else.}
    \item \rubricfont\textsf{The word "open" (for open-ended questions) and nothing else.}
\end{itemize}
\bigskip

\rubricfont\textsf{Output:}

\end{tcolorbox}

\end{document}